%% file: iclr2026_conference.tex
\documentclass{article} 
\usepackage{iclr2026_conference,times}

\input{math_commands.tex}

\usepackage{hyperref}
\usepackage{url}
\usepackage[table]{xcolor}
\usepackage{graphicx}
\usepackage{subfigure}
\usepackage{booktabs}
\usepackage{calc}      
\usepackage{booktabs}  

\usepackage{algorithm}
\usepackage{algorithmic}
\usepackage{amsmath}
\usepackage{amssymb} 
\usepackage{mathtools}
\usepackage{amsthm}

\usepackage[capitalize,noabbrev]{cleveref}

\theoremstyle{plain}
\newtheorem{theorem}{Theorem}[section]
\newtheorem{proposition}[theorem]{Proposition}

\theoremstyle{definition}

\theoremstyle{remark}

\usepackage[textsize=tiny]{todonotes}

\newcommand{\OUTPUT}{\STATE \textbf{Output:} }

\title{Adaptive Conformal Anomaly Detection with Time Series Foundation Models for Signal Monitoring}

\author{Natalia Martinez Gil \quad
Fearghal O'Donncha \quad
Wesley M.~Gifford \quad
\textbf{Nianjun Zhou} \\
\textbf{Dhaval C. Patel} \quad
\textbf{Roman Vaculin} \\
\\
IBM Research \\
\texttt{natalia.martinez.gil@ibm.com}
}

\iclrfinalcopy 
\begin{document}

\maketitle

\begin{abstract}
We propose a post-hoc adaptive conformal anomaly detection method for monitoring time series that leverages predictions from pre-trained foundation models without requiring additional fine-tuning. Our method yields an interpretable anomaly score directly interpretable as a false alarm rate (p-value), facilitating transparent and actionable decision-making. It employs weighted quantile conformal prediction bounds and adaptively learns optimal weighting parameters from past predictions, enabling calibration under distribution shifts and stable false alarm control, while preserving out-of-sample guarantees. As a model-agnostic solution, it integrates seamlessly with foundation models and supports rapid deployment in resource-constrained environments. This approach addresses key industrial challenges such as limited data availability, lack of training expertise, and the need for immediate inference, while taking advantage of the growing accessibility of time series foundation models. Experiments on both synthetic and real-world datasets show that the proposed approach delivers strong performance, combining simplicity, interpretability, robustness, and adaptivity. \footnote{Code: \url{https://github.com/ibm-granite/granite-tsfm/tree/main/notebooks/hfdemo/adaptive_conformal_tsad}}
\end{abstract}

\input{ICLR2026/sections/introduction}
\input{ICLR2026/sections/related_work}
\input{ICLR2026/sections/problem_statement}

\input{ICLR2026/sections/method}
\input{ICLR2026/sections/experiments}
\input{ICLR2026/sections/conclusion}




\bibliography{iclr2026_conference}
\bibliographystyle{iclr2026_conference}

\appendix

\include{ICLR2026/sections/appendix_experiments}

\end{document}

%% file: math_commands.tex
\usepackage{amsmath,amsfonts,bm}

\def\eqref#1{equation~\ref{#1}}

\def\1{\bm{1}}

\DeclareMathAlphabet{\mathsfit}{\encodingdefault}{\sfdefault}{m}{sl}
\SetMathAlphabet{\mathsfit}{bold}{\encodingdefault}{\sfdefault}{bx}{n}



%% file: ICLR2026/sections/introduction.tex
\section{Introduction}

A common challenge in industrial applications such as predictive maintenance and signal monitoring is the scarcity of sufficient quality data and infrastructure to train robust models  \cite{cook2019anomaly,ajami2012data,kanawaday2017machine,beghi2016data,shah2018anomaly,moghaddass2017hierarchical}. This limitation can hinder the ability to make accurate and reliable predictions, which are essential to detect anomalies and ensure operational efficiency. 
Foundation models, particularly in the time series domain \cite{liang2024foundation}, offer a promising solution. These models excel at leveraging prior knowledge and historical observations, enabling them to provide good enough initial estimates of expected values and statistical characteristics of monitored signals, even in data-scarce environments. This capability is invaluable for industries aiming to enhance their monitoring systems without the need for extensive datasets.


In the context of time series anomaly detection, an adaptive approach is crucial for monitoring and maintaining the reliability of signals. Anomalies, or deviations from expected behavior, can manifest in different forms, such as point anomalies, where an individual observation significantly deviates from normal patterns, and contextual anomalies, where a value is only considered anomalous within a specific temporal context \cite{boniol2024dive}. Detecting these  effectively requires models that capture underlying temporal dependencies and adapt to non-stationary data distributions.

A prominent class of anomaly detection methods relies on predictive modeling, where a forecasting model learns normal time series behavior, and deviations between predicted and actual values could indicate anomalies in operations or shifts in operational modes that require expert attention \cite{basseville1993detection,choudhary2017runtime,gama2014survey,saurav2018online}. However, many existing approaches assume access to large amounts of training data, making them impractical in settings where only a few samples are initially available. This motivates the use of pretrained Time Series Foundation Models (TSFMs) \cite{rasul2023lag, rasul2024lag, ansari2024chronos, liang2024foundation}, which have been trained on large-scale datasets and can generalize to new time series with minimal adaptation. Furthermore, existing anomaly detection systems often lack interpretability, relying on thresholding mechanisms that assume a fixed data distribution \cite{schmidl2022anomaly, paparrizos2022tsb, goswami2022unsupervised}, which limits their adaptability to evolving time series data. In this setting, a robust system must balance sensitivity and adaptability, minimizing false alarms while effectively detecting significant behavioral transitions. This ensures timely identification of suspicious patterns without overwhelming experts with noise, fostering a more efficient and reliable monitoring framework \cite{cook2019anomaly}.

To address these limitations, we propose a conformal-based anomaly detection method that integrates the predictions of pretrained TSFMs with conformal prediction techniques \cite{vovk2005algorithmic, angelopoulos2021gentle} to produce an interpretable, adaptive anomaly score directly linked to a desired alarm rate. Conformal methods offer model-agnostic and distribution-free uncertainty quantification with finite-sample guarantees, making them highly suitable for real-world anomaly detection. However, standard conformal approaches rely on the assumption of exchangeability, which is often violated in time series due to temporal dependencies. Furthermore, existing conformal methods for anomaly detection primarily focus on thresholding arbitrary anomaly scores derived from non-anomalous data while assuming exchangeability \cite{angelopoulos2021gentle, guan2019conformal, bates2023testing}, limiting their applicability in dynamic, non-stationary settings.

\begin{figure*}[t]
  \centering

  \begin{minipage}[t]{0.35\textwidth}
    \centering
    \subfigure[Adaptive Anomaly Score Pipeline.\label{fig:triad-a}]{
      \includegraphics[width=\linewidth]{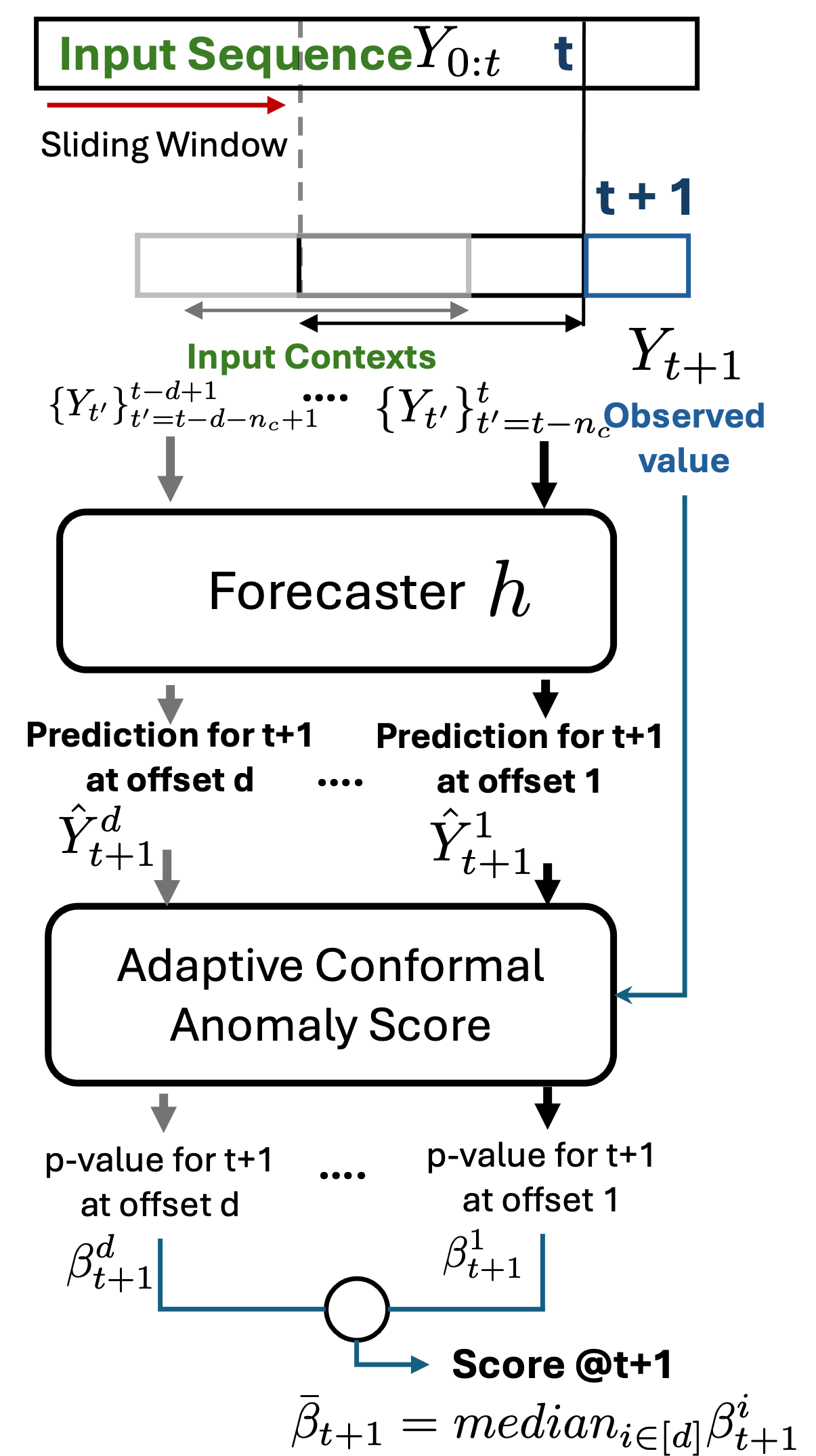}
    }
  \end{minipage}
  \begin{minipage}[t]{0.6\textwidth}
    \centering
    \subfigure[Anomaly Detection Example.\label{fig:triad-c}]{
      \includegraphics[width=0.95\linewidth]{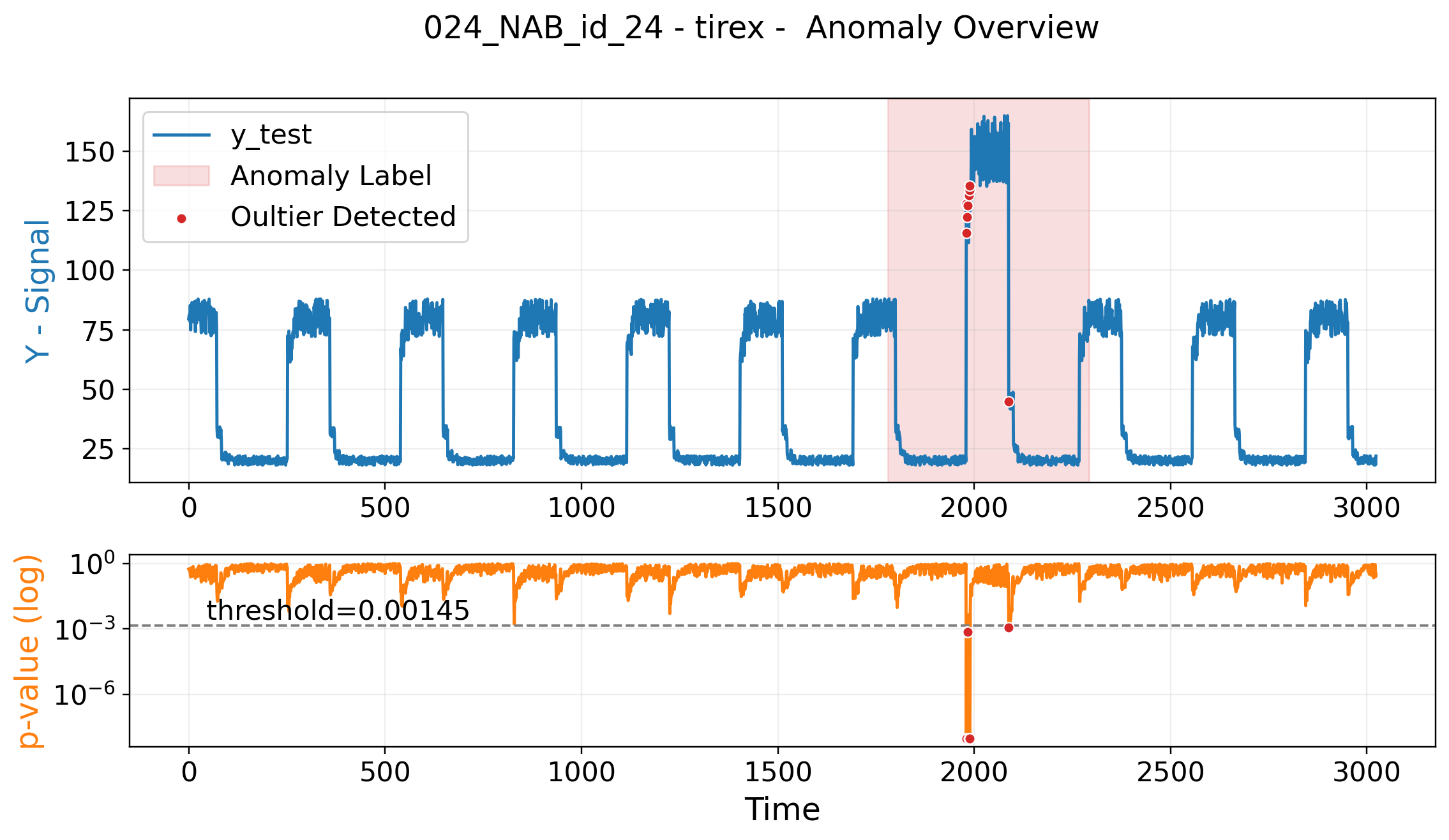}
    } \\
    \subfigure[Non-conformity Scores vs Conformal Weights.\label{fig:triad-b}]{
      \includegraphics[width=0.95\linewidth]{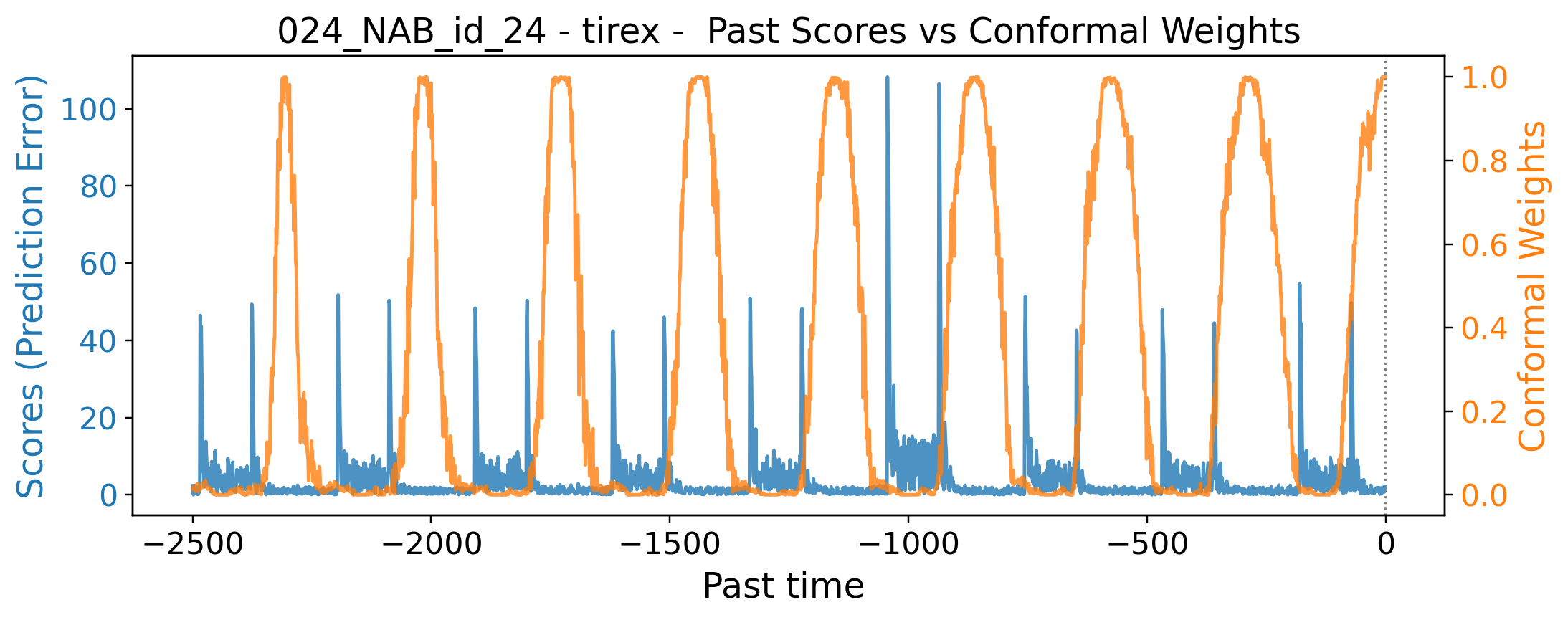}
    } 
  \end{minipage}

\caption{Illustration of our proposed $\mathcal{W}_1$\textsc{-ACAS} method. 
(a) Anomaly scoring pipeline: conformal $p$-values are computed across forecast horizons from forecaster errors and aggregated. The mapping is adapted online by weighting past nonconformity scores, with weights evolving to capture distributional shifts or recurring patterns. 
(b) Example signal (blue) with ground-truth anomaly labels, where detected outliers (red dots) occur when adaptive $p$-values (orange) fall below a threshold. 
(c) Converged adaptive weights (orange) over past errors (blue), averaged across horizons, shows how $\mathcal{W}_1$\textsc{-ACAS} captures error patterns with similar distributions, here reflecting its periodic behavior.}
\label{fig:triad}
\end{figure*}

\paragraph{Main Contributions}
We propose $\mathcal{W}_1$\textsc{-ACAS}, a post-hoc adaptive conformal anomaly detection framework that leverages predictions from pretrained forecasters (e.g., TSFMs) to monitor signals without requiring fine-tuning. This is particularly valuable in industrial settings, where users often lack sufficient data, data-cleaning pipelines, or specialized expertise \cite{cook2019anomaly}. Our approach provides a practical solution for immediate anomaly monitoring. Figure~\ref{fig:triad} illustrates the method: (a) anomaly scores are derived as conformal $p$-values from forecaster errors across multiple horizons and aggregated into a single score; (b) anomalies are flagged when adaptive $p$-values fall below a threshold on real signals; and (c) the learned adaptive weights emphasize past errors with similar distributions, capturing recurring patterns such as periodicity, thereby improving detection while offering direct control over the alarm rate. Our framework offers the following properties:



\begin{itemize}
    \item \textbf{Interpretability:} The anomaly score corresponds directly to an alarm rate ($p$-value), providing a transparent and probabilistic basis for decisions.  
    \item \textbf{Distribution-Agnostic:} Built on quantile conformal prediction, the method is robust to heavy-tailed and complex error distributions.  
    \item \textbf{Adaptivity:} By weighting past nonconformity scores via the Wasserstein distance, the framework adapts online to distribution shifts, reducing false alarms while preserving calibration \cite{barber2023conformal}.
    \item \textbf{Post-Hoc and Model-Agnostic:} The method applies directly to pretrained TSFMs or any anomaly score, requiring no retraining while inheriting the guarantees of weighted conformal prediction. Its effectiveness is proved through integration with TSFM forecasters.

\end{itemize}

%% file: ICLR2026/sections/related_work.tex
\section{Related Work}

\paragraph{Time Series Anomaly Detection}
Prediction-based methods detect anomalies by comparing observed values against forecasts \citep{giannoni2018anomaly,boniol2024dive}. Recent TSFMs \citep{rasul2023lag,rasul2024lag,ansari2024chronos,liang2024foundation} are well suited for online detection in data-scarce scenarios, offering accurate zero-shot forecasting performance. Recent benchmark studies \citep{paparrizos2022tsb,liu2024elephant} show that classical distance- and density-based methods \citep{li2007unifying,ramaswamy2000efficient,aggarwal2017introduction,paparrizos2015k,paparrizos2017fast,boniol2021sand} often outperform more complex models, but they typically require access to the full dataset (non-causal), lack robustness across temporal patterns, and are unsuitable for streaming settings. Moreover, many anomaly scores lack clear probabilistic meaning, and common thresholding strategies rely on full-dataset statistics \citep{ahmad2017unsupervised}, limiting real-time applicability. In practice, anomaly detection systems must not only achieve high accuracy but also provide interpretable confidence scores while maintaining low false alarm rates \citep{cook2019anomaly}. Our work addresses these challenges by combining TSFMs with adaptive conformal scoring, yielding interpretable and calibrated thresholds for reliable streaming anomaly detection.

\paragraph{Conformal Prediction.} 
Conformal prediction provides distribution-free uncertainty quantification with finite-sample guarantees \citep{vovk2005algorithmic,shafer2008tutorial,angelopoulos2021gentle}. A widely used variant, split conformal prediction (SCP) \citep{papadopoulos2002inductive}, is post-hoc and model-agnostic, relying only on model predictions and a calibration set. While effective under exchangeability \footnote{informally, a sequence of observations is exchangeable if any permutation of the observations has the same joint probability}, this assumption is often violated in time series settings, motivating adaptive extensions. Recent works \citep{gibbs2021adaptive,zaffran2022adaptive,gibbs2024conformal} adjust conformal quantiles online to handle distribution shifts, but typically optimize for a single error rate.  Weighted conformal methods offer adaptation by reweighting calibration or past scores based on some notion of similarity to new observations \citep{lei2014distribution,guan2019conformal,tibshirani2019conformal,sesia2021conformal,han2022split,guan2023localized,ghosh2023improving,mao2024valid} improving local coverage. Bounds for non-exchangeable sequences \citep{barber2023conformal} further suggest emphasizing calibration samples that are nearly exchangeable with the test point. This motivates our approach, which leverages weighted adaptive conformal quantiles to remain calibrated across time. Conformal prediction has also been applied to anomaly detection by thresholding arbitrary anomaly scores under exchangeability \citep{angelopoulos2021gentle,guan2019conformal,bates2023testing}. However, existing methods do not simultaneously provide interpretable, distribution-agnostic anomaly scores, directly control alarm rates, and adapt robustly to non-exchangeable time series. Our work addresses this gap by developing a conformal anomaly detection framework that is both interpretable and resilient to real-world distribution shifts.

%% file: ICLR2026/sections/problem_statement.tex
\section{Background}
\label{sec:background}

Consider \( S \in \mathbb{R} \) a nonconformity score variable that quantifies the performance of a predictive model \( h: \mathcal{X} \to \hat{\mathcal{Y}} \) on a joint distribution \( P_{X, Y} \) using a nonconformity function \( e: \mathcal{Y} \times \hat{\mathcal{Y}} \to \mathbb{R} \). The input \( X \in \mathcal{X} \) represents the model's input space, \( Y \in \mathcal{Y} \) denotes the true target variable, and \( \hat{\mathcal{Y}} \) corresponds to the output space of the model, which may include predictions or derived statistics over \( Y \). The nonconformity function \( e \) measures the degree of disagreement between the true target and the model's predictions, enabling $S = e(Y,h(X))$ to capture how atypical a prediction is within the given distribution. An example of a nonconformity function for a point prediction model is  absolute error $e(Y,\hat{Y}) = |Y - \hat{Y}|$.

\subsection{Conformal Outlier Detection.} In the context of anomaly detection we characterize the distribution of the non-conformity score variable $S \sim  P_{S}$ where $S = e(Y,h(X)) \in \mathbb{R}$ under non-anomalous conditions $X,Y \sim P_{X,Y}$. \footnote{Although 
$\mathcal{X}$ and $\mathcal{Y}$ are treated as separate spaces, they may overlap, as in reconstruction-error-based scores where $Y = X$.} Observations are flagged as outliers (or anomalies) when the composition of the nonconformity function $e$ and the predictive model $h$ produces unusually high scores.\footnote{Unusually low scores can be handled similarly, nonconformity scores need not be positive} Given a significance level 
$\alpha$, which controls the tolerated false positive rate, an anomaly detection function 
$C_{\alpha}:\mathcal{X,Y} \rightarrow \{0,1\}$ should satisfy the following property:
\begin{equation}
    \mathbb{P}(C_{\alpha}(X_{n+1},Y_{n+1}) = 1) \le \alpha
\label{ad_false_alarm_bound}
\end{equation}
where $\mathbb{P}$ is the probability over unseen test data sampled from the non-anomalous distribution, $X_{n+1},Y_{n+1} \sim P_{X,Y}$. In the standard split-conformal setting, we observe $\mathbf{s} = S_{1},\dots,S_{n}$ nonconformity scores derived from non-anomalous data, $S_i = e(Y_i,h(X_i))$ with $X_i,Y_i \sim P_{X,Y}$. Non-conformity scores need not be independent of each other; the following conformal anomaly detection function satisfies, under echangeability conditions\footnote{The sequence $S_{1},\dots, S_{n+1}$ is exchangeable if $P(S_{1},\dots, S_{n+1})=P(S_{\sigma(1)},\dots, S_{\sigma(n+1)})$ for any permutation $\sigma$}, the false positive bound in \eqref{ad_false_alarm_bound}:
\begin{equation}
\begin{array}{l}
    C_{\alpha}(X_{n+1},Y_{n+1}) = \mathbf{1}[S_{n+1} > \hat{q}_{\alpha}] , \quad 
    
    \hat{q}_{\alpha} = Q_{1-\alpha}(\sum^n_{i=1}\frac{1}{n+1}\delta_{S_i} + \frac{1}{n+1}\delta_{\infty} ).
\end{array}
\label{conformal_anomaly_detector}
\end{equation}
Here $\hat{q}_{\alpha}$ is the empirical conformal quantile, conservatively adjusted with a point mass at infinity. 
\paragraph{Conformal Outlier Detection Beyond Exchangeability}
To account for heterogeneity in the nonconformity scores across the input space or potential temporal drift, we consider the generalized weighted conformal quantile estimate $\hat{q}^{w}_{\alpha} = \mathbb{Q}_{1-\alpha}(\mathbf{s},\mathbf{w})$ defined as:
\begin{equation}
    \mathbb{Q}_{1-\alpha}(\mathbf{s},\mathbf{w}) =  Q_{1-\alpha}(\sum^n_{i=1}\frac{w_i}{||\mathbf{w}||_1 + 1}\delta_{S_i} + \frac{1}{||\mathbf{w}||_1 + 1}\delta_{\infty} ).
\label{weighted_conformal_quantile}
\end{equation}
where $\mathbf{w} = \{w_{i} \in [0,1]\}^n_{i=1}$ is a weighting vector applied to the calibration points. The standard result in Eq.~\ref{conformal_anomaly_detector} is recovered when $w_i=1, \forall i=1,\dots,n$. 

This weighted conformal quantile estimate produces a generalization of the conformal anomaly detector from \eqref{conformal_anomaly_detector}. This conformal anomaly detection has false alarm rate guarantees even in non-exchangeable settings as described in the following proposition \ref{prop:coverage_weighted_exchangeability}.

\begin{proposition}
    \textit{(Direct application of Theorem 2 and 3 in \cite{barber2023conformal}}
    Given $\alpha \in (0,1)$, $\mathbf{s}= \{S_{i}\}^{n+1}_{i=1}$ a set of non-conformity scores where $S_{n+1}$ corresponds to the test point, and a vector of weights $\mathbf{w} = \{w_{i} \in [0,1]\}^n_{i=1}$ for the previous $n$ observations the detector
    \begin{equation}
        A_{n+1} = C_{\alpha,\mathbf{w}}(X_{n+1},Y_{n+1}) = \mathbf{1}[S_{n+1} >\hat{q}^{\textbf{w}}_{\alpha}]
    \label{eq:anomaly_weighted_function}
    \end{equation}
    based on the weighted conformal quantile estimate in Eq.\ref{weighted_conformal_quantile} satisfies the false alarm rate guarantees 
    \begin{equation}
    \begin{array}{rl}
         \mathbb{P}(A_{n+1} = 1) & \le  \alpha + \sum^n_{i=1} \frac{w_i}{||\mathbf{w}||_1 + 1} d_{TV}(\textbf{s},{\textbf{s}}^i) \\
         &  > \alpha - \sum^n_{i=1} \frac{w_i}{||\mathbf{w}||_1 + 1} d_{TV}(\textbf{s},{\textbf{s}}^i) - \frac{1}{||\mathbf{w}||_1 + 1}.
    \end{array}
    \label{eq:weighted_coverage_bounds}
    \end{equation}
    Here $d_{TV}(\textbf{s},{\textbf{s}}^i)$ is the distance in total variation between the sequence ${\textbf{s}}$ ($n$ previously observed point and the test point $n+1$) and ${\textbf{s}}^i$ which denotes the sequence of non-conformity scores after swapping the test point $n+1$ with the $i$-th previously observation. The lower bound is valid under the assumption that the non-conformity scores take equal values with probability 0.
\label{prop:coverage_weighted_exchangeability}
\end{proposition}


Intuitively, Proposition \ref{prop:coverage_weighted_exchangeability} indicates that one would like to assign higher weights to previous observations that are, pairwise, most exchangeable with the test sample (i.e., $P(S_1,\dots, S_i, \dots, S_{n+1})\simeq P(S_1,\dots, S_{n+1}, \dots, S_{i})$, and lower weights otherwise. Additionally, the lower bound encourages the maximization of $||\mathbf{w}||_1$ and therefore keeping the weights as close to one as possible. One could decide $\mathbf{w}$ if given access to prior knowledge about the values or reasonable upper bounds of $d_{TV}(\textbf{s},{\textbf{s}}^i)$. In the context of time series, previous works such as \citet{barber2023conformal} have set $\mathbf{w}$ to exponentially decay with time ($w_{i}=\gamma^{n-i}$); in non-time-series settings, other works such as \citep{lei2014distribution,guan2019conformal,sesia2021conformal,han2022split,guan2023localized,ghosh2023improving,mao2024valid} decide the weights based on criteria such as distance in covariate space, or optimize them to guarantee a particular false positive rate coverage $\alpha$, \citep{han2022split,amoukou2023adaptive}. Next, we present our adaptive conformal score method, which learns $\mathbf{w}$ with the objective of providing scores that are calibrated for every feasible false alarm across time.


%% file: ICLR2026/sections/method.tex
\section{Adaptive Conformal Anomaly Score}
\label{sec:Adaptive Conformal Anomaly Score}
The conformal outlier detection framework provides a principled way to define a binary anomaly decision variable based on a preselected $\alpha$ with generalization guarantees. However, the underlying nonconformity score $S$ may not itself be an interpretable indicator of anomaly, particularly in sequential settings where its distribution may drift over time. To address this, we aim to learn an adaptive mapping that assigns each score an approximate probability of observing a more extreme value under prior (ideally normal) conditions, yielding a distribution-agnostic $p$-value estimate. Formally, we consider a time series setting with a sequence of nonconformity scores $S_{1}, \ldots, S_{t}$. In prediction-based anomaly detection, these are derived from a forecasting model $h:\mathbb{R}^{n_c \times n_f} \rightarrow \mathcal{Y}^{d}$, which maps a context of length $n_c$ with $n_f$ features to a $d$-step-ahead forecast $\hat{Y}^d_{t+1} = h_{d}(X_{t-n_c-d:t-d+1})$. The nonconformity score for sample $t+1$ at horizon $d$ is $S^{d}_{t+1} = |Y_{t+1} - \hat{Y}^d_{t+1}|$. For clarity, we omit the index $d$ in the following section, since the analysis applies independently to each prediction horizon, and reintroduce it later when needed.

\subsection{Conformal Anomaly Score}
We wish to learn a parametric mapping $\beta_{\textbf{w}}:\mathbb{R}, \mathbb{R}^{t} \rightarrow [0,1]$ of the previous nonconformity scores $\textbf{s} = \{S_{i}\}^t_{i=1}$ and the test sample $S_{t+1}$; this mapping $\beta_{\textbf{w}}$ should be such that it can be directly compared to any $\alpha$ threshold to produce an anomaly detector with the same false alarm rate guarantees as the one described in equations \eqref{ad_false_alarm_bound} and \eqref{conformal_anomaly_detector}. Given a set of non-conformity scores derived from past, ideally non-anomalous data \footnote{For sequences containing a known fraction of anomalous samples below some upper bound $\alpha'$, the derivation follows similarly, but the interpretation of $\beta_{\textbf{w}}(S_{t+1})$ is $\alpha+\alpha'$ where $\alpha$ is the lower bound of the p-value of the sample.}, their associated weights $\vec{w} = \{w_{i} \in [0,1]\}^t_{i=1}$, and a non-conformity score test sample $S_{t+1}$ we propose the following score normalization 
\begin{equation}
 \beta_{\textbf{w}}(S_{t+1}) = \sup\{ \alpha \in [0,1] : S_{t+1} \le \mathbb{Q}_{1-\alpha}(\textbf{s},\textbf{w})\}.
\label{conformal_p_value_transformation}
\end{equation}
Here $\beta_{\textbf{w}}(S_{t+1})$ can be interpreted as the weighted, conformalized p-value, $\beta_{\textbf{w}}(S_{t+1}) =\beta_{t+1}$ (we omit the explicit dependence on $\textbf{s}$ for clarity). The proposed function automatically maps an anomaly score $S$, which can take arbitrary real values, into a normalized score that directly relates to the desired false alarm rate. The decision of an anomaly detection threshold becomes interpretable for the end user (it directly translates into the desired false alarm level) and preserves the guarantees of the original conformal outlier detector as shown in Proposition \ref{prop:conformal_detector_marginal_guarantees}. 


\begin{proposition}
Given $\alpha \in [0,1]$, $\{S_{i}\}^{t+1}_{i=1}$ a set of exchangeable non-conformity scores, and their weights $\textbf{w} = \{w_{i} = \in [0,1]\}^t_{i=1}$ the detector $C_{\beta_{\textbf{w}}}(X_{t+1},Y_{t+1}) = \mathbf{1}[\beta_{\textbf{w}}(S_{t+1}) < \alpha]$ based on the $\beta_{\textbf{w}}(\cdot)$ mapping defined in \eqref{conformal_p_value_transformation} is equivalent to \eqref{eq:anomaly_weighted_function} and therefore satisfies the conformal false alarm rate guarantees presented in \eqref{eq:weighted_coverage_bounds} in Proposition \ref{prop:coverage_weighted_exchangeability}
\label{prop:conformal_detector_marginal_guarantees}. Proof in Appendix \ref{sec:appendix_proofs}.
\end{proposition}


\subsection{Adaptive weighted anomaly scores under non-exchangeability}

Our proposed conformal anomaly score mapping $\beta_{\textbf{w}}(\cdot)$ in \eqref{conformal_p_value_transformation} depends on the weights $\textbf{w}$ assigned to the previously observed scores. Therefore, given a new observation $S_{t+1}$ the mapping can be directly expressed as a function of $\textbf{w}$, $\beta_{\textbf{w}}(S_{t+1}) = \beta_{t+1}(\textbf{w})$ such that
\begin{equation}
\begin{array}{r}
     \beta_{t+1}(\textbf{w}) \coloneqq \frac{1+\sum\limits^{n}_{k={j_{t+1}}}w_{\pi^{-1}(k)} }{|\textbf{w}|+1},  \quad
 {j_{t+1}} = \sum_{i=1}^{t}\mathbf{1}[S_{t+1}\le S_i].
\end{array}
\label{eq:scores_wrt_weights}
\end{equation}
Where $\pi:[n] \rightarrow [n]$ represents a sorted mapping of the previous $n$ nonconformity scores such that $\pi(i)=k \in [n], \forall i \in [n]$ where $\pi(i)<\pi(j)$ if $S_{i}\le S_{j}, \forall i\not=j$. $\pi^{-1}(k)$ is the inverse sorting operation, mapping $k$ to the index of the observation corresponding to the $k$ largest value.



We want our proposed conformal score to be well calibrated across time, meaning  \( \mathbb{P}(\beta_{\textbf{w}}(S_{t+1}) \le \alpha) \approx \alpha \), for all \( \alpha \in [0,1] \) and \( t \). In lieu of that, we require $\beta_{\textbf{w}}(S_{t+1})$ to be a conservative estimate such that \( \mathbb{P}(\beta_{\textbf{w}}(S_{t+1}) \le \alpha) \le \alpha \). Such calibration ensures that the conformalized scores adapts effectively to distributional shifts over time.
The ideal condition under non-anomalous distributions for $S_{t+1}$, $\mathbb{P}(\beta_{\textbf{w}}(S_{t+1}) \le \alpha) =\alpha, \forall \alpha \in [0,1]$ is achieved when $\beta_{\textbf{w}}(S_{t+1}) \sim U_{[0,1]}$. We also note that $\beta_{\textbf{w}}(S_{t+1})$ cannot produce non-trivial quantile estimates below its effective sample size $\alpha_c = \frac{1}{|\textbf{w}|+1}$. We therefore seek to learn a set of feasible weights $\textbf{w}$ satisfying these conditions by minimizing the 1-Wasserstein distance ($\mathcal{W}_1$) between the cumulative density function (CDF) of the proposed score variable $F_{\beta_{t+1}(\mathbf{w})}$, where $\beta_{t+1}(\mathbf{w}) = \beta_{\mathbf{w}}(S_{t+1})$ as in \eqref{eq:scores_wrt_weights}, and the CDF of the uniform distribution $F_{U}$, subject to an effective sample size constraint determined by our critical false alarm rate $\alpha_c$. Namely
\begin{equation}
    \begin{array}{c}
         \min_{\textbf{w}} \mathcal{W}_1(F_{\beta_{t+1}(\textbf{w})}, F_U) \quad
         s.t. \quad |\textbf{w}|> \frac{1}{\alpha_{c}} - 1, w_{i} \in [0,1], \forall i \in [n].
    \end{array}
\label{eq:weights_objective}
\end{equation}
Here $\alpha_c$ is the user-defined critical false alarm rate. From the dual definition of $\mathcal{W}_1$ we have
\begin{equation}
    \begin{array}{rl}
        \mathcal{W}_1(F_{\beta_{t+1}(\textbf{w})}, F_U) & = \int^1_{0}|F^{-1}_{\beta_{t+1}(\textbf{w})}(p) - F^{-1}_{U}(p)| dp \\
         & = \int^1_{0}|F_{\beta_{t+1}(\textbf{w})}(\alpha) - F_{U}(\alpha)| d\alpha \\
         & = \mathbb{E}_{\alpha \sim U_{[0,1]}}| \mathbb{P}(\beta_{t+1}(\mathbf{w}) \le \alpha) - \alpha|,
    \end{array}
\label{eq:wasserstain_to_calibration}
\end{equation}
which indicates that minimizing $\mathcal{W}_1(F_{\beta_{t+1}(\textbf{w})}, F_U)$ is equivalent to minimizing the calibration gap $| \mathbb{P}(\beta_{t+1}(\mathbf{w}) \le \alpha) - \alpha| $ uniformly across all false alarm rates. We next approximate the objective in ~\eqref{eq:weights_objective} using finite samples and give the corresponding algorithm.
\section{Optimization}

In practice, we need to approximate $F_{\beta_{t+1}(\textbf{w})}(\alpha)$ in \eqref{eq:weights_objective} with a finite number of samples $n_b$, which results in the following empirical CDF based on the scores $\{\beta_{t+j}\}^{n_b}_{j=1}$
\begin{equation}
    \hat{F}_{\beta_{t+1}(\textbf{w})}(\alpha) = \frac{1}{n_b}\sum^{n_b}_{j = 1}\mathbf{1}[\beta_{t+j}(\textbf{w}) \le \alpha].
\end{equation}
Then, the  $\mathcal{W}_1$  objective in \eqref{eq:weights_objective} can be empirically approximated as follows
\begin{equation}
     \mathcal{W}_1(\hat{F}_{\beta_{t+1}(\textbf{w})}, F_U) = \sum^{n_b}_{k=1}\int^{\frac{k}{n_b}}_{\frac{k-1}{n_b}}|\beta_{t+\hat{\pi}^{-1}(k)}(\textbf{w}) - \alpha| d\alpha, 
\label{eq:empirical_wass1_uniform}
\end{equation}
where $\hat{\pi}$ is the sort mapping of $\{\beta_{t+j}(\textbf{w})\}^{n_b}_{j=1}$ scores such that $\beta_{t+\hat{\pi}^{-1}(k)} (\textbf{w})\le \beta_{t+\hat{\pi}^{-1}(k+1)}(\textbf{w})$. Note that the expression in \eqref{eq:empirical_wass1_uniform} is a sum of integrals of piecewise linear functions. Therefore, it is differentiable w.r.t. to each $\beta_{t+j}(\textbf{w})$, and consequenlty w.r.t. to each $\textbf{w}$ (see \eqref{eq:anomaly_weighted_function}) and also computable in closed form. Then the weights can be updated using projected gradient descent 
\begin{equation}
\begin{array}{c}
    \textbf{w}_{t+n_b+1} = \textbf{w}_{t+n_b} - \gamma \Big \{\sum^{n_b}_{i=1}\frac{\partial\mathcal{W}_1} {\partial \beta_{t+i}} \frac{\partial \beta_{t+i}(\textbf{w}_{t+n_b})}{\partial w_{k}} \Big\}^{n}_{k=1} \\
     \textbf{w}_{t+n_b+1} = \prod_{\textbf{w} \in [0,1]^n, |\textbf{w}|> \frac{1}{\alpha_c} - 1} \Big[\textbf{w}_{t+n_b+1}\Big] 
\end{array}
\label{eq:weight_projection}
\end{equation}

Note that here $\textbf{w}_{t}$ denotes our current estimate of the entire weighting vector $\textbf{w}$ at time $t$. The partial derivatives can be expressed in closed form as 
\begin{equation}
\frac{\partial\mathcal{W}_1}{\partial \beta_{t+i}}=
\begin{cases}
-\tfrac{1}{n_b}, 
 & \text{if }\, \beta_{t+i} < \tfrac{\hat{\pi}(i)-1}{n_b},\\
2\,\beta_{t+i} \;-\;\tfrac{2\,\hat{\pi}(i) - 1}{n_b}, 
 & \text{if }\, \tfrac{\hat{\pi}(i)-1}{n_b} \le \beta_{t+i} \le \tfrac{\hat{\pi}(i)}{n_b},\\[6pt]
+\tfrac{1}{n_b}, 
 & \text{if }\, \beta_{t+i} > \tfrac{\hat{\pi}(i)}{n_b}.
\end{cases}
\label{eq:wass1_derivatives_wrt_beta}
\end{equation}
and
\begin{equation}
\frac{\partial \beta_{t+i}(\textbf{w})}{\partial w_{k}}= \frac{-\beta_{t+i}(\textbf{w}) + \mathbf{1}[j_{t+i} \le \pi(k)]}{|| \textbf{w}||_1 + 1}
\label{eq:beta_derivatives_wrt_w}
\end{equation}
The derivatives themselves have a simple interpretation. The derivative of $\frac{\partial\mathcal{W}_1}{\partial \beta_{t+i}}$ pushes a normalized score $\beta_{t+i}$ to lie within the ranges of its empirical quantile bucket $[\tfrac{\hat{\pi}(i)-1}{n_b},\tfrac{\hat{\pi}(i)}{n_b}]$, and is minimized when $\beta_{t+i}=\tfrac{2\,\hat{\pi}(i) - 1}{2n_b}$. The derivative $\frac{\partial \beta_{t+i}(\textbf{w})}{\partial w_{k}}$ establishes that one can increase $\beta_{t+i}$ by decreasing the weight of scores higher than the currently-observed score $S_{t+1}$ or by globally decreasing the overall sample size $||\mathbf{w}||_1$.


\begin{algorithm}[h!]
\caption{1-Wasserstein Adaptive Conformal Anomaly Score }
\label{alg:acas}
\begin{algorithmic}
\REQUIRE $\{S_t\}^T_{t=1}$: Scores, $\alpha_c$: min false alarm rate, $n$ max past samples, $n_b$ min batch size
\OUTPUT: $\boldsymbol{\beta} \in [0,1]^{T-n_c}$ normalized score vector
\STATE $n_c = \frac{1}{\alpha_c} - 1$, $\textbf{w} = \{w_i = \textbf{1}[i \le n_c]\}^{n}_{i=1}$. {\small\# Compute critical samples and init weights}
\STATE $\mathbf{J}\boldsymbol{\beta}(\textbf{w}) \leftarrow \{0\}^{n_b \times n}$,  ${i}_b =0$, $\boldsymbol{\beta} \leftarrow \{\}$ {\small\# Initialize score Jacobian, batch counter and output}
\FOR{$t=n_c:T-n_c$ }
\STATE $\mathbf{s} = \{S_{i}\}^{t}_{i=\max(t-n,1)} $ , $\hat{\mathbf{w}} = \{\hat{w}_{i} = w_{|\mathbf{s}|+1-i}\}^{|\mathbf{s}|}_{i=1}${\small \# Get past scores and corresponding weights}
\STATE $\pi \leftarrow \textsc{argsort}(\mathbf{s})$ {\small \# sort past scores in ascending order}
\STATE $j_{t+1} = \sum_{s \in \textbf{s}} \mathbf{1}[S_{t+1} < s]$ ,  $\beta_{t+1} = \frac{\sum^{|\textbf{s}|}_{k=j_{t+1}} \hat{w}_{\pi^{-1}(k)} + 1}{||\hat{\textbf{w}}||_1 + 1}$ {\small\# Compute p-value score for $S_{t+1}$}
\STATE $\boldsymbol{\beta} \leftarrow\boldsymbol{\beta} \cup \beta_{t+1}$, $i_b \leftarrow i_b + 1$
\STATE $\mathbf{J}\boldsymbol{\beta}(\textbf{w})_{i_b,n-k} = \{\frac{\partial \beta_{t+1}}{\partial \hat{w}_k}\} $ for $k=1,...,|s|$, using \eqref{eq:beta_derivatives_wrt_w} {\small\# Compute partial derivatives }
\IF{$i_b = n_b$}
\STATE $\hat{\pi} \leftarrow\textsc{argsort}(\boldsymbol{\beta}_{t+1-n_b:t+1})$ {\small\#Sort last $n_b$ normalized scores and compute gradient}
\STATE Compute $\{\frac{\partial\mathcal{W}_1}{\partial \beta_{t+i}}\}^{n_b}_{i=1}$ using $\hat{\pi}$, \eqref{eq:wass1_derivatives_wrt_beta},  $\nabla \mathcal{W}_1(\textbf{w}) = \{\sum^{n_b}_{i=1}\frac{\partial\mathcal{W}_1}{\partial \beta_{t+i}}\mathbf{J}\boldsymbol{\beta}(\mathbf{w})_{i,k}\}^{n}_{k=1}$
\STATE $\mathbf{w}\leftarrow \prod_{\textbf{w} \in [0,1]^n, |\textbf{w}|> n_c} \Big[ \mathbf{w} - \gamma \nabla \mathcal{W}_1(\textbf{w})\Big] $ , $i_b \leftarrow 0$
\ENDIF
\ENDFOR 

\end{algorithmic}
\end{algorithm}

We propose $\mathcal{W}_1$\textsc{-ACAS} (Algorithm~\ref{alg:acas}), which operates by sequentially estimating normalized scores $\beta_t$ using the current weight estimates. The weights $\mathbf{w}$ are then periodically updated to minimize the objective in Eq.~\ref{eq:weights_objective}, based on the online sample buffer and the update rules in Eqs.~\ref{eq:weight_projection}, \ref{eq:wass1_derivatives_wrt_beta} and \ref{eq:beta_derivatives_wrt_w}.

\paragraph{Aggregation Across Multiple Forecast Horizons}  
We extend Algorithm~\ref{alg:acas} to operate across multiple forecast horizons. Specifically, we run $D$ parallel instances of the algorithm, each associated with a $d$-step ahead prediction error,  
\(S^d_{t+1} = \big| Y_{t+1} - \hat{Y}^d_{t+1} \big|\), with \( 
\hat{Y}^d_{t+1} = h_d(Y_{t-n_c-d:t-d+1}), \; d \in [D].
\)  
This produces a set of $D$ conformal $p$-values for each observation $t+1$, denoted $\{\beta^d_{t+1}\}_{d \in [D]}$. The final anomaly score is the median across horizons,  
\begin{equation}
    \bar{\beta}_{t+1} = \operatorname{median}_{d \in [D]} \beta^d_{t+1}, \qquad 
\beta^d_{t+1} = \beta_{\mathbf{w}^d}(S^d_{t+1}).
\label{eq:beta_aggregation}
\end{equation}
This requires an observation to be identified as a significant outlier by more than half of the horizon-specific detectors. In the streaming setting, we maintain a buffer of forecasts at different horizons. When a new sample $Y_{t+1}$ is observed, we collect its aligned forecasts $\{\hat{Y}^d_{t+1}\}_{d \in [D]}$, compute the corresponding errors $\{S^d_{t+1}\}_{d \in [D]}$, and update each horizon-specific instance of Algorithm~\ref{alg:acas} to obtain the adaptive $p$-values, $\{\beta^d_{t+1}\}_{d \in [D]}$. {In Appendix~\ref{subsec:appendix_extension_mv} we describe how Algorithm~\ref{alg:acas} extends to multivariate time series anomaly detection in a similar manner. We also outline several standard p-value combination techniques, which can also be applied to aggregate the horizon-specific p-values.}

%% file: ICLR2026/sections/experiments.tex
\section{Experiments}

We evaluate the proposed conformalized anomaly score $\mathcal{W}_1$\textsc{-ACAS} (Algorithm~\ref{alg:acas}) by analyzing its calibration and anomaly detection performance on time series data. Synthetic experiments (Appendix~\ref{sec:appendix_simulexamples}) validate its ability to remain calibrated under both gradual and abrupt distribution shifts, where ground-truth $p$-values are available. Our main empirical study focuses on real-world anomaly detection datasets, where we assess detection accuracy using both threshold-independent and threshold-dependent metrics.

\paragraph{Anomaly Detection Datasets.}
We evaluated the performance of our proposed method ($\mathcal{W}_1$\textsc{-ACAS}, Algorithm~\ref{alg:acas}) for unsupervised univariate time series anomaly detection when applied to a pre-trained time series foundation model. Experiments are conducted on seven benchmark datasets: 
YAHOO~\citep{laptev2015s5}, 
NEK~\citep{si2024timeseriesbench}, 
NAB~\citep{ahmad2017unsupervised}, 
MSL~\citep{lai2021revisiting}, 
IOPS~\citep{iops_dataset}, 
STOCK~\citep{tran2016distance}, 
and WSD~\citep{zhang2022efficient}, 
all part of the curated anomaly detection benchmark of \citet{liu2024elephant}. {For the multivariate experiments, we additionally use the curated subsets of TAO~\citep{tao_noaa_dataset}, GECCO~\citep{rehbach2018gecco}, LTDB~\citep{goldberger2000physiobank}, and Genesis~\citep{von2018anomaly} released as part of the benchmark in \citet{liu2024elephant}.} Each dataset consists of an initial segment without anomalies used for training or calibration, followed by a test split that may contain anomalies. 










\paragraph{$\mathcal{W}_1$\textsc{-ACAS} + TSFM.}We integrate $\mathcal{W}_1$\textsc{-ACAS} with three pre-trained TSFMs: 
Tiny Time Mixers (TTM)~\citep{ekambaram2024ttms}, 
Chronos-Bolt-Small (Chronos)~\citep{ansari2024chronos}, 
and TiRex~\citep{auer2025tirex}. 
All models use a context length of 52 and a forecast horizon of $D=15$. 
For Algorithm~\ref{alg:acas}, we set the critical false alarm rate to $\alpha_c = 0.01$, batch size $n_b=10$, and learning rate $\gamma=0.001$. We use ADAM ~\citep{kingma2015adam} to perform an adaptive gradient descent on the weights $\mathbf{w}$. 
Appendix~\ref{sec:appendix_additional_results}, Fig.~\ref{fig:w1-acas-steps-grid}, analyzes the impact of aggregating forecast horizons, showing that $D=15$ provides a reasonable balance between performance and sample efficiency. {Figures~\ref{fig:w1-acas-steps-grid-lr},~\ref{fig:w1-acas-steps-grid-nbu}, and~\ref{fig:w1-acas-steps-grid-alarm} show the sensitivity of $\mathcal{W}_1$\textsc{-ACAS} to the learning rate $\gamma$, batch size $n_b$, and $\alpha_c$. The method shows low variability for small $\gamma$ and $n_b$. The parameter $\alpha_c$ controls the maximum acceptable $p$-value resolution: smaller values require a larger number of in-distribution past observations $n_c$, but do not impose a lower bound on the detectable anomaly level.}

\paragraph{Baseline Methods.}
We compare $\mathcal{W}_1$\textsc{-ACAS} against two TSFM-based baselines: a \textbf{Gaussian} model that fits the mean absolute forecast error across $d$ steps using calibration data, and a \textbf{Conformal} offline approach that learns $p$-value mappings per horizon and aggregates them by the median. We also include top-performing classical methods from \citet{liu2024elephant}: \textbf{KShape}~\citep{paparrizos2015k, paparrizos2017fast, boniol2021sand}, \textbf{POLY}~\citep{li2007unifying}, 
\textbf{Sub-PCA}~\citep{aggarwal2017introduction},
\textbf{Sub-KNN}~\citep{ramaswamy2000efficient},
and \textbf{SAND}~\citep{boniol2021sand}. {We further include strong semi-supervised deep learning–based anomaly detection methods~\citep{audibert2022deep}, namely \textbf{CNN}~\citep{munir2018deepant}, \textbf{USAD}~\citep{audibert2020usad}, and \textbf{OmniAnomaly}~\citep{su2019robust}, as well as the recent general purpose TSFM \textbf{MOMENT}~\citep{goswami2024moment}, which provides zero-shot anomaly scoring.} Additional details are provided in Appendix \ref{sec:appendix_baseline_methods}.

\paragraph{Evaluation Metrics.} 
We report both point-wise (AUC, PA-F1)~\citep{wu2022timesnet,wang2024deep,liu2024elephant} and range-wise metrics (VUS~\citep{paparrizos2022volume}, Affiliation-F1~\citep{huet2022local}). For threshold-dependent scores (PA-F1, Affiliation-F1), we follow the oracle strategy of \citet{liu2024elephant}, selecting the best threshold in $[0,1]$ and reporting the associated False Positive Rate (FPR) and calibration error (CalErr). Further details are in Appendix~\ref{sec:appendix_metrics}.



\begin{table*}[h!]
\centering
\scriptsize
\setlength{\tabcolsep}{5pt}
\caption{\textbf{Performance Summary across univariate datasets.}
Entries indicate the mean $\pm$ standard deviation computed by first averaging within each dataset group, then averaging across groups (equal weight).
Higher numbers are better for PA-F1, Affiliation-F, AUC-PR, VUS-PR; lower numbers are better for FPR, and calibration error (CalErr). Underlined results indicate best post-hoc methods applied to the same base forecaster, while bold indicate best results overall. Methods marked with * denote deep learning semi-supervised approaches.
}
\label{tab:ad_summary_group_v0}
\begin{tabular}{llcccccc}
\toprule
Forecaster & AD Method & PA-F1 \(\uparrow\) & Affiliation-F \(\uparrow\) & FPR \(\downarrow\) & CalErr \(\downarrow\) & AUC-PR \(\uparrow\) & VUC-PR \(\uparrow\) \\
\midrule
Chronos & $\mathcal{W}_1$-ACAS & \underline{0.912 ± 0.066} & \underline{0.893 ± 0.060} & \textbf{0.077 ± 0.114} & \textbf{0.025 ± 0.029} & \textbf{0.355 ± 0.261} & \underline{0.440 ± 0.272} \\
Chronos & Conformal & 0.863 ± 0.109 & 0.891 ± 0.063 & 0.111 ± 0.130 & 0.038 ± 0.055 & 0.310 ± 0.240 & 0.420 ± 0.248 \\
Chronos & Gaussian & 0.716 ± 0.260 & 0.842 ± 0.066 & 0.123 ± 0.109 & 0.075 ± 0.061 & 0.265 ± 0.250 & 0.438 ± 0.245 \\
\midrule
TTM & $\mathcal{W}_1$-ACAS & \underline{0.889 ± 0.108} & \underline{0.886 ± 0.058} & \underline{0.082 ± 0.120} & \underline{0.029 ± 0.031} & \underline{0.342 ± 0.261} & 0.449 ± 0.245 \\
TTM & Conformal & 0.851 ± 0.124 & 0.885 ± 0.062 & 0.120 ± 0.145 & 0.044 ± 0.056 & 0.317 ± 0.247 & 0.448 ± 0.250 \\
TTM & Gaussian & 0.733 ± 0.240 & 0.849 ± 0.067 & 0.128 ± 0.115 & 0.081 ± 0.065 & 0.270 ± 0.261 & \underline{0.450 ± 0.249} \\
\midrule
TiRex & $\mathcal{W}_1$-ACAS & \textbf{0.925 ± 0.048} & \textbf{0.897 ± 0.064} & \underline{0.084 ± 0.113} & \underline{0.025 ± 0.031} & \underline{0.344 ± 0.269} & \underline{0.438 ± 0.272} \\
TiRex & Conformal & 0.878 ± 0.085 & 0.890 ± 0.063 & 0.107 ± 0.137 & 0.038 ± 0.055 & 0.308 ± 0.257 & 0.429 ± 0.256 \\
TiRex & Gaussian & 0.714 ± 0.264 & 0.837 ± 0.068 & 0.119 ± 0.103 & 0.090 ± 0.071 & 0.270 ± 0.264 & 0.432 ± 0.250 \\
\midrule
- & POLY & 0.527 ± 0.276 & 0.848 ± 0.072 & 0.334 ± 0.269 & 0.282 ± 0.130 & 0.044 ± 0.031 & 0.377 ± 0.207 \\
- & Sub-KNN & 0.479 ± 0.291 & 0.786 ± 0.074 & 0.451 ± 0.276 & 0.174 ± 0.124 & 0.118 ± 0.106 & 0.321 ± 0.234 \\
- & KShape & 0.533 ± 0.299 & 0.789 ± 0.096 & 0.508 ± 0.291 & 0.176 ± 0.132 & 0.125 ± 0.135 & 0.303 ± 0.262 \\
- & PCA & 0.536 ± 0.332 & 0.826 ± 0.097 & 0.374 ± 0.297 & 0.248 ± 0.131 & 0.100 ± 0.093 & 0.417 ± 0.274 \\
- & SAND & 0.460 ± 0.309 & 0.790 ± 0.079 & 0.511 ± 0.296 & 0.134 ± 0.048 & 0.101 ± 0.117 & 0.289 ± 0.190 \\
\midrule
- & CNN & 0.858 ± 0.138 & 0.881 ± 0.059 & 0.083 ± 0.103 & 0.643 ± 0.227 & 0.269 ± 0.292 & 0.423 ± 0.289 \\
- & OmniAnomaly & 0.674 ± 0.282 & 0.855 ± 0.068 & 0.209 ± 0.171 & 0.571 ± 0.187 & 0.166 ± 0.087 & 0.429 ± 0.317 \\
- & USAD & 0.498 ± 0.333 & 0.809 ± 0.099 & 0.425 ± 0.298 & 0.324 ± 0.161 & 0.088 ± 0.088 & 0.398 ± 0.262 \\
- & MOMENT\_ZS & 0.596 ± 0.305 & 0.867 ± 0.088 & 0.261 ± 0.292 & 0.417 ± 0.198 & 0.110 ± 0.075 & \textbf{0.461 ± 0.162} \\

\bottomrule
\end{tabular}
\end{table*}

\begin{figure}[!ht]
    \centering
    \subfigure[PA-F1 \(\uparrow\) per dataset]{\includegraphics[width=0.49\textwidth]{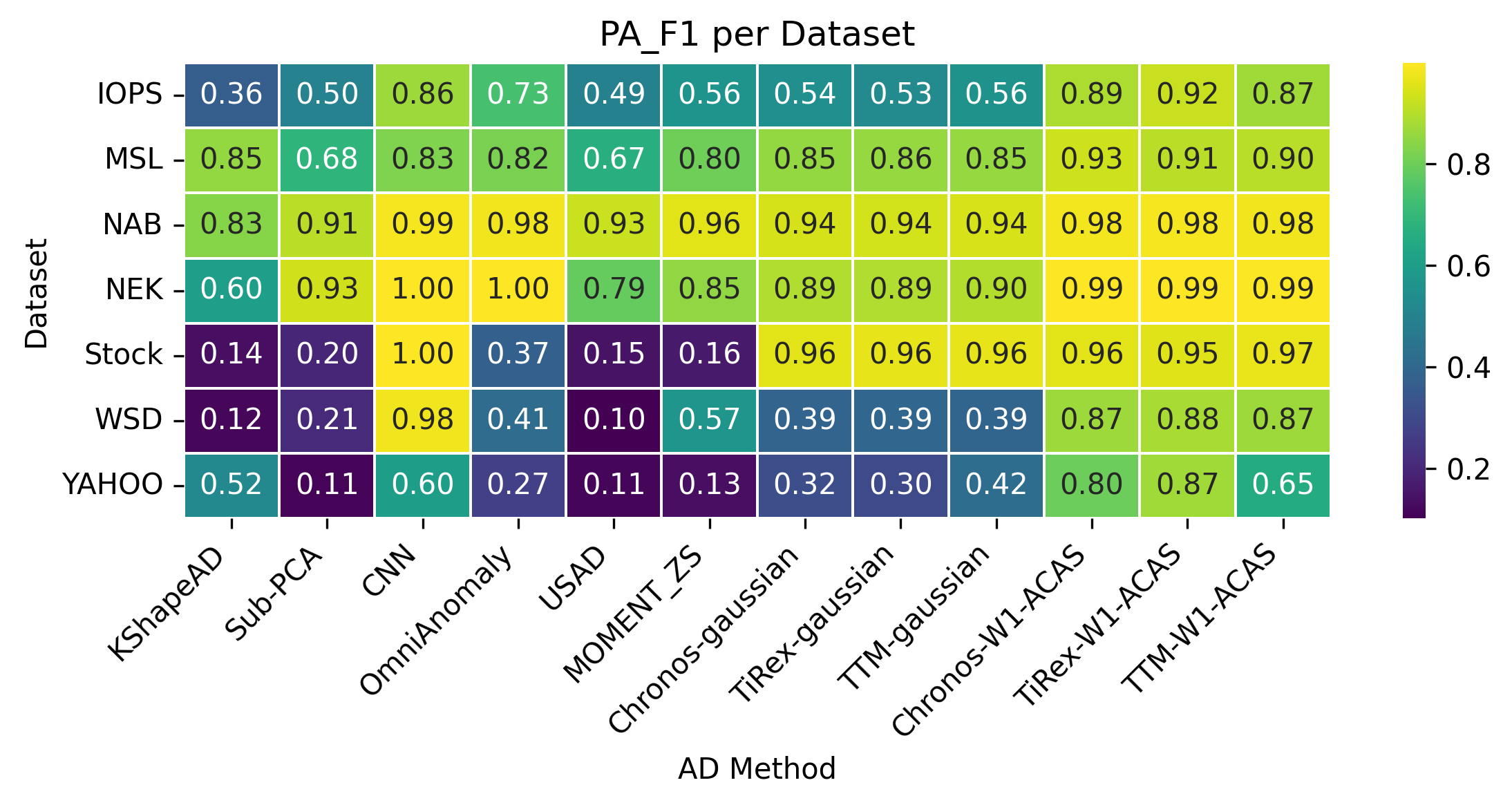}}
    \subfigure[CalErr \(\downarrow\) per dataset]{\includegraphics[width=0.49\textwidth]{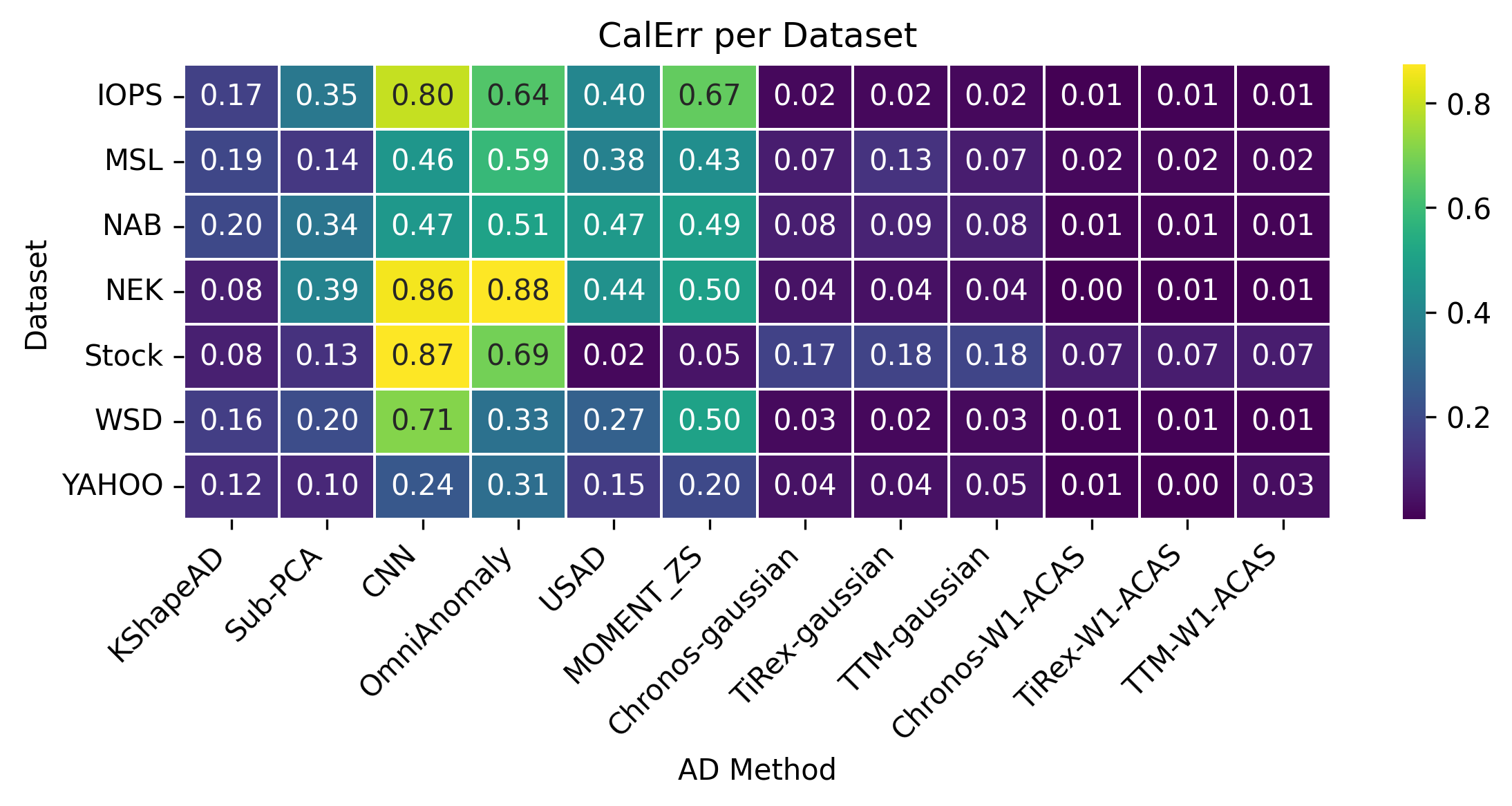}}

    \subfigure[Affiliation-F \(\uparrow\) per dataset]{\includegraphics[width=0.49\textwidth]{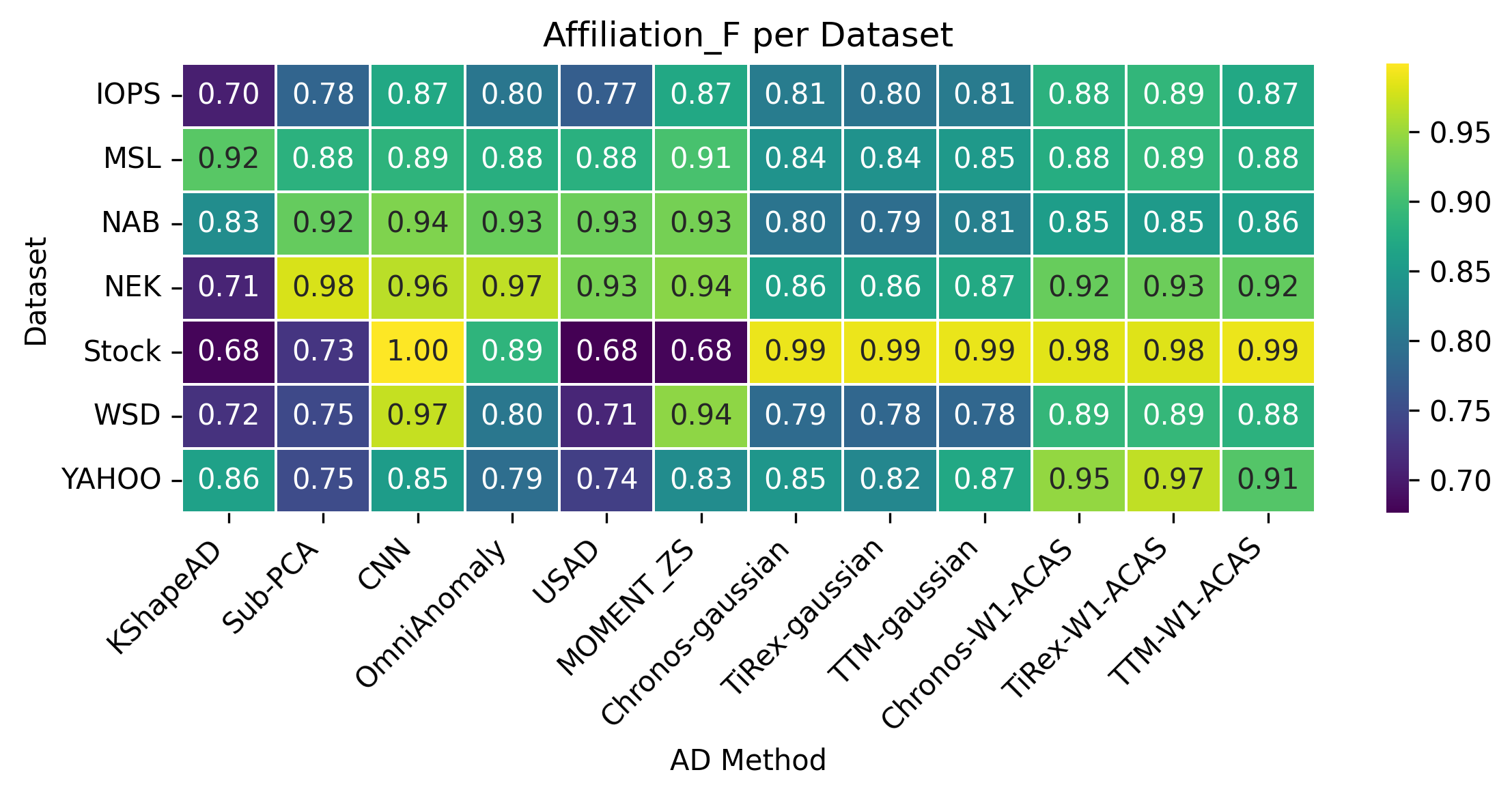}}
    \subfigure[AUC-PR \(\uparrow\) per dataset]{\includegraphics[width=0.49\textwidth]{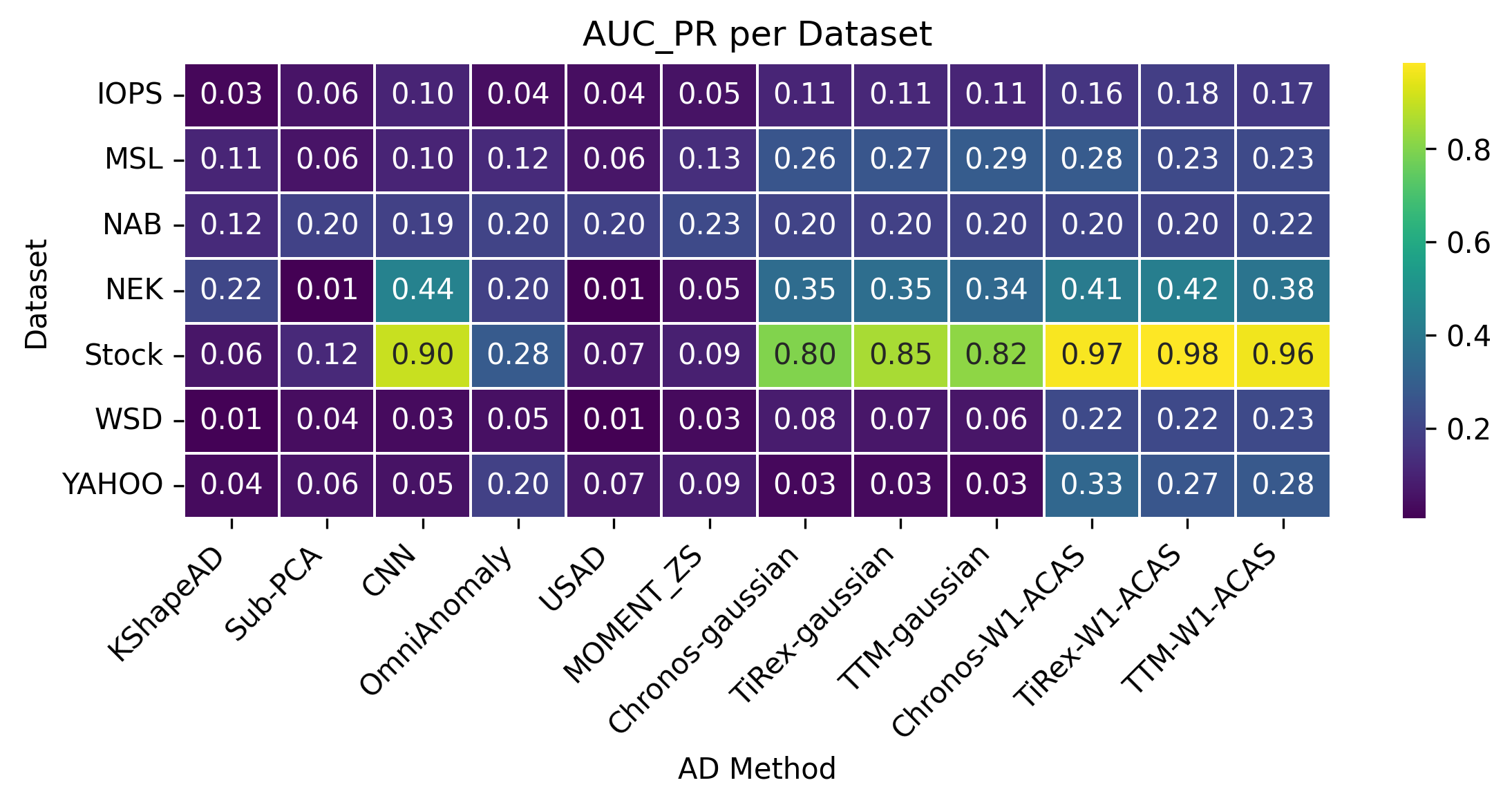}}

    \caption{{ \textbf{Performance across univariate datasets for a subset of anomaly detection methods.}
Heatmaps show the average per-dataset performance for PA-F1, Affiliation-F, AUC-PR, and Calibration Error (CalErr) across a selected subset of methods. Higher values indicate better performance for PA-F1, Affiliation-F, and AUC-PR, while lower values are preferred for CalErr. Overall, the proposed $\mathcal{W}_1$\textsc{-ACAS} combined with Chronos, TiRex or TTM yields consistently low calibration error while remaining among the top-performing approaches. Note that CNN, OmniAnomaly and USAD are semi-supervised methods trained on the non-anomalous training datasplit.}}
    \label{fig:per_dataset_heatmaps}
\end{figure}

\paragraph{Results} 
Table~\ref{tab:ad_summary_group_v0} reports the average performance of $\mathcal{W}_1$\textsc{-ACAS}, applied to different TSFM models, compared against the described baselines { on the univariate datasets}. Our method achieves the strongest performance on threshold-dependent metrics (PA-F1, Affiliation-F),{ including when compared with semi-supervised methods such as CNN, USAD, and OmniAnomaly, }while remaining competitive on threshold-independent metrics (AUC, VUS). When conditioned on the same TSFM model, $\mathcal{W}_1$\textsc{-ACAS} shows clear improvements over the Gaussian and Conformal baselines. { Figure~\ref{fig:per_dataset_heatmaps} shows the average performance per univariate dataset for a subset of the methods, extended per-dataset results are provided in Tables~\ref{tab:placeholder},~\ref{tab:placeholder_ext} and \ref{tab:placeholder-v2} in Appendix~\ref{sec:appendix_additional_results}. Table~\ref{tab:uv_tsfm_forecasting_error} shows that TSFM models have similar prediction errors across datasets, consistent with their comparable anomaly detection performance.} {Results for the multivariate datasets are presented in Table~\ref{tab:mv_results_ext} in Appendix~\ref{subsec:appendix_extension_mv}, where we demonstrate how our approach naturally extends to the multivariate setting via $p$-value aggregation, achieving top performance relative to the corresponding baselines.}

\begin{figure*}[!ht]
    \centering
    \subfigure[NEK]{\includegraphics[width=0.24\textwidth]{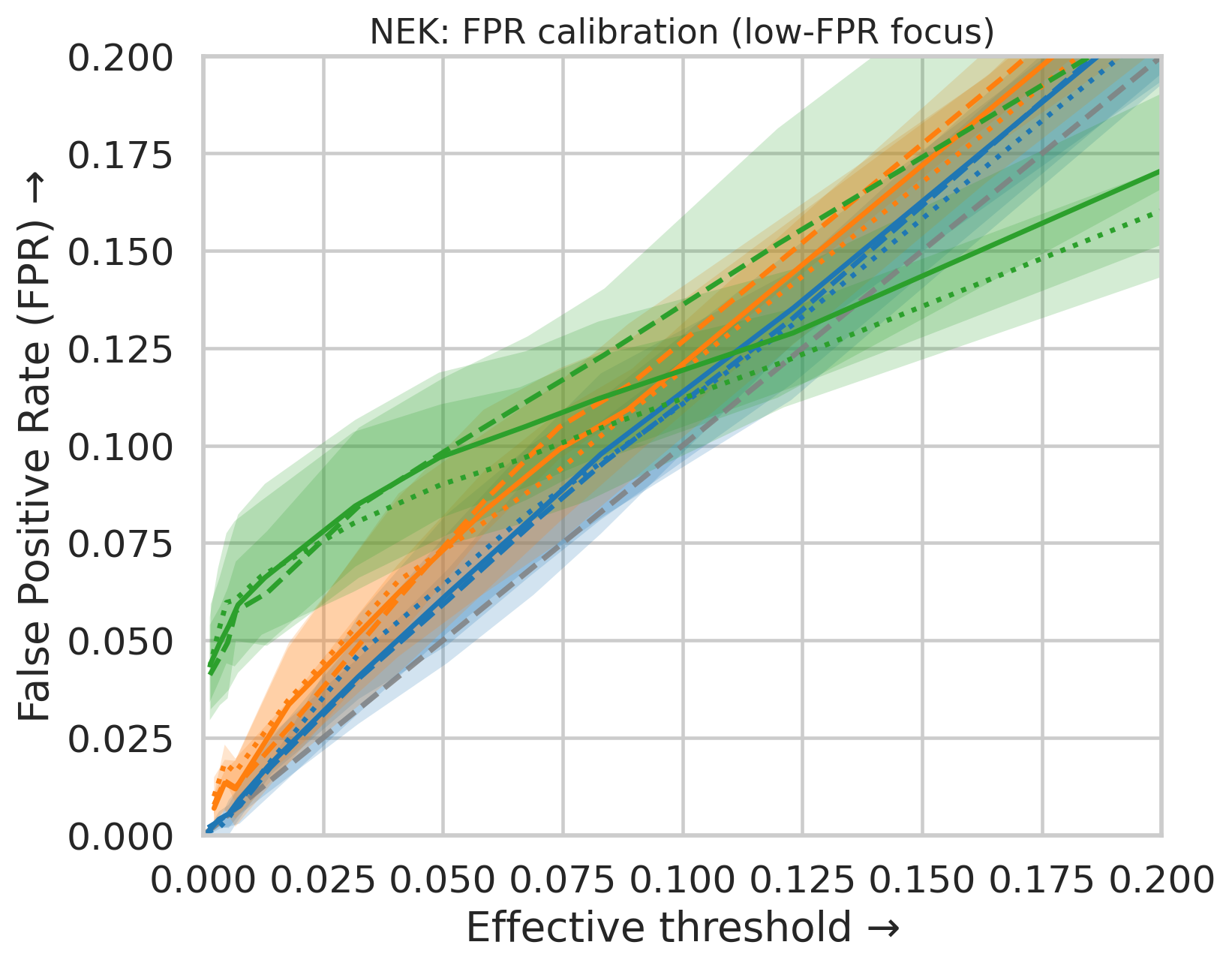}}
    \subfigure[YAHOO]{\includegraphics[width=0.24\textwidth]{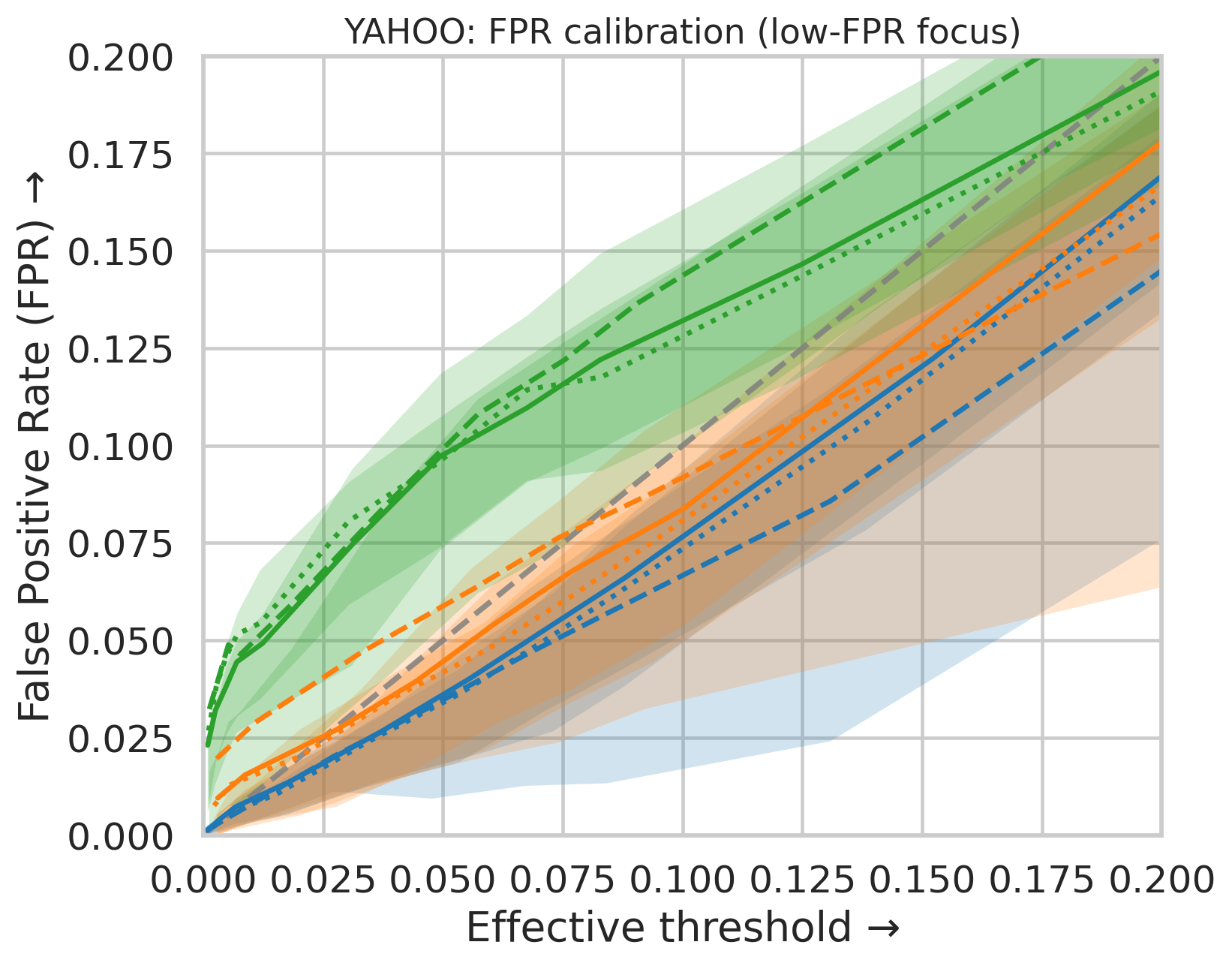}}
    \subfigure[IOPS]{\includegraphics[width=0.24\textwidth]{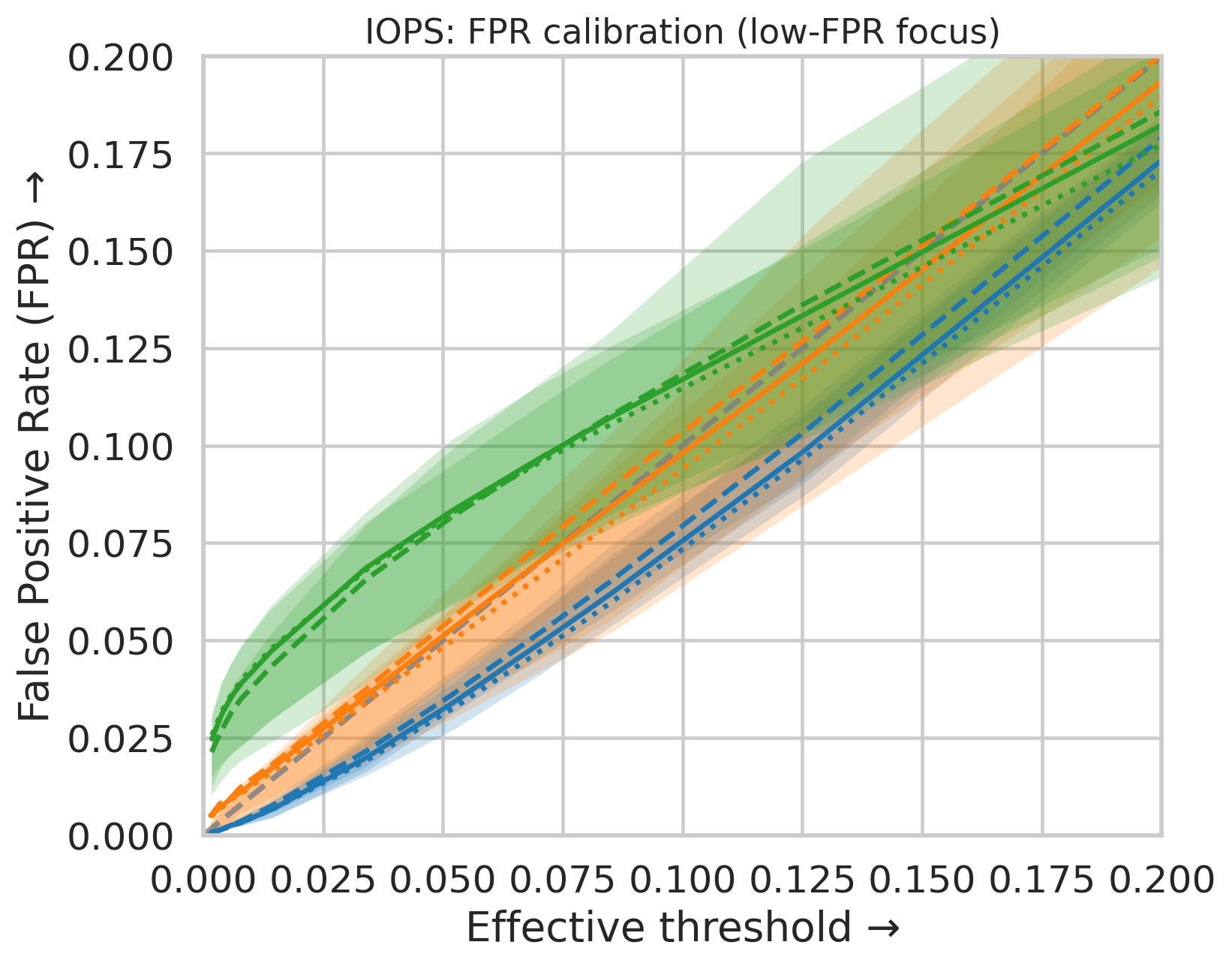}}
    \subfigure[NAB]{\includegraphics[width=0.24\textwidth]{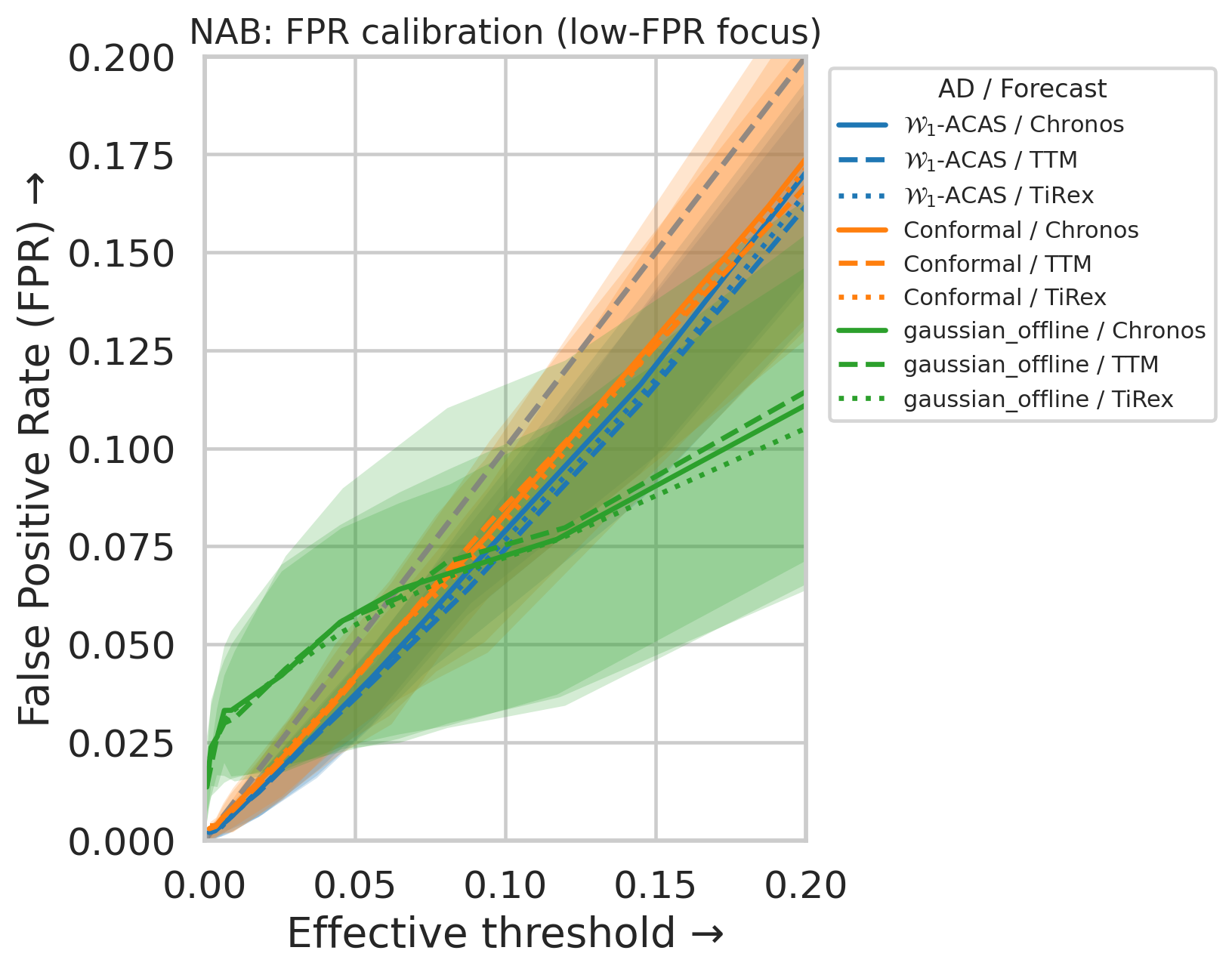}}
    
    
    \caption{\textbf{FPR vs.\ threshold in the low-FPR regime.} 
    Curves shows the mean false positive rate (FPR) across datasets for a given method, with shaded inter-quartile range (IQR) bands.
    The dashed gray line indicates ideal calibration ($FPR=\beta$). 
    Curves above the line reflect over-confident scoring (FPR larger than threshold), while curves below the line reflect conservative scoring.
    In most cases, $\mathcal{W}_1$\textsc{-ACAS} (blue) yields the most conservative thresholds, staying closer to or below the identity line compared to competing methods, while also having the lowest variance.}
    
    \label{fig:fpr-threshold-cal}
\end{figure*}

\begin{figure}[!ht]
    \centering
    \subfigure[NEK-Detection]{\includegraphics[width=0.30\textwidth]{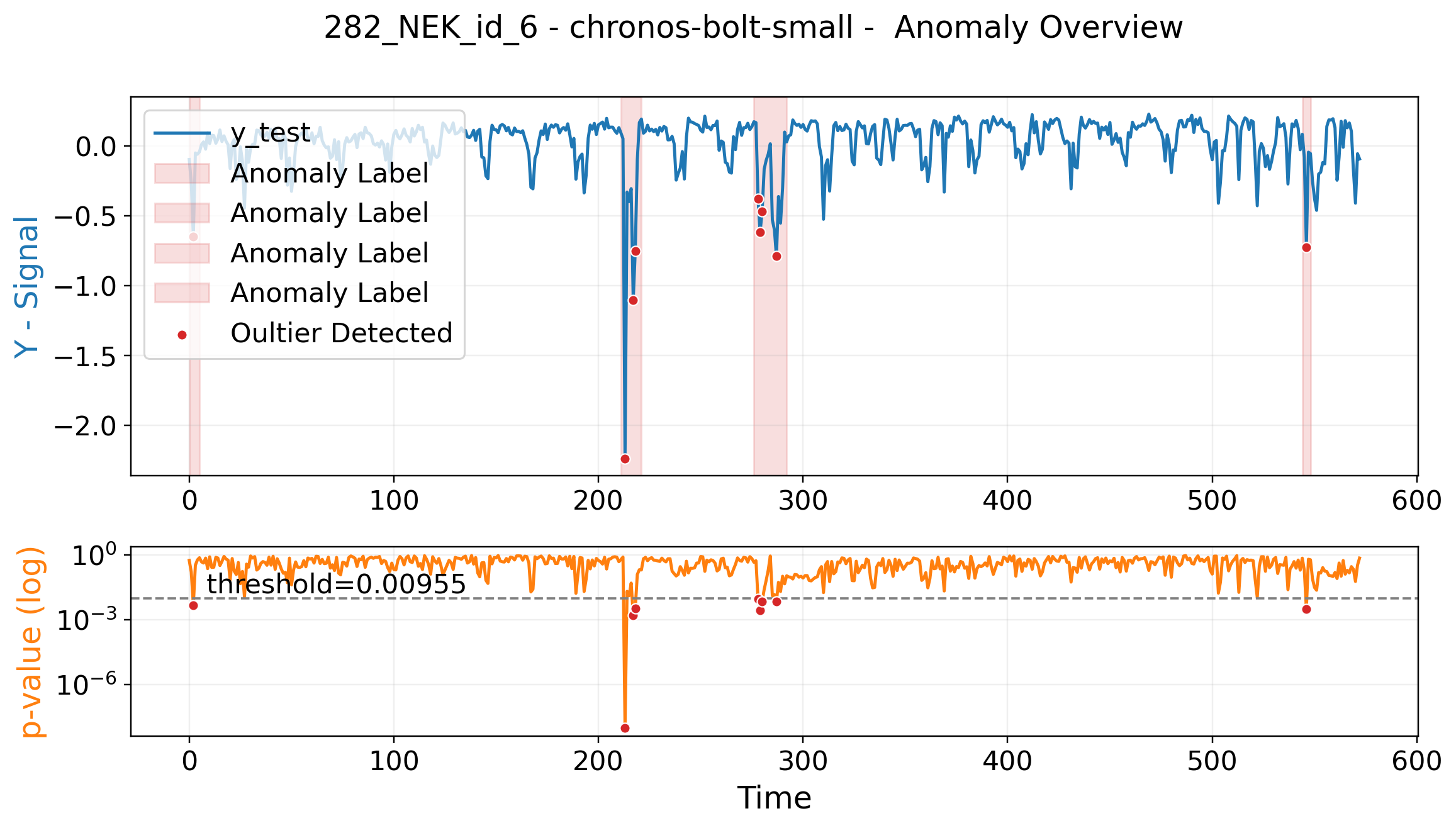}}
    \subfigure[WSD-Detection]{\includegraphics[width=0.30\textwidth]{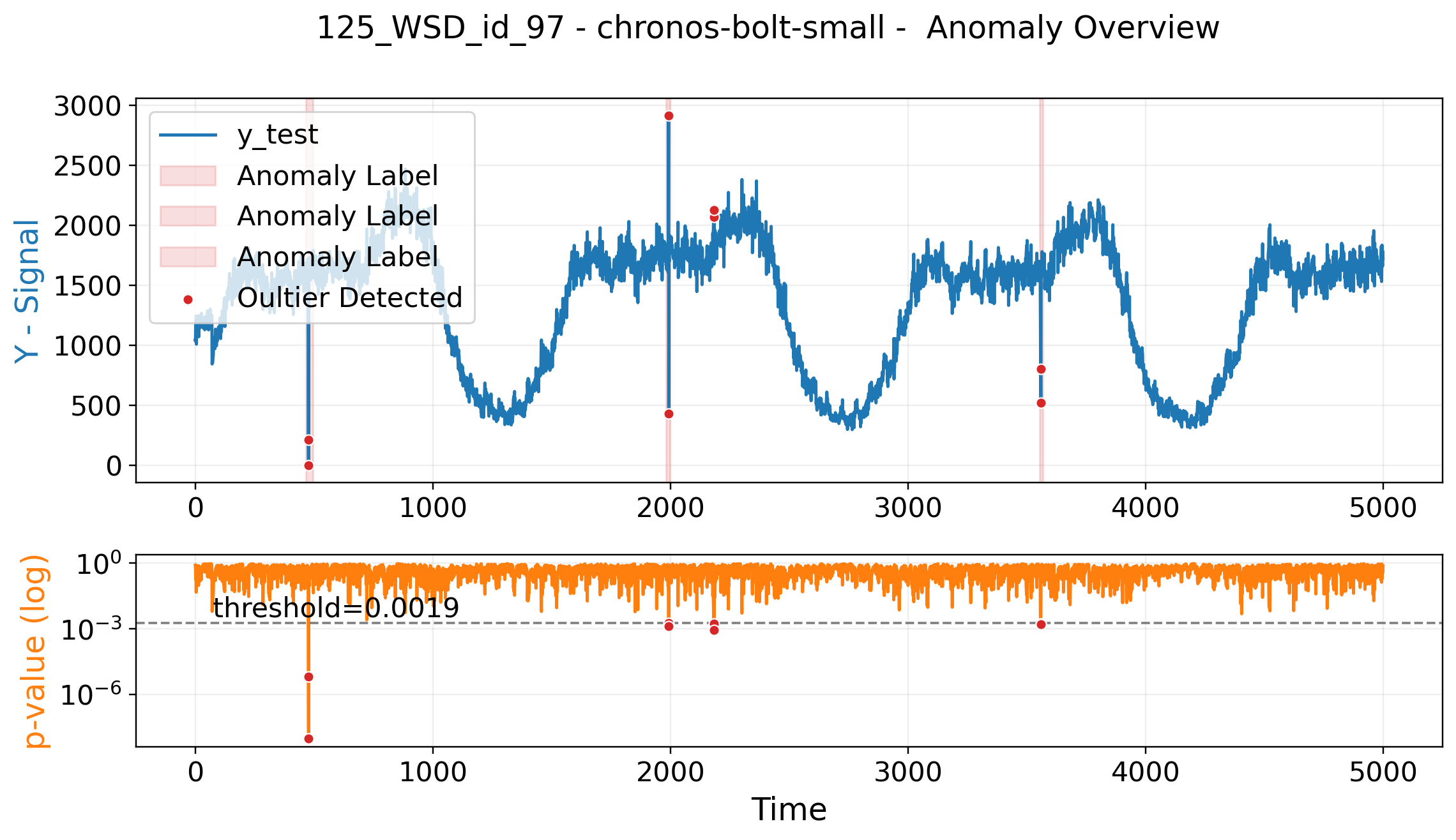}}
    \subfigure[YAHOO-Detection]{\includegraphics[width=0.30\textwidth]{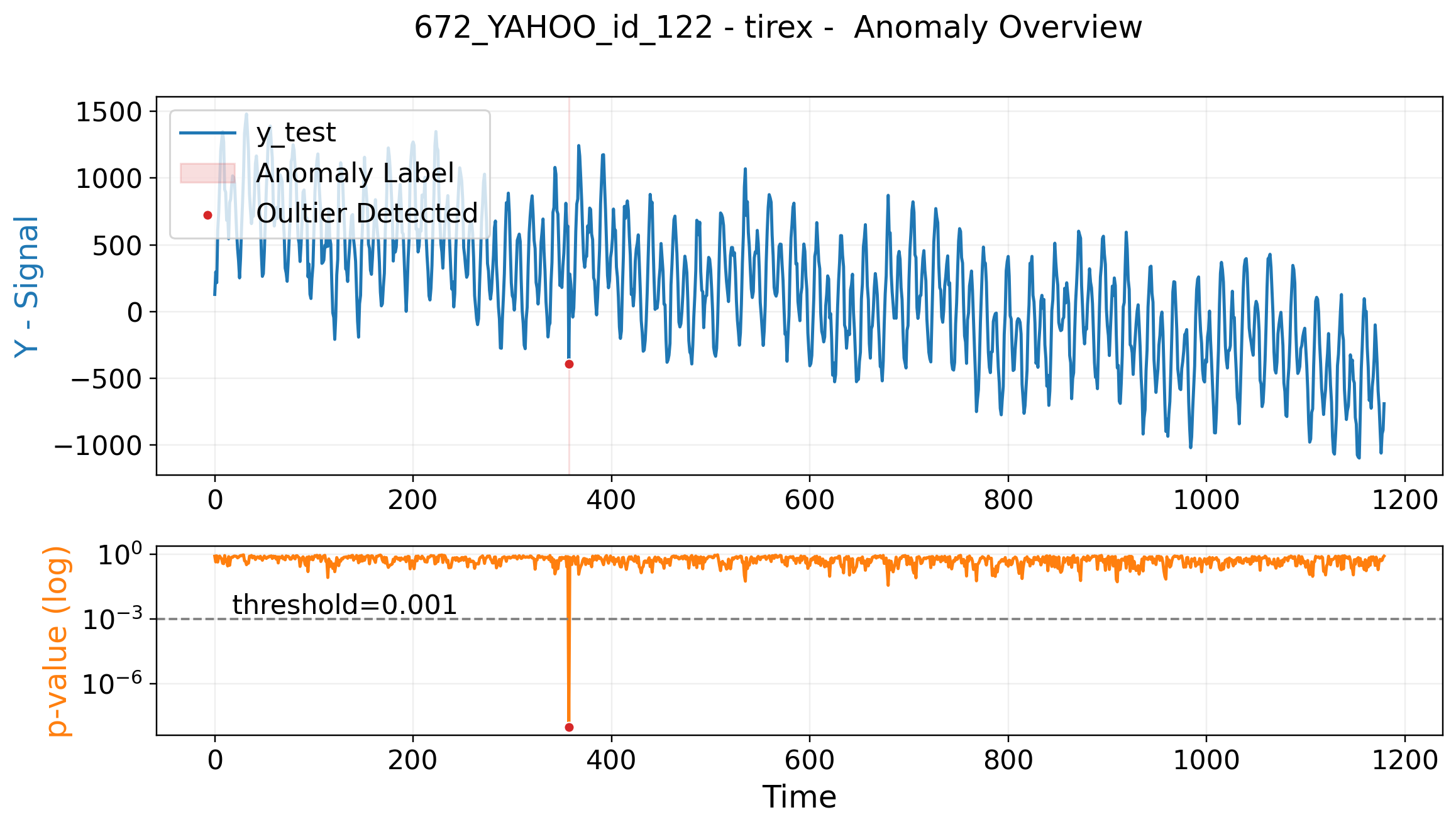}}
        
    \subfigure[NEK-W1-ACAS-Weights]{\includegraphics[width=0.30\textwidth]{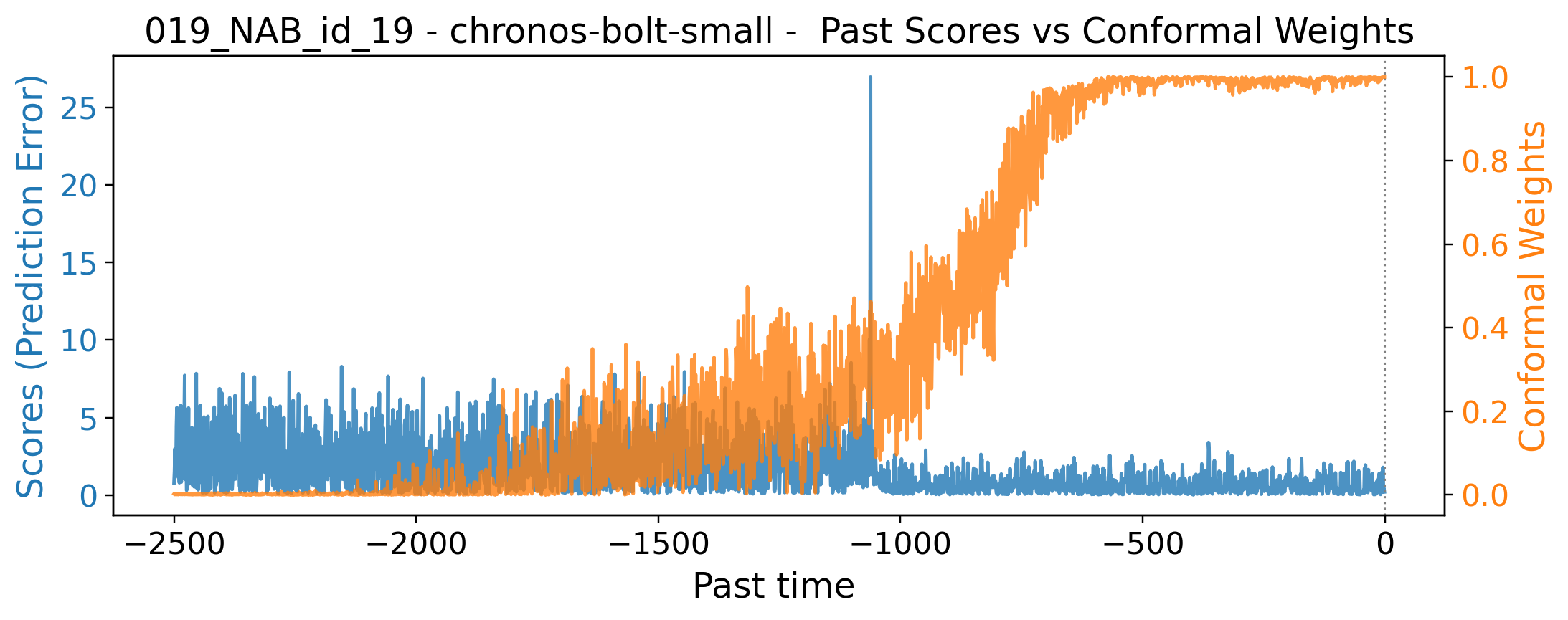}}
    \subfigure[WSD-W1-ACAS-Weights]{\includegraphics[width=0.30\textwidth]{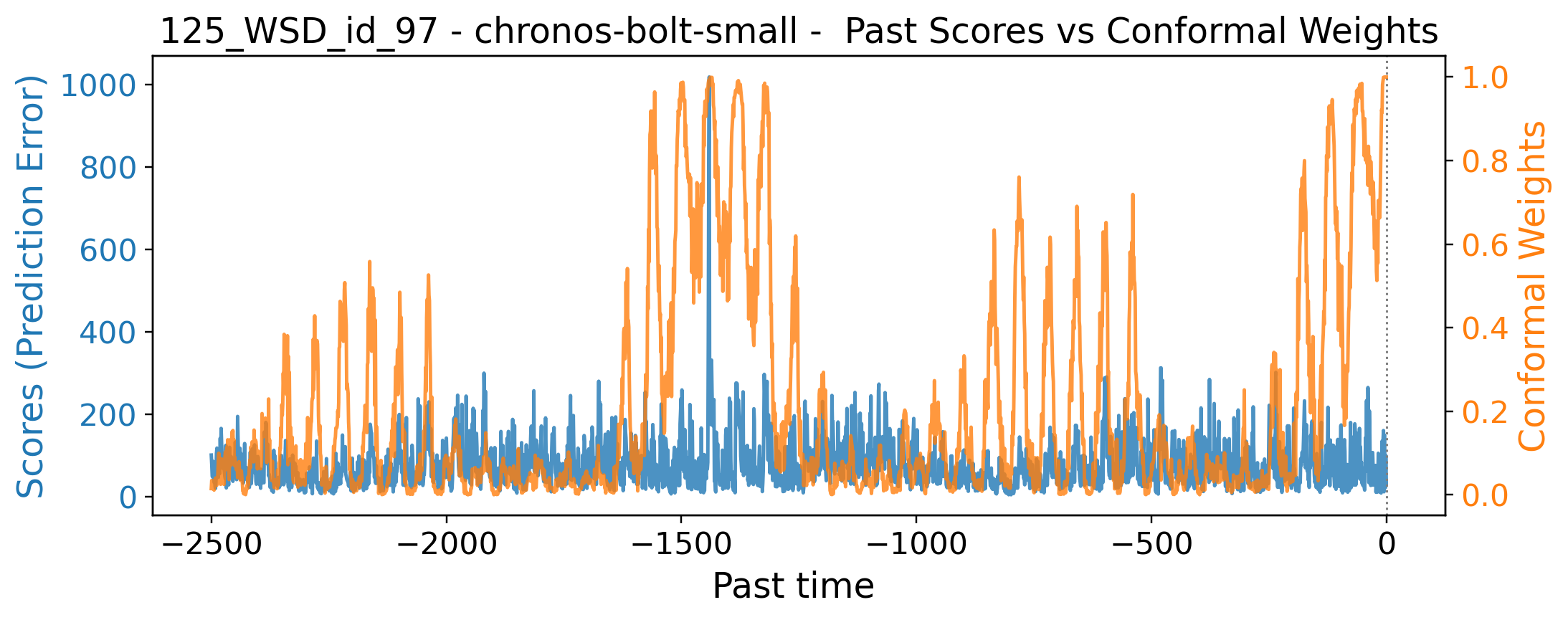}}
    \subfigure[YAHOO-W1-ACAS-Weights]{\includegraphics[width=0.30\textwidth]{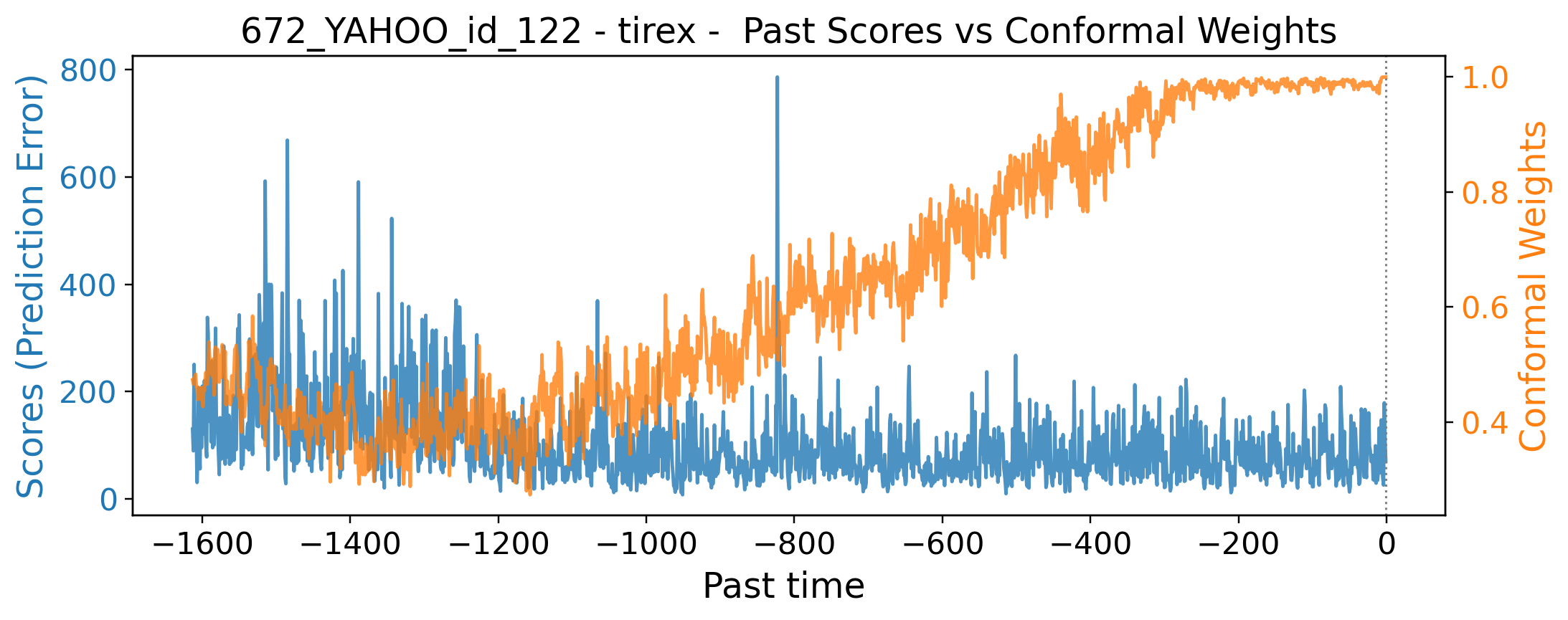}}
    
    \caption{Example signals (blue) with ground-truth anomaly labels (red shading) are shown in the first row, where detected outliers (red dots) occur when adaptive $p$-values (orange) fall below a threshold under our proposed $\mathcal{W}_1$\textsc{-ACAS} method. The second row shows the final adaptive weights (orange) over past errors (blue), averaged across horizons, illustrating how $\mathcal{W}_1$\textsc{-ACAS} adapts to and captures underlying error patterns}
    \label{fig:examples_detection}
\end{figure}

Figure~\ref{fig:fpr-threshold-cal} shows the FPR–threshold curves in the low-FPR regime, where $\mathcal{W}_1$\textsc{-ACAS} (blue) yields the most conservative thresholds, staying closer to or below the identity line compared to competing methods, while also exhibiting the lowest variance. Figure~\ref{fig:examples_detection} shows representative detection examples along with the final learned weights. We observe that $\mathcal{W}_1$\textsc{-ACAS} is adapted to capture underlying temporal patterns in errors if present. Moreover, our method effectively identifies a transition in score distributions (e.g., in the vicinity of an anomalous region) but then quickly adapts to the new anomalous distribution;  this helps minimize the number of alarms in the end-to-end system.

Additional examples are provided in Appendix~\ref{sec:appendix_additional_results}: Figure~\ref{fig:detection_examples_with_weights} shows more detection cases, and Figure~\ref{fig:pa-aff-tradeoff} illustrates the trade-offs between FPR and F1 scores (PA-F1, Affiliation-F) across datasets. The operating points of $\mathcal{W}_1$\textsc{-ACAS} (blue), in most cases, achieve both the highest F1 score and lowest FPR, especially for PA-F1. Within each TSFM model, our method dominates its Gaussian (green) and Conformal (orange) counterparts in nearly all cases. 
Furthermore, it produces better-calibrated scores (low CalErr), making threshold selection more reliable in practice. 

%% file: ICLR2026/sections/conclusion.tex
\section{Conclusion} 

In this paper, we presented $\mathcal{W}_1$\textsc{-ACAS}, a post-hoc adaptive conformal anomaly detection framework that leverages predictions from pretrained TSFMs to provide interpretable, distribution-agnostic, and well-calibrated anomaly scores without requiring retraining or large datasets. Experiments on benchmark datasets show that our method consistently outperforms competing baselines. $\mathcal{W}_1$\textsc{-ACAS} yields more conservative and stable thresholds, its a principled and easily applicable approach that adapts online to temporal error patterns, and minimizes false alarms by adjusting to distributions shifts. These properties make it especially suited for online monitoring in industrial and data-scarce environments. Future work will explore refining conformal weighting with contextual features, with straightforward extensions to multivariate anomalies via horizon-style aggregation.

%% file: ICLR2026/sections/appendix_experiments.tex
\section{Related Work Extended}
\label{appendix_related_work}

\paragraph{Time Series Anomaly Detection}
A key class of anomaly detection methods is prediction-based \cite{giannoni2018anomaly}, where anomalies are identified by deviations between predicted and observed values. These approaches assume that a well-trained forecaster captures normal temporal patterns, and significant prediction errors indicate potential anomalies \cite{boniol2024dive}. Such methods can in principle capture both point anomalies, where individual values deviate sharply, and contextual anomalies, where deviations only emerge relative to surrounding context \cite{boniol2024dive}. Given our focus on unsupervised settings with limited historical data, we build on pretrained forecasting models. Recent Time Series Foundation Models (TSFMs), trained at scale for forecasting, are particularly well suited for online detection in data-scarce scenarios \cite{rasul2023lag,rasul2024lag,ansari2024chronos,liang2024foundation}. In this work, we leverage three representative TSFMs: 
Tiny Time Mixers (TTM)~\citep{ekambaram2024ttms}, based on the TSMixer architecture; 
Chronos-Bolt-Small (Chronos)~\citep{ansari2024chronos}, a transformer-based model; 
and TiRex~\citep{auer2025tirex}, which leverages an xLSTM architecture. 


Recent benchmarks have evaluated the effectiveness of time series anomaly detection methods. The study by \citet{liu2024elephant} found that in unsupervised settings, classical distance-based and density-based approaches \cite{li2007unifying,ramaswamy2000efficient,aggarwal2017introduction,paparrizos2015k, paparrizos2017fast, boniol2021sand} often outperform more complex models. However, these methods typically require access to the entire dataset (i.e., anomaly detections are non-causal and occur after the fact) and are not inherently designed for streaming applications \cite{boniol2024dive}. They may also struggle to capture richer temporal structures in the data, which limits their effectiveness in dynamic environments. Another critical challenge concerns the interpretability of anomaly scores and the choice of thresholds. Many evaluation studies emphasize threshold-independent metrics \cite{schmidl2022anomaly,paparrizos2022tsb,goswami2022unsupervised}, yet the scores themselves often lack clear probabilistic meaning. Common thresholding strategies, such as standard deviation-based rules, depend on statistics computed over the entire dataset, making them impractical for streaming scenarios \cite{ahmad2017unsupervised}.

In real-world deployments, an anomaly detection system must not only detect anomalies but also provide interpretable confidence scores while minimizing false alarms \cite{cook2019anomaly}. A high false alarm rate can overwhelm monitoring systems, reducing their practical utility. Our work addresses these challenges by developing an approach that enables adaptive thresholding in streaming environments while ensuring reliable anomaly detection, regardless of whether the anomalies are point-based or contextual.

\paragraph{Conformal Prediction}
Conformal prediction methods \cite{vovk2005algorithmic} have gained significant attention for their ability to provide distribution-free uncertainty quantification with finite-sample generalization guarantees \cite{shafer2008tutorial,angelopoulos2021gentle}. Among these, split conformal prediction (SCP) \cite{papadopoulos2002inductive} is a particularly appealing post-hoc, model-agnostic technique that requires only the model’s predictions and a calibration dataset. SCP estimates an empirical quantile of a nonconformity score measuring how well the model’s predictions align with the data to construct prediction sets that achieve the desired coverage. However, these guarantees rely on the exchangeability assumption \footnote{informally, a sequence of observations is exchangeable if any permutation of the observations has the same joint probability} between calibration and test observations, which often does not hold in time series settings.

For non-exchangeable data, particularly time series, several adaptive conformal prediction methods have been proposed \cite{gibbs2021adaptive,zaffran2022adaptive,gibbs2024conformal}. These approaches dynamically adjust the estimated quantile to correct for distribution shifts and achieve the target coverage level. However, they are typically designed for a single error rate objective, often optimizing the pinball loss or a surrogate function. In contrast, our work focuses on an adaptive method that remains effective across all error rates and desired alarm rate.

Weighted conformal quantile estimation \cite{gibbs2023conformal}, where the calibration or past non-conformity scores are weighted differently has been used to achieve local coverage when the distribution of the error differs across the input space. Essentially, for any given observation, scores of samples that are similar to that observation get up-weighted, usually based on some metric (e.g., proximity in the covariate space) \citep{lei2014distribution,guan2019conformal,tibshirani2019conformal,sesia2021conformal,han2022split,guan2023localized,ghosh2023improving,mao2024valid}, weights can also be optimized to capture the variance of the non-conformity score across the input space \citep{han2022split,amoukou2023adaptive}.  In the context of non-exchangeable data, \cite{barber2023conformal} derived a coverage bound linking the weights associated with a calibration sample and the total variation distance between the observed sequence and one where the calibration sample is swapped with the test sample. This bound suggests one should up-weight samples that are `nearly exchangeable' with the new observation on a pairwise basis. This is the main inspiration for our proposed approach. 

Conformal prediction has been explored for anomaly detection by setting thresholds on arbitrary anomaly scores from non-anomalous data while assuming exchangeability \cite{angelopoulos2021gentle,guan2019conformal,bates2023testing}. However, to the best of our knowledge, there is no existing method that simultaneously (i) seamlessly applies these techniques to generate interpretable, distribution-agnostic anomaly scores, (ii) directly translates scores into a desired alarm rate, and (iii) is inherently adapted to operate under non-exchangeability assumptions. Our work aims to bridge this gap by developing a conformal anomaly detection framework that is both interpretable and robust to real-world time series shifts.

\section{Proofs}
\label{sec:appendix_proofs}
\paragraph{Proof Proposition \ref{prop:conformal_detector_marginal_guarantees}}
We show the equivalence of the detector $C_{\beta_{\textbf{w}}}$ and the conformal outlier detector in \eqref{conformal_anomaly_detector} over the non-conformity scores $S$ by proving the following 
\begin{equation}
\begin{array}{cl}
     C_{\beta_{\textbf{w}}}(X_{t+1},Y_{t+1}) &= \mathbf{1}[\beta_{\textbf{w}}(S_{t+1}) > 1 - \alpha] \\
     &= \mathbf{1}[S_{t+1} > \mathbb{Q}_{1-\alpha}(\textbf{s},\textbf{w})]
\end{array}
\end{equation}
which involves proving that the events $\beta_{\textbf{w}}(S_{t+1}) < \alpha$ and $S_{t+1} > \mathbb{Q}_{1-\alpha}(\textbf{s},\textbf{w})$ are equivalent.

If $\beta_{\textbf{w}}(S_{t+1}) = \beta_{t+1} < \alpha$ then $S_{t+1} > \mathbb{Q}_{1-\alpha}(\textbf{s},\textbf{w})$ since by definition of $\beta_{\textbf{w}}(\cdot)$ in \eqref{conformal_p_value_transformation} then $\beta_{t+1}$ is the maximum value in [0,1] that satisfies the quantile upper bound.

If $\beta_{\textbf{w}}(S_{t+1}) = \beta_{t+1} \ge \alpha$ and since $S_{n+1} \le \mathbb{Q}_{1-\beta_{t+1}}(\textbf{s},\textbf{w}) \le  \mathbb{Q}_{1-\alpha'}(\textbf{s},\textbf{w}), \forall \alpha' \le \beta_{t+1}$ we have that  $S_{t+1} \le \mathbb{Q}_{1-\alpha}(\textbf{s},\textbf{w})$.

\section{Additional Experiments}
\label{sec:appendix_additonal_experiments}

\subsection{Simulated Examples}
\label{sec:appendix_simulexamples}

We consider a similar simulated setting as \cite{gibbs2023conformal} to empirically evaluate the performance of the proposed method across time. We analyze a simple scenario where we observe a sequence of random variables \(\{Y_t\}_{t=1}^T\), where \(Y_t \sim \mathcal{N}(\mu_t, 1)\). We assume that our predictive model $h$ outputs a constant $\hat{Y}_{t} =0, \forall t$. Then the error is $\epsilon_t = Y_t  - \hat{Y}_t  \sim \mathcal{N}(\mu_t, 1)$ and its distribution changes across time based on $\mu_t$.  The nonconformity score is $s_t = |\epsilon_t|, \forall t$.  
We consider two different settings for the sequence of means \(\{\mu_t\}_{t=1}^T\):  
\begin{itemize}
    \item \textbf{Random shift setting:} \(\mu_t\) drifts continuously across time. Specifically, we set \(\mu_0 = 0\) and  
        \begin{equation}
           \mu_{t+1} = \mu_t + \frac{1}{2}(\mu_t - \mu_{t-1}) + \frac{1}{2}\epsilon_t, \{\epsilon_t\} {\sim} \mathcal{N}(0, 0.05), \forall t .
    \end{equation}

    \item \textbf{Jump shift setting:} \(\mu_t\) undergoes abrupt discontinuities every 500 time steps where \(\mu_t\) increases by one step 15 times, and then starts decreasing by 1, 
    \begin{equation}
        \begin{array}{c}
            \mu_t =  \lfloor t/500 \rfloor \mathbf{1}[\lfloor t/500 \rfloor< 15]  
              +[15 - \lfloor t/500 \rfloor] \mathbf{1}[\lfloor t/500 \rfloor\ge 15].
        \end{array}
    \end{equation}
\end{itemize}
Given an observed non-conformity score $s_t = |\epsilon_t|$ we can compute its corresponding p-value $\alpha_t$ such that $\mathbb{P}_{\boldsymbol\epsilon_t \sim \mathcal{N(\mu_t,1)}}(|\boldsymbol\epsilon_t| > s_t) =  \alpha_t = 1 - \Phi(s_t - \mu_t) + \Phi(- s_t-\mu_t)$ and compare it with the one estimated by the proposed normalized anomaly score $\beta_\mathbf{w}(s_t)$. 

\begin{figure}[!ht]
    \centering
    \subfigure[Random Shift Signal]{\includegraphics[width=0.30\textwidth]{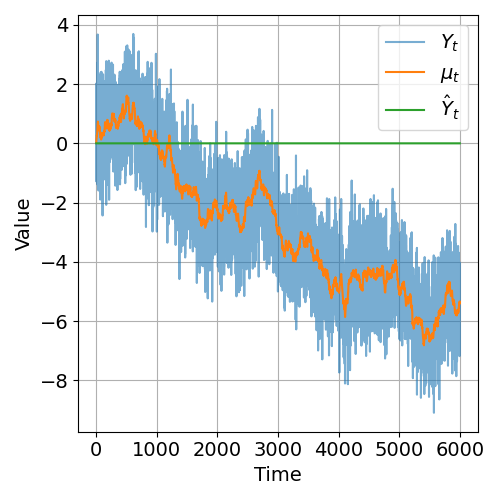}}
    \subfigure[Random Shift Calibration]{\includegraphics[width=0.30\textwidth]{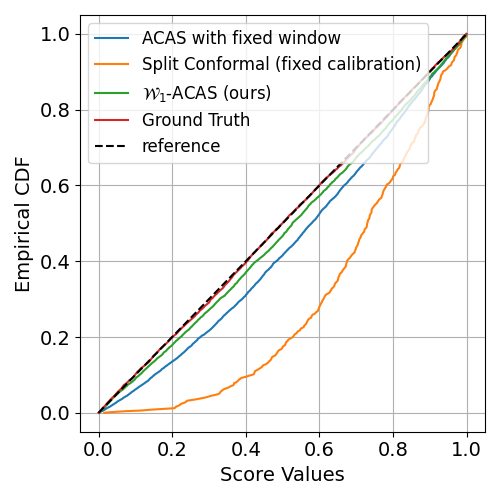}}
    \subfigure[Random Shift Error]{\includegraphics[width=0.30\textwidth]{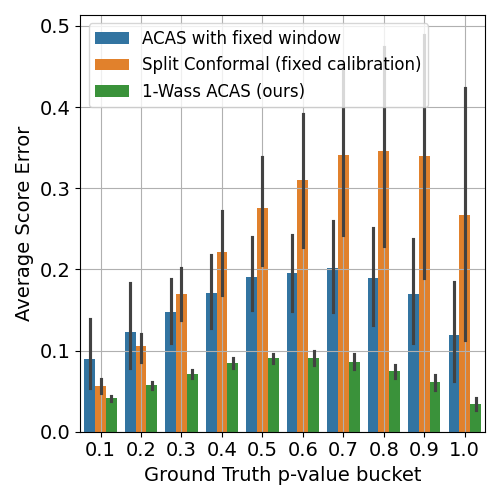}}
        
    \subfigure[Jump Shift Signal]{\includegraphics[width=0.30\textwidth]{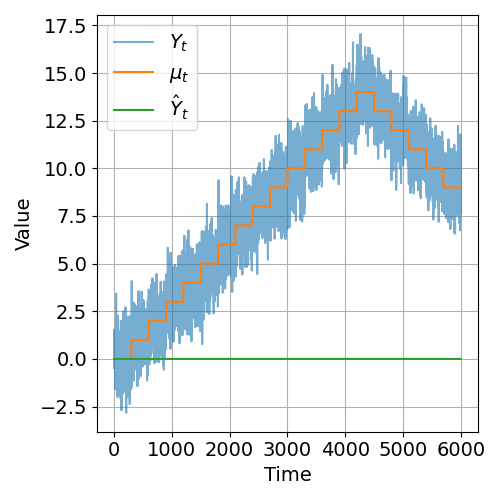}}
    \subfigure[Jump Shift Calibration]{\includegraphics[width=0.30\textwidth]{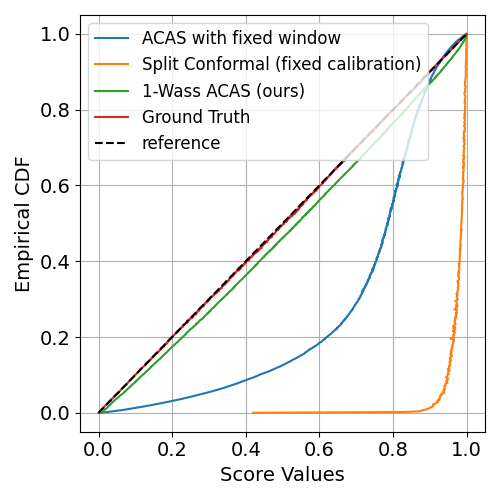}}
    \subfigure[Jump Shift Error]{\includegraphics[width=0.30\textwidth]{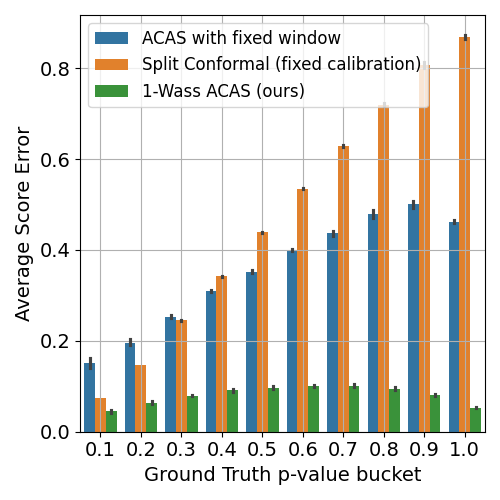}}
    
    \caption{ Figures (a) and (d) show an example of a generated signal under the random shift and jump shift settings with a sequence length of \(T = 6000\). $\mu_t$ is the expected value of the observed signal, $Y_t$ is the observed signal $Y_t \sim \mathcal{N}(\mu_t,1)$, and $\hat{Y}_t = 0$ the predicted value of a naive constant forecaster. Figures (b) and (e) show the empirical cumulative distribution functions (CDFs) of the various calibration approaches compared to the ground truth p-values (Ground Truth), which aligns with the idealized uniform CDF (reference). Results are averaged over $15$ realizations. $\mathcal{W}_1\textsc{-ACAS}$ demonstrates superior calibration, closely matching the ground truth distribution and improving upon the reference split conformal method (computed over calibration samples) and a fixed window ACAS method.  Figures (c) and (f) show the average absolute error of the scores of the different methods with respect to the ground truth p-values, evaluated across bucket ranges of size $0.1$ in \([0,1]\). $\mathcal{W}_1\textsc{-ACAS}$ consistently achieves lower estimation errors, highlighting the effectiveness of its adaptive weighting strategy with minimum parameters.}
    \label{fig:synthetic_data}
\end{figure}

Figures \ref{fig:synthetic_data}.a and \ref{fig:synthetic_data}.b illustrate a sample of the generated signals under the Random Shift and Jump Shift settings. Each sequence consists of \(T=6000\) time steps, and our results are averaged over 15 independent realizations. We assess the performance of our proposed approach, Algorithm \ref{alg:acas}, referred to as $\mathcal{W}_1\textsc{-ACAS}$, with parameters \(n = 2000\), \(\alpha_c = 0.01\), and \(n_b = n_c = \lceil\frac{1}{\alpha_c}-1\rceil\). We compare it against two baseline methods: (i) an adaptive conformal approach that assigns equal weights of 1 to the most recent 2000 samples (\textsc{ACAS} with a fixed window) and (ii) a naive split conformal approach that computes scores using only the initial 100 samples (Split Conformal with fixed calibration).  

In Figures \ref{fig:synthetic_data}.c and \ref{fig:synthetic_data}.d, we compare the empirical CDFs of each method against the empirical CDF of the ground truth p-values (denoted as Ground Truth), which naturally aligns with the identity line (reference). Notably, $\mathcal{W}_1\textsc{-ACAS}$ demonstrates superior calibration, consistently aligning closely with the ground truth CDF and outperforming the other approaches.  

Figures \ref{fig:synthetic_data}.e and \ref{fig:synthetic_data}.f present the average error of the scores of each method with respect to the ground truth p-values, across different bucket ranges of size 0.1 within \([0,1]\). Specifically, we evaluate \(\mathbb{E}[|\beta_{\mathbf{w}}(s_{t+1}) - \alpha_{t+1}| \mid \alpha_t \in [\alpha_l,\alpha_u]]\), where \(\alpha_{t+1}\) represents the ground truth p-value for observation \(t+1\). The results indicate that $\mathcal{W}_1\textsc{-ACAS}$ consistently outperforms the baseline methods, highlighting the advantages of an adaptive approach that dynamically learns how to weight past observations in a principled manner, rather than relying on a fixed number of past samples.

\subsection{Anomaly Detection Real Datasets}
\label{sec:appendix_experiments_realdatasets}

\subsubsection{Baseline Methods}
\label{sec:appendix_baseline_methods}
We compare $\mathcal{W}_1$\textsc{-ACAS} against two TSFM-based baselines. The first fits a \textbf{Gaussian} distribution to the mean absolute forecast error across $d$ steps using the calibration portion and assigns anomaly scores via the resulting $p$-values. The second applies a \textbf{Conformal} offline approach that learns a $p$-value mapping from the calibration split for each $d$ and aggregates the scores by the median. These baselines provide simple references built directly on TSFM errors.

We additionally consider several classic anomaly detection methods reported as top-performing in \citet{liu2024elephant}: 

\begin{itemize}
    \item \textbf{KShape}~\citep{paparrizos2015k, paparrizos2017fast, boniol2021sand}, which clusters subsequences via the k-Shape algorithm and scores anomalies by their distance to cluster centroids;
    \item \textbf{POLY}~\citep{li2007unifying}, which fits a polynomial to the series and applies a GARCH model to residuals to estimate volatility; 
    \item \textbf{Sub-PCA}~\citep{aggarwal2017introduction}, which projects subsequences onto a lower-dimensional hyperplane and scores deviations; 
    \item \textbf{Sub-KNN}~\citep{ramaswamy2000efficient}, which scores each instance by its distance to the $k$-th nearest neighbor; 
    \item \textbf{SAND}~\citep{boniol2021sand}, an online method that adaptively down-weights older subsequences. 
    \item {\textbf{CNN} ~\citep{munir2018deepant}, is a causal convolutional forecasting model, the anomaly score is the prediction error. It is trained on non-anomalous data.}
    \item  {\textbf{USAD}~\citep{audibert2020usad} is an adversarially trained dual–autoencoder model learned on non-anomalous data, where anomaly scores are computed from a weighted reconstruction loss.}
    \item {\textbf{OmniAnomaly} ~\citep{su2019robust} is a stochastic recurrent VAE that incorporates GRU dynamics, planar normalizing flows, and temporal latent stochasticity; anomaly scores are derived from reconstruction probabilities. It is trained on non-anomalous data.}
    \item {\textbf{MOMENT}~\citep{goswami2024moment} is a general-purpose TSFM based on a T5-style encoder trained via masked time-series modeling. It supports zero-shot anomaly scoring using masked-token reconstruction error and is pretrained on a broad corpus including anomaly detection datasets~\citep{liu2024elephant}.}

\end{itemize}
For these approaches we adopt the implementations from \citet{liu2024elephant} with the best reported hyperparameters and their default $[0,1]$ min–max normalization fitted on the full dataset. 
For consistency with our $p$-value scoring (where lower values indicate greater anomaly), we take one minus the reported score. 
Unlike our method, these baselines require access to the full test set, whereas ours supports adaptive, causal anomaly detection without full-dataset access.

\subsubsection{Metrics}
\label{sec:appendix_metrics}
\textbf{Threshold-dependent metrics.}  
We follow the evaluation pipeline provided in \cite{liu2024elephant}. Given anomaly scores $\{\beta_i \in [0,1]\}_{i=1}^t$ (interpreted as $p$-values, where smaller values indicate stronger outliers) and ground-truth labels $\{\ell_i \in \{0,1\}\}_{i=1}^t$, we evaluate metrics $M(\{\ell_i\}, \{\hat{\ell}_i\}) \in [0,1]$ where larger is better. Examples include Affiliation-F and PA-F1. For a family of thresholds $\{\alpha_j \in [0,1]\}_{j=1}^k$, we select the best score
\[
j^* = \arg\max_{j \in [k]} M\Big(\{\ell_i\}, \{ \mathbf{1}[\beta_i \le \alpha_j]\}\Big), 
\quad M^* = M\Big(\{\ell_i\}, \{ \mathbf{1}[\beta_i \le \alpha_{j^*}]\}\Big),
\]
and report the corresponding false positive rate
\(
FPR(\alpha_{j^*}) = \mathbb{P}[\beta_i \le \alpha_{j^*} \mid \ell_i=0],
\)
as well as the calibration error
\(
CalErr(\alpha_{j^*}) = \big| FPR(\alpha_{j^*}) - \alpha_{j^*} \big|.
\)
Thresholds are evaluated on a uniform grid (linspace) with finer resolution at small $p$-values: 21 values in $[0.001,0.01]$, 21 values in $[0.02,0.1]$, and 21 values in $[0.2,1]$.

\textbf{Threshold-independent metrics.}  
We also report AUC and VUS-PR. In both cases, integration is performed using 250 quantiles of each method’s calibration score distribution.

\begin{figure}[!ht]
    \centering
    \subfigure[NAB-Detection]{\includegraphics[width=0.45\textwidth]{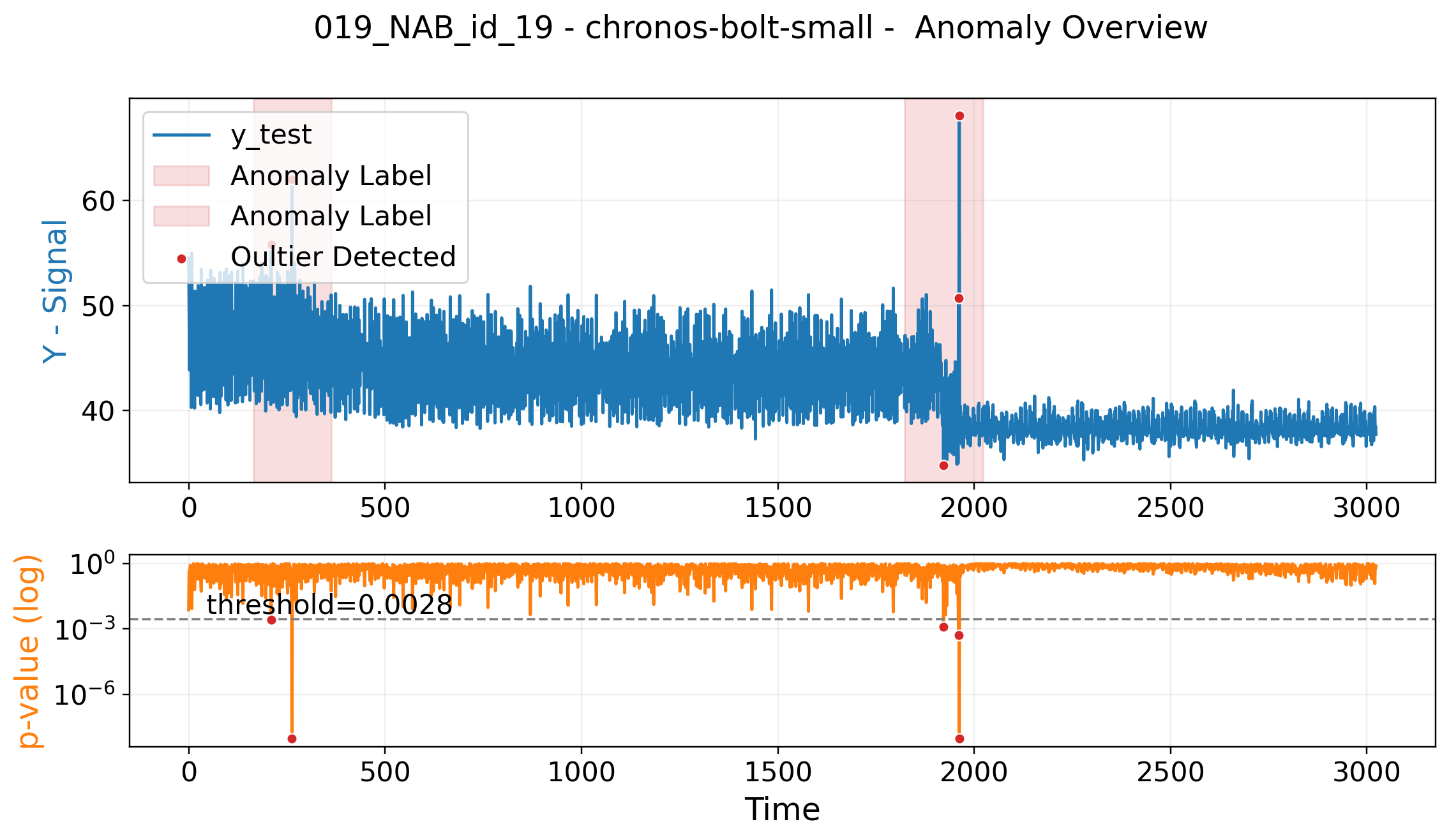}}
    \subfigure[YAHOO-Detection]{\includegraphics[width=0.45\textwidth]{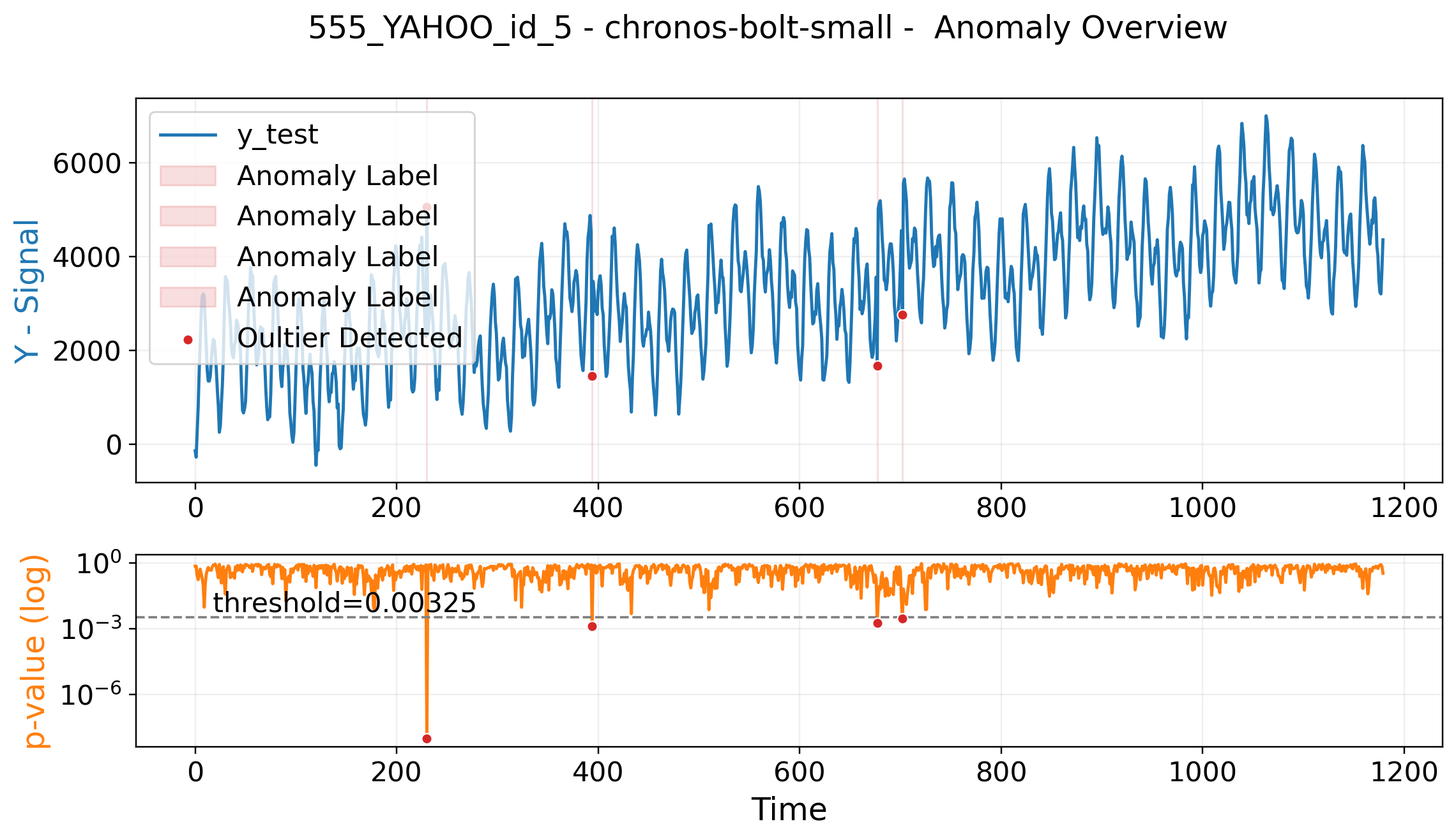}}
    
    \subfigure[STOCK-Detection]{\includegraphics[width=0.45\textwidth]{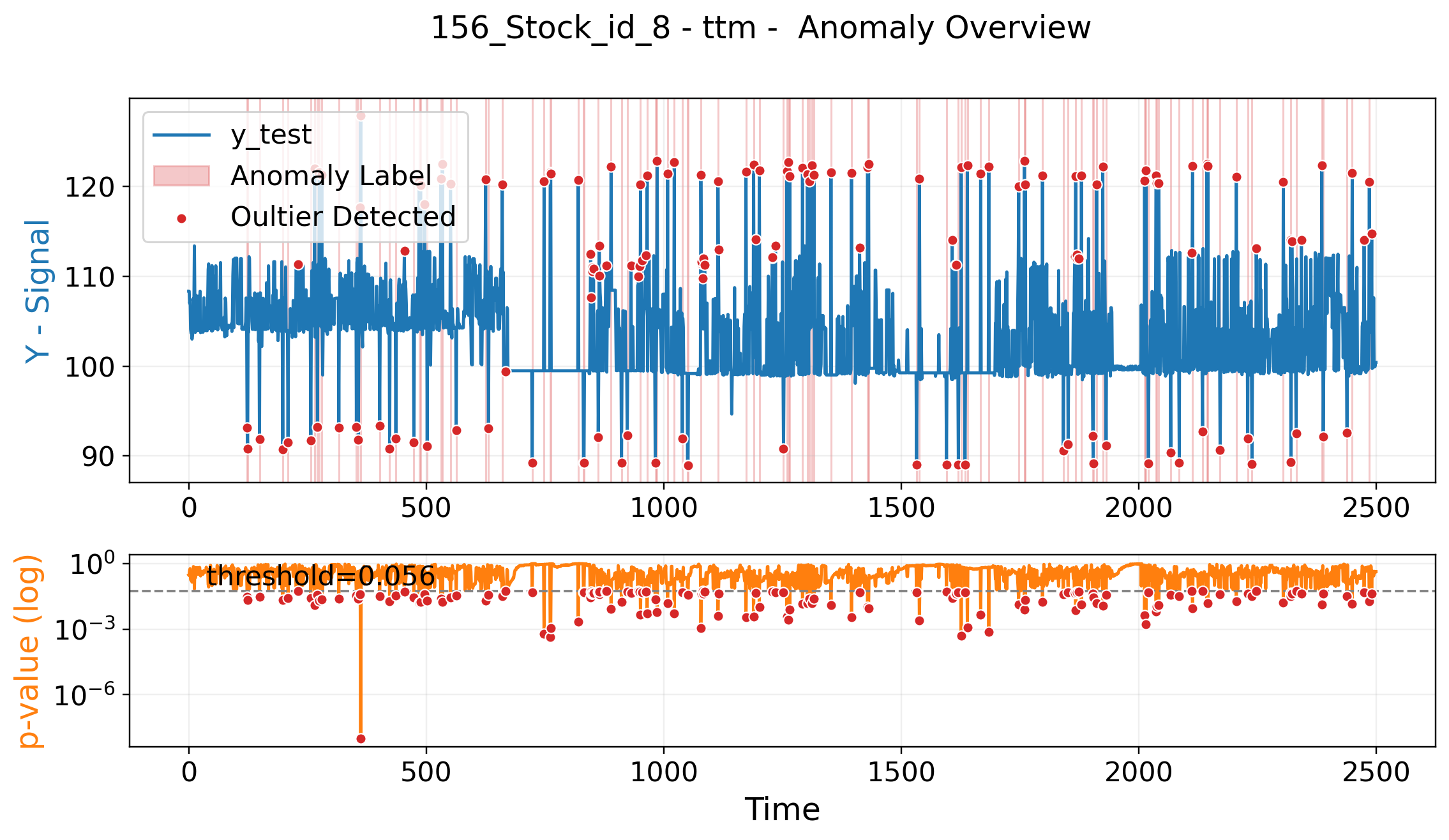}}
    \subfigure[NEK-Detection]{\includegraphics[width=0.45\textwidth]{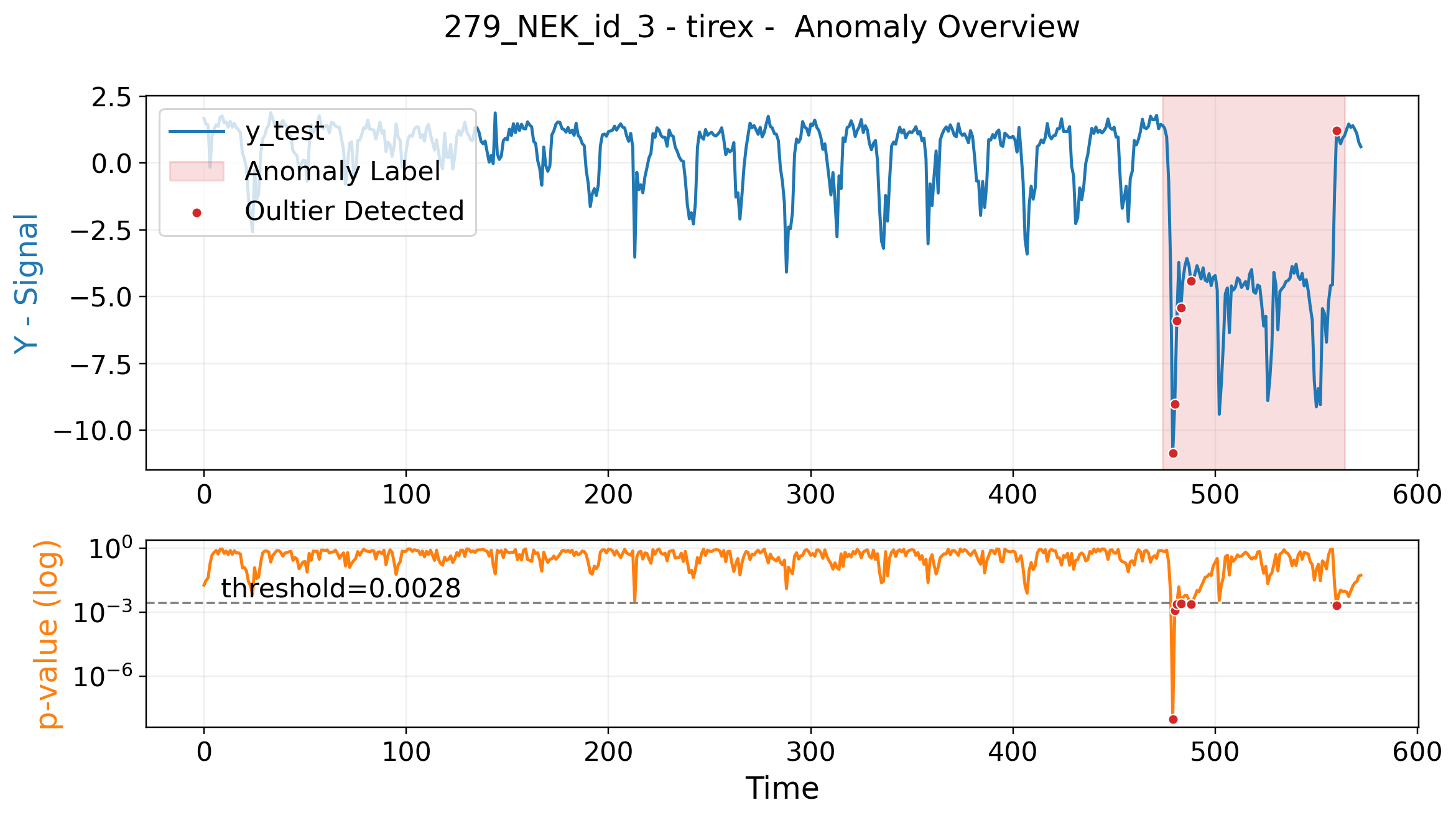}}
    
    \subfigure[WSD-Detection]{\includegraphics[width=0.45\textwidth]{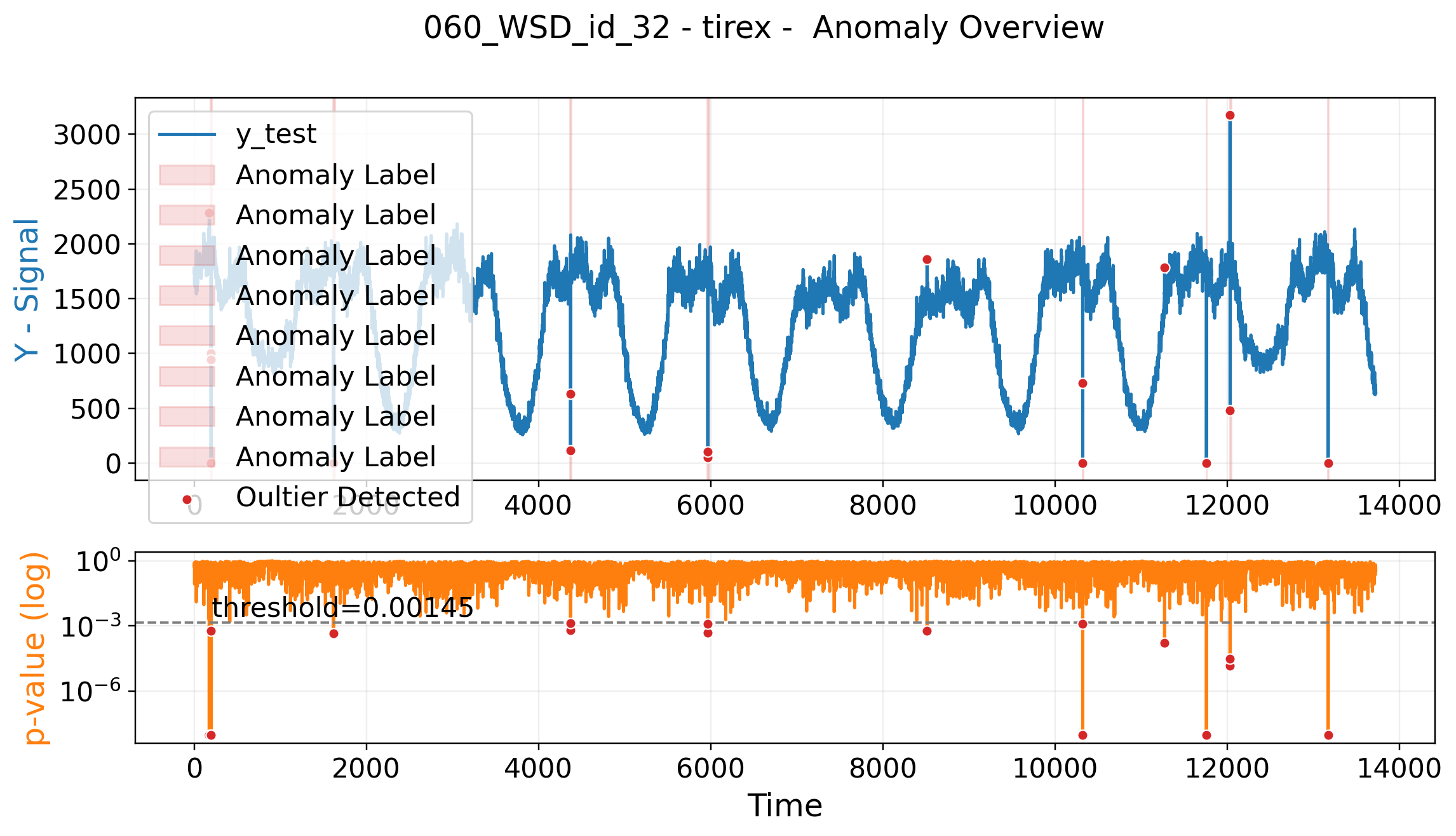}}
    \subfigure[MSL-Detection]{\includegraphics[width=0.45\textwidth]{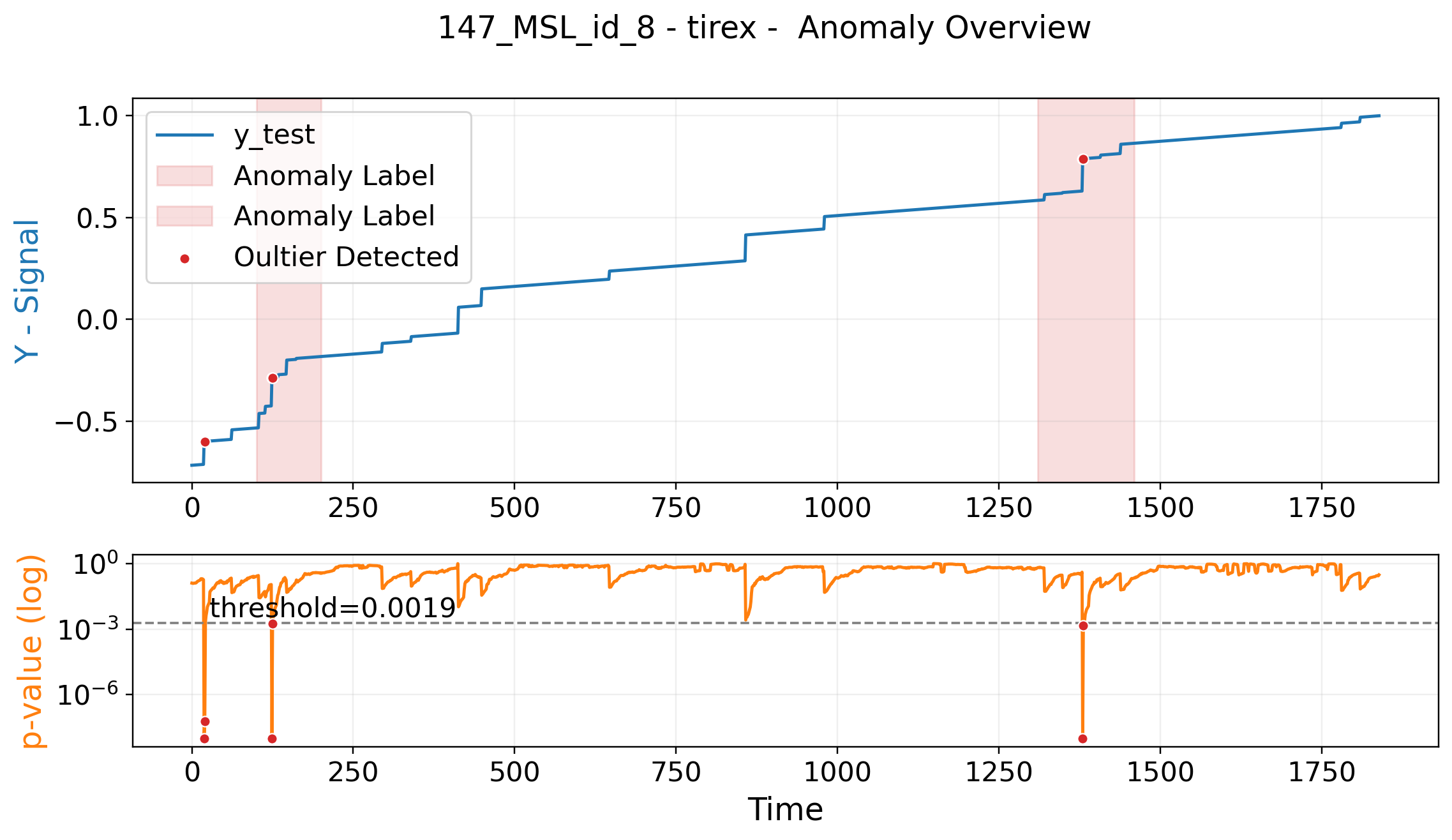}}

    \subfigure[WSD-Detection]{\includegraphics[width=0.45\textwidth]{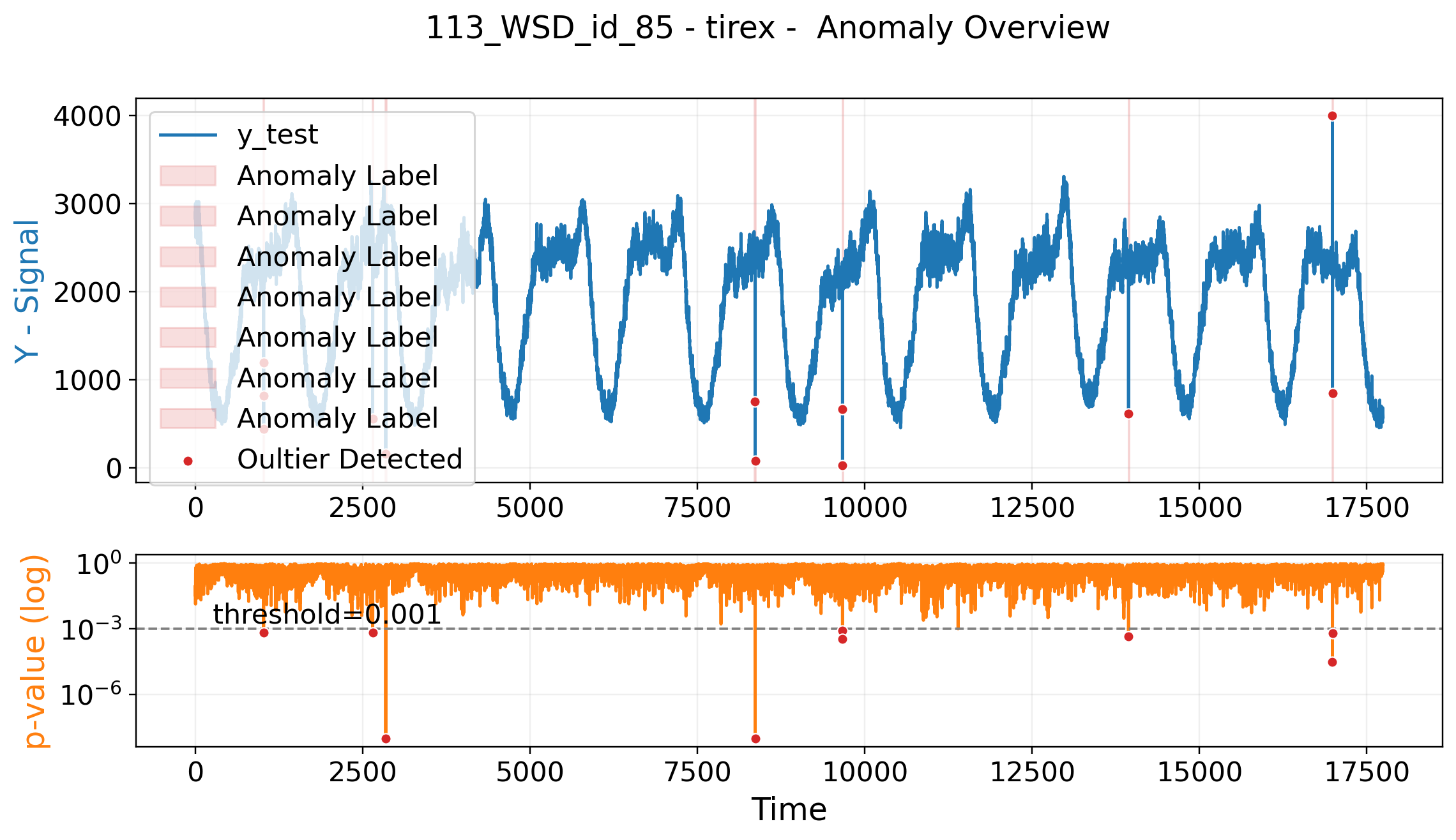}}
    \subfigure[IOPS-Detection]{\includegraphics[width=0.45\textwidth]{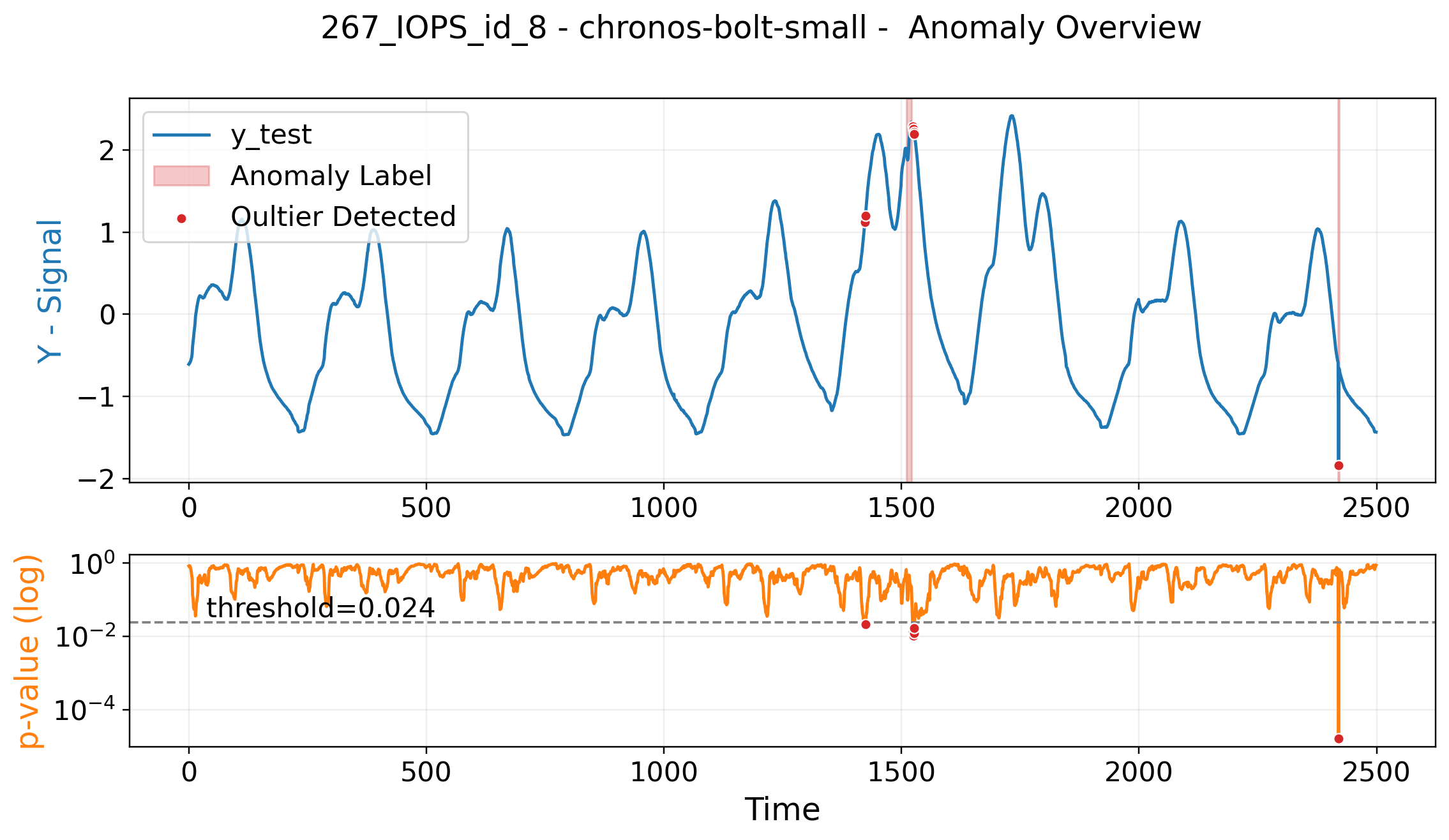}}
    
    
    \caption{Example signals (blue) with ground-truth anomaly labels (red areas), detected outliers (red dots) occur when adaptive $p$-values (orange) fall below a threshold under our proposed $\mathcal{W}_1$\textsc{-ACAS} method.}
    \label{fig:detection_examples_with_weights}
\end{figure}

\begin{figure*}[!ht]
    \centering
    
    \subfigure[IOPS]{\includegraphics[width=0.36\textwidth]{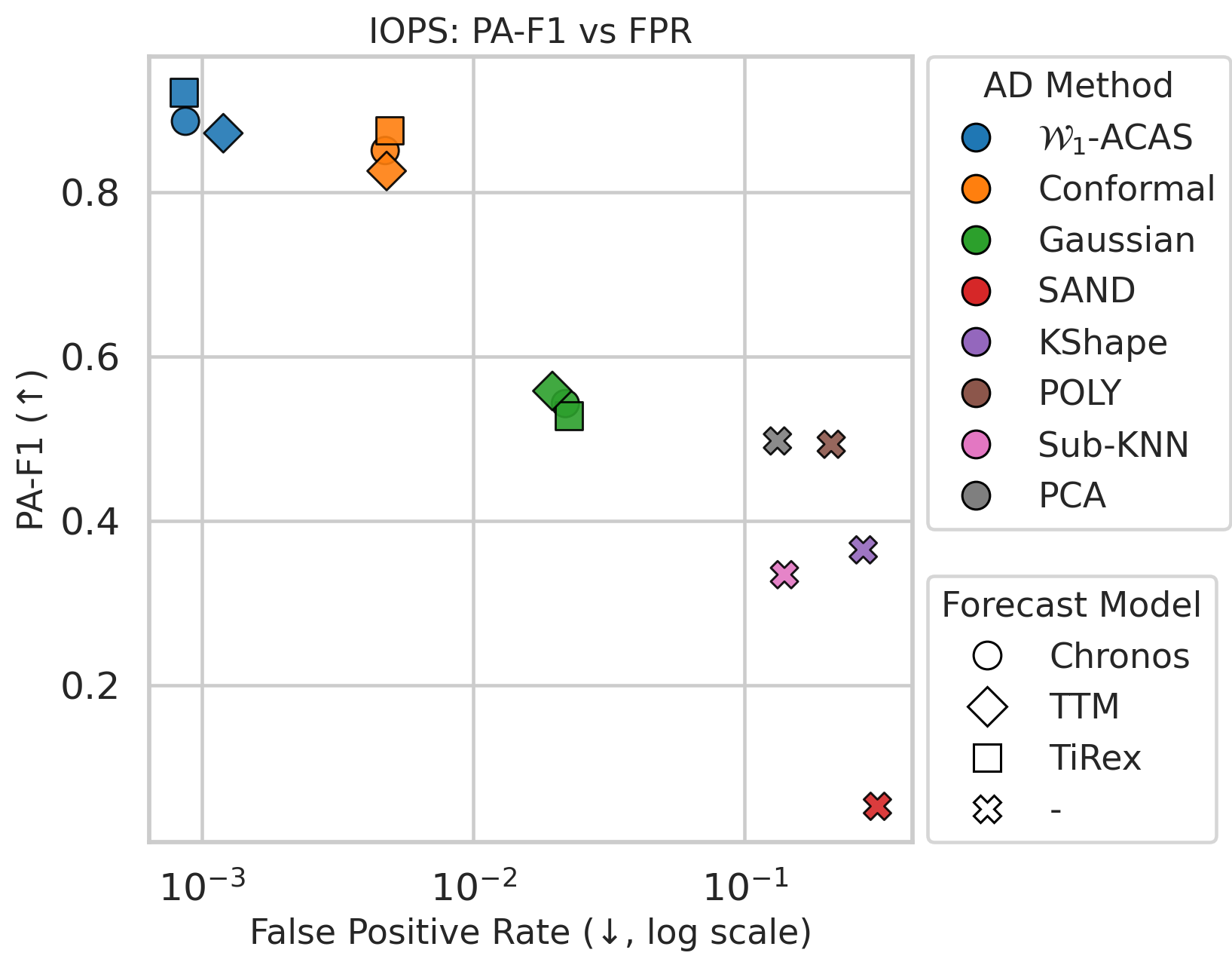}}
    \subfigure[IOPS]{\includegraphics[width=0.36\textwidth]{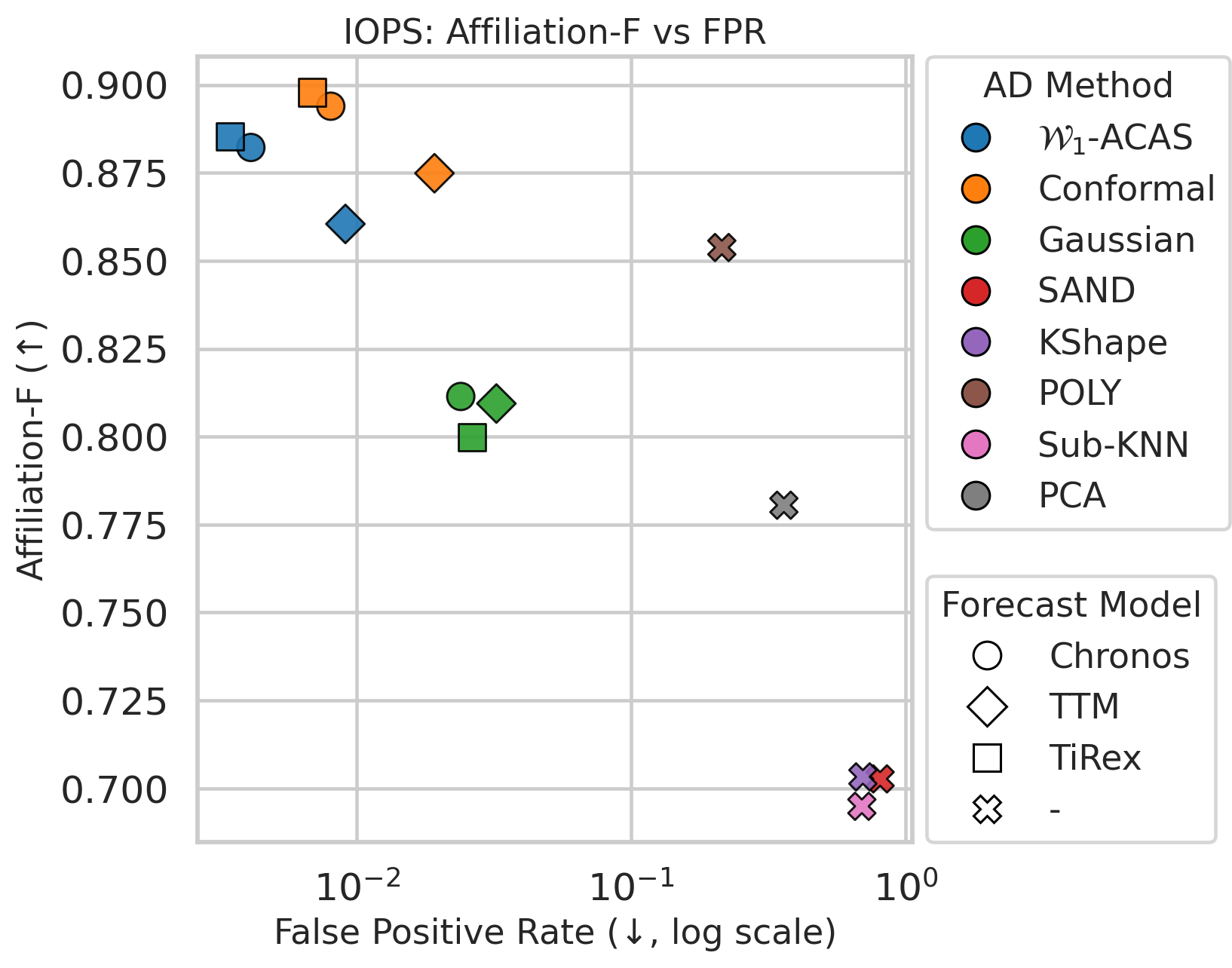}}

    \subfigure[NEK]{\includegraphics[width=0.36\textwidth]{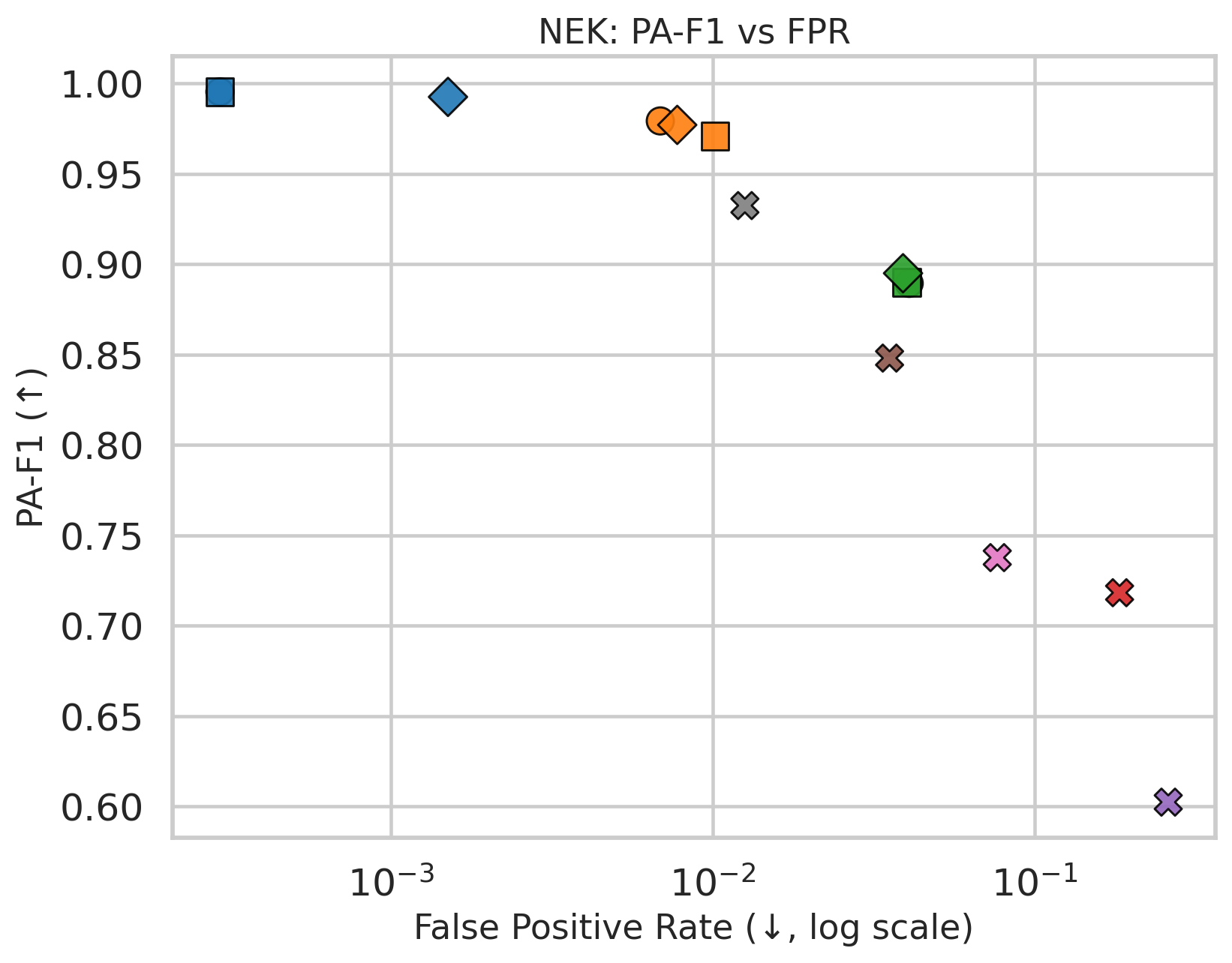}}
    \subfigure[NEK]{\includegraphics[width=0.36\textwidth]{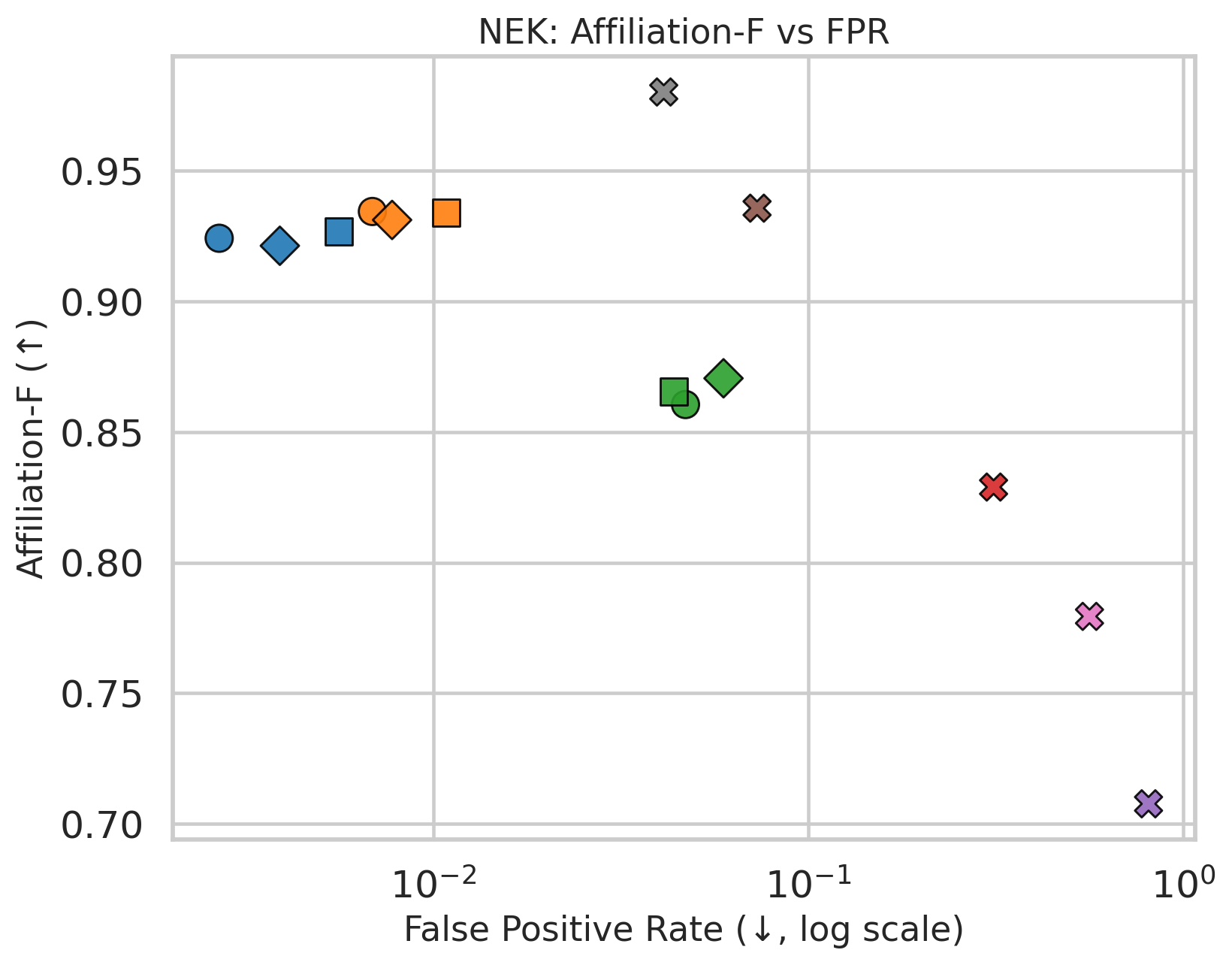}}
    
    \subfigure[YAHOO]{\includegraphics[width=0.36\textwidth]{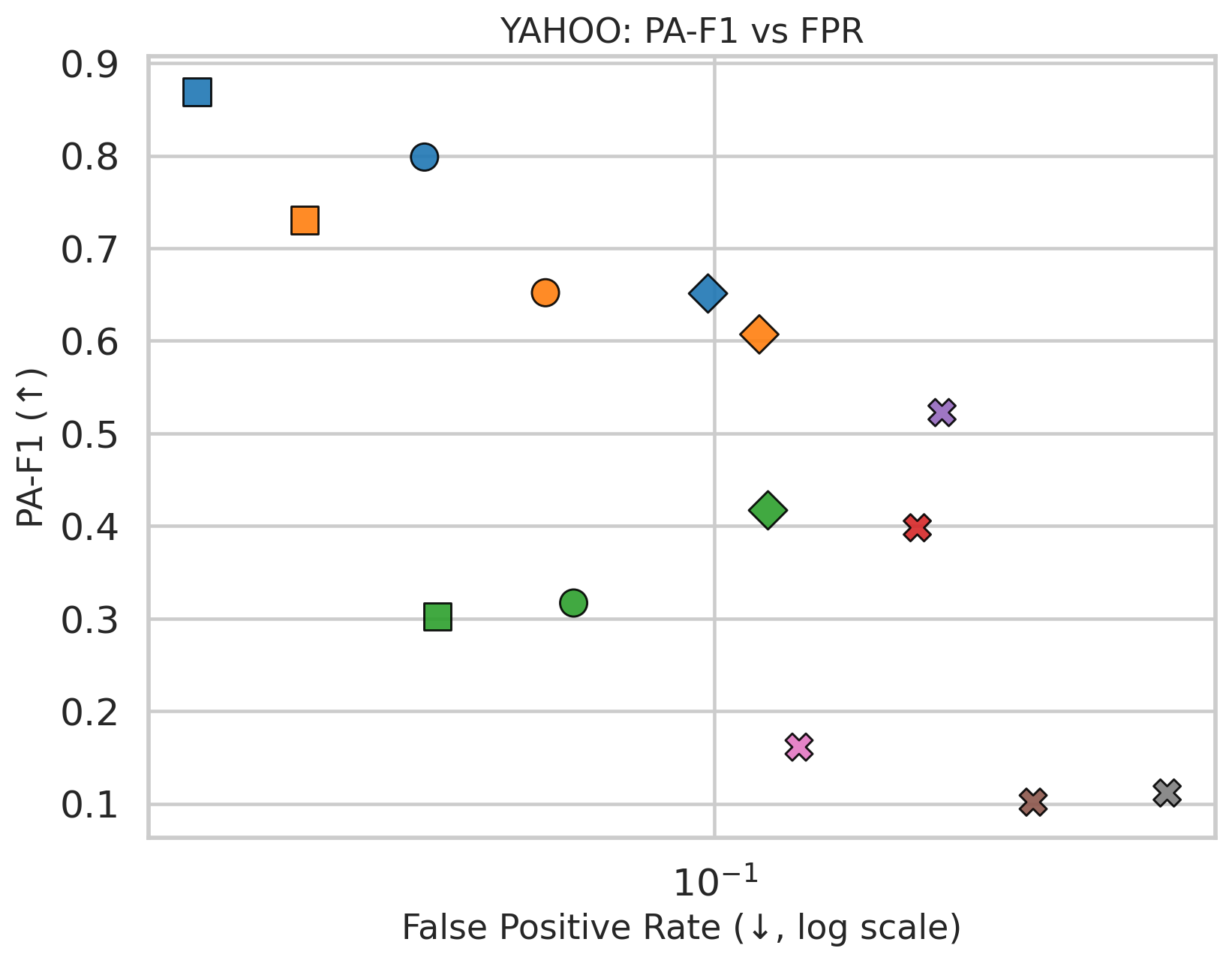}}
    \subfigure[YAHOO]{\includegraphics[width=0.36\textwidth]{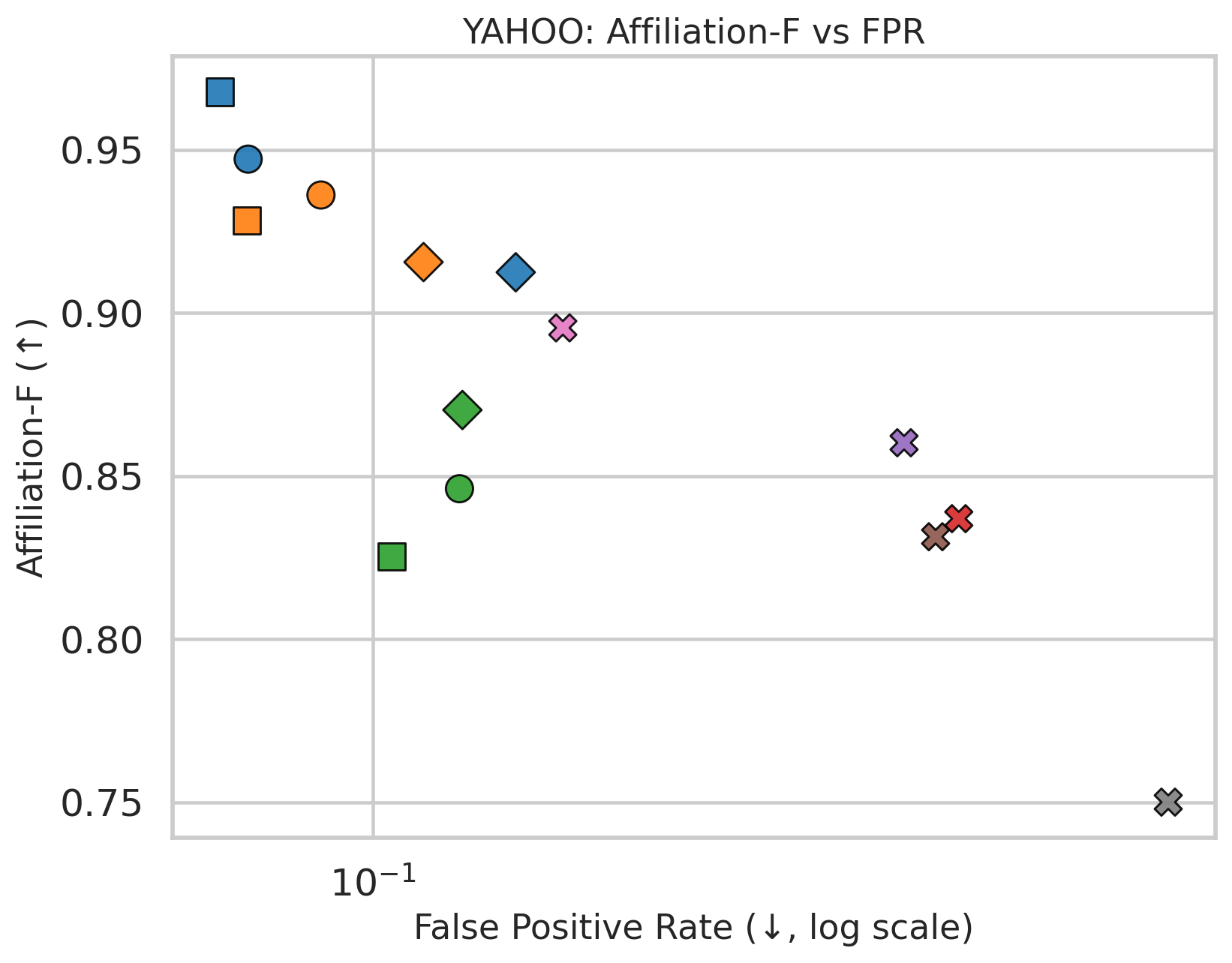}}
    
    \subfigure[NAB]{\includegraphics[width=0.36\textwidth]{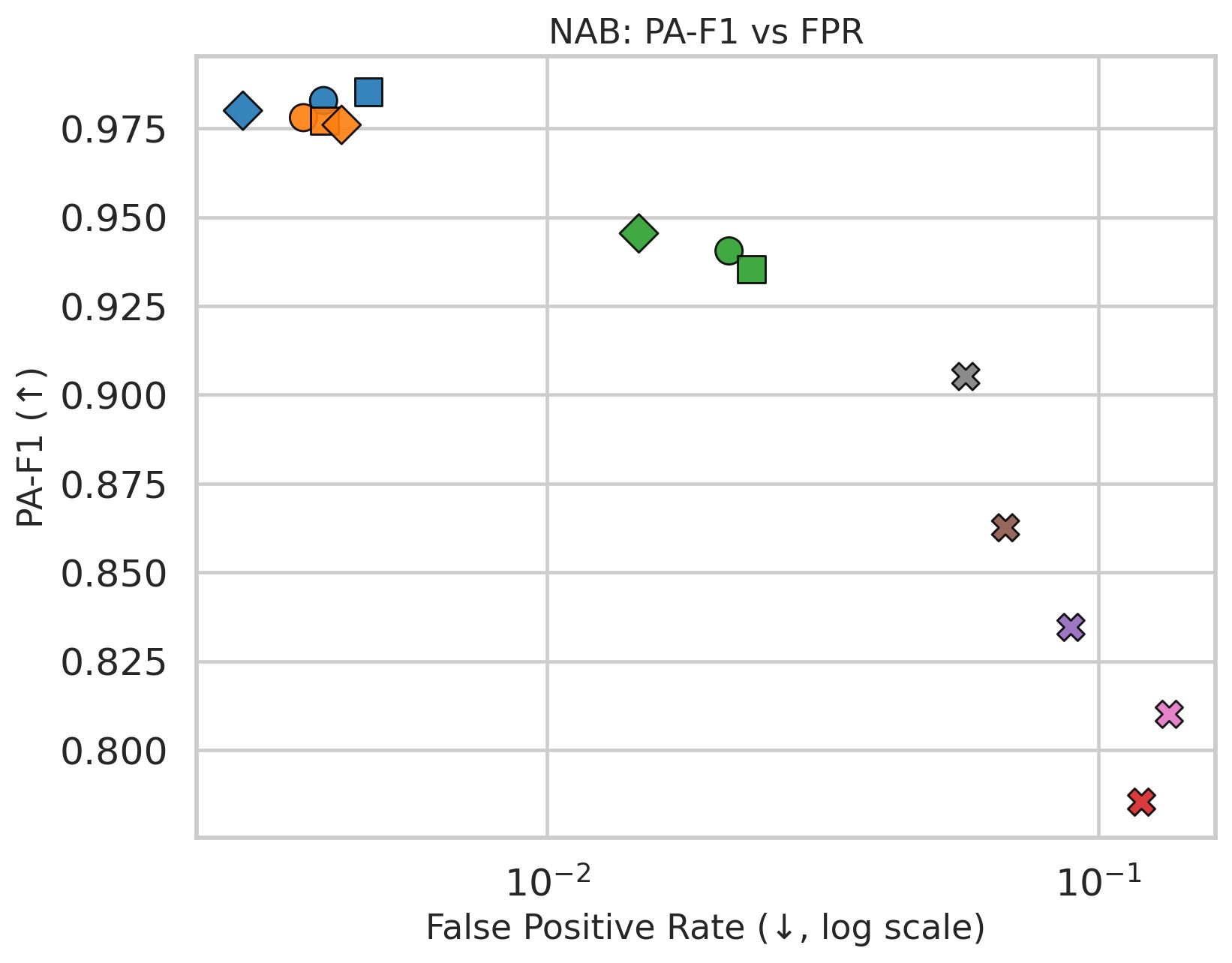}}
    \subfigure[NAB]{\includegraphics[width=0.36\textwidth]{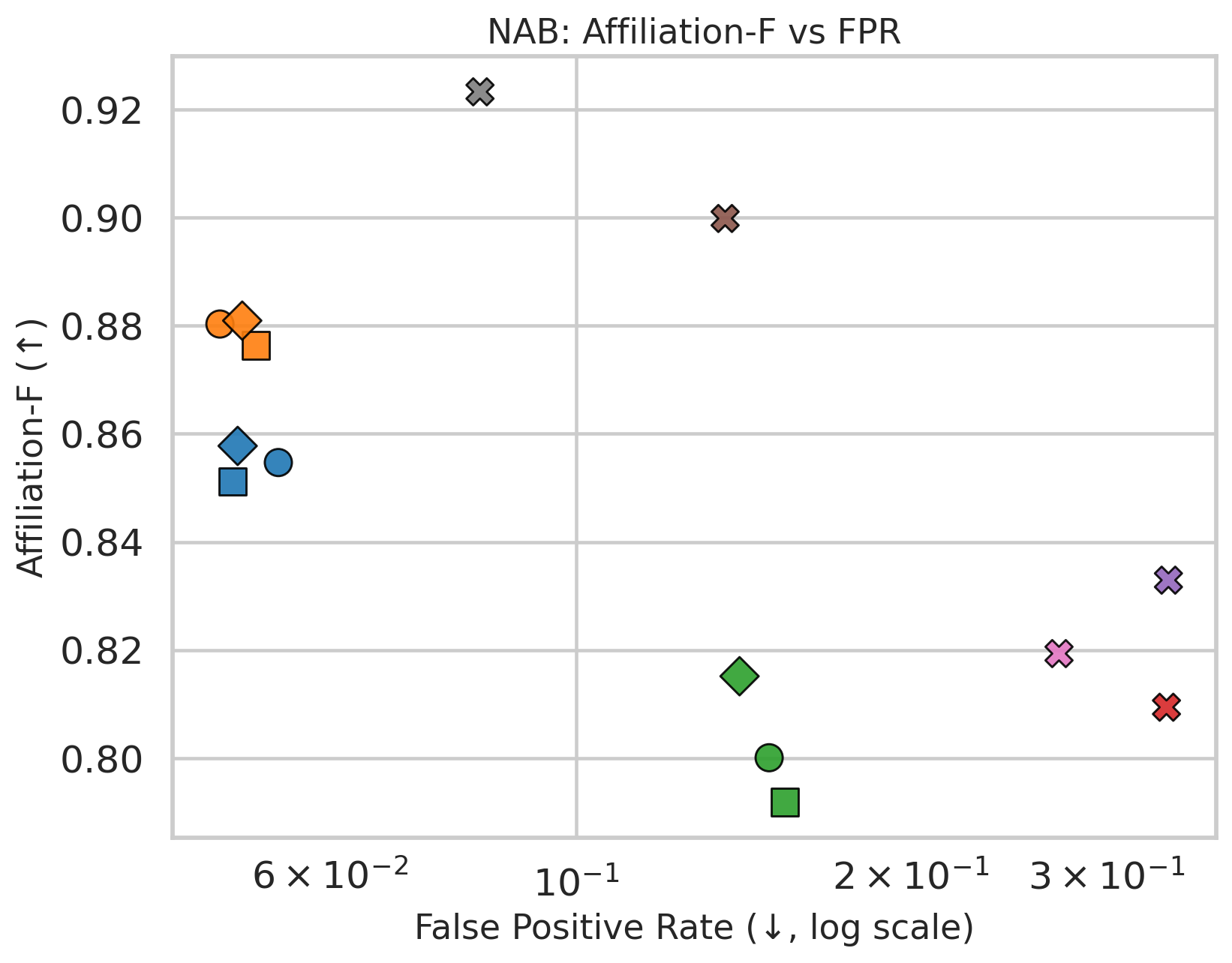}}
    

    \caption{\textbf{Trade-offs between false positive rate and detection performance across datasets.} 
    Left column: PA-F1 vs FPR (log scale). Right column: Affiliation-F vs FPR (log scale). 
    Each point uses color for AD method and marker for forecast model. The operating points of $\mathcal{W}_1$\textsc{-ACAS} (blue), in most cases, achieve both the highest F1 score and lowest FPR, especially for PA-F1. Within the same TSFM model, $\mathcal{W}_1$\textsc{-ACAS} is better than the alternatives, and in general dominate most of the alternatives.}
    \label{fig:pa-aff-tradeoff}
\end{figure*}

\begin{figure*}[t]
  \centering

  \subfigure[NAB — PA-F1]{\includegraphics[width=0.24\textwidth]{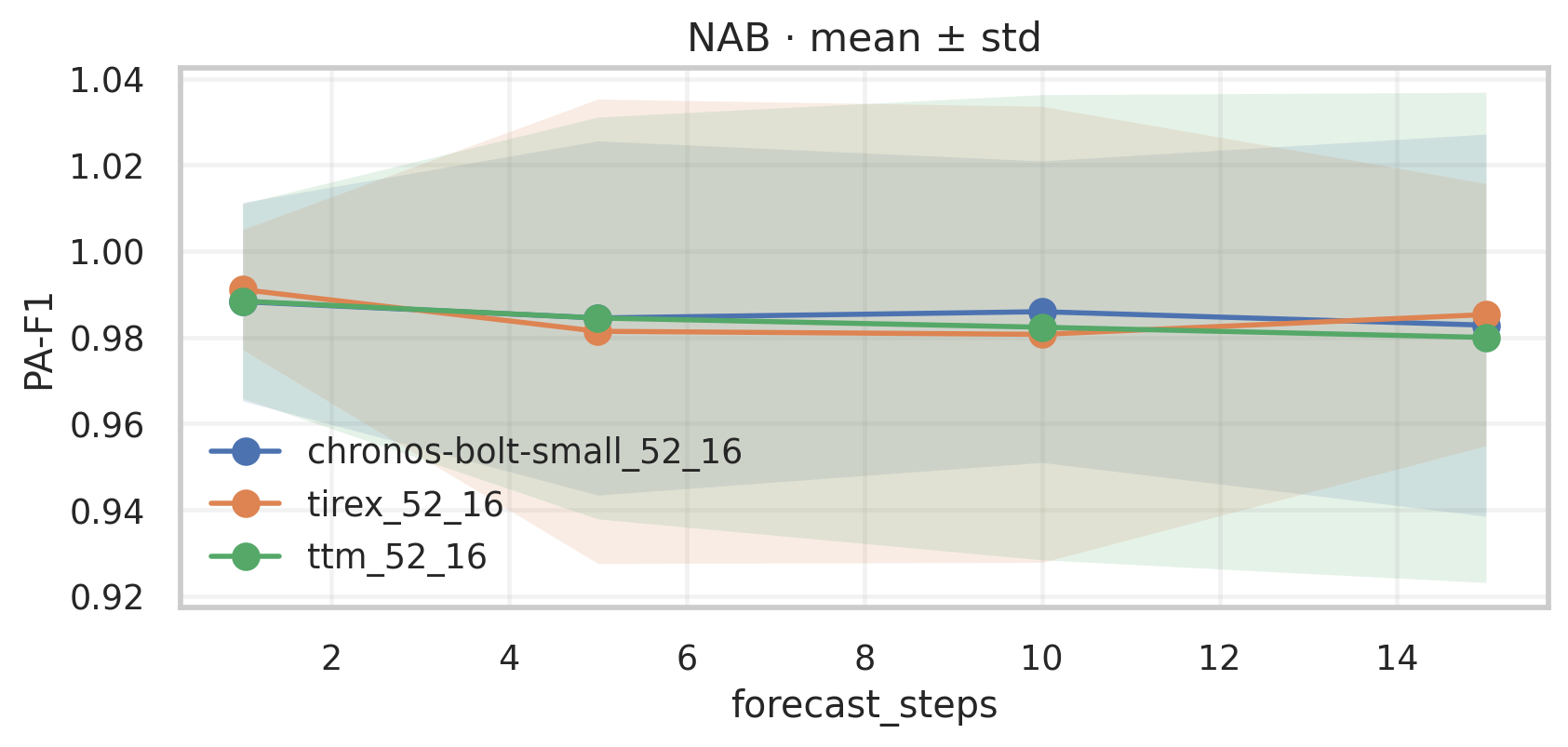}}
  \subfigure[NAB — Affiliation-F]{\includegraphics[width=0.24\textwidth]{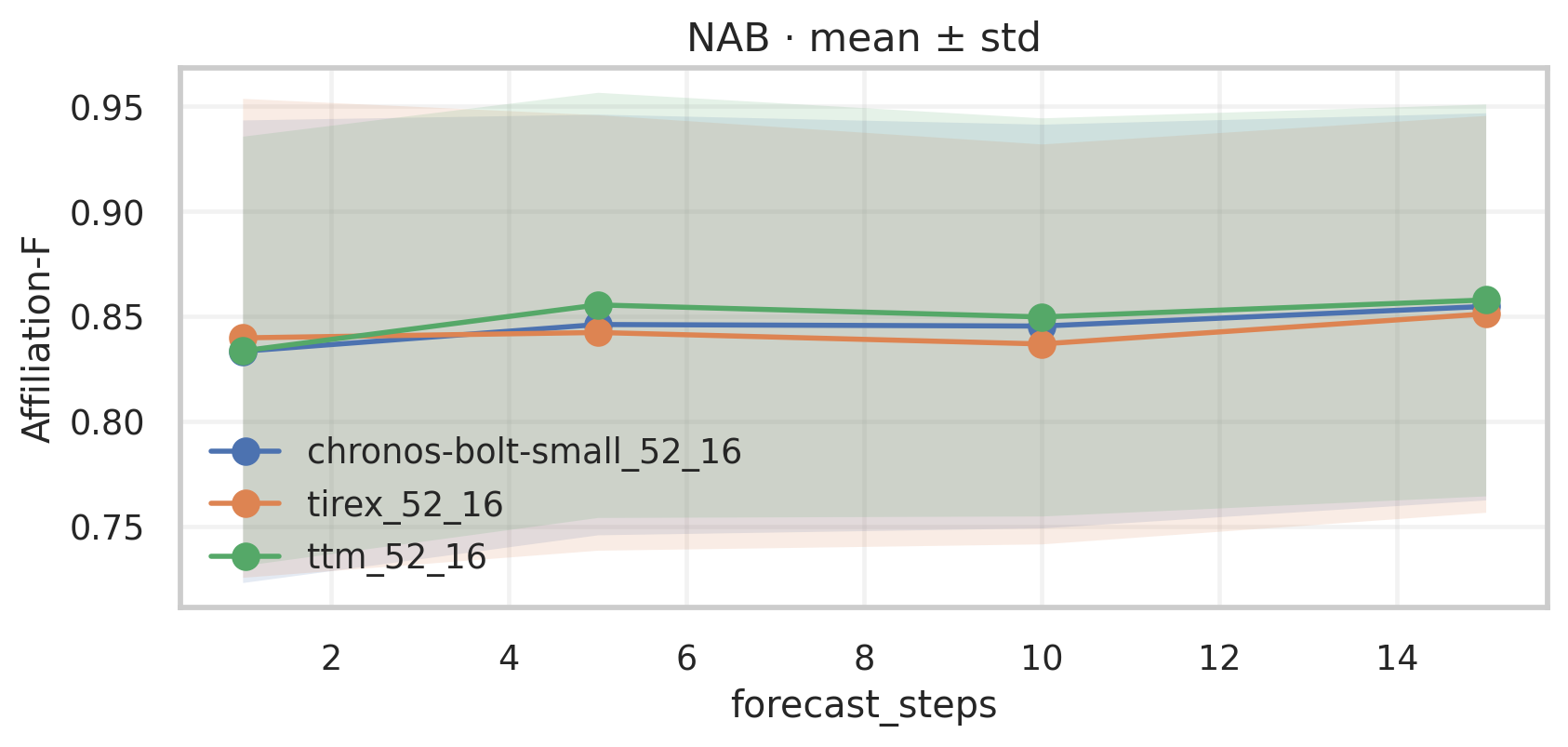}}
  \subfigure[NAB — AUC-PR]{\includegraphics[width=0.24\textwidth]{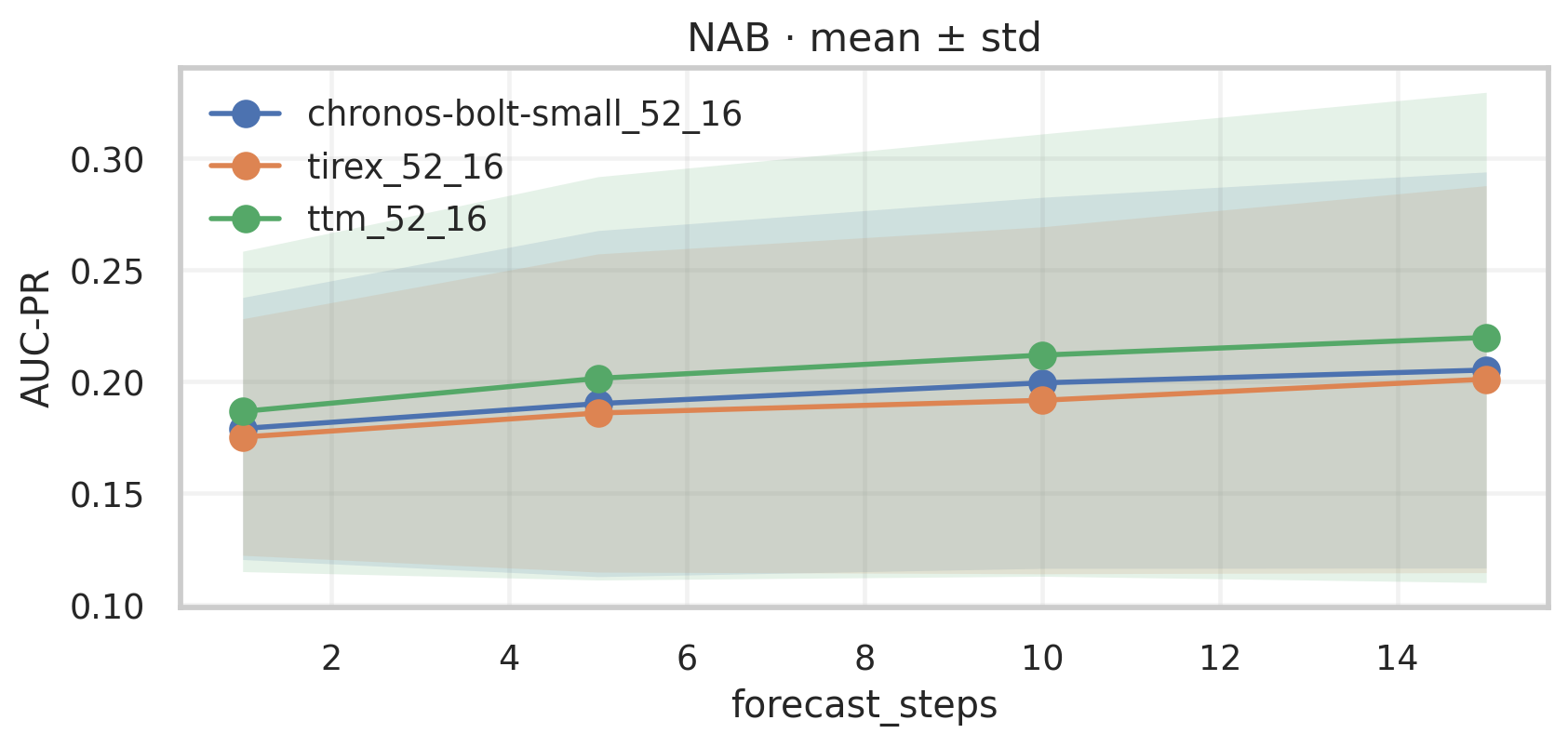}}
  \subfigure[NAB — VUS-PR]{\includegraphics[width=0.24\textwidth]{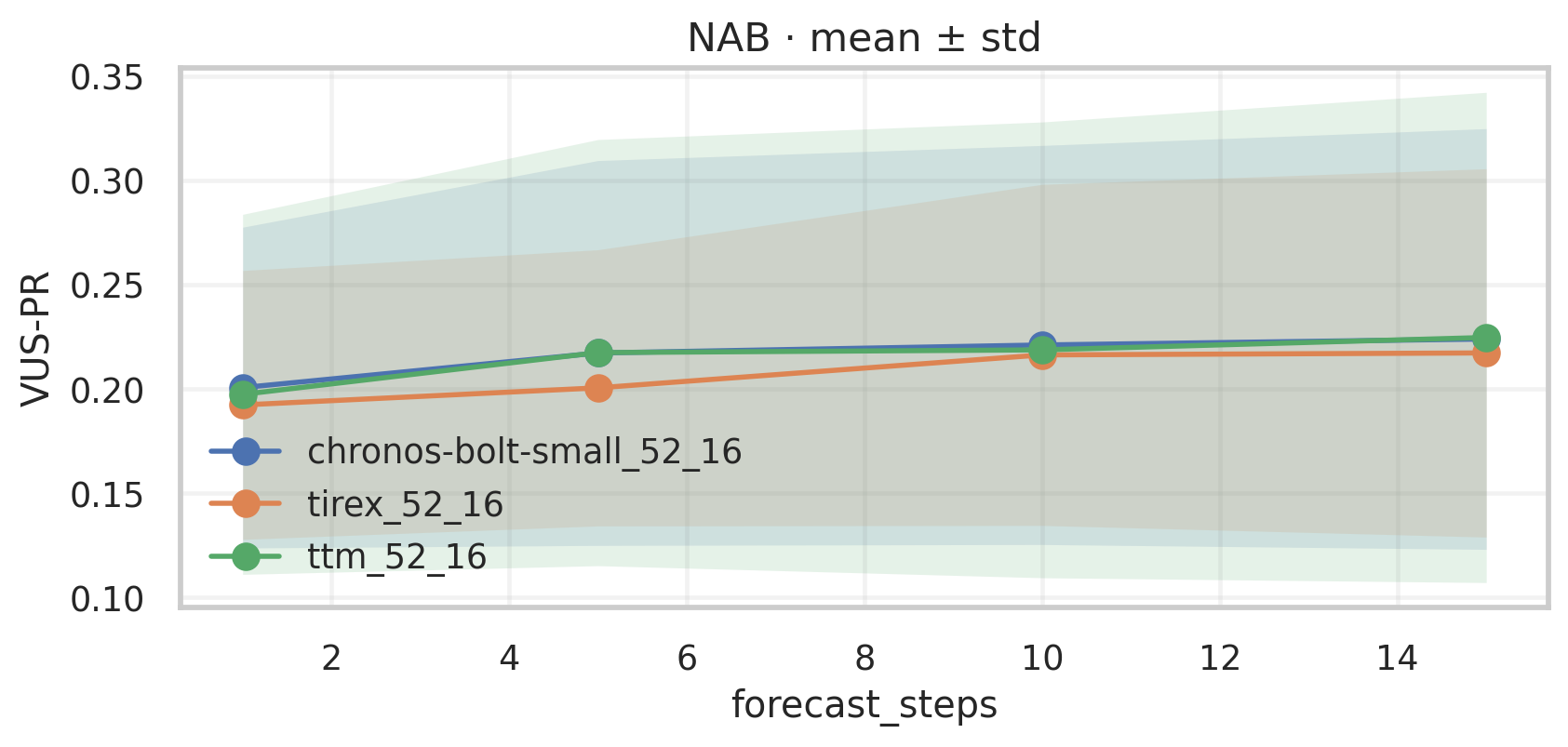}}

  \subfigure[NEK — PA-F1]{\includegraphics[width=0.24\textwidth]{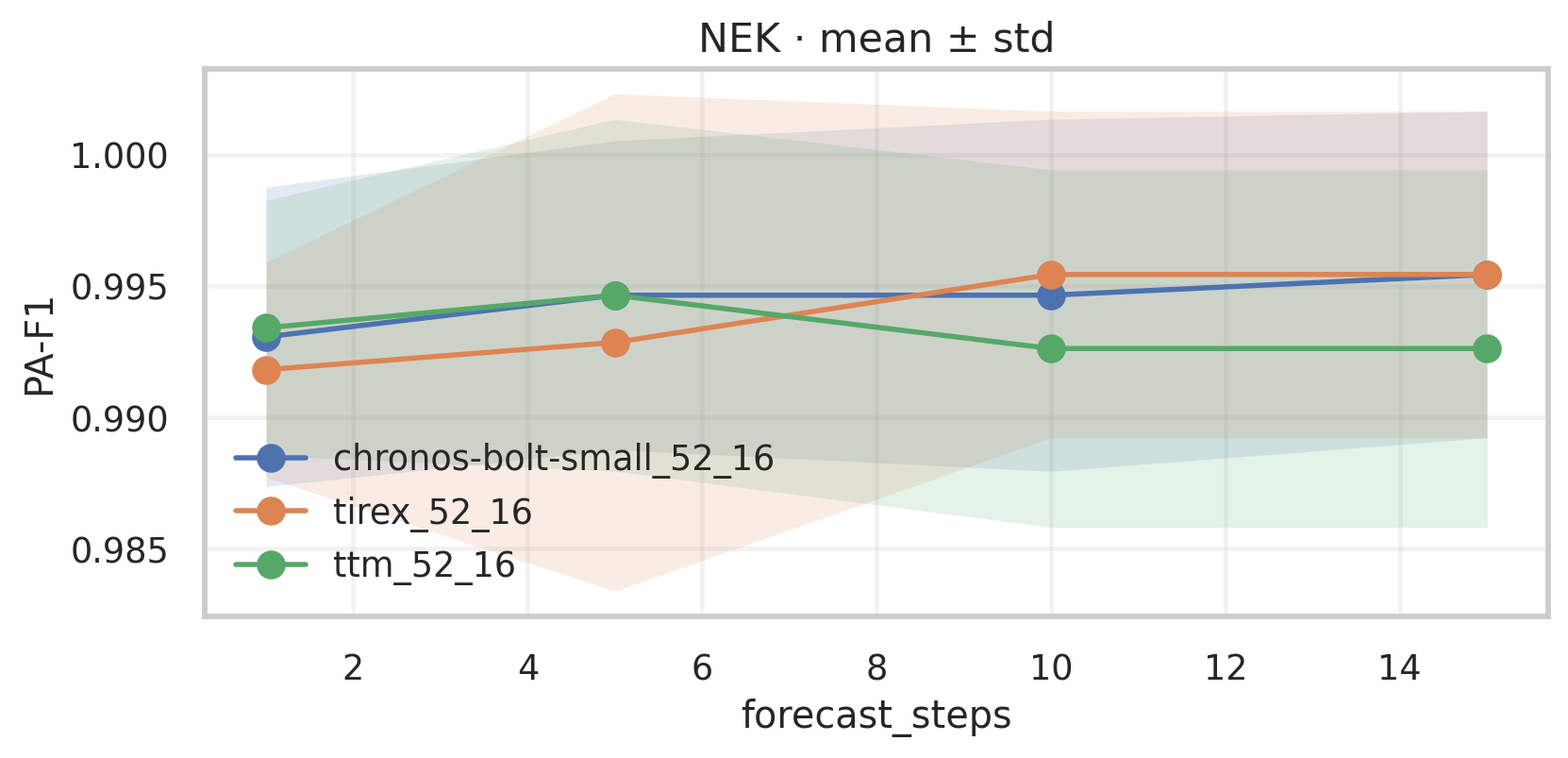}}
  \subfigure[NEK — Affiliation-F]{\includegraphics[width=0.24\textwidth]{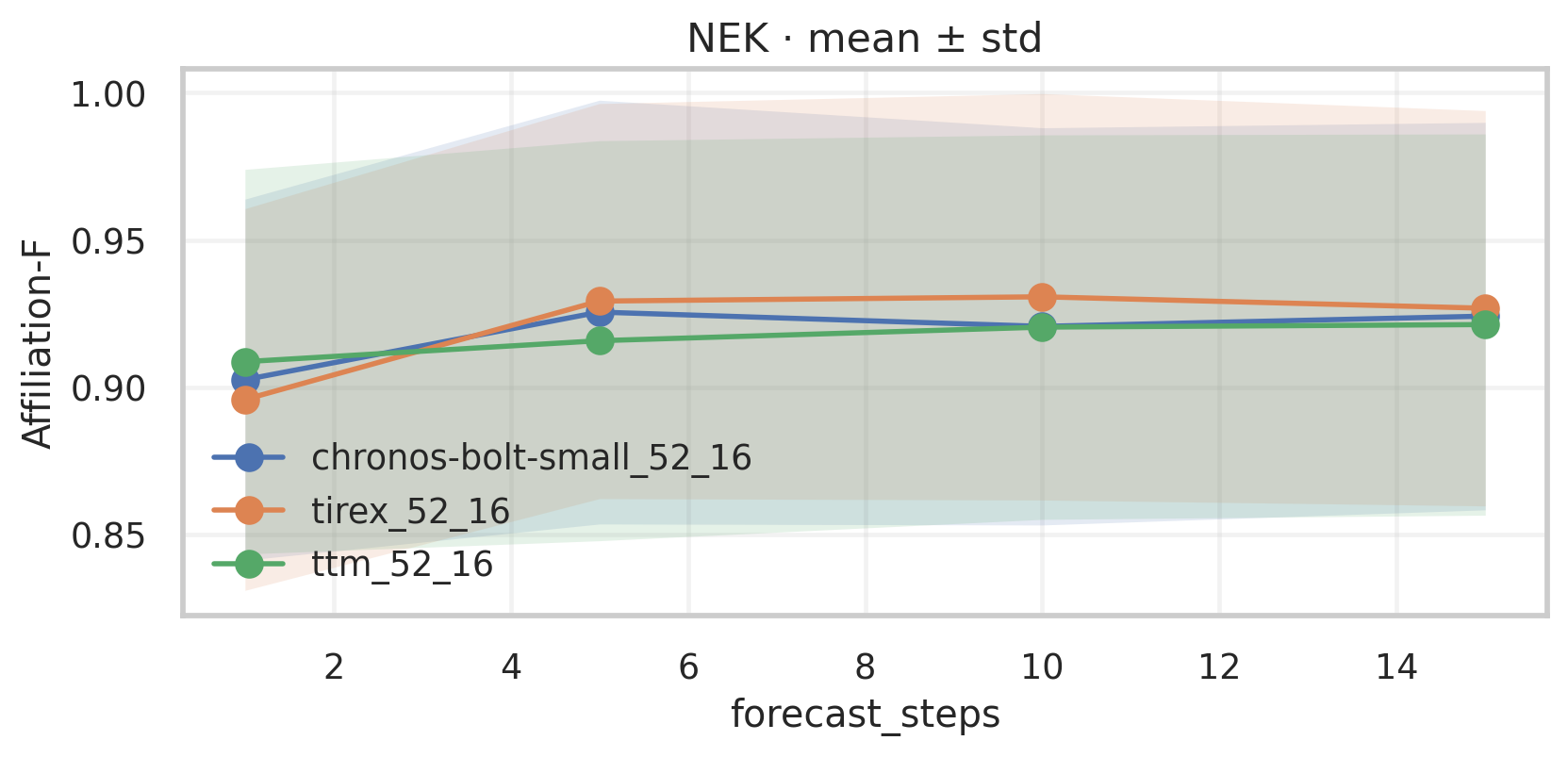}}
  \subfigure[NEK — AUC-PR]{\includegraphics[width=0.24\textwidth]{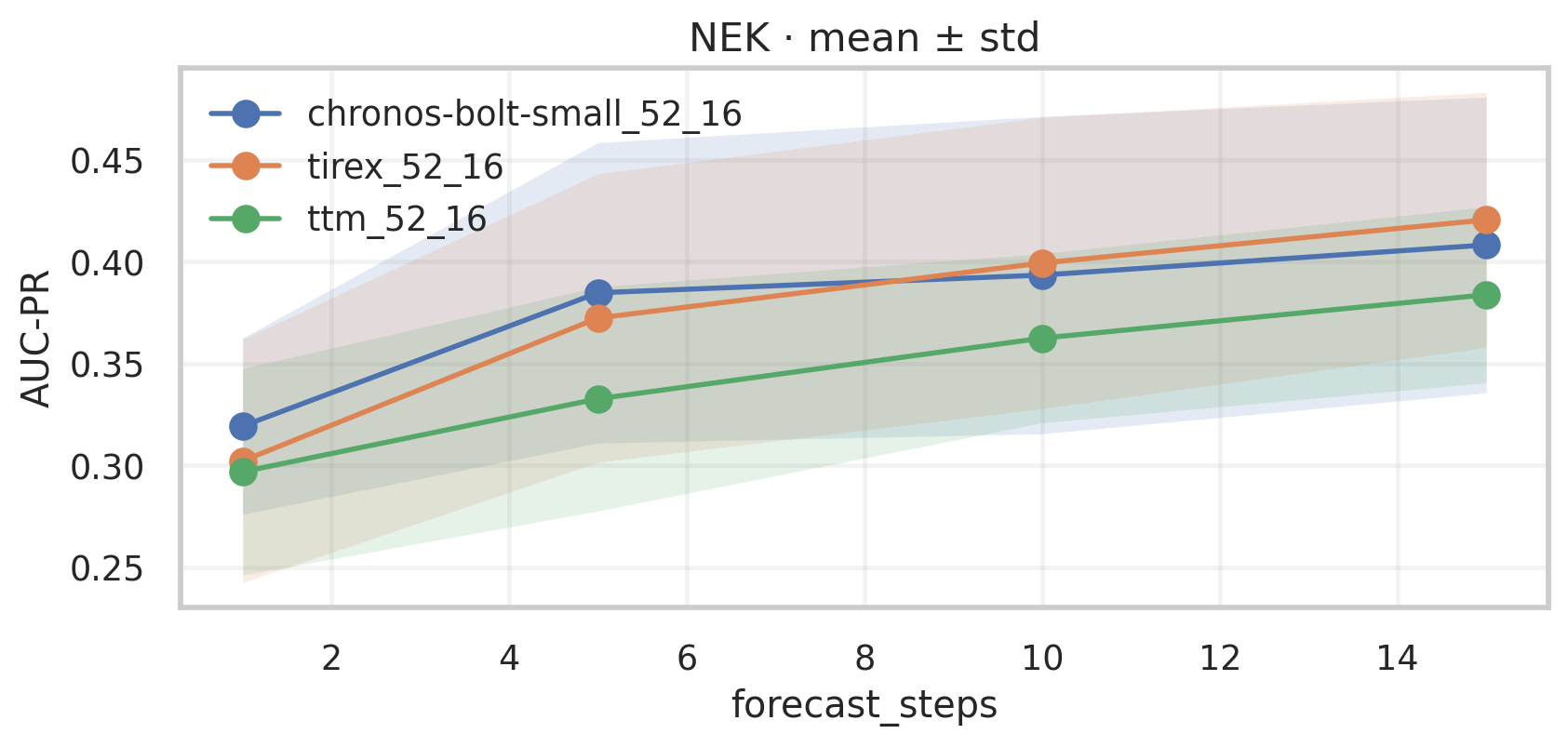}}
  \subfigure[NEK — VUS-PR]{\includegraphics[width=0.24\textwidth]{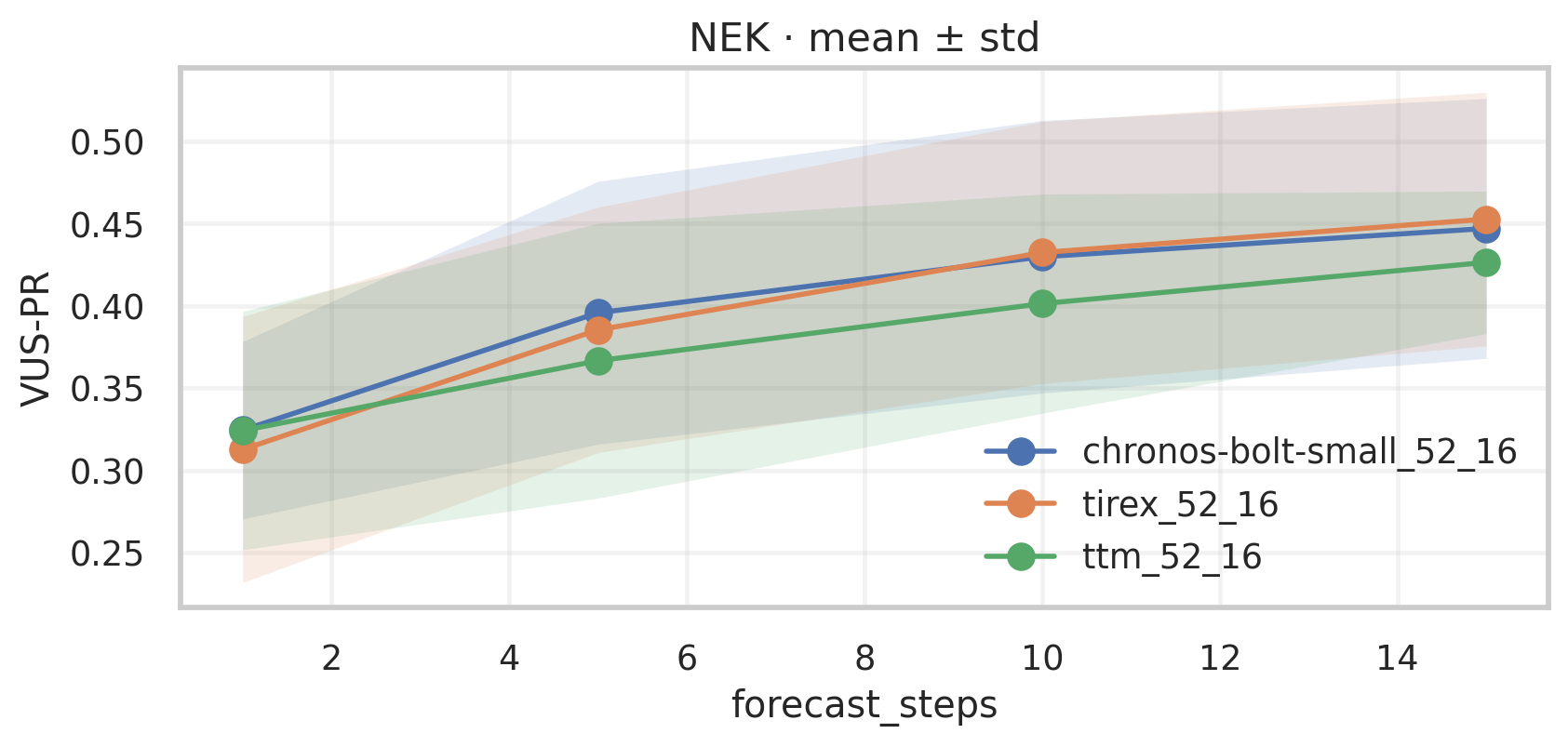}}

  \subfigure[MSL — PA-F1]{\includegraphics[width=0.24\textwidth]{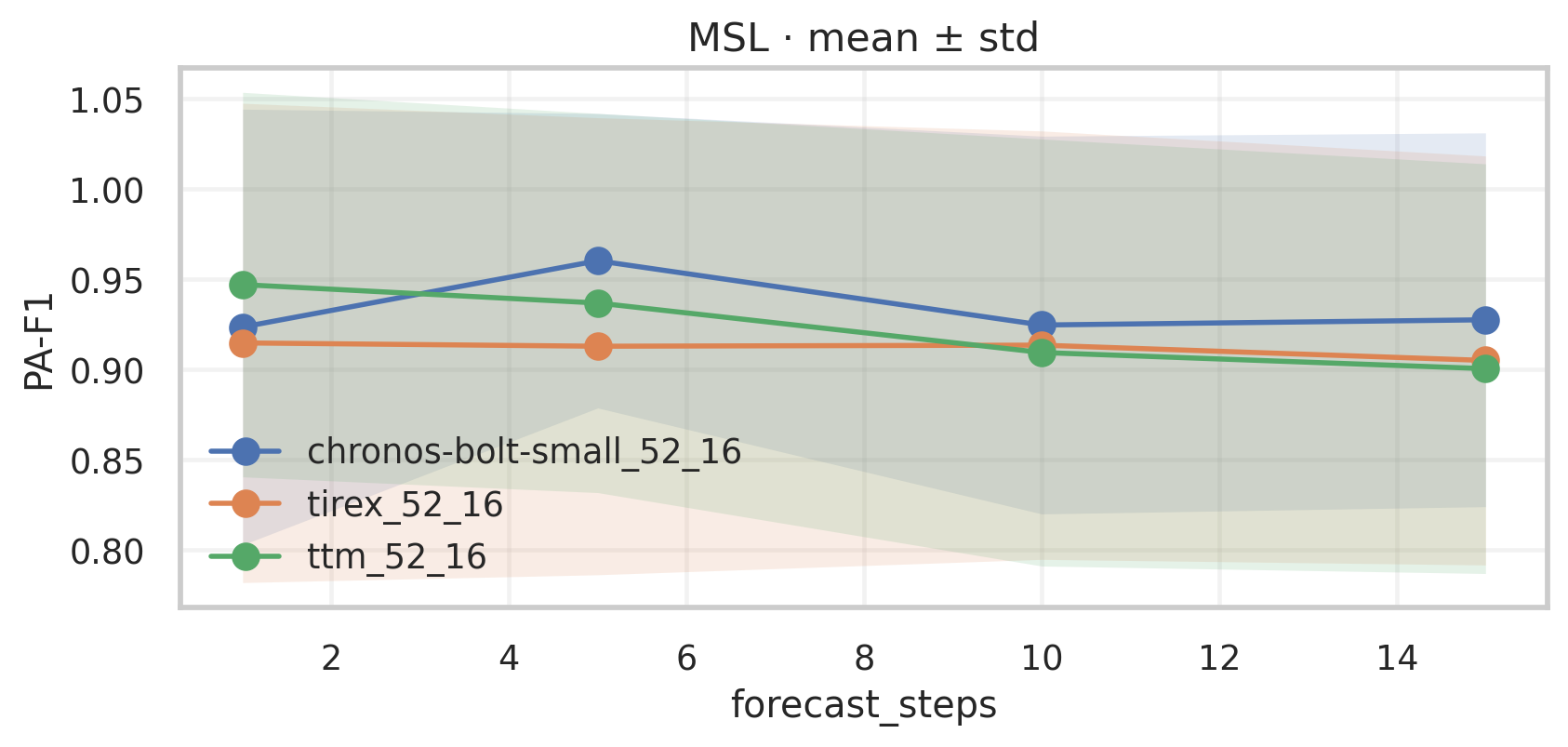}}
  \subfigure[MSL — Affiliation-F]{\includegraphics[width=0.24\textwidth]{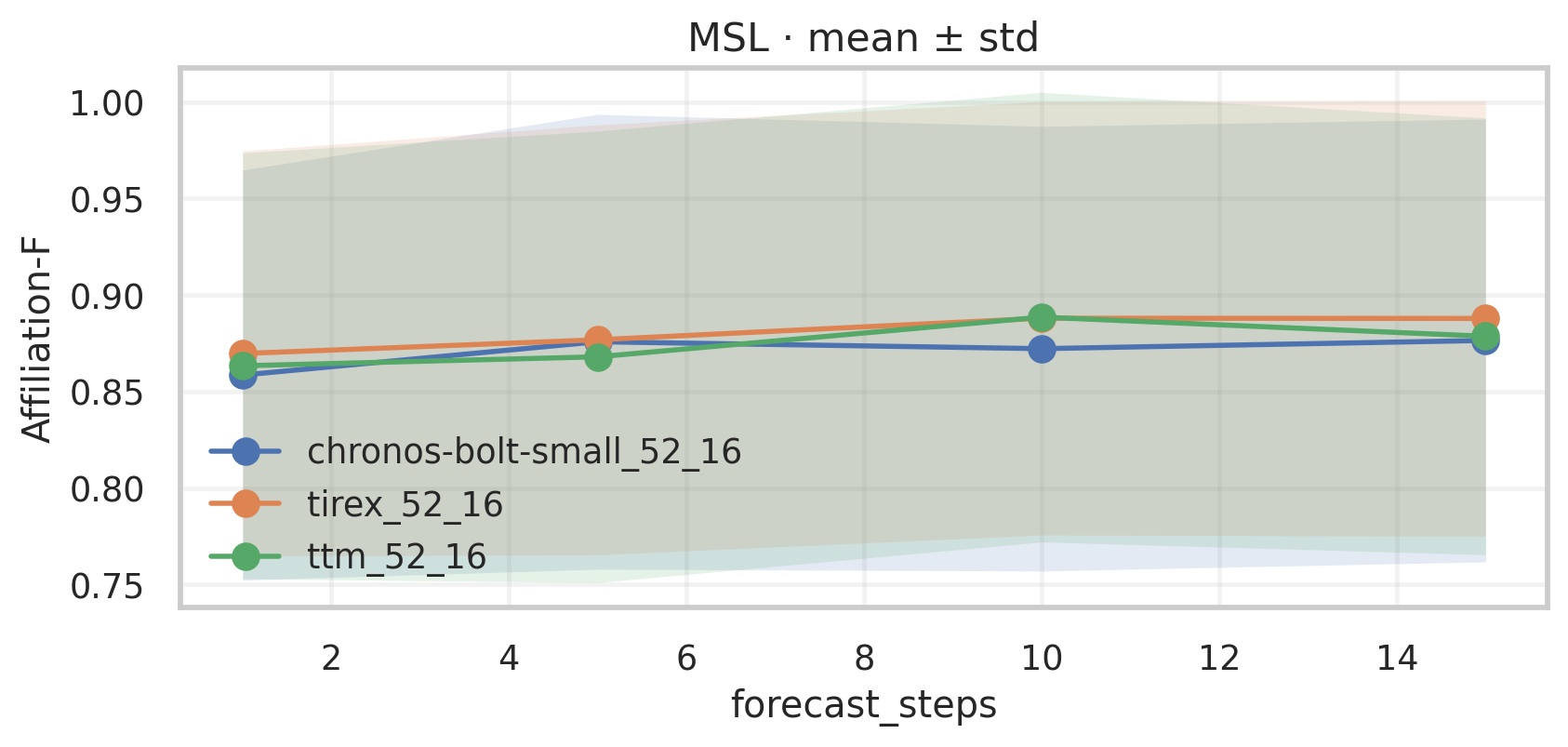}}
  \subfigure[MSL — AUC-PR]{\includegraphics[width=0.24\textwidth]{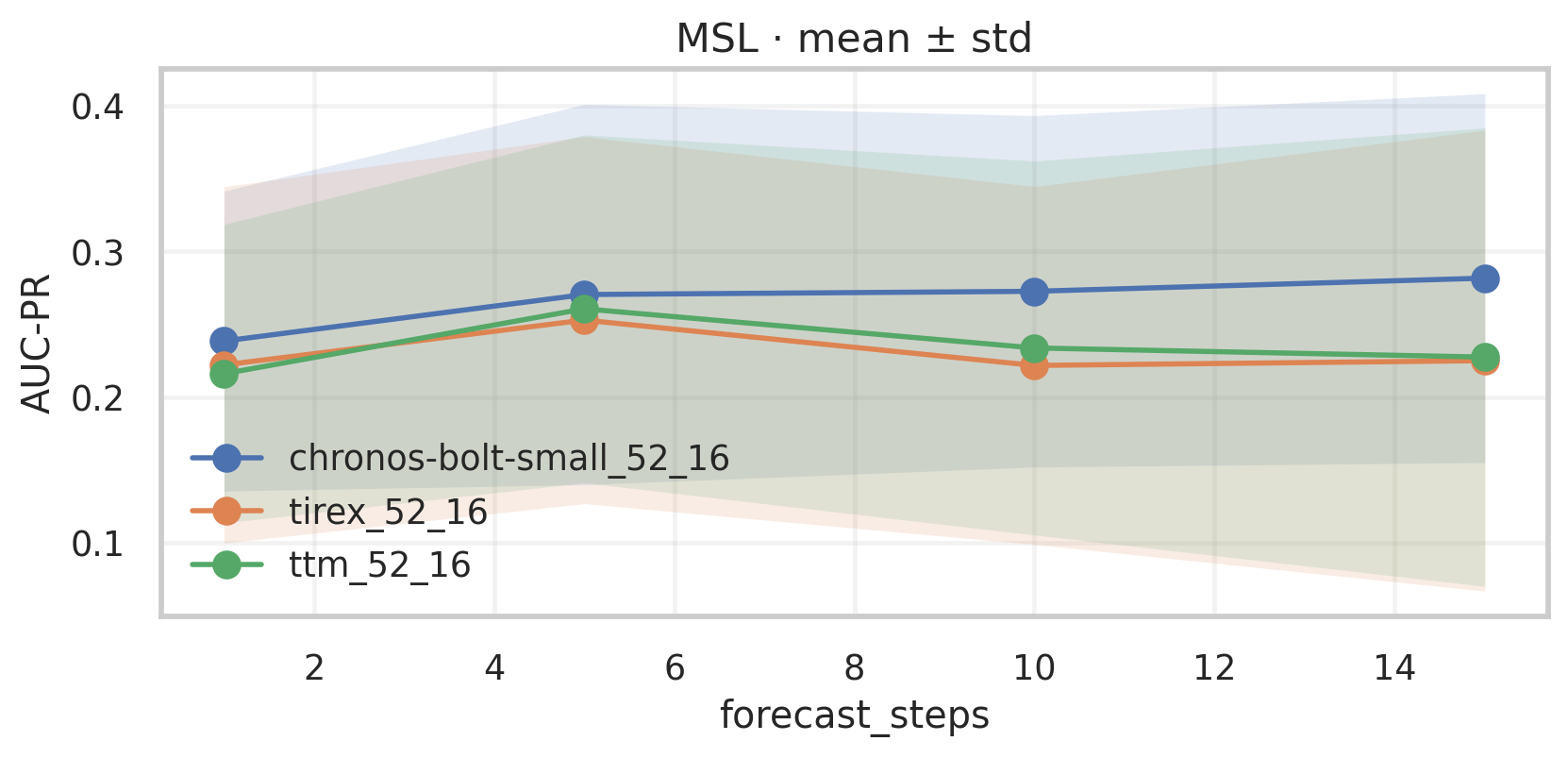}}
  \subfigure[MSL — VUS-PR]{\includegraphics[width=0.24\textwidth]{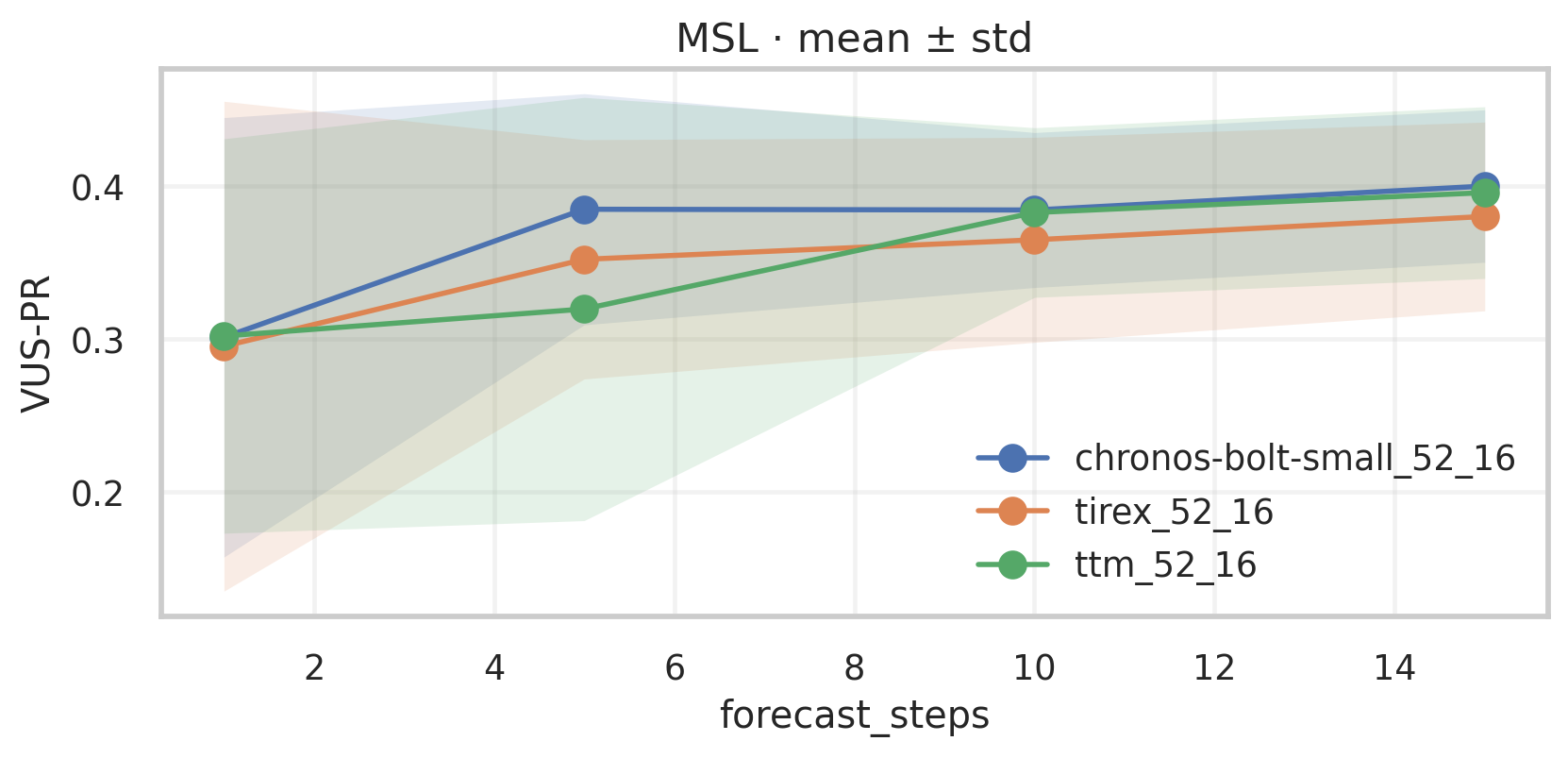}}

  \subfigure[YAHOO — PA-F1]{\includegraphics[width=0.24\textwidth]{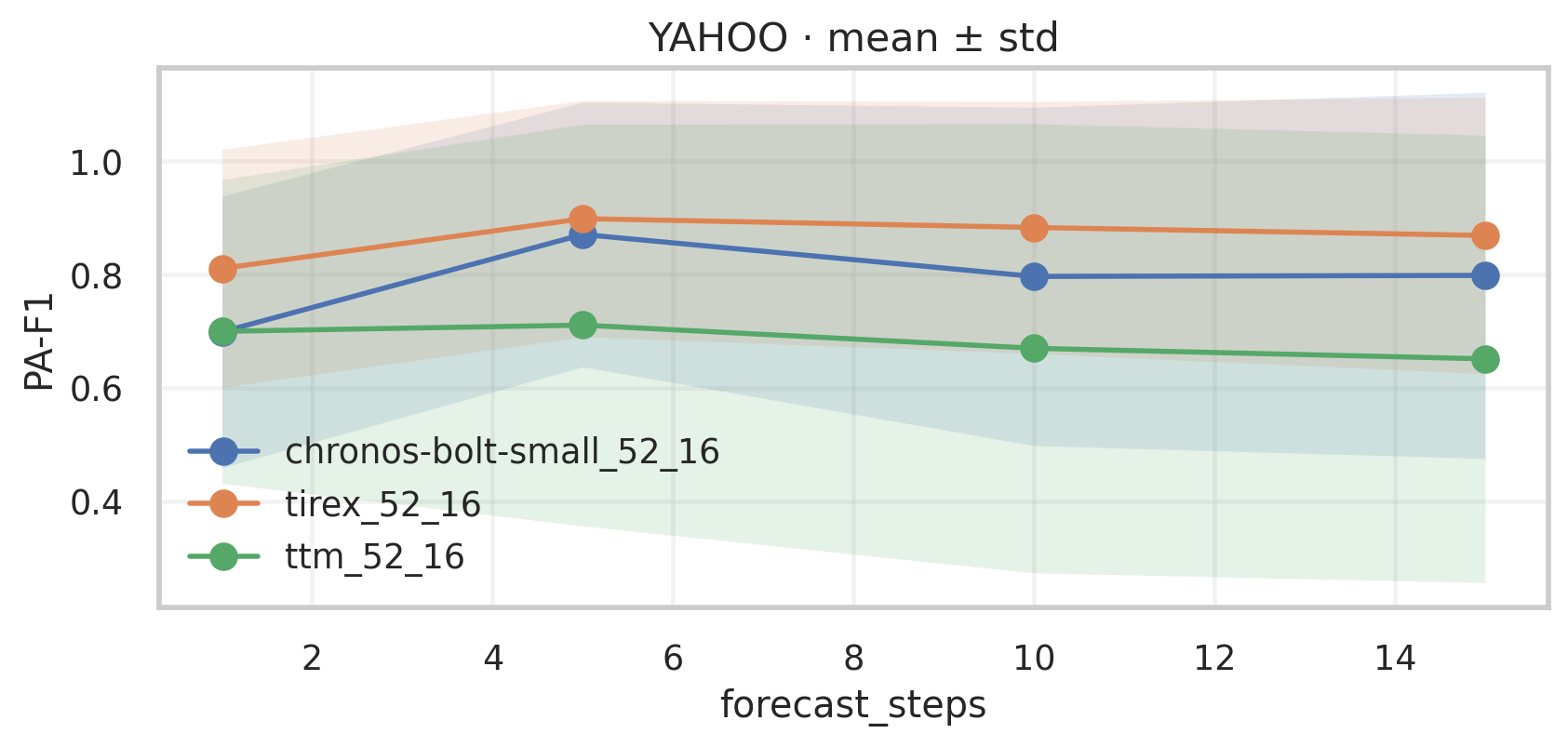}}
  \subfigure[YAHOO-Affiliation-F]{\includegraphics[width=0.24\textwidth]{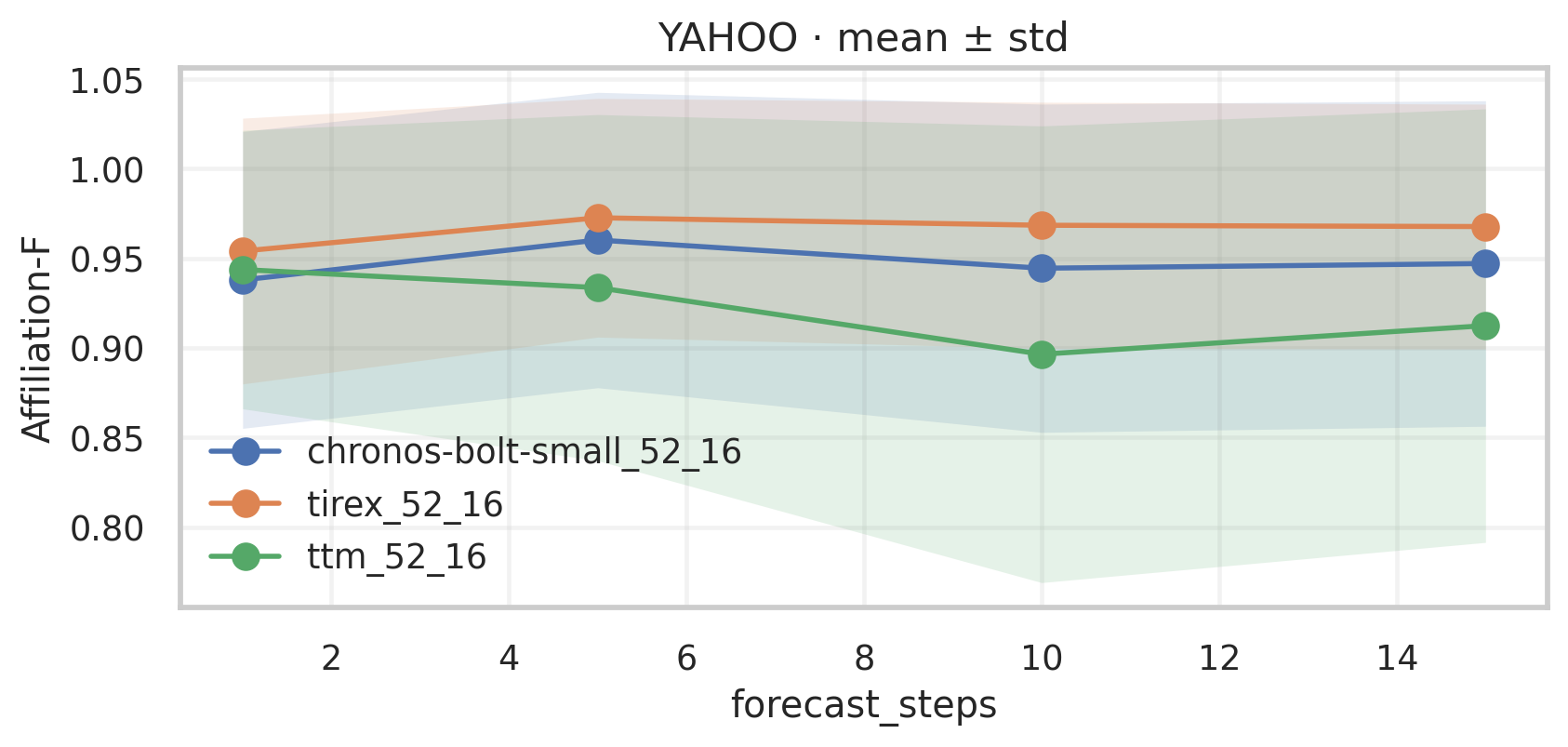}}
  \subfigure[YAHOO — AUC-PR]{\includegraphics[width=0.24\textwidth]{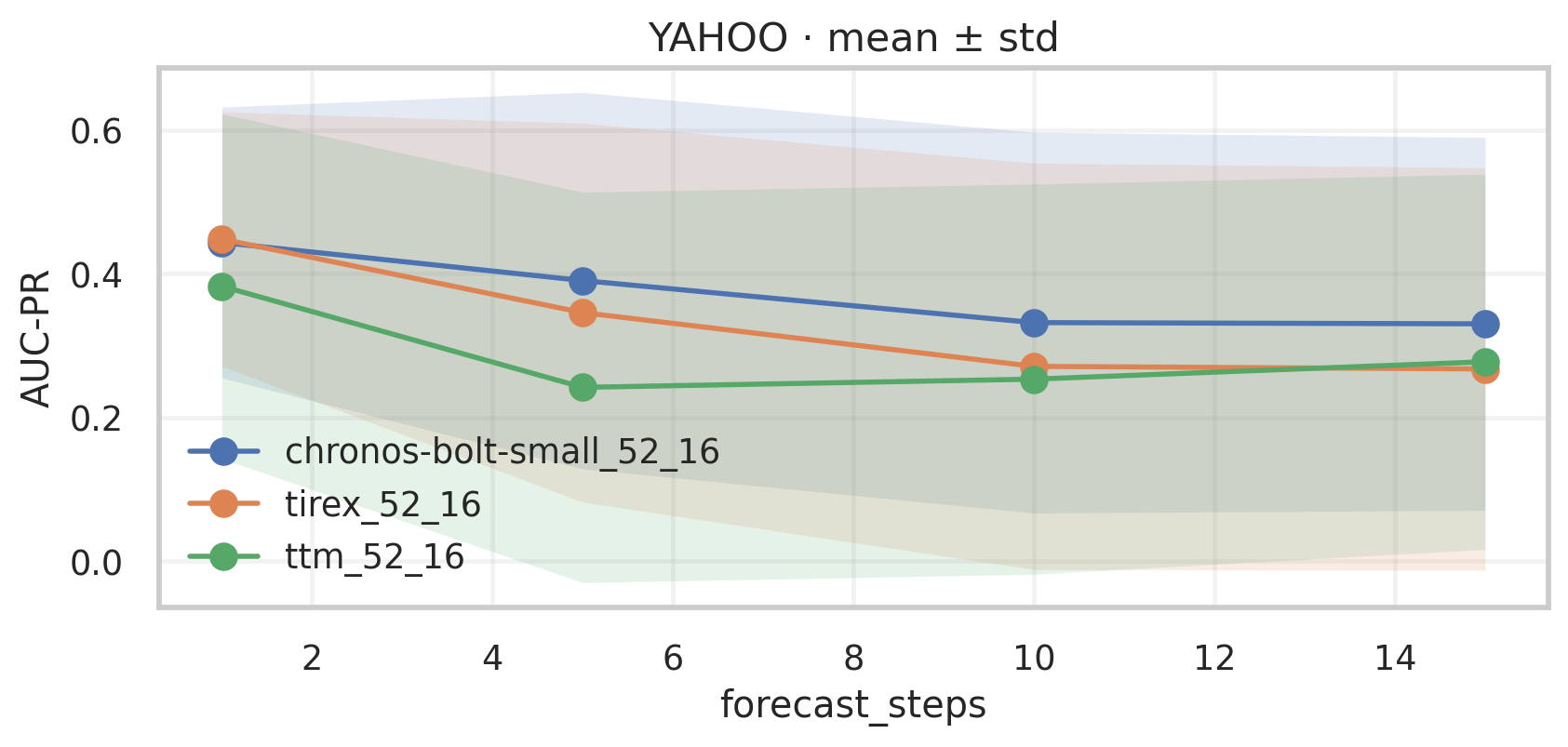}}
  \subfigure[YAHOO — VUS-PR]{\includegraphics[width=0.24\textwidth]{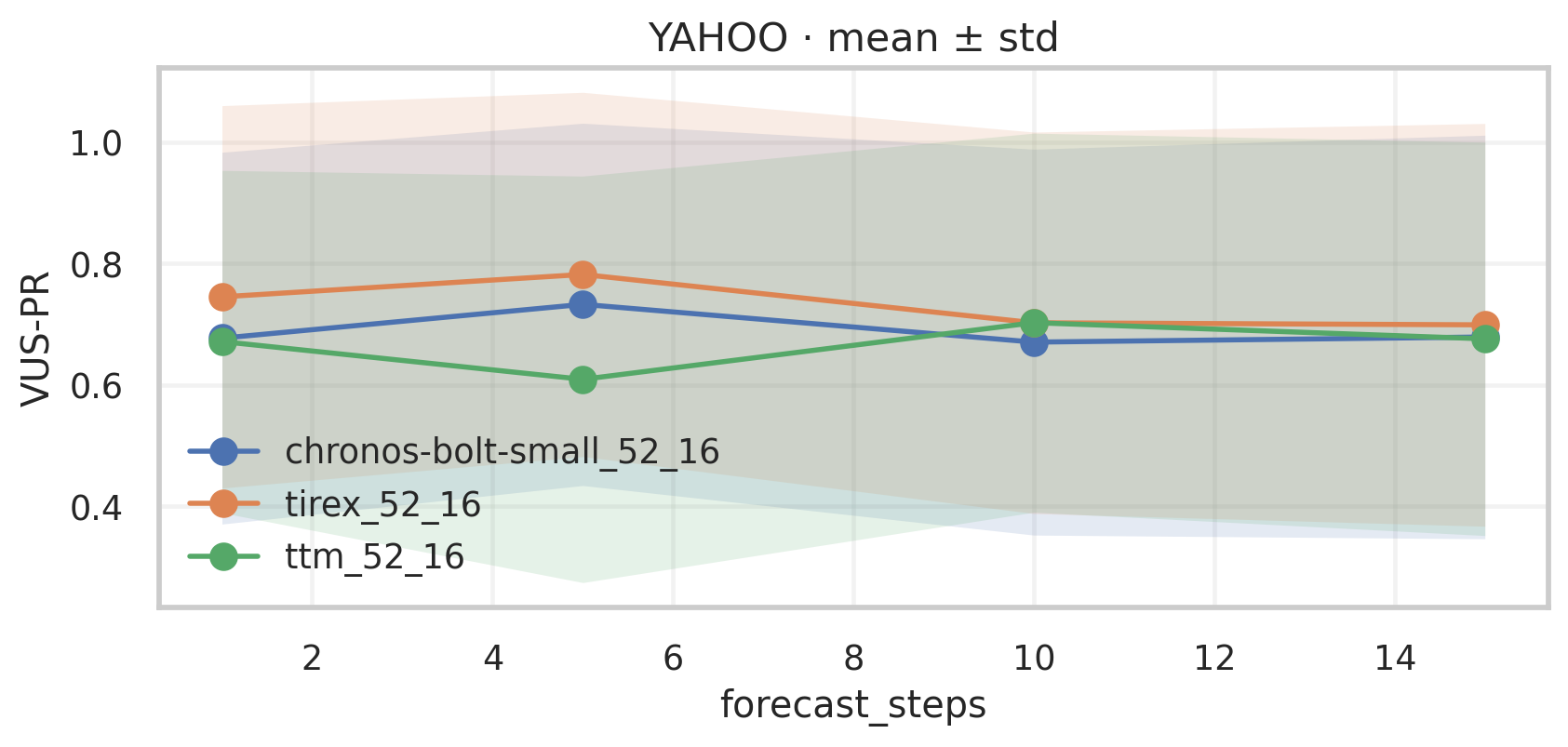}}

  \subfigure[Stock — PA-F1]{\includegraphics[width=0.24\textwidth]{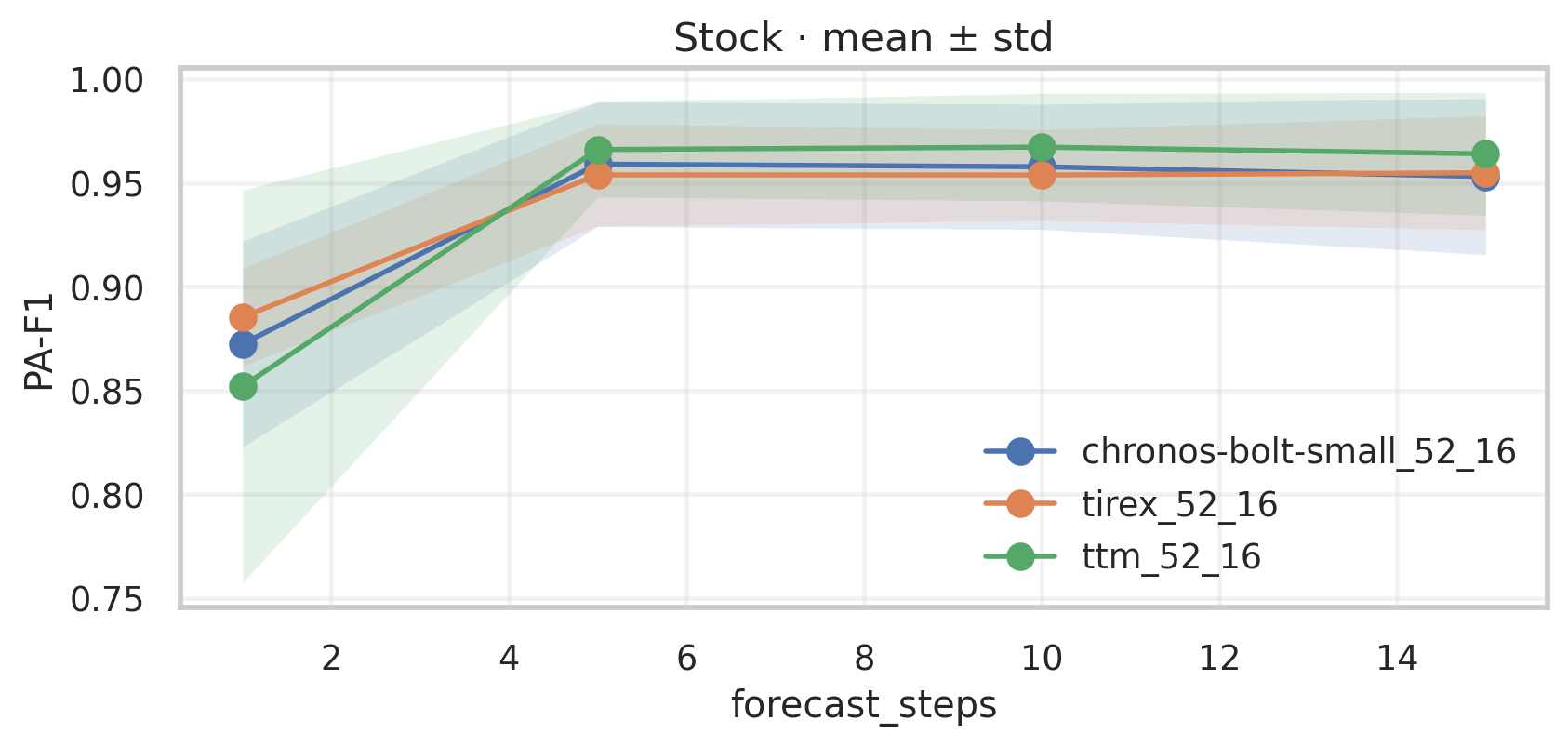}}
  \subfigure[Stock — Affiliation-F]{\includegraphics[width=0.24\textwidth]{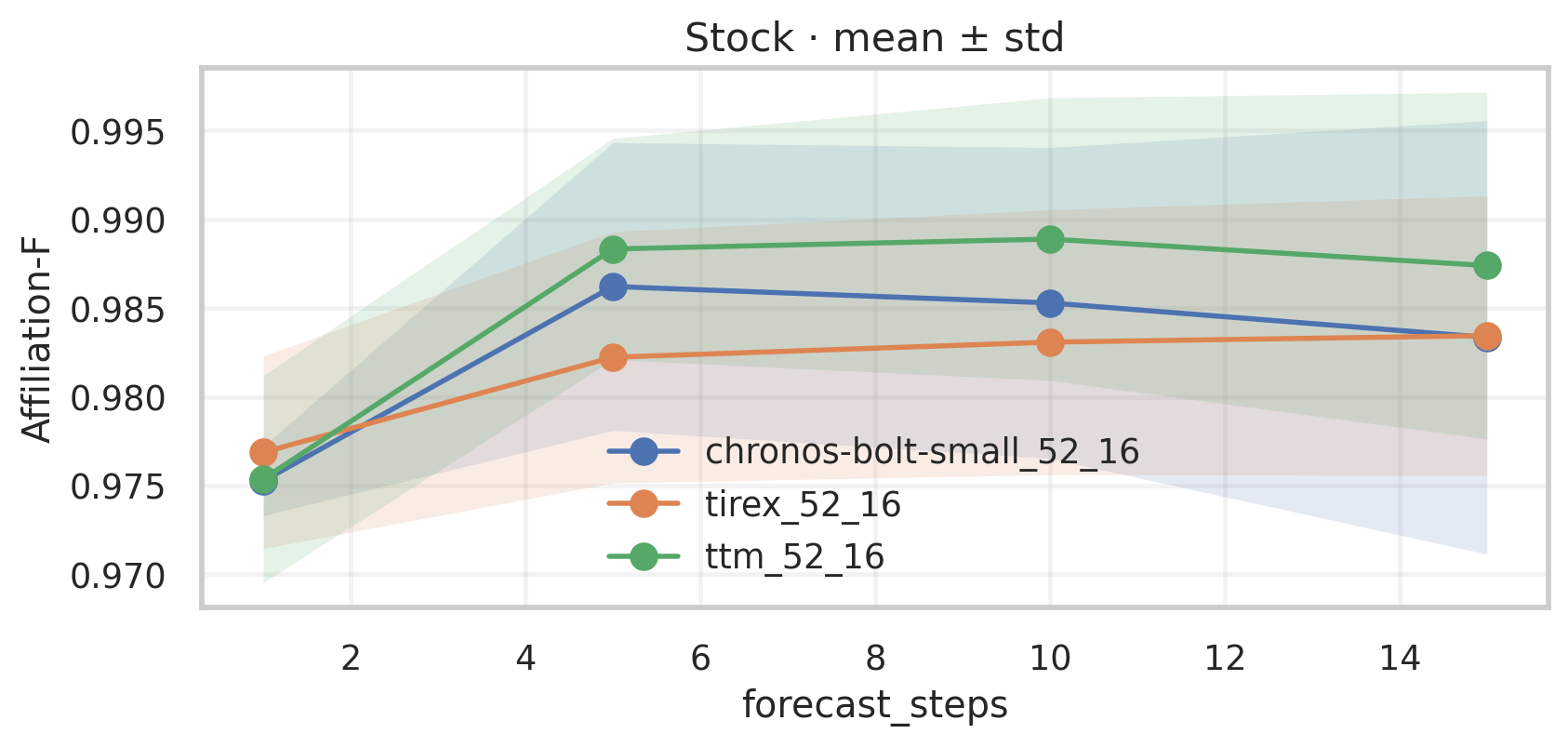}}
  \subfigure[Stock — AUC-PR]{\includegraphics[width=0.24\textwidth]{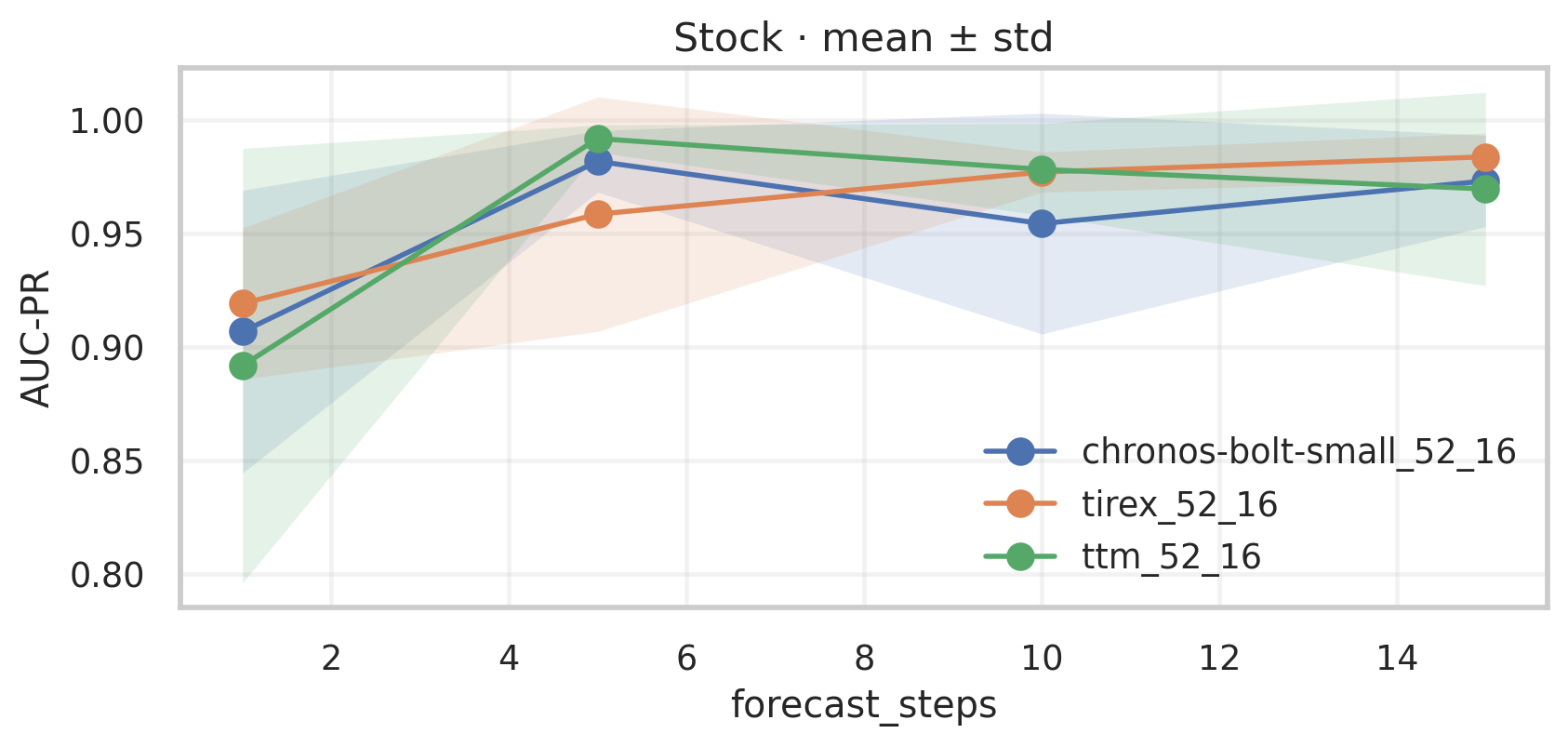}}
  \subfigure[Stock — VUS-PR]{\includegraphics[width=0.24\textwidth]{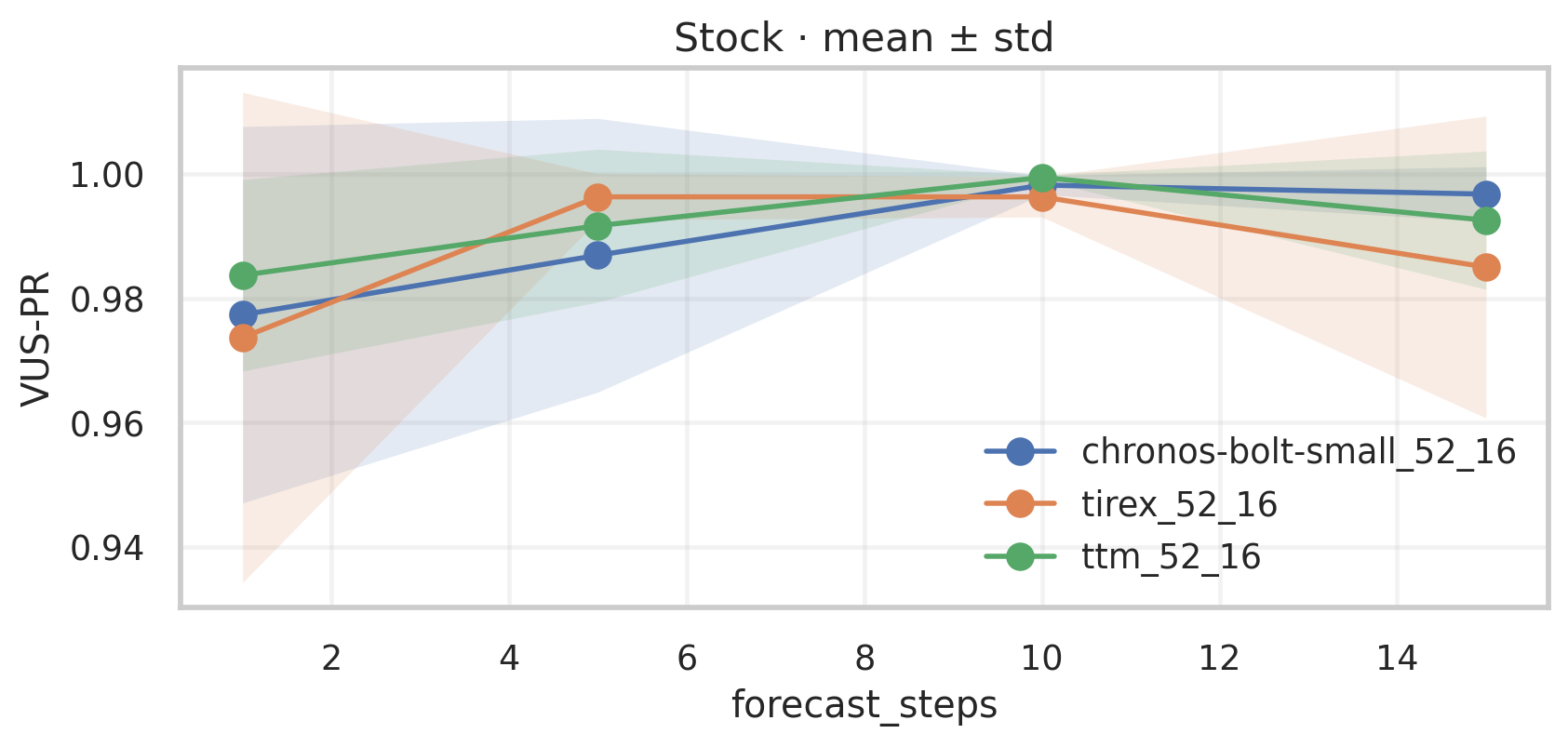}}

  \subfigure[WSD — PA-F1]{\includegraphics[width=0.24\textwidth]{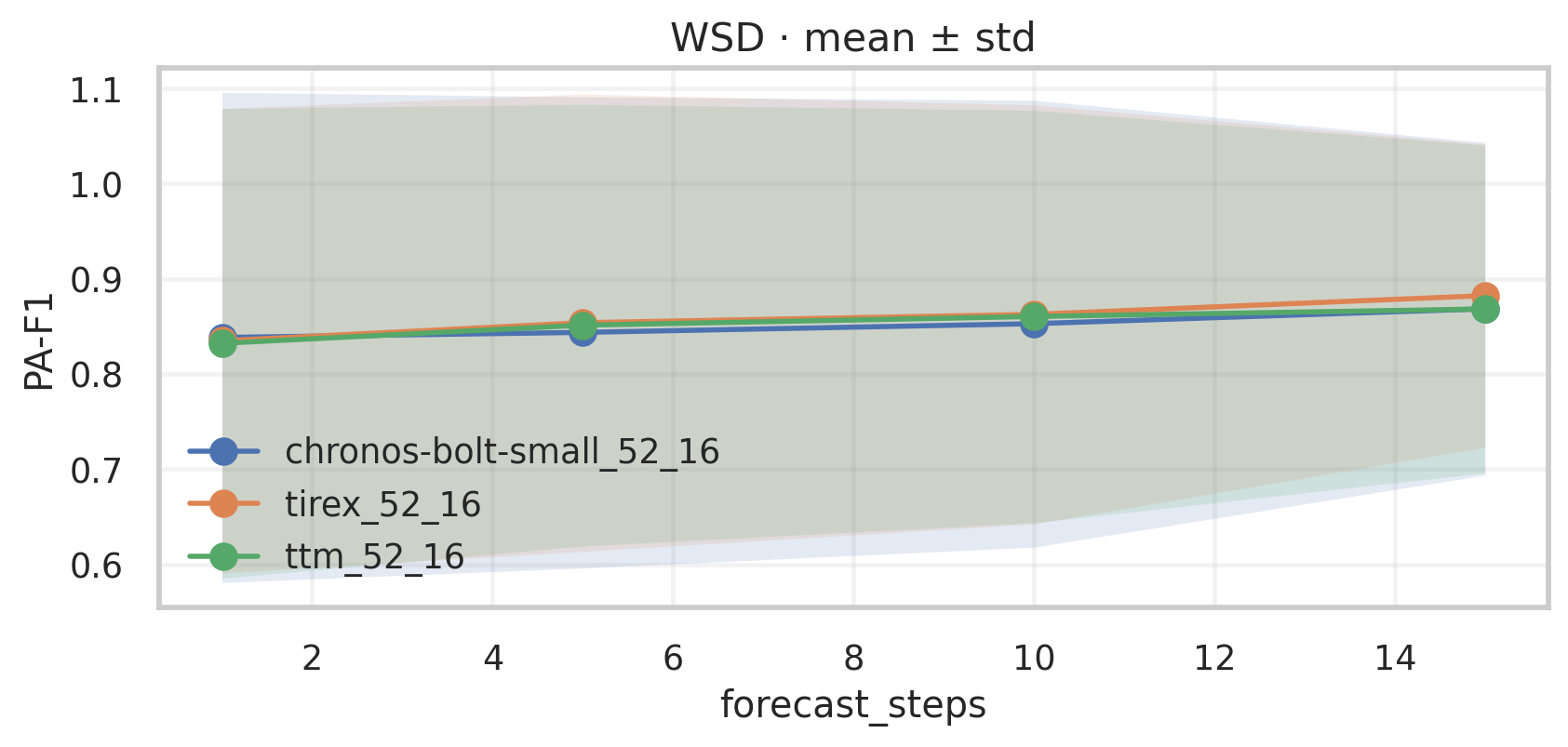}}
  \subfigure[WSD — Affiliation-F]{\includegraphics[width=0.24\textwidth]{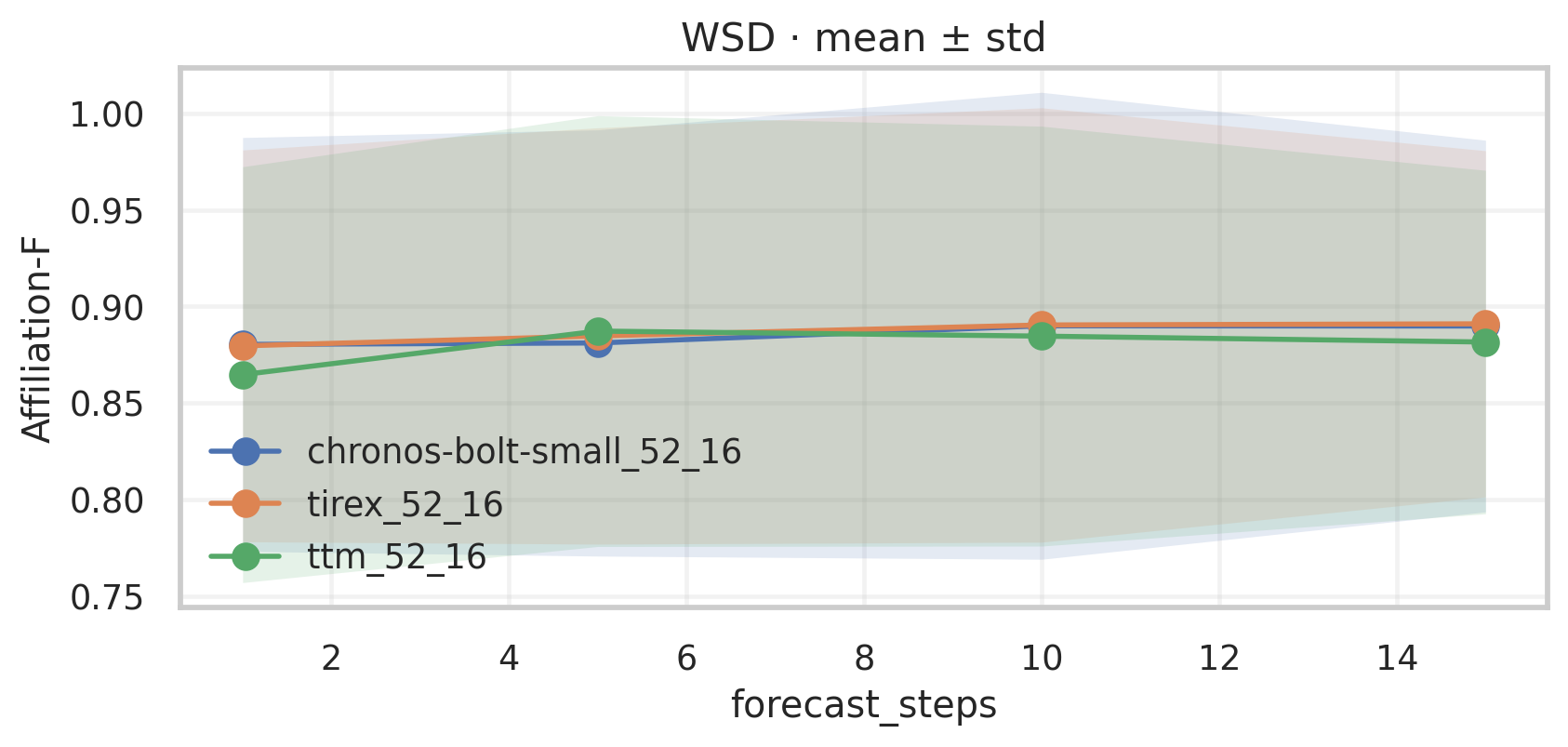}}
  \subfigure[WSD — AUC-PR]{\includegraphics[width=0.24\textwidth]{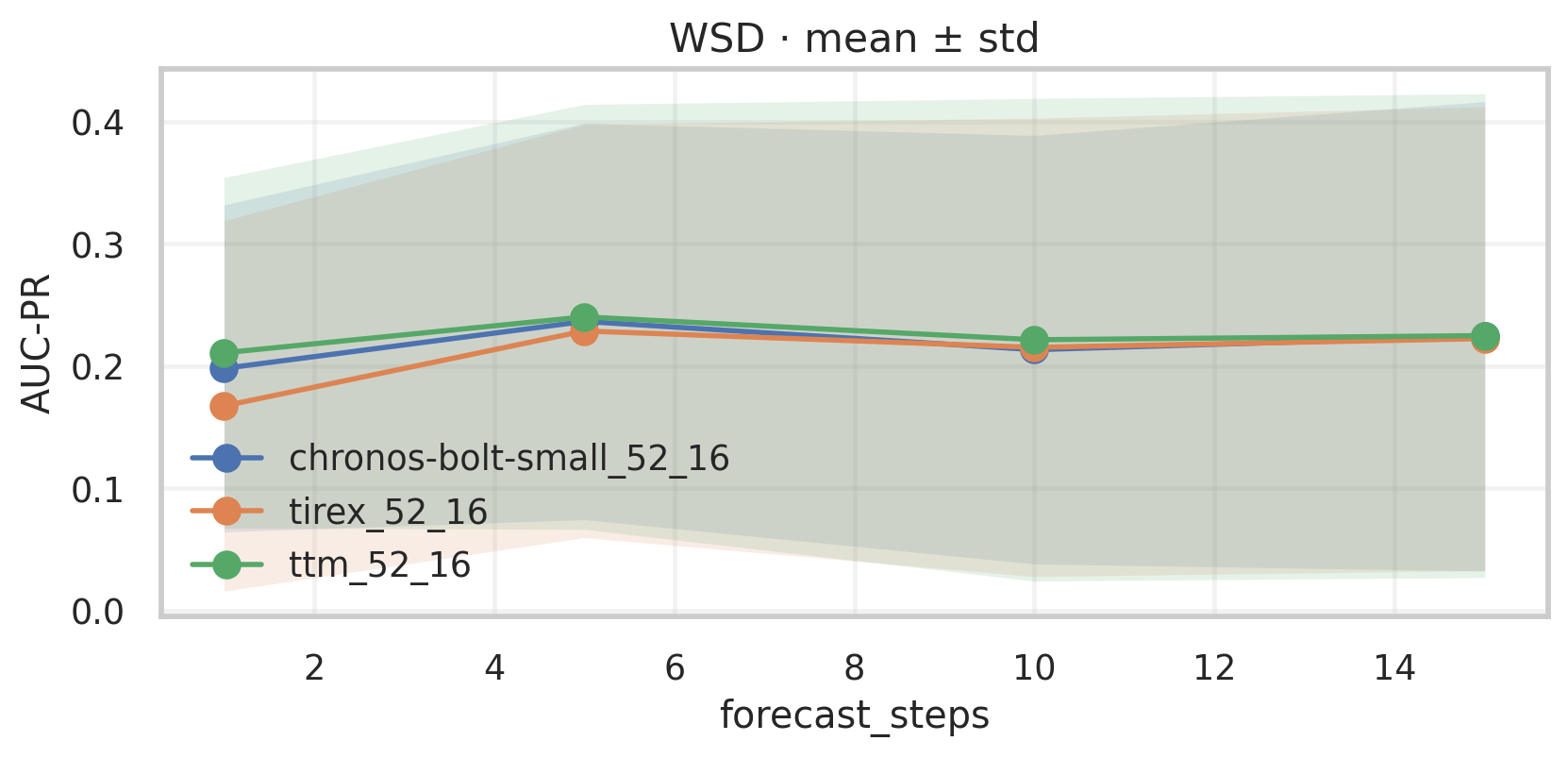}}
  \subfigure[WSD — VUS-PR]{\includegraphics[width=0.24\textwidth]{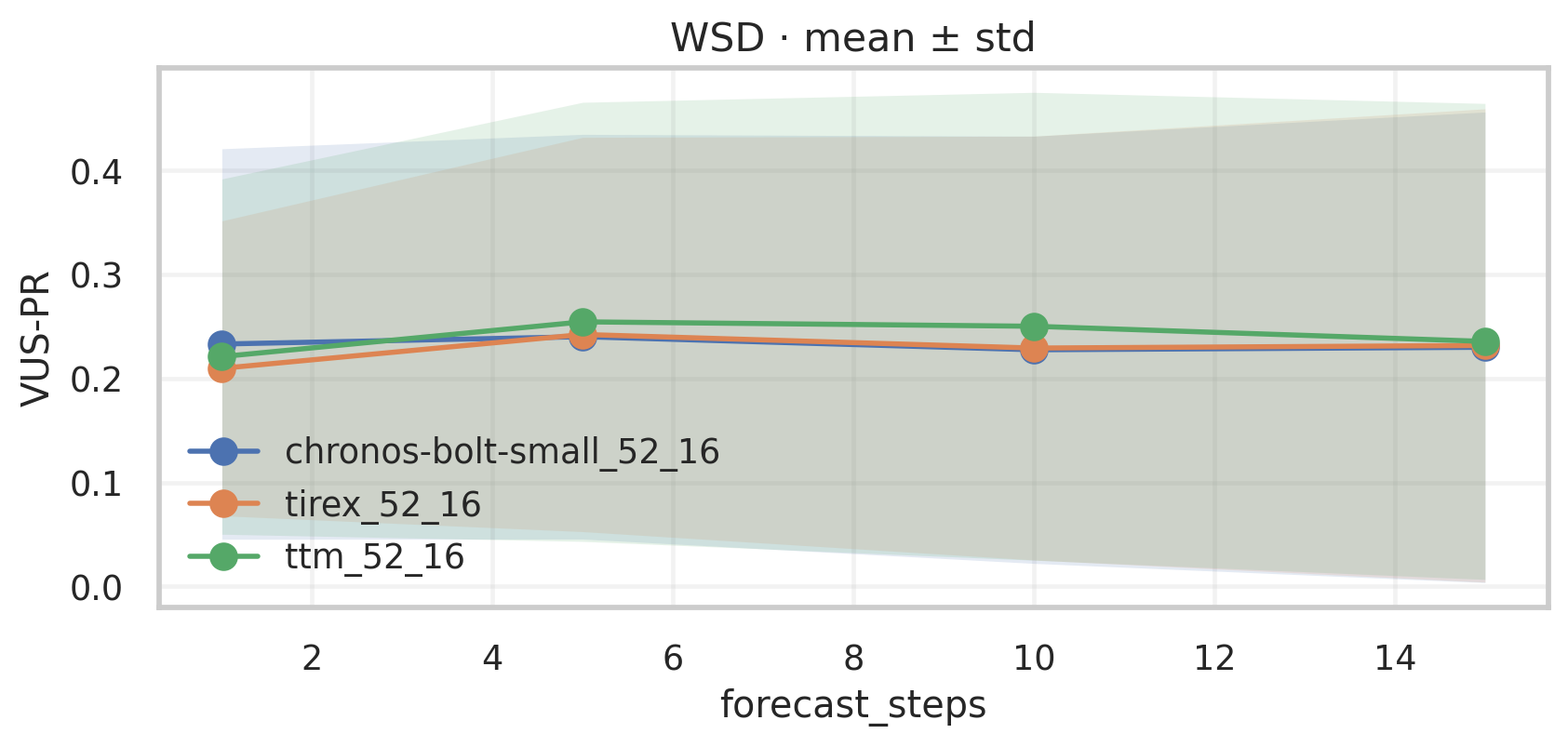}}

  \caption{Performance of $\mathcal{W}_1$-ACAS when aggregating different forecast steps. Rows correspond to datasets (NAB, NEK, MSL, YAHOO, Stock, WSD) and columns to metrics (PA-F1, Affiliation-F, AUC-PR, VUS-PR).}
  \label{fig:w1-acas-steps-grid}
\end{figure*}

\input{ICLR2026/sections/appendix_figures_ablation}
\input{ICLR2026/sections/appendix_tables}

\subsubsection{Additional Results}
\label{sec:appendix_additional_results}
Figure~\ref{fig:detection_examples_with_weights} provides additional detection examples, while Figure~\ref{fig:pa-aff-tradeoff} illustrates the trade-offs between FPR and F1 scores (PA-F1 and Affiliation-F) at the operating points that maximize the respective F1 metric, as defined in Appendix~\ref{sec:appendix_metrics}. For PA-F1, $\mathcal{W}_1$\textsc{-ACAS} consistently dominates competing approaches. For Affiliation-F, $\mathcal{W}_1$\textsc{-ACAS} yields operating points that are rarely dominated and is the top-performing method in several datasets.  

{
\paragraph{Hyperparameter Sensitivity.} Figure~\ref{fig:w1-acas-steps-grid} examines the effect of aggregating different numbers of forecast horizons. Performance generally stabilizes once more than 10 horizons are included, with limited gains beyond this point. 
Figure~\ref{fig:w1-acas-steps-grid-lr} shows the sensitivity of $\mathcal{W}_1$\textsc{-ACAS} to the learning rate $\gamma$ (with $\alpha_c = 0.01$ and $n_b = 10$). Since the weights are updated using ADAM, $\gamma$ must remain sufficiently small; empirically, the method exhibits no significant variability for small learning rates. Figure~\ref{fig:w1-acas-steps-grid-nbu} illustrates the effect of the batch size~$n_b$ (with $\gamma = 0.001$ and $\alpha_c = 0.01$). This parameter controls the number of samples used in the Wasserstein distance computation: if the distribution of nonconformity scores changes over time, $n_b$ should not be too large. In practice, the method is only mildly sensitive to~$n_b$, with smaller values performing slightly better on some datasets. Finally, Figure~\ref{fig:w1-acas-steps-grid-alarm} examines the sensitivity to the critical alarm rate~$\alpha_c$. This parameter determines the maximum acceptable $p$-value resolution: smaller values require a larger number of in-distribution past observations~$n_c$ for stable quantile estimation, but do not impose a lower bound on the smallest detectable anomaly level.}

\paragraph{Per-dataset performance}Tables~\ref{tab:placeholder},~\ref{tab:placeholder_ext}, and~\ref{tab:placeholder-v2} report per-dataset metrics, which align with and reinforce the trends discussed in the main Experimental section. {Table~\ref{tab:uv_tsfm_forecasting_error} summarizes the forecasting performance of the TSFM models across datasets. Overall, the models exhibit broadly similar MAE/RMSE values, which aligns with their comparable anomaly-detection performance once forecast errors are properly calibrated online using $\mathcal{W}_1$\textsc{-ACAS}. Notably, the slightly higher forecasting error of TTM on YAHOO corresponds to its lower anomaly-detection performance in Table~\ref{tab:placeholder}, suggesting a consistent relationship between forecast quality and downstream AD results.

\paragraph{Computation Time.} The average per-sample computation time of $\mathcal{W}_1$\textsc{-ACAS} with a 15-step forecast is 0.025 ± 0.012 seconds per sample per feature on a single V100 32 GB GPU. Note that this implementation updates weights for all 15 predictors serially, these updates are independent and can be parallelized to further reduce runtime.

}
\input{ICLR2026/sections/appendix_forecast_tables}

{
\subsubsection{Extension to Multivariate Time Series Anomaly Detection.}
\label{subsec:appendix_extension_mv}
\paragraph{$\mathcal{W}_1$\textsc{-ACAS} via $p$-value aggregation.}.
Lets consider a multivariate time series with features $f \in [n_f]$, we can run Algorithm~\ref{alg:acas} independently on each dimension to obtain per-feature $p$-values $\bar{\beta}^{f}{t+1}$ at time $t+1$ (as defined in Eq.~\ref{eq:beta_aggregation}). These are then combined into a single anomaly score using standard $p$-value combination methods \cite{heard2018choosing}:

\begin{itemize}
    \item Fisher’s Method~\citep{fisher1970statistical}: Combined p-value is $\rho_{t+1}=1-F^{-1}_{\chi^2_{2n_f}}(Z_{t+1})$ with $Z_{t+1} = -2 \sum_{f}\bar{\beta}^f_{t+1}$.
    \item {Harmonic Mean $p$-value (HMP)}~\citep{wilson2019harmonic}: Combined p-value is $\rho_{t+1} = \frac{n_f}{\sum_{f} 1/\bar{\beta}^{f}_{t+1}}$.  

\end{itemize}
We refer to these variants as $\mathcal{W}_1$\textsc{-ACAS}-F and $\mathcal{W}_1$\textsc{-ACAS}-H, respectively.

\paragraph{Experiments and Results.}
We adopt the curated subsets from the TSB-AD benchmark~\citep{liu2024elephant}: TAO~\citep{tao_noaa_dataset} (13 curated series, each with $\sim$10k samples and 3 features, containing both sequential and point anomalies), GECCO~\citep{rehbach2018gecco} (a single long sequence with 9 features and over 138k samples), Genesis~\citep{von2018anomaly} (1 sequence with 18 features and over 16k samples), and LTDB~\citep{goldberger2000physiobank} (5 curated sequences, each with 2 features and approximately 100k samples). 

We evaluate our multivariate extensions, $\mathcal{W}_1$\textsc{-ACAS}-F and $\mathcal{W}_1$\textsc{-ACAS}-H, combined with Chronos and TiRex forecasters that leverage all available historical context (up to their maximum context window, with a minimum of 52 past points). These are compared against strong semi-supervised deep anomaly detection baselines~\citep{liu2024elephant}: CNN~\citep{munir2018deepant}, OmniAnomaly~\citep{su2019robust}, and USAD~\citep{audibert2020usad}, which benefit from being trained directly on non-anomalous segments. As reported in Table~\ref{tab:mv_results_ext}, both $\mathcal{W}_1$\textsc{-ACAS}-F and $\mathcal{W}_1$\textsc{-ACAS}-H achieve the best or highly competitive performance across all multivariate datasets, demonstrating the effectiveness of our $p$-value aggregation extension in this setting.

\input{ICLR2026/sections/appendix_mv_tables}
}

\section{The Use of Large Language Models (LLMs)}
\label{sec:appendix_use_llm}
We used large language models (LLMs) to assist with improving the readability and clarity of the manuscript. LLMs were used to improve and summarize the language in certain paragraphs, and to refine code for generating plots.

%% file: ICLR2026/sections/appendix_figures_ablation.tex
\begin{figure*}[t]
  \centering

  \subfigure[NAB — PA-F1]{\includegraphics[width=0.24\textwidth]{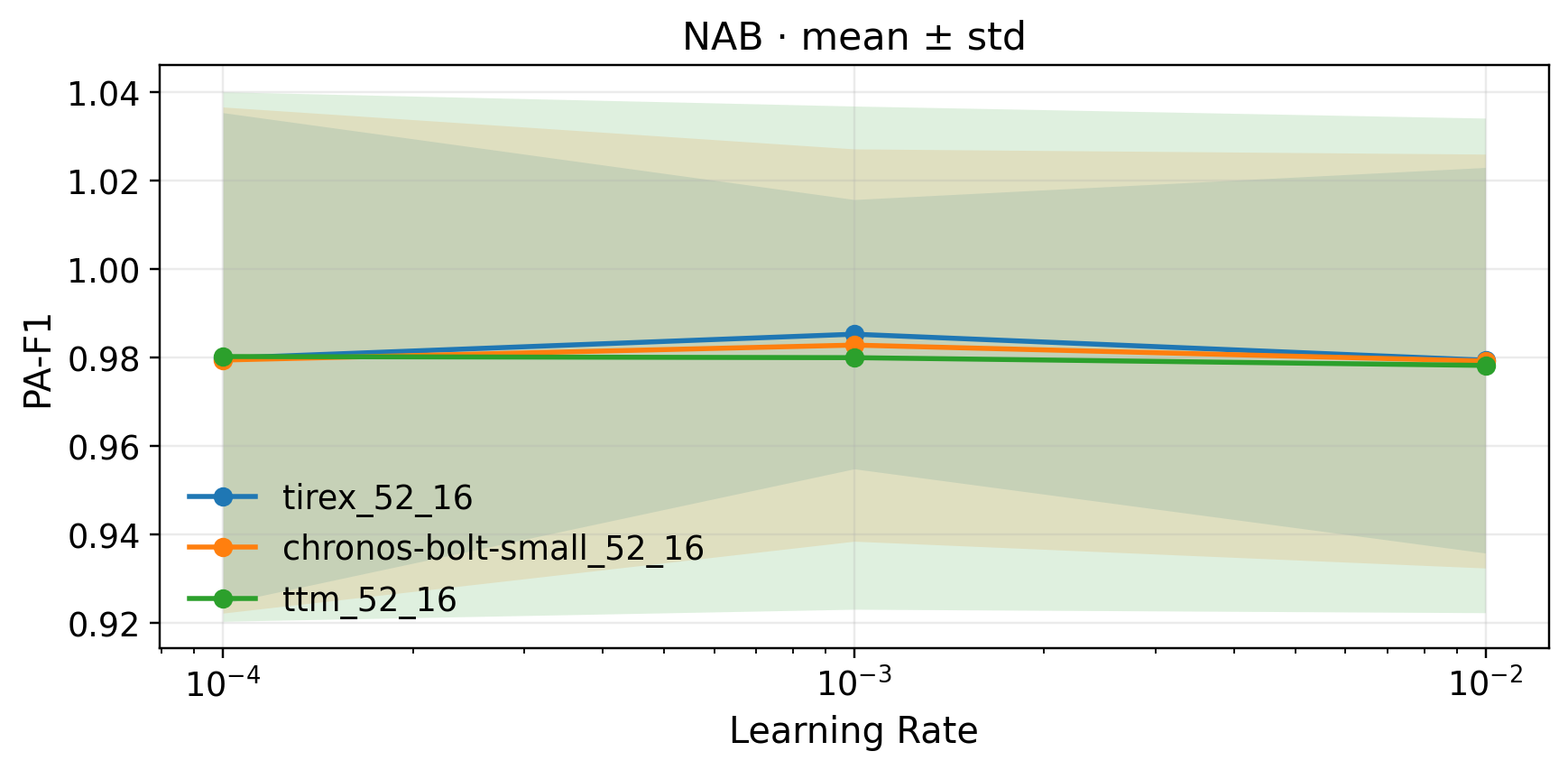}}
  \subfigure[NAB — Affiliation-F]{\includegraphics[width=0.24\textwidth]{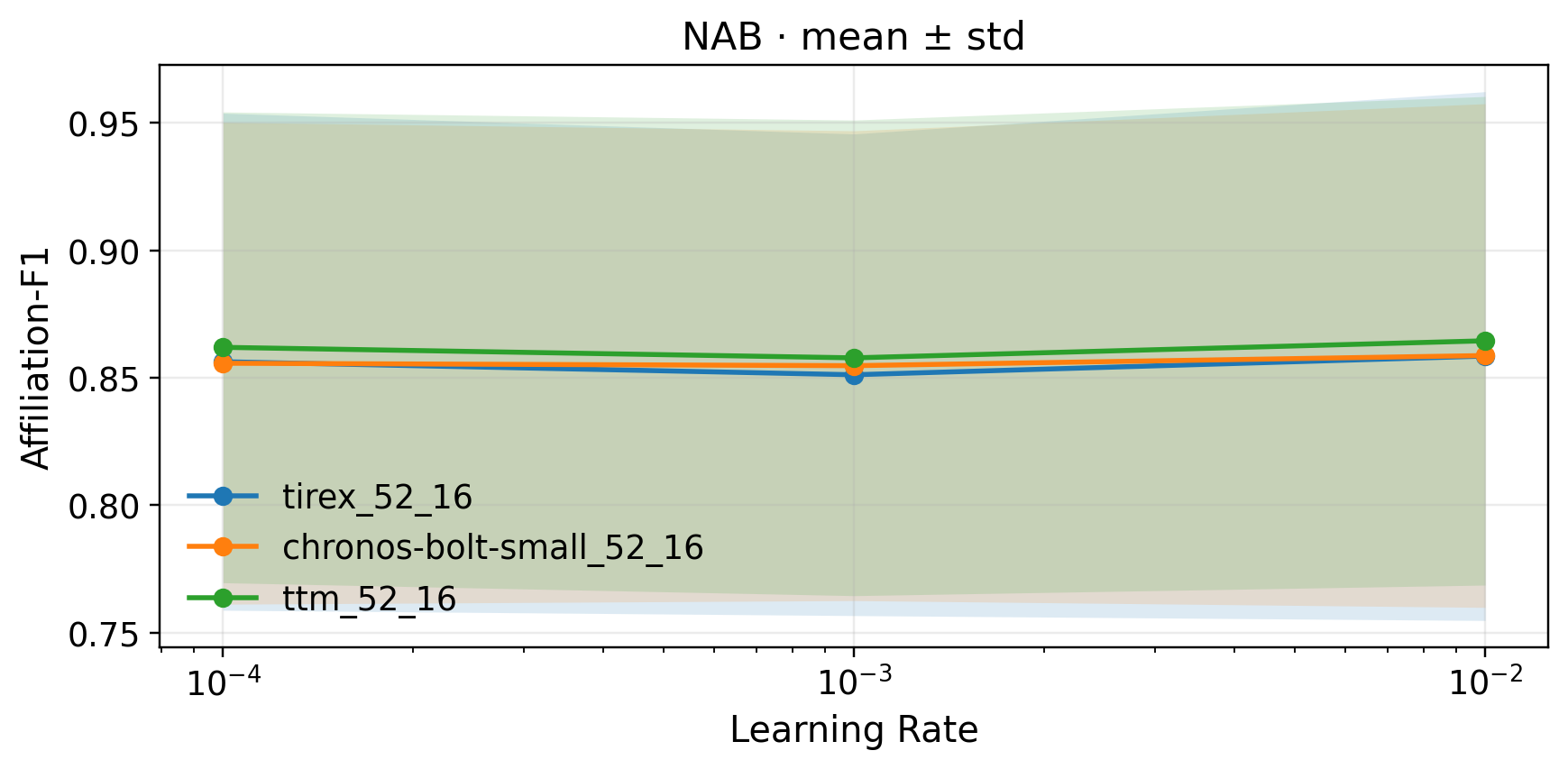}}
  \subfigure[NAB — AUC-PR]{\includegraphics[width=0.24\textwidth]{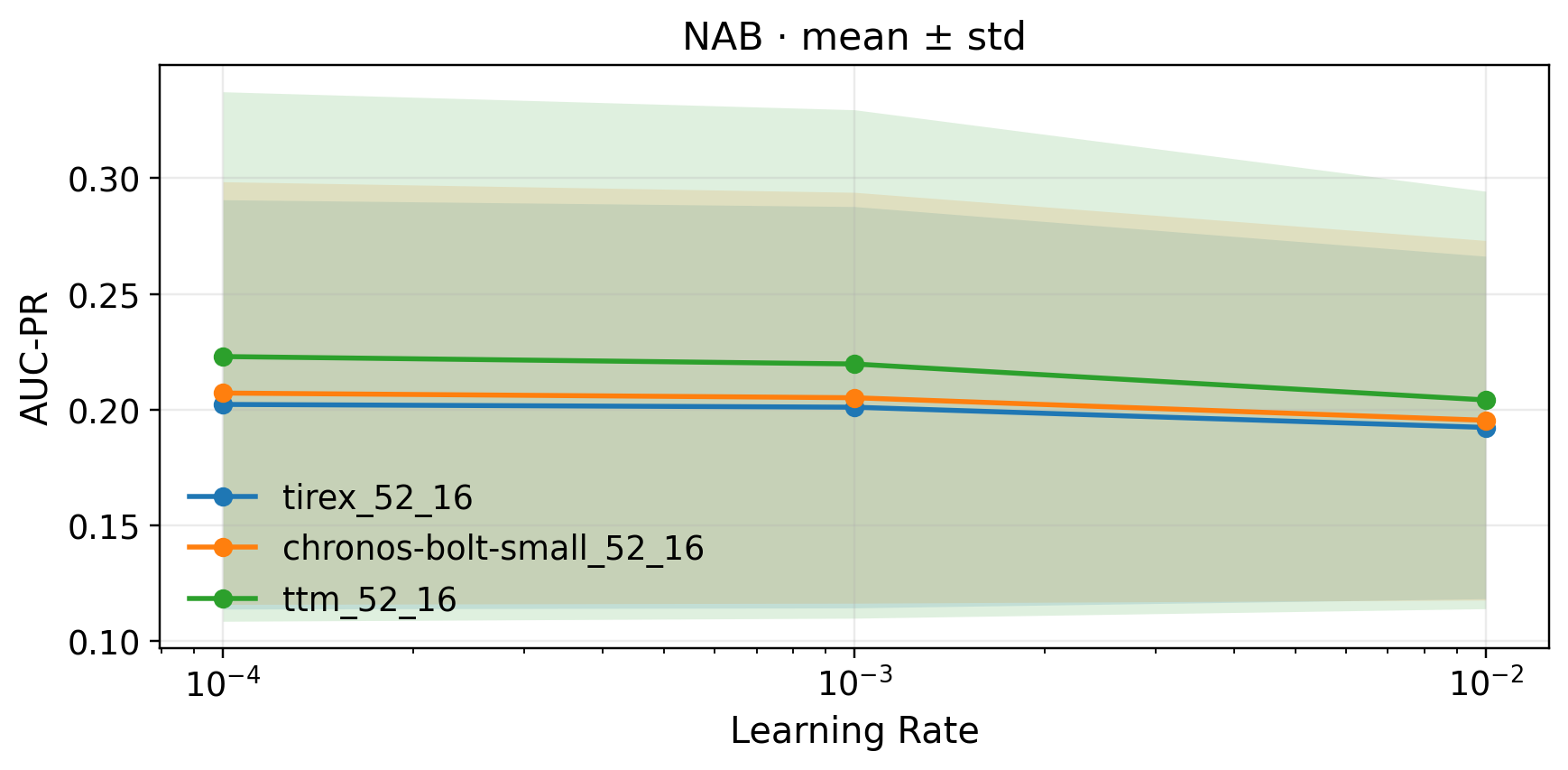}}
  \subfigure[NAB — VUS-PR]{\includegraphics[width=0.24\textwidth]{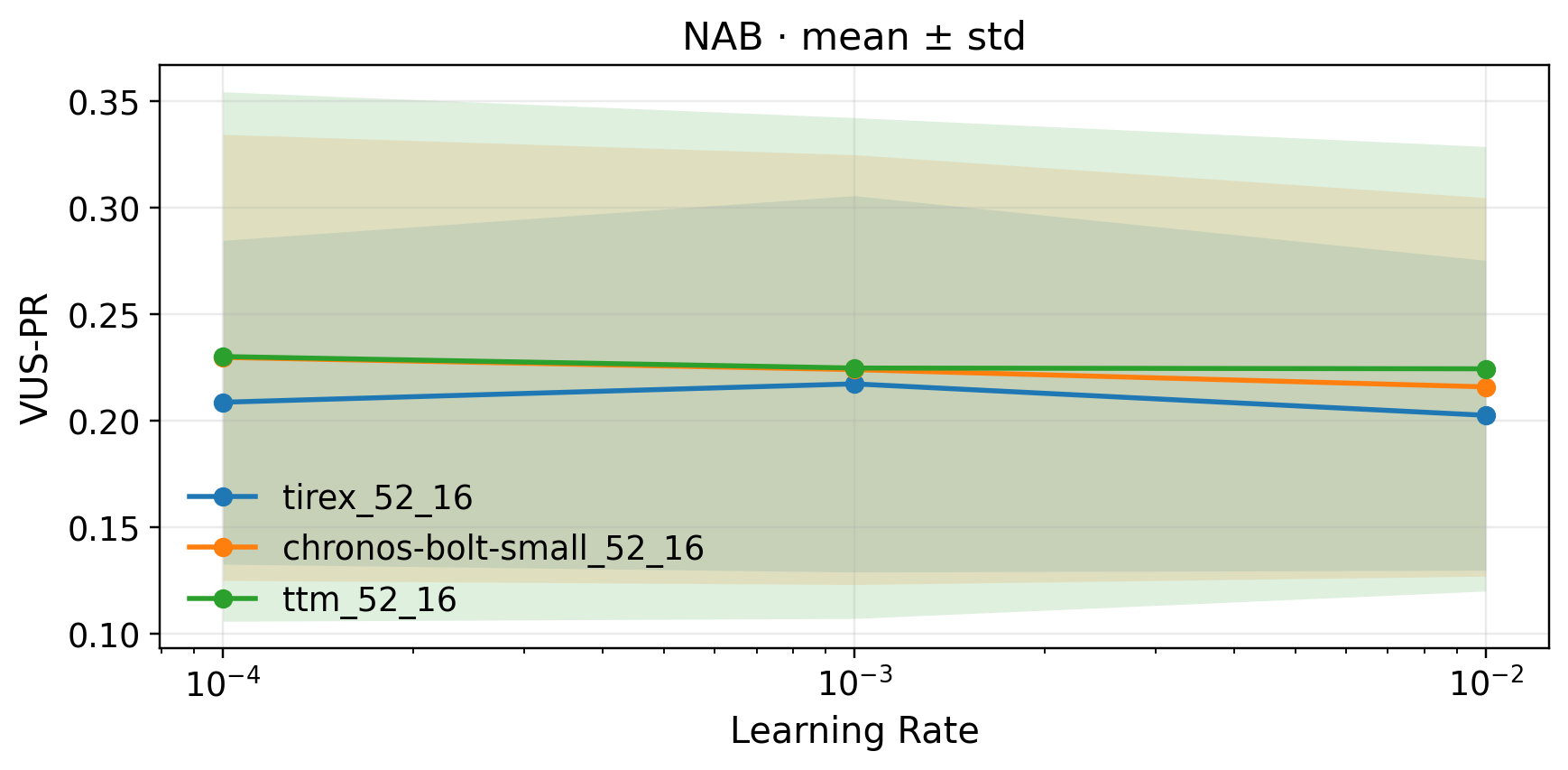}}

  \subfigure[NEK — PA-F1]{\includegraphics[width=0.24\textwidth]{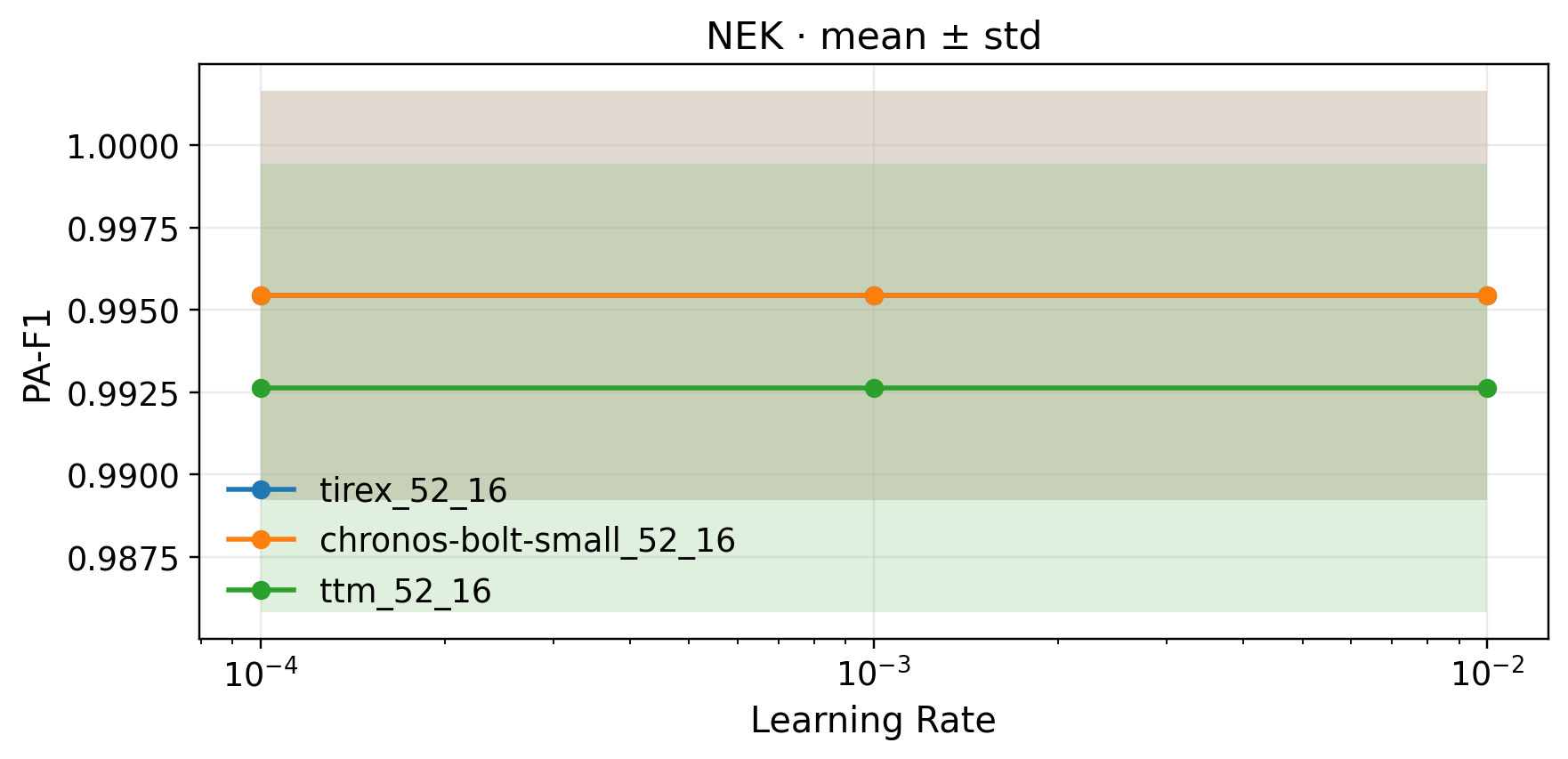}}
  \subfigure[NEK — Affiliation-F]{\includegraphics[width=0.24\textwidth]{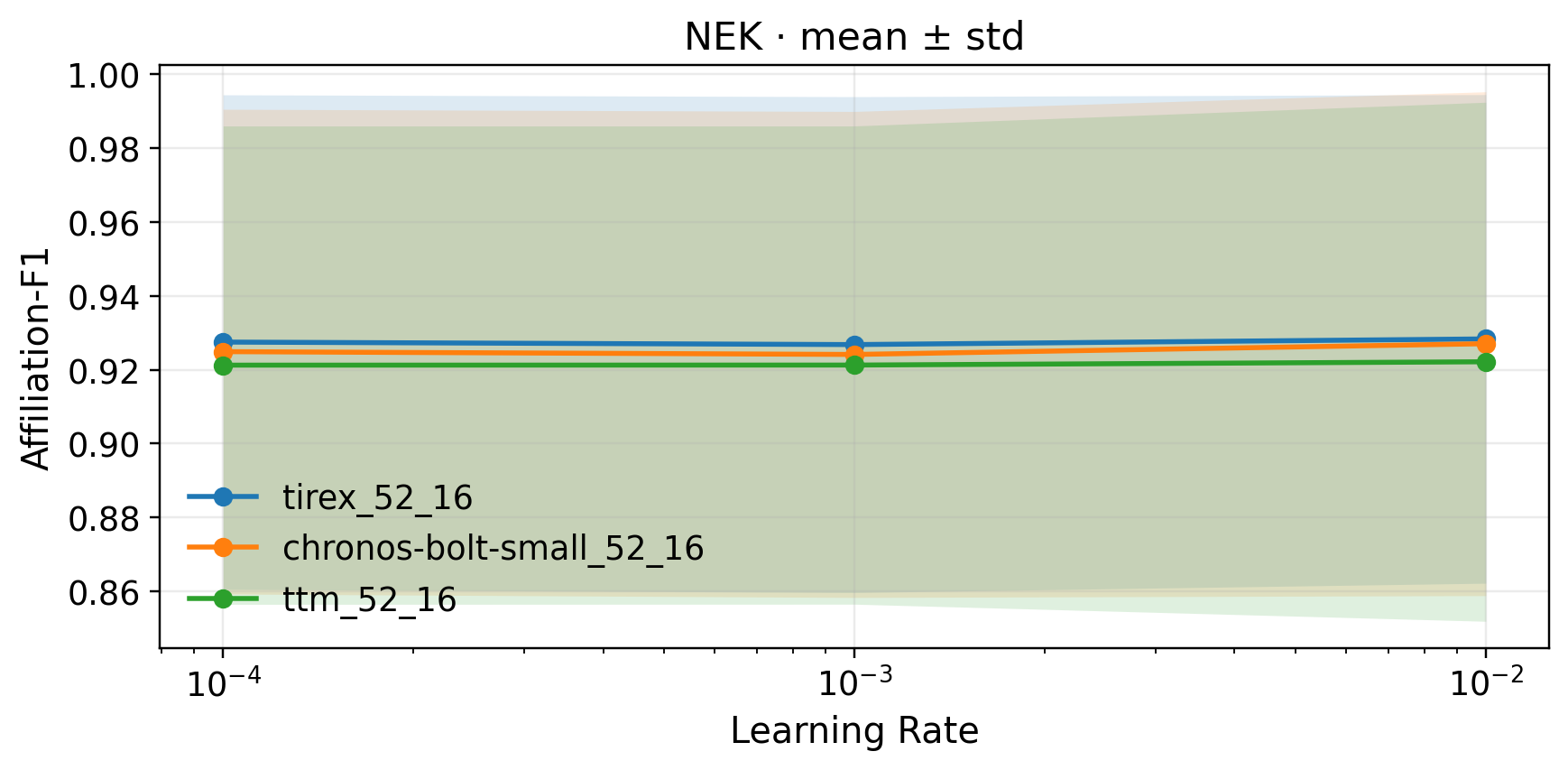}}
  \subfigure[NEK — AUC-PR]{\includegraphics[width=0.24\textwidth]{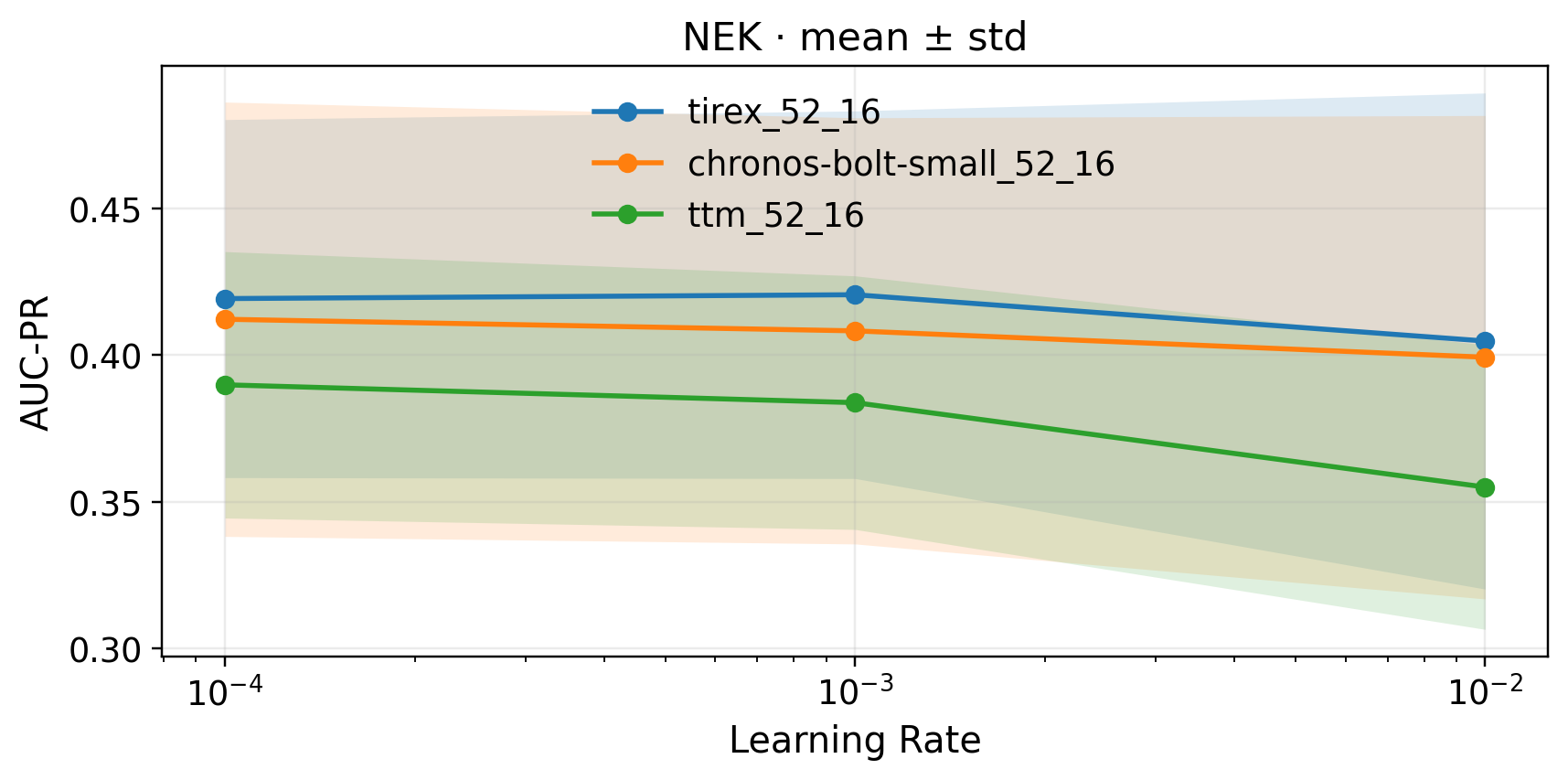}}
  \subfigure[NEK — VUS-PR]{\includegraphics[width=0.24\textwidth]{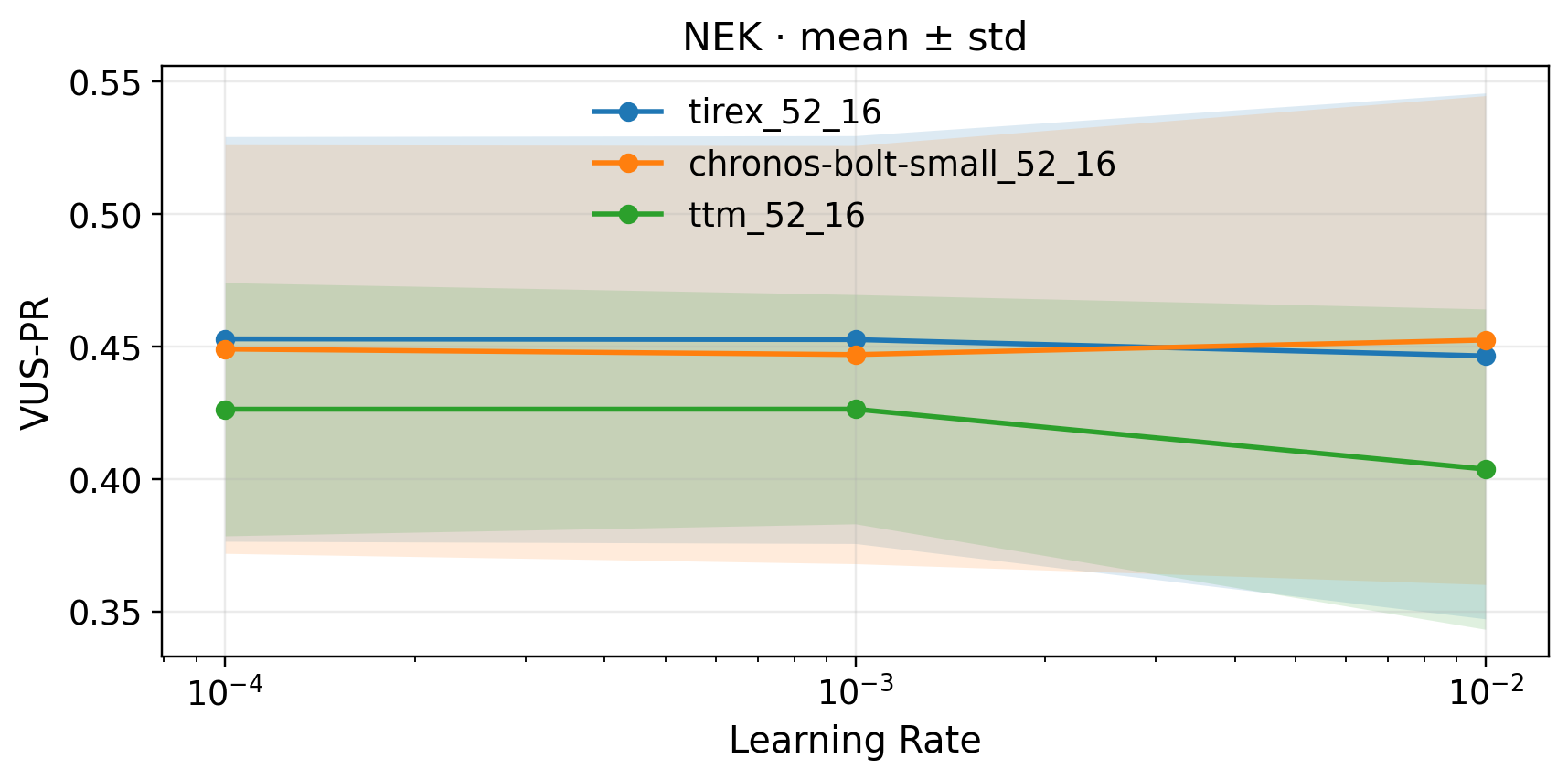}}

  \subfigure[MSL — PA-F1]{\includegraphics[width=0.24\textwidth]{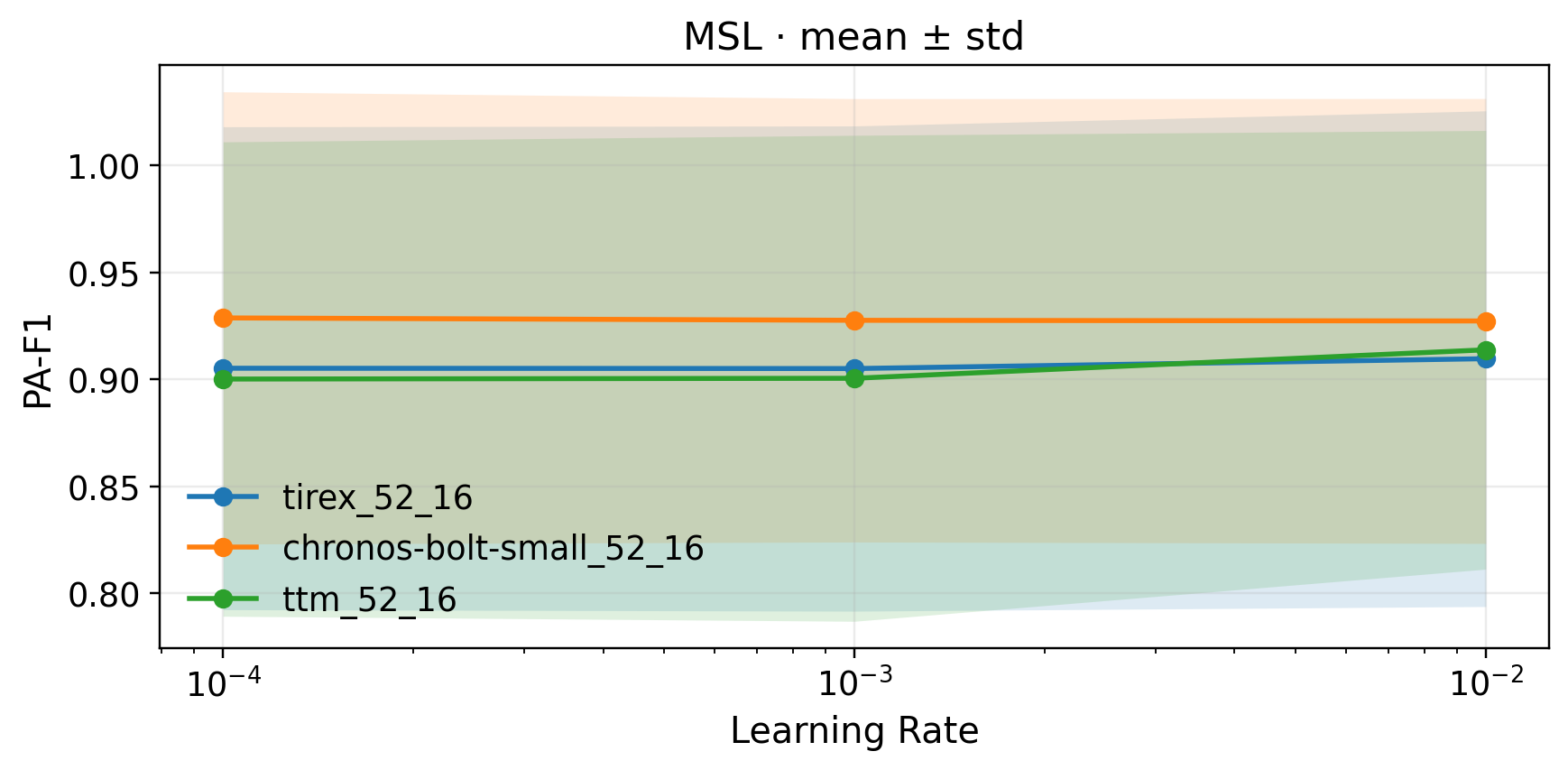}}
  \subfigure[MSL — Affiliation-F]{\includegraphics[width=0.24\textwidth]{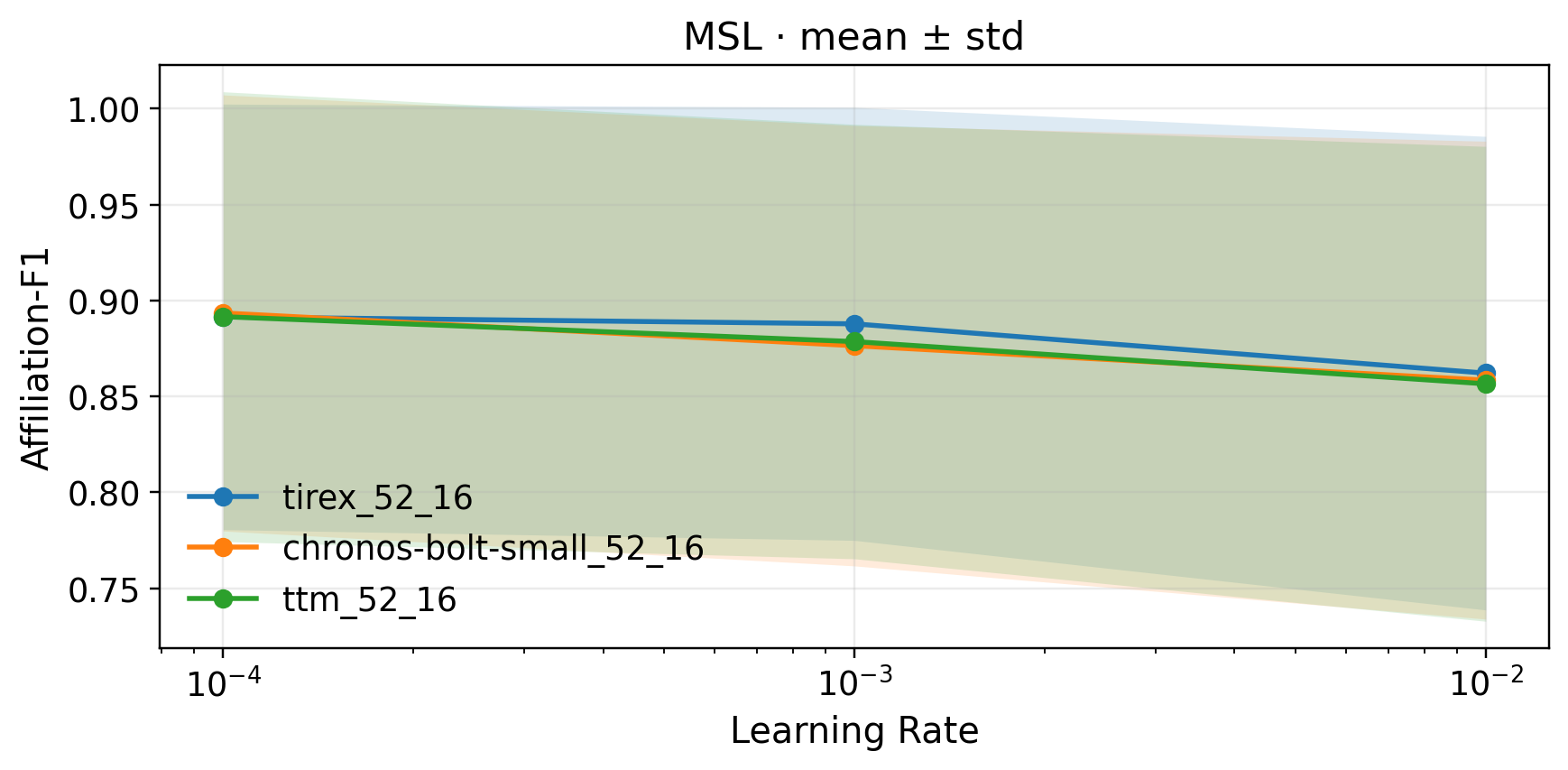}}
  \subfigure[MSL — AUC-PR]{\includegraphics[width=0.24\textwidth]{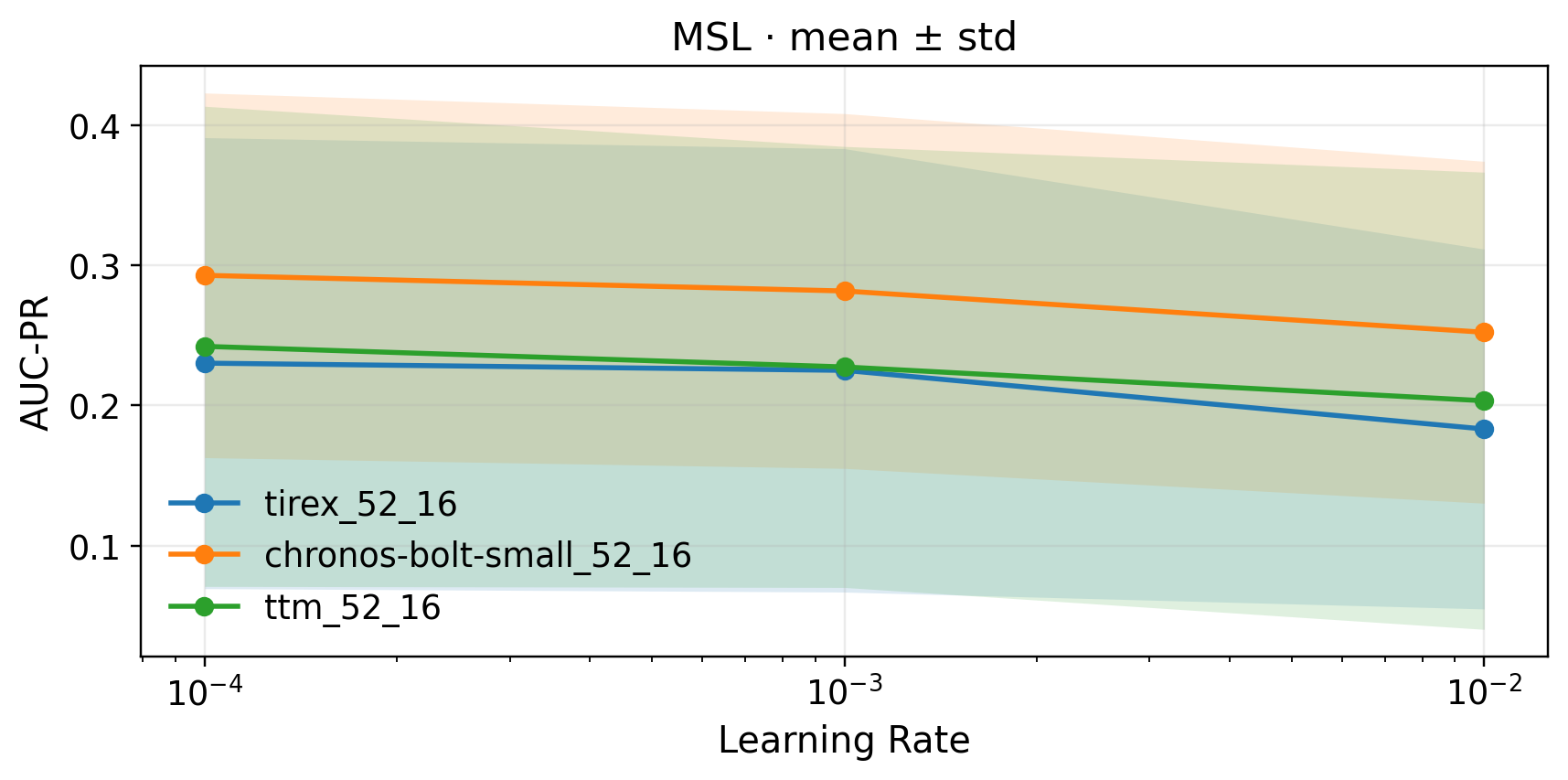}}
  \subfigure[MSL — VUS-PR]{\includegraphics[width=0.24\textwidth]{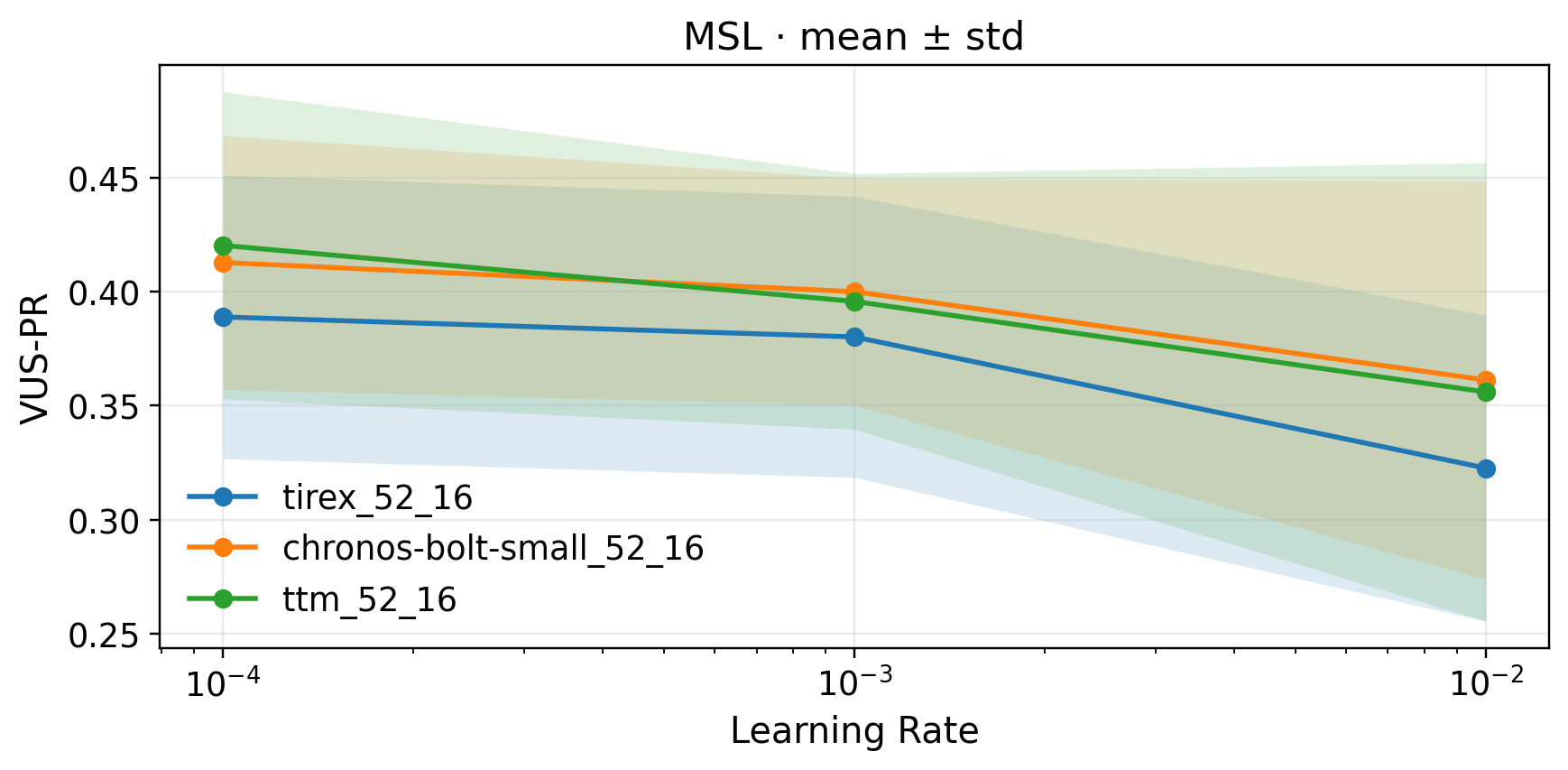}}

  \subfigure[YAHOO — PA-F1]{\includegraphics[width=0.24\textwidth]{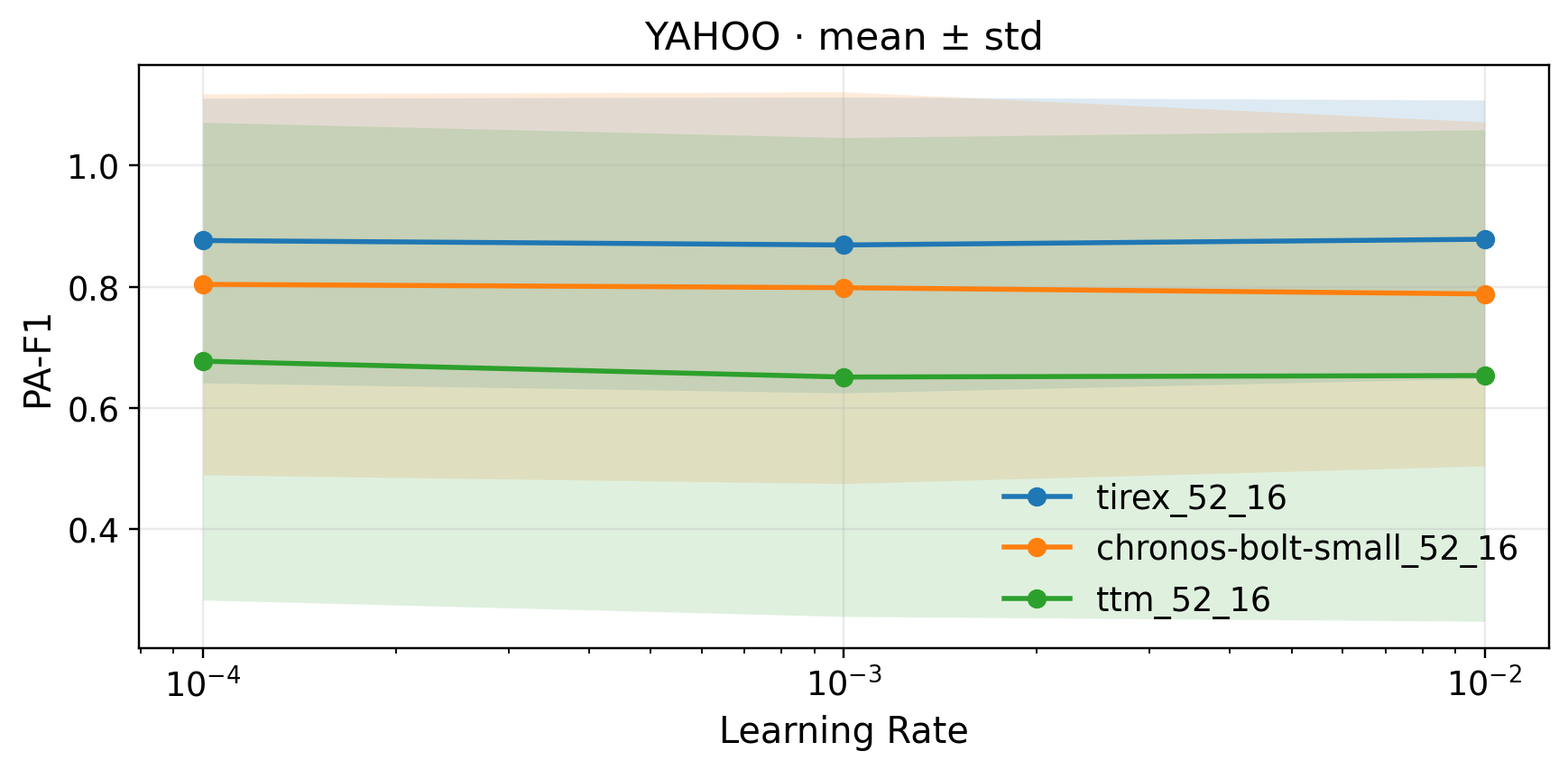}}
  \subfigure[YAHOO-Affiliation-F]{\includegraphics[width=0.24\textwidth]{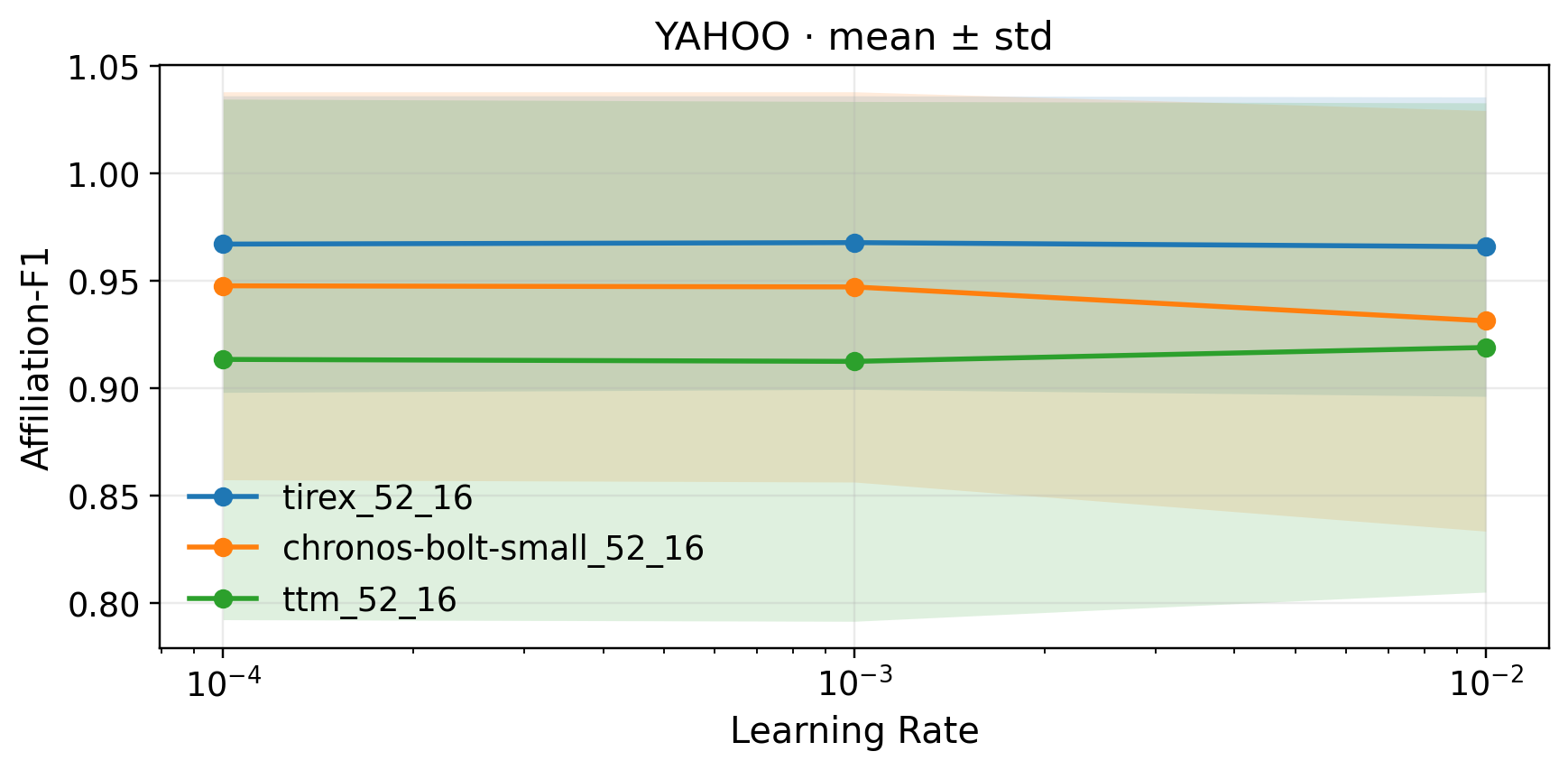}}
  \subfigure[YAHOO — AUC-PR]{\includegraphics[width=0.24\textwidth]{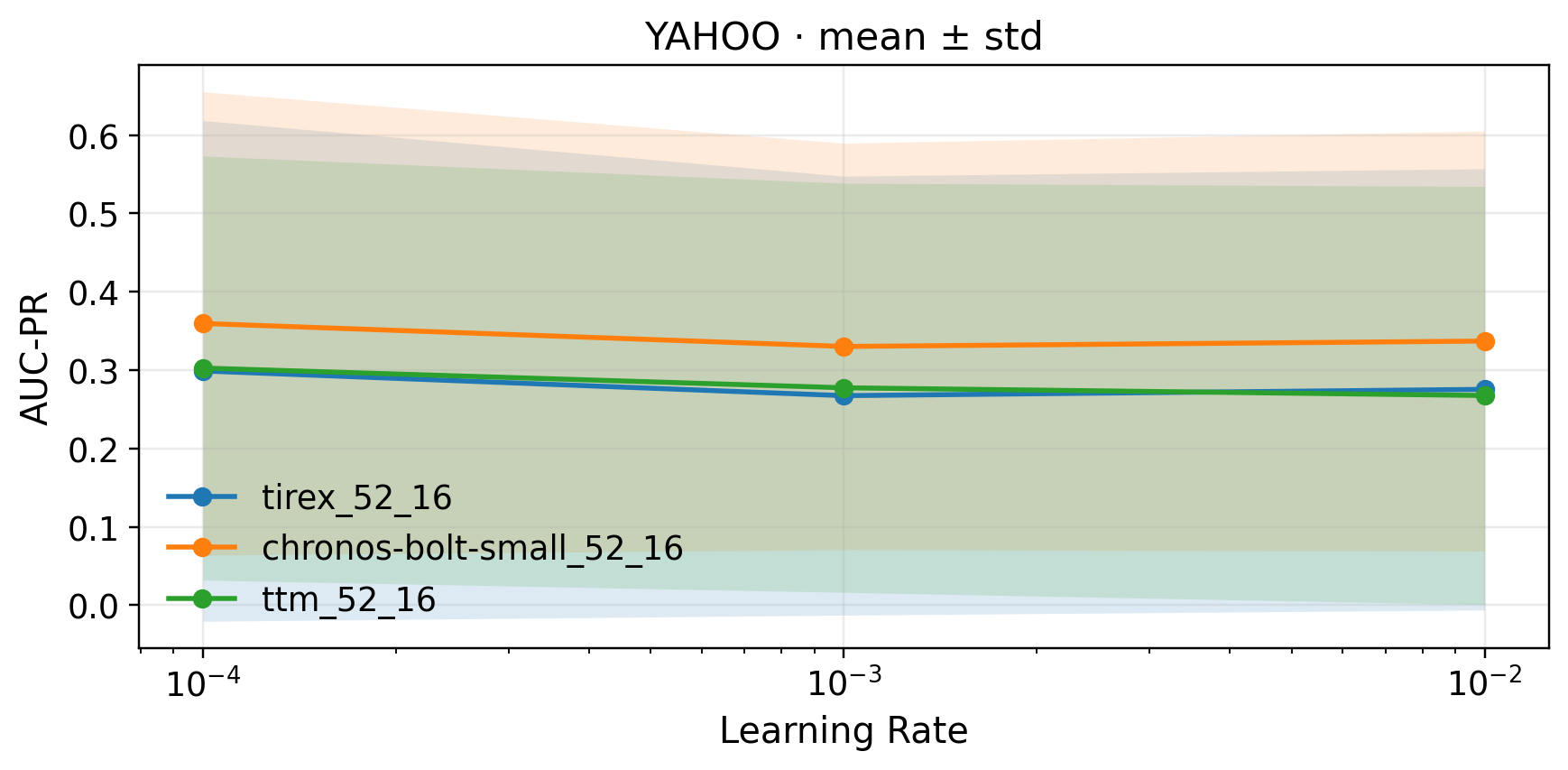}}
  \subfigure[YAHOO — VUS-PR]{\includegraphics[width=0.24\textwidth]{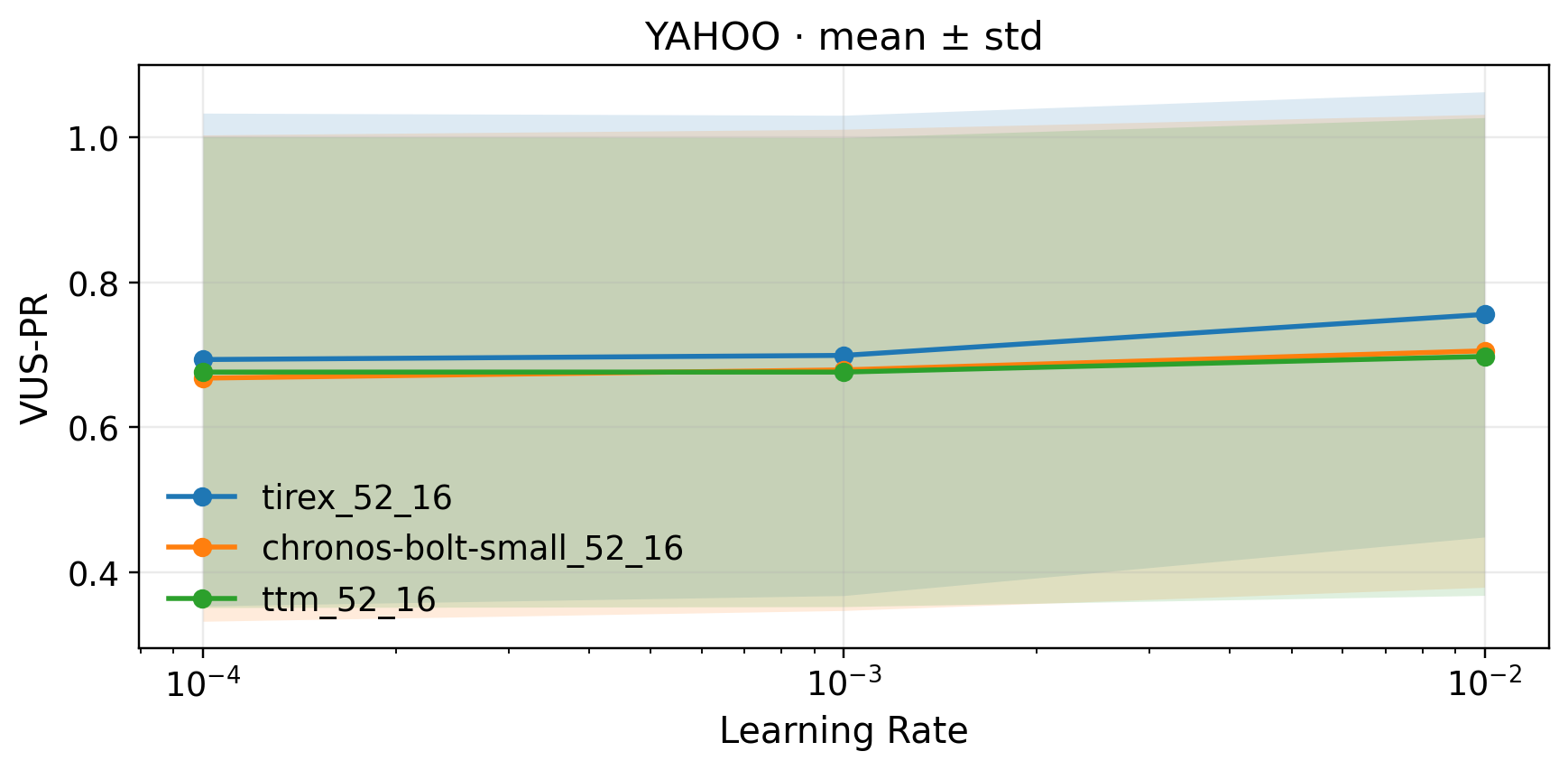}}



  \caption{{Performance of $\mathcal{W}_1$-ACAS when aggregating different learning rate. Rows correspond to datasets (NAB, NEK, MSL, YAHOO, Stock, WSD) and columns to metrics (PA-F1, Affiliation-F, AUC-PR, VUS-PR).}}
  \label{fig:w1-acas-steps-grid-lr}
\end{figure*}

\begin{figure*}[t]
  \centering

  \subfigure[NAB — PA-F1]{\includegraphics[width=0.24\textwidth]{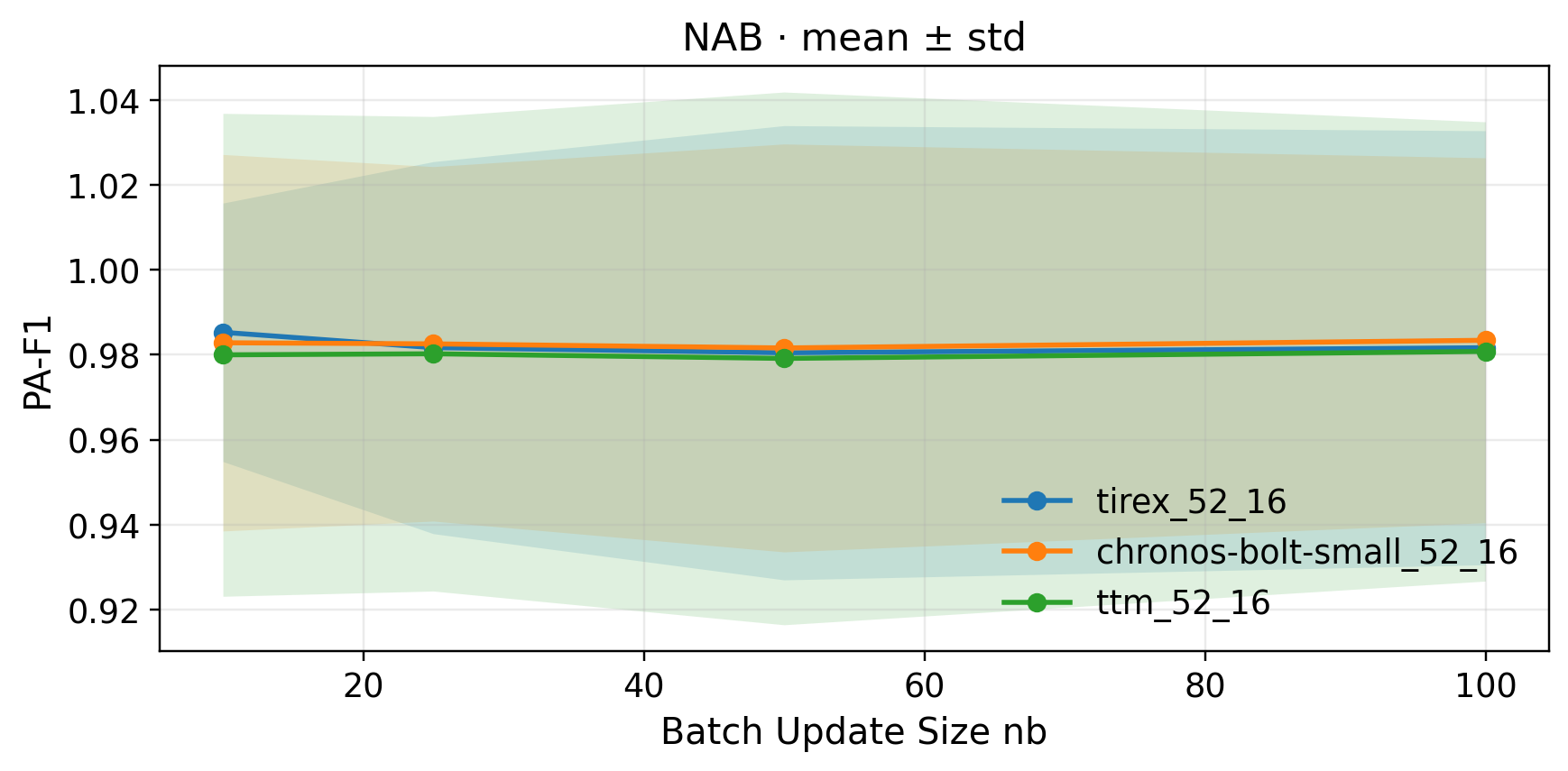}}
  \subfigure[NAB — Affiliation-F]{\includegraphics[width=0.24\textwidth]{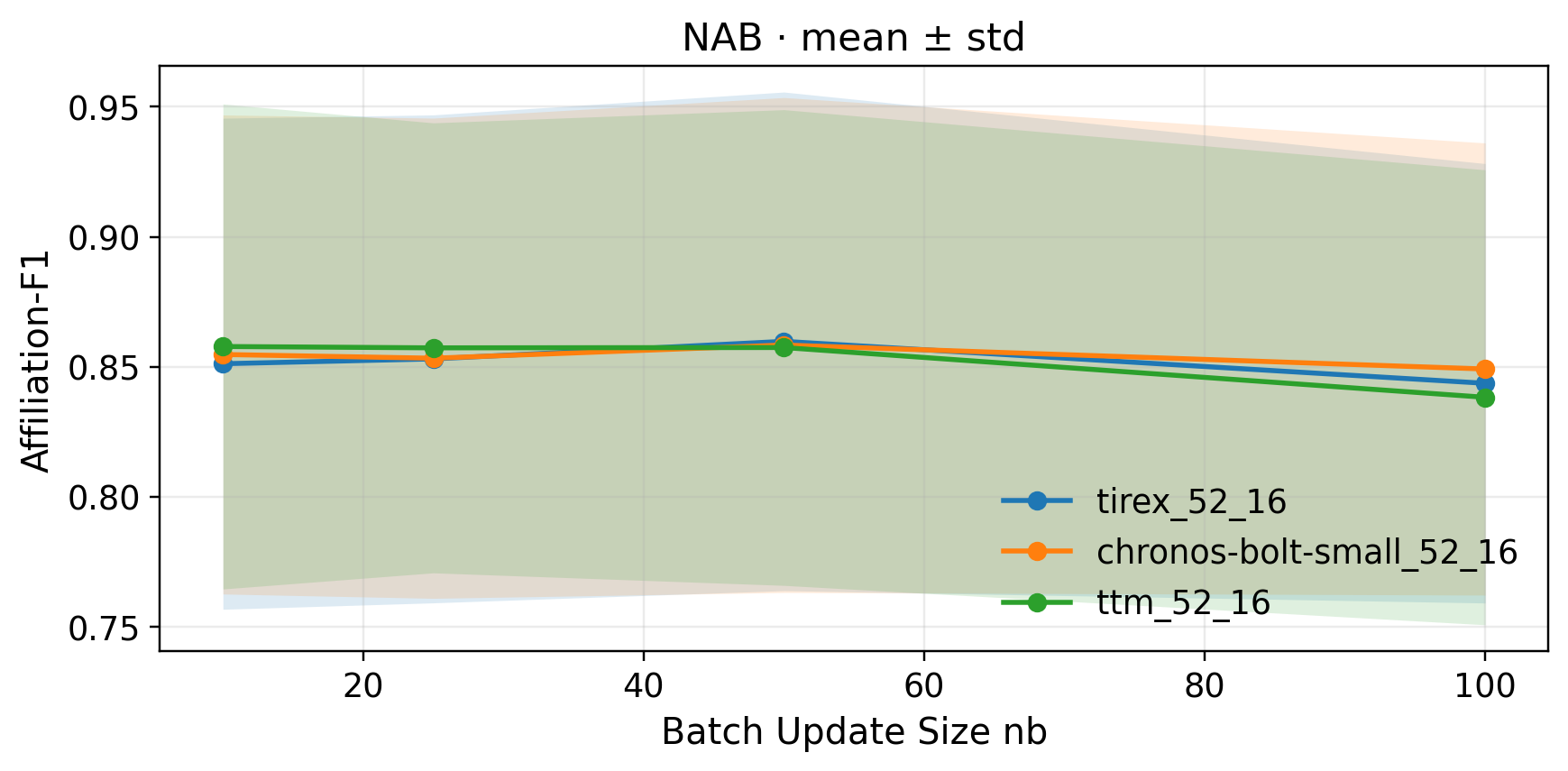}}
  \subfigure[NAB — AUC-PR]{\includegraphics[width=0.24\textwidth]{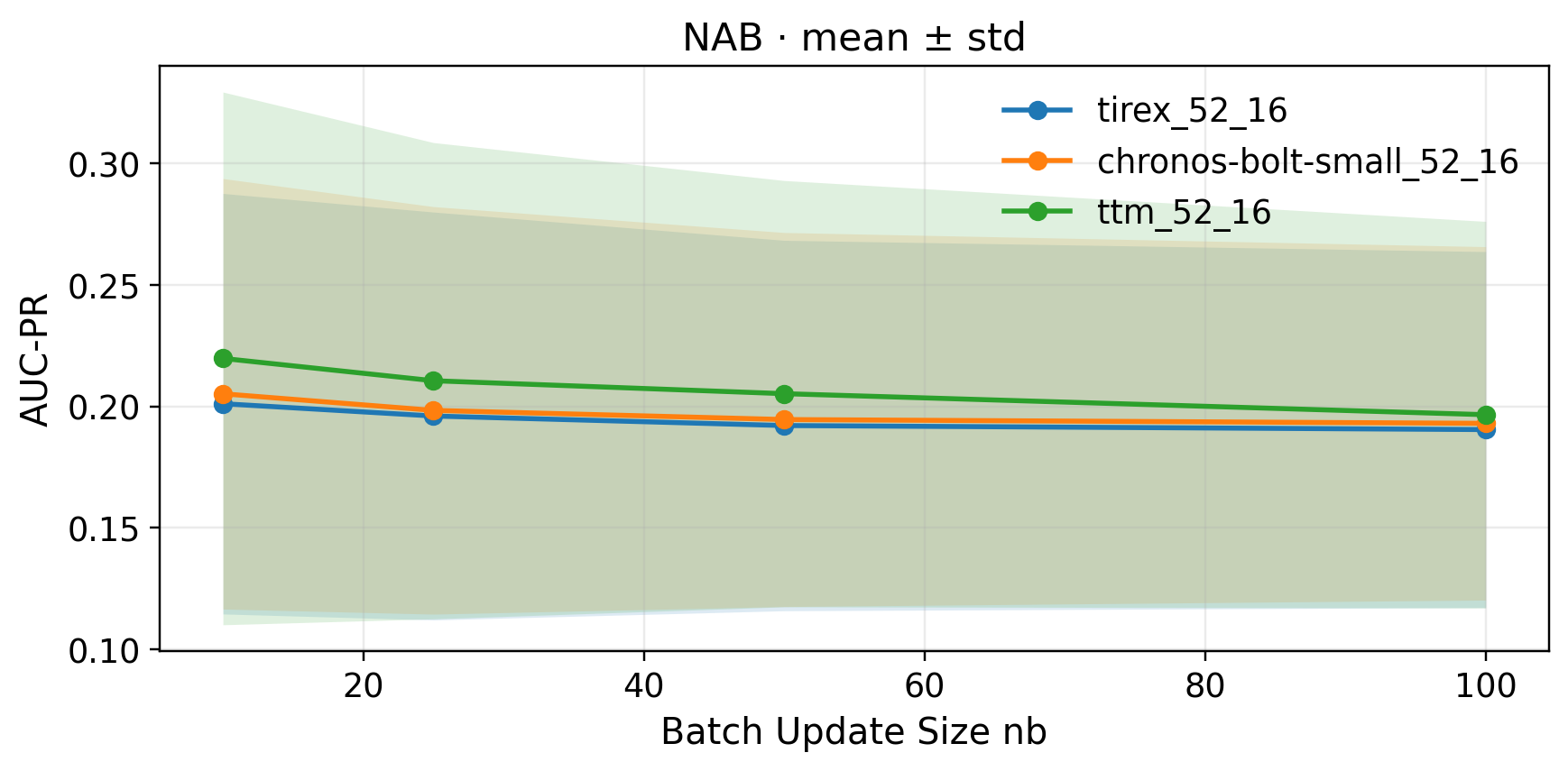}}
  \subfigure[NAB — VUS-PR]{\includegraphics[width=0.24\textwidth]{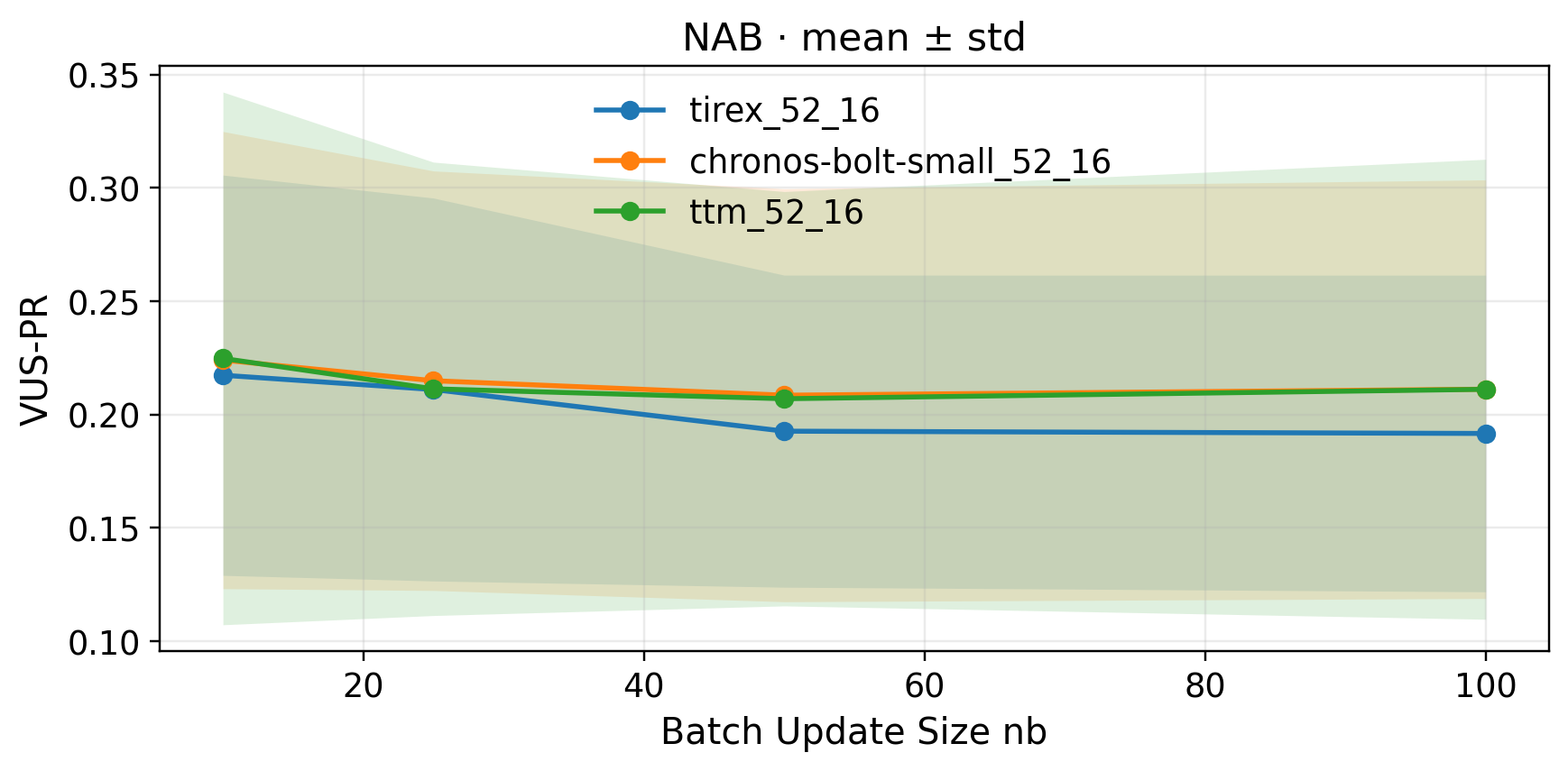}}

  \subfigure[NEK — PA-F1]{\includegraphics[width=0.24\textwidth]{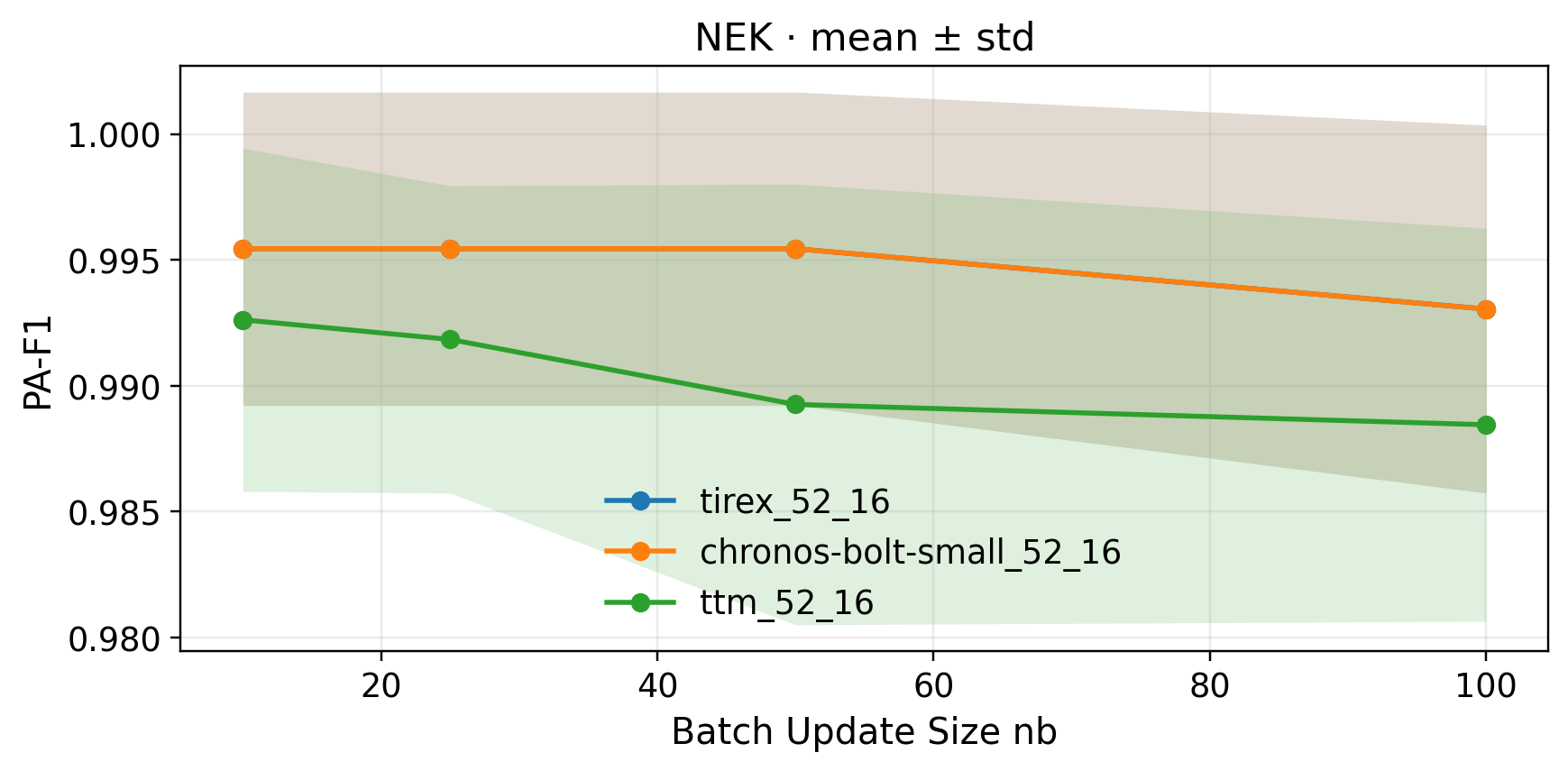}}
  \subfigure[NEK — Affiliation-F]{\includegraphics[width=0.24\textwidth]{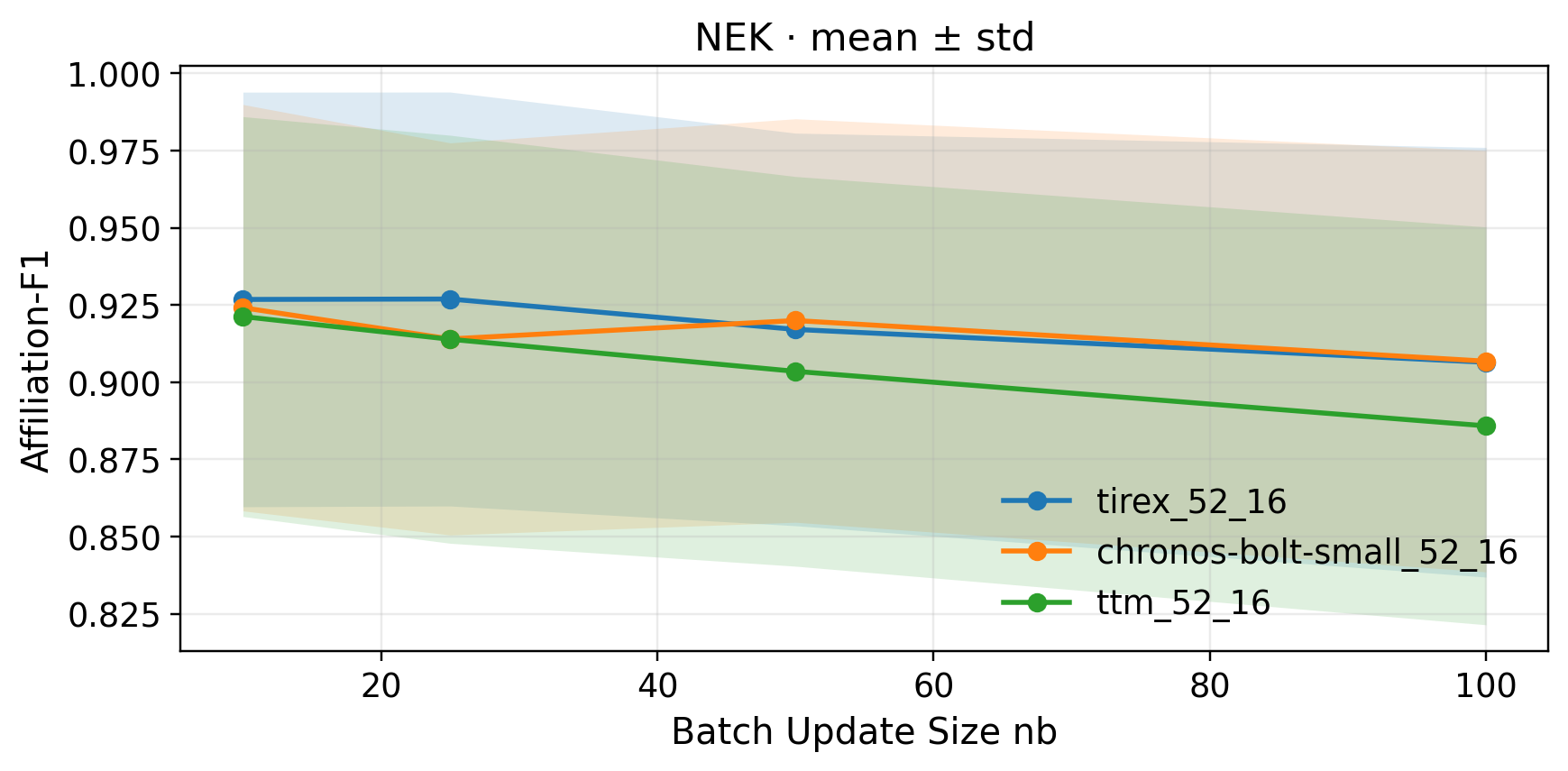}}
  \subfigure[NEK — AUC-PR]{\includegraphics[width=0.24\textwidth]{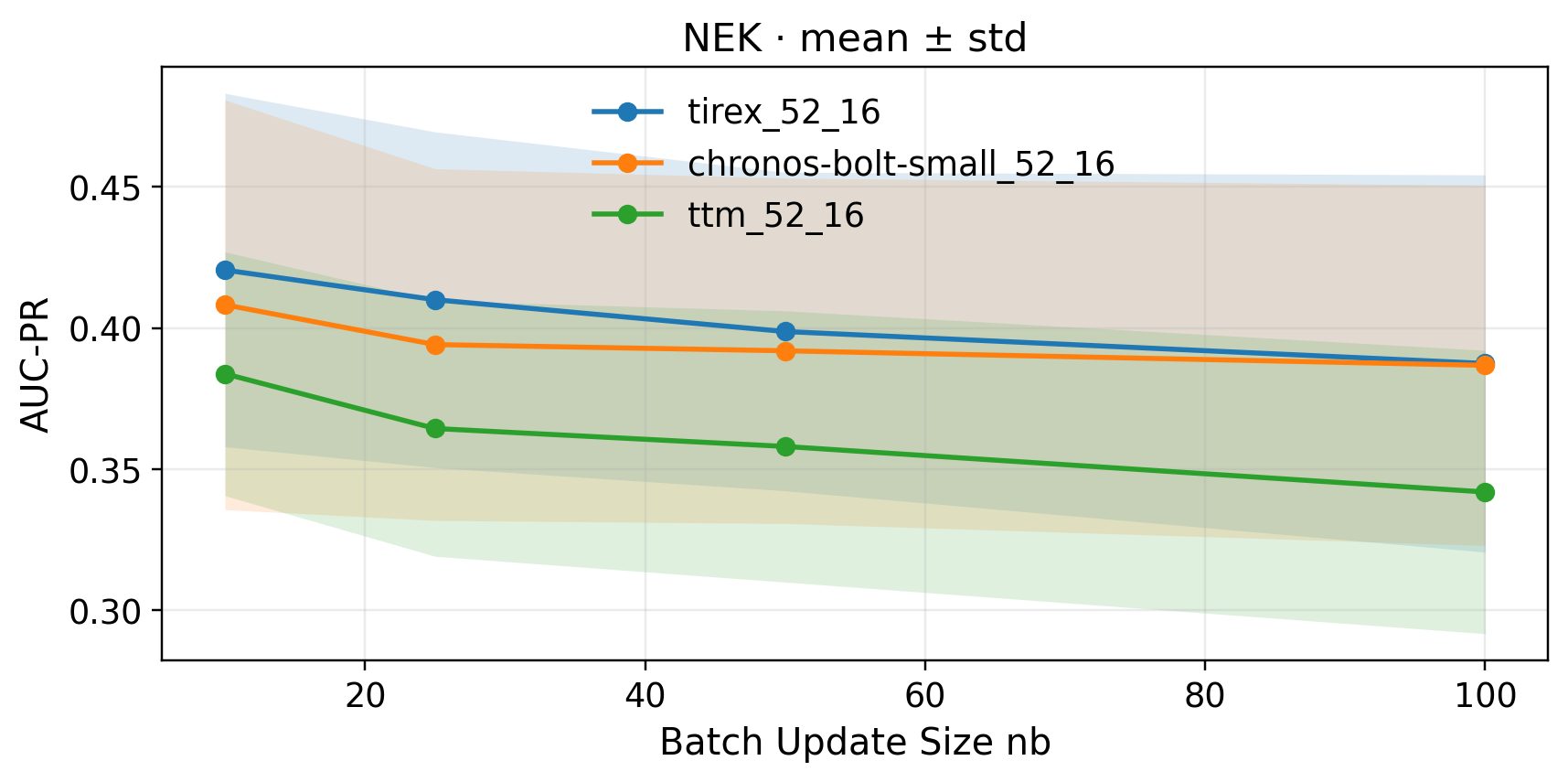}}
  \subfigure[NEK — VUS-PR]{\includegraphics[width=0.24\textwidth]{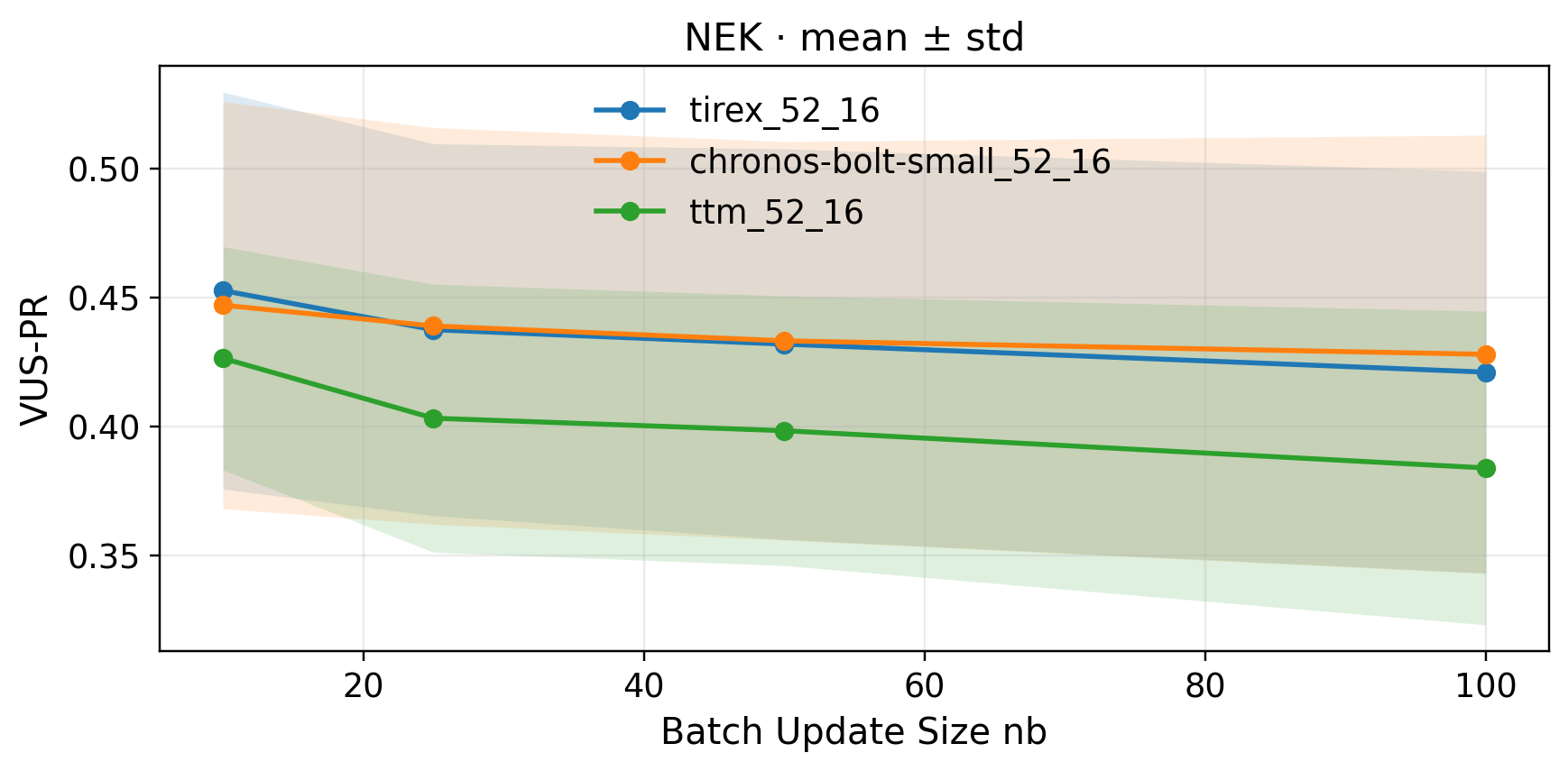}}

  \subfigure[MSL — PA-F1]{\includegraphics[width=0.24\textwidth]{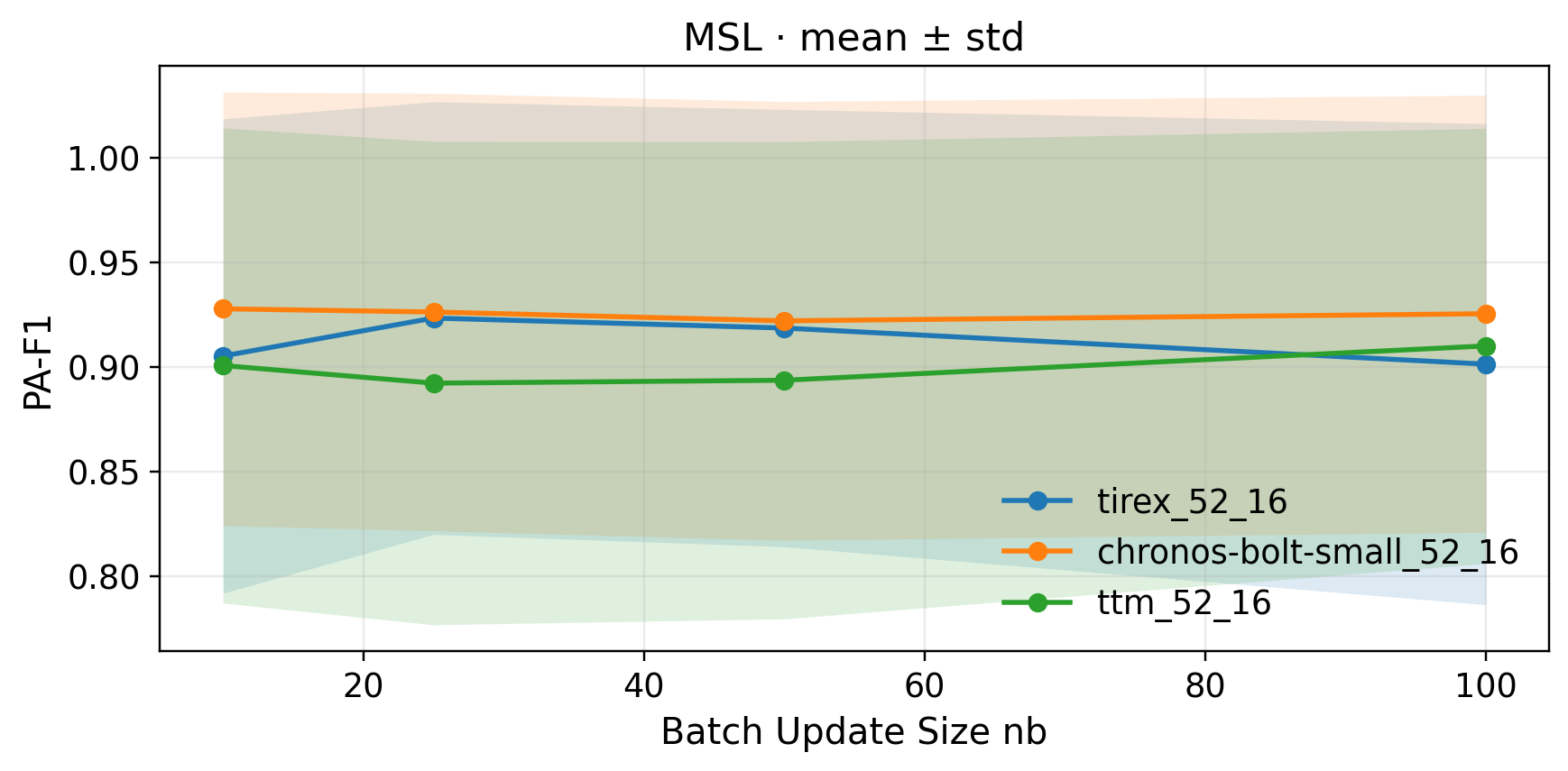}}
  \subfigure[MSL — Affiliation-F]{\includegraphics[width=0.24\textwidth]{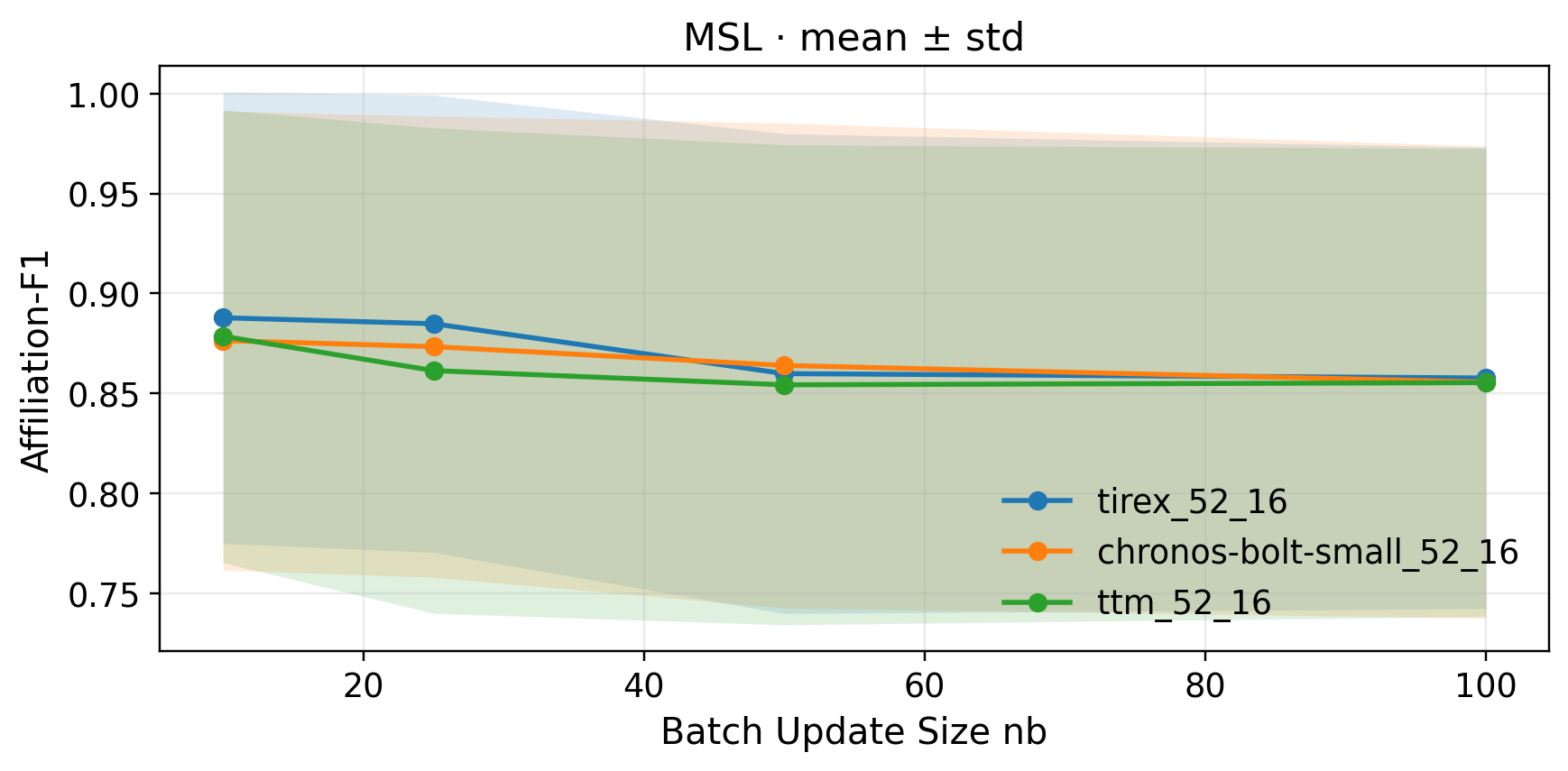}}
  \subfigure[MSL — AUC-PR]{\includegraphics[width=0.24\textwidth]{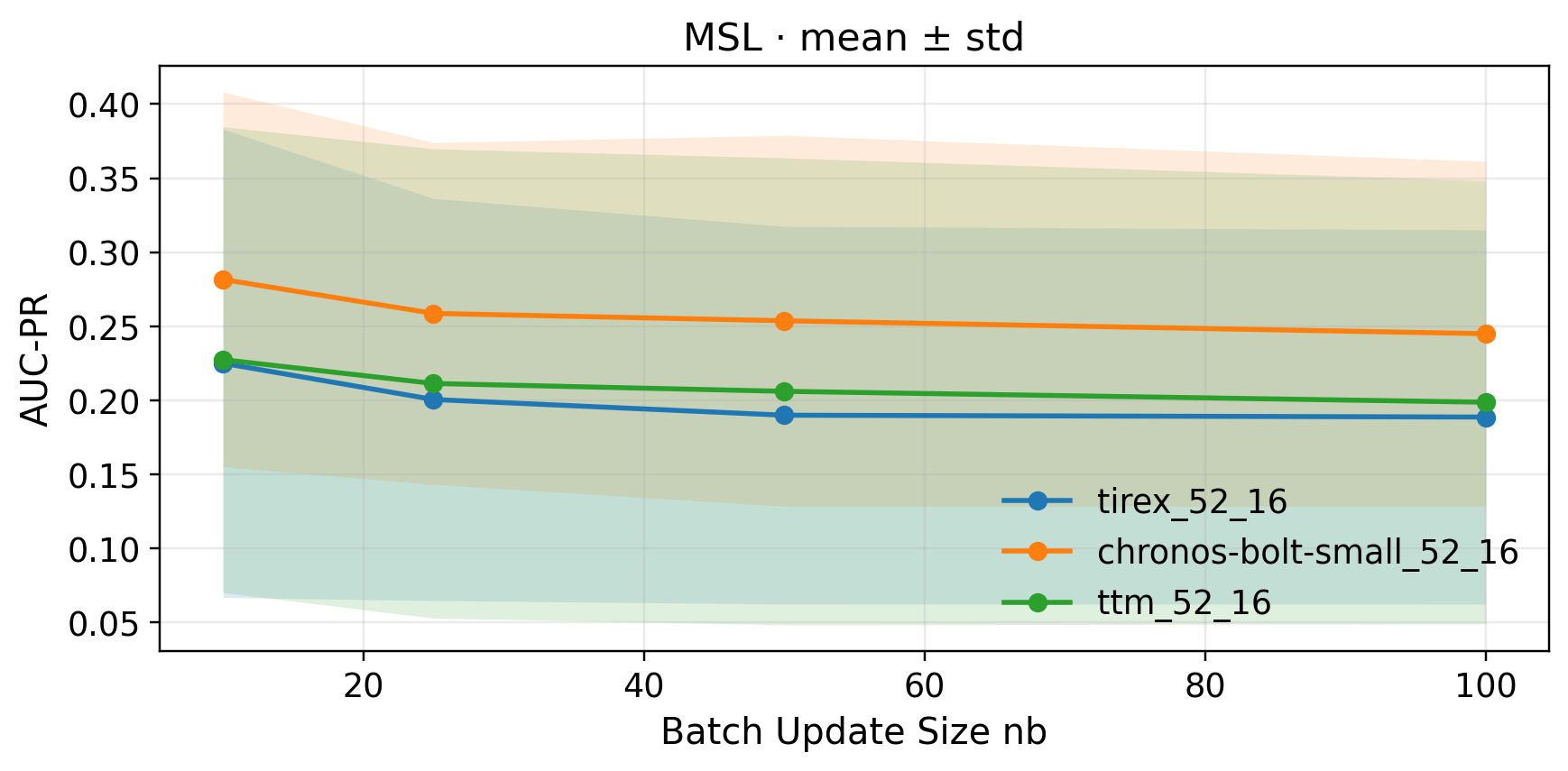}}
  \subfigure[MSL — VUS-PR]{\includegraphics[width=0.24\textwidth]{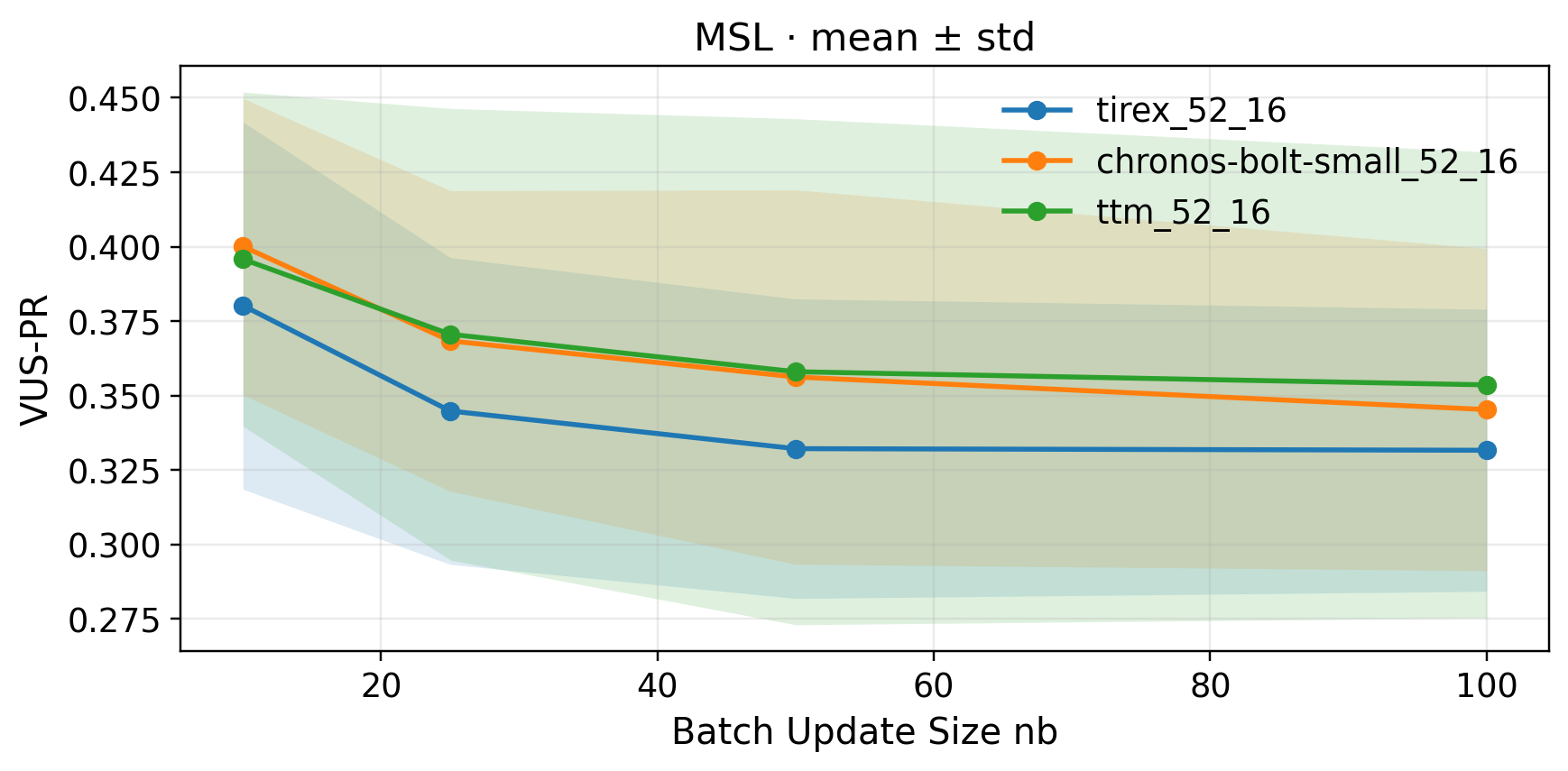}}

  \subfigure[YAHOO — PA-F1]{\includegraphics[width=0.24\textwidth]{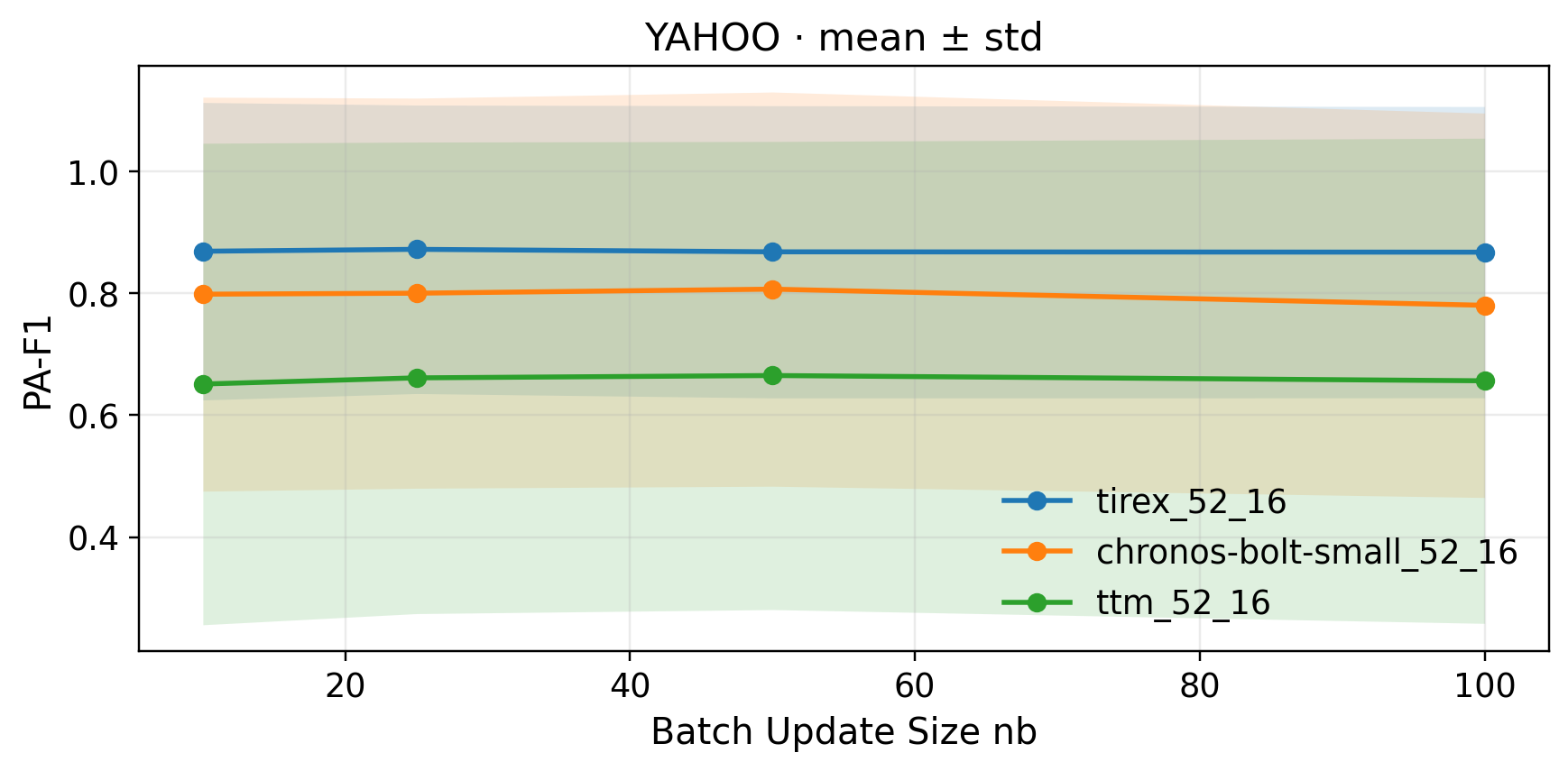}}
  \subfigure[YAHOO-Affiliation-F]{\includegraphics[width=0.24\textwidth]{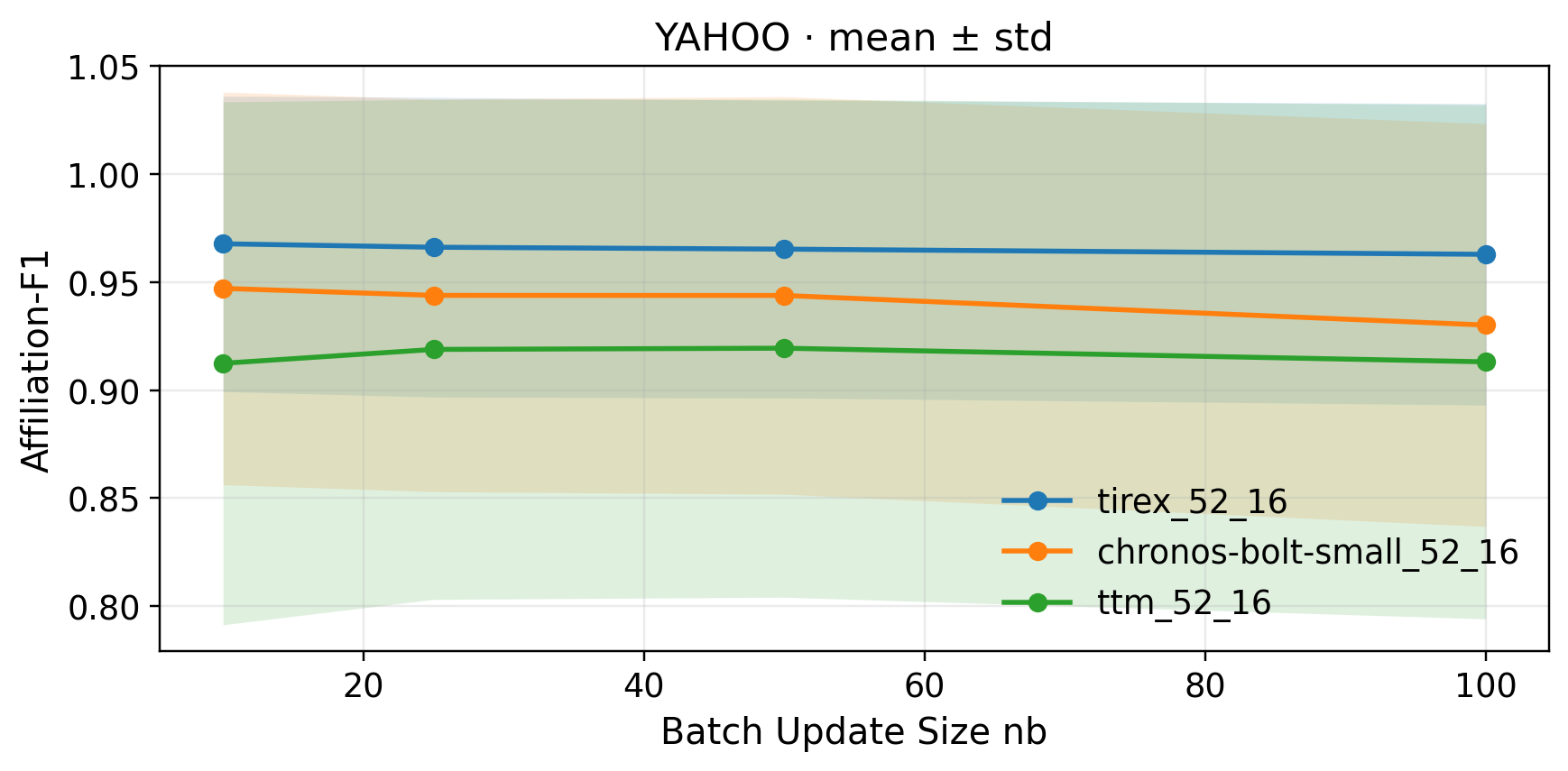}}
  \subfigure[YAHOO — AUC-PR]{\includegraphics[width=0.24\textwidth]{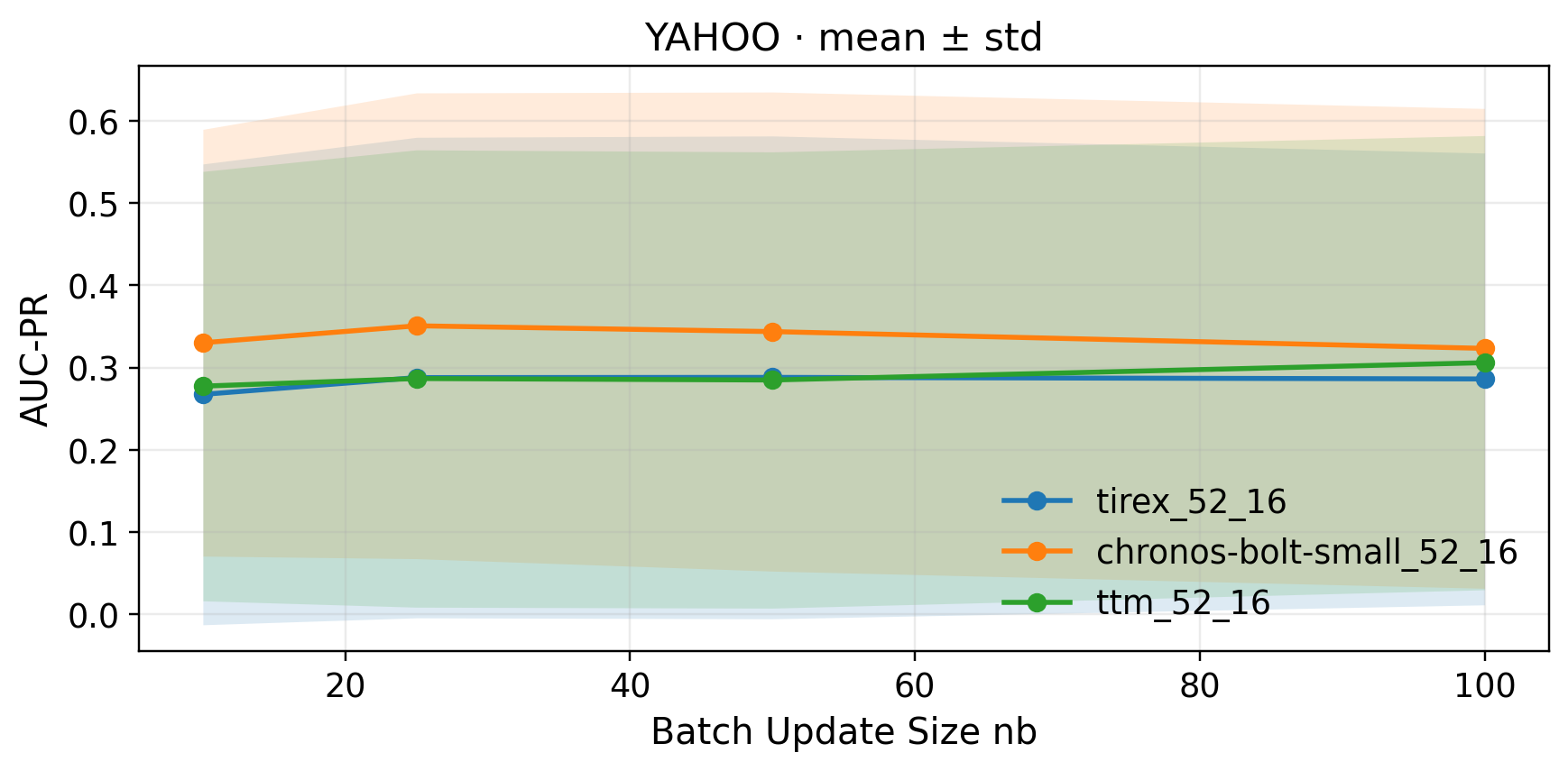}}
  \subfigure[YAHOO — VUS-PR]{\includegraphics[width=0.24\textwidth]{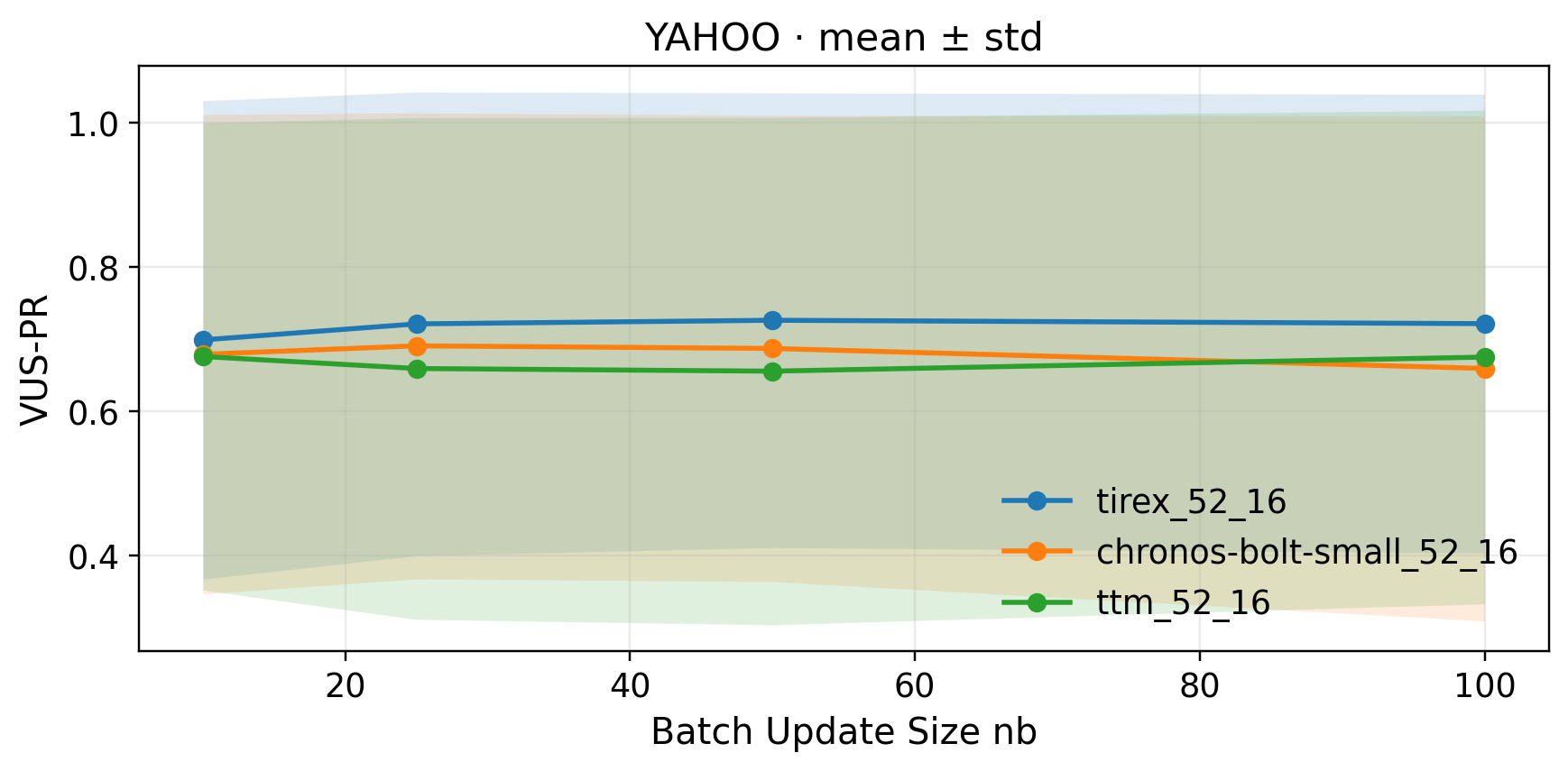}}



  \caption{{Performance of $\mathcal{W}_1$-ACAS when aggregating different batch size update $n_b$. Rows correspond to datasets (NAB, NEK, MSL, YAHOO, Stock, WSD) and columns to metrics (PA-F1, Affiliation-F, AUC-PR, VUS-PR).}}
  \label{fig:w1-acas-steps-grid-nbu}
\end{figure*}

\begin{figure*}[t]
  \centering

  \subfigure[NAB — PA-F1]{\includegraphics[width=0.24\textwidth]{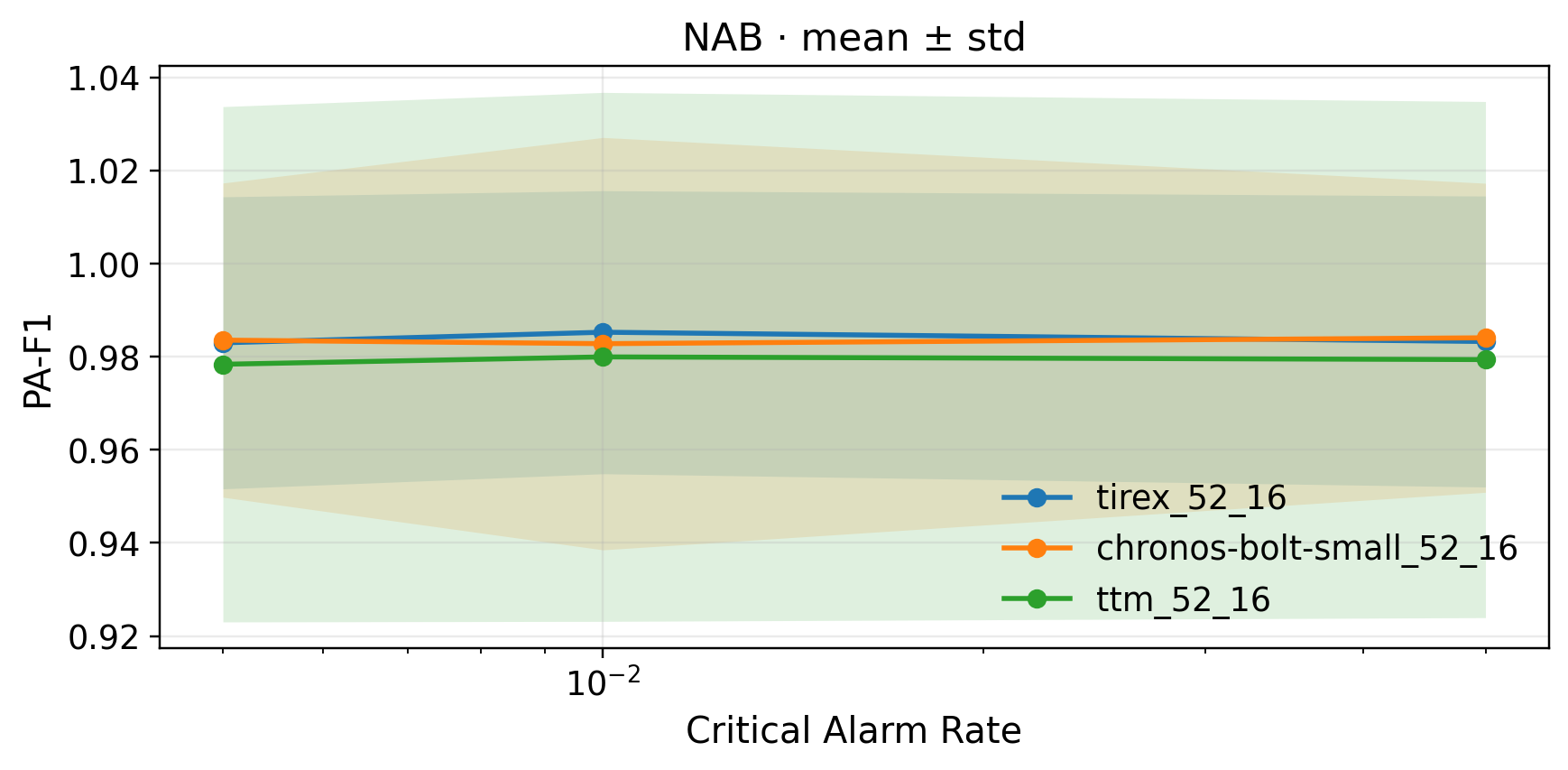}}
  \subfigure[NAB — Affiliation-F]{\includegraphics[width=0.24\textwidth]{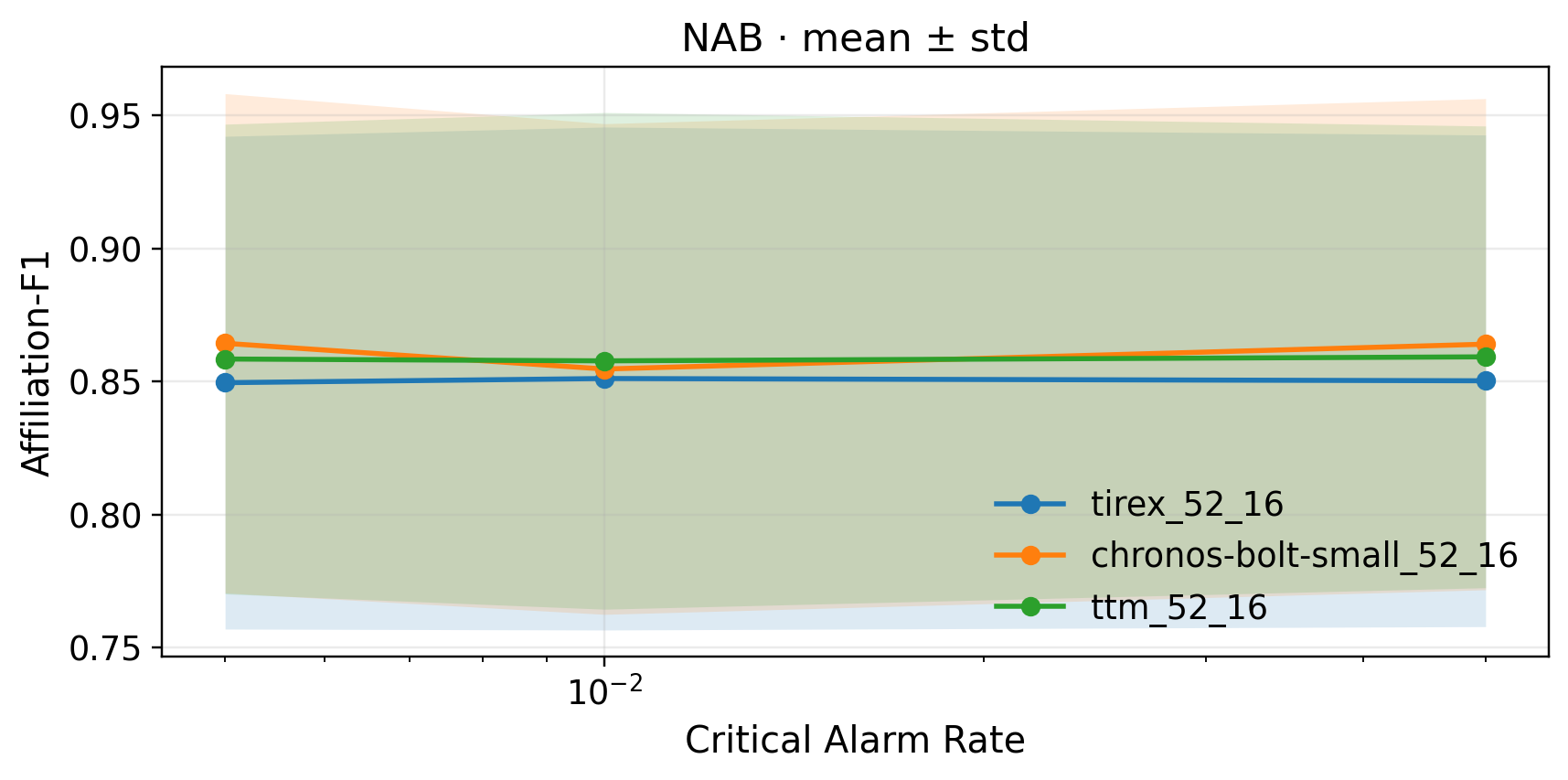}}
  \subfigure[NAB — AUC-PR]{\includegraphics[width=0.24\textwidth]{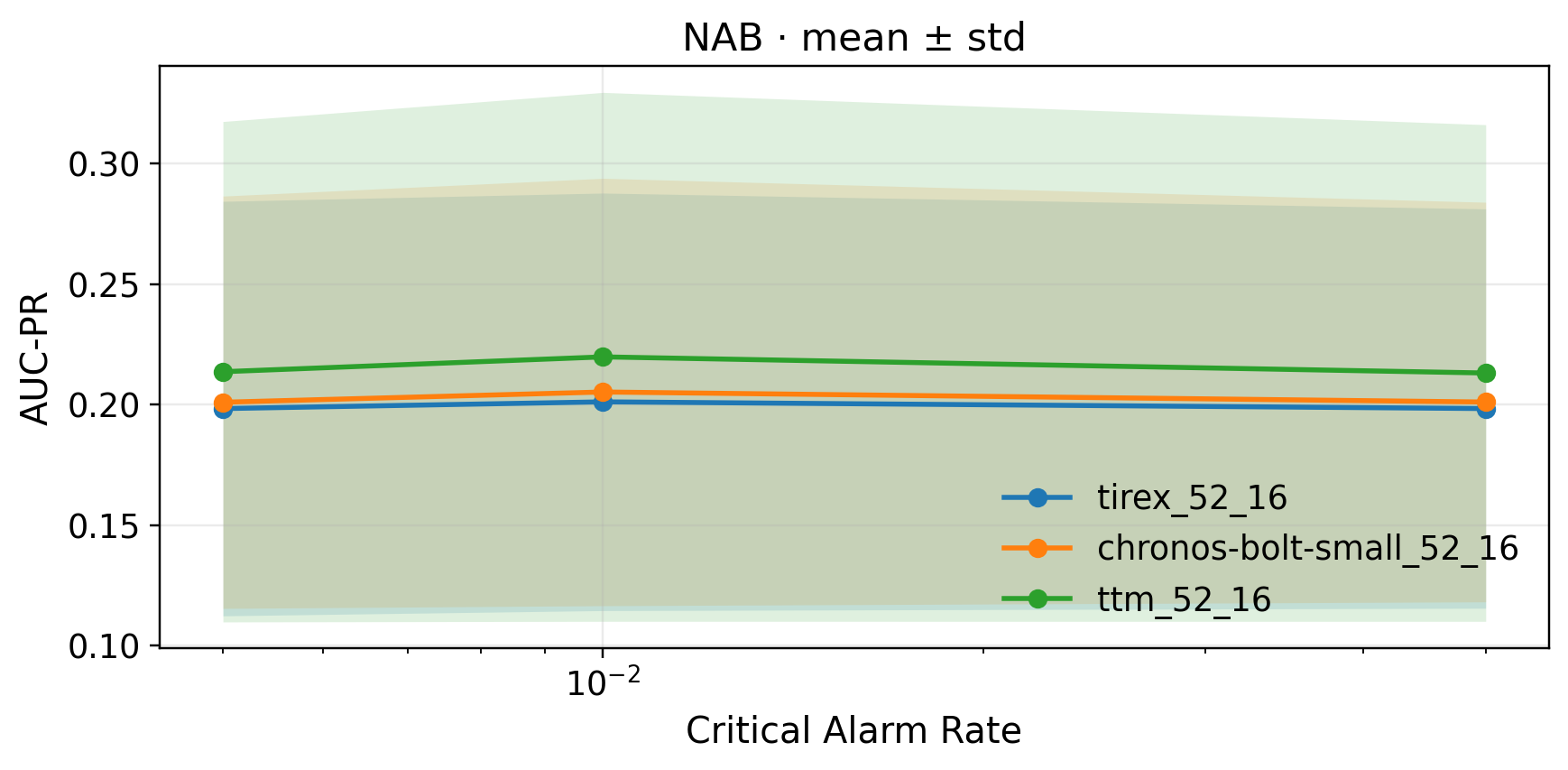}}
  \subfigure[NAB — VUS-PR]{\includegraphics[width=0.24\textwidth]{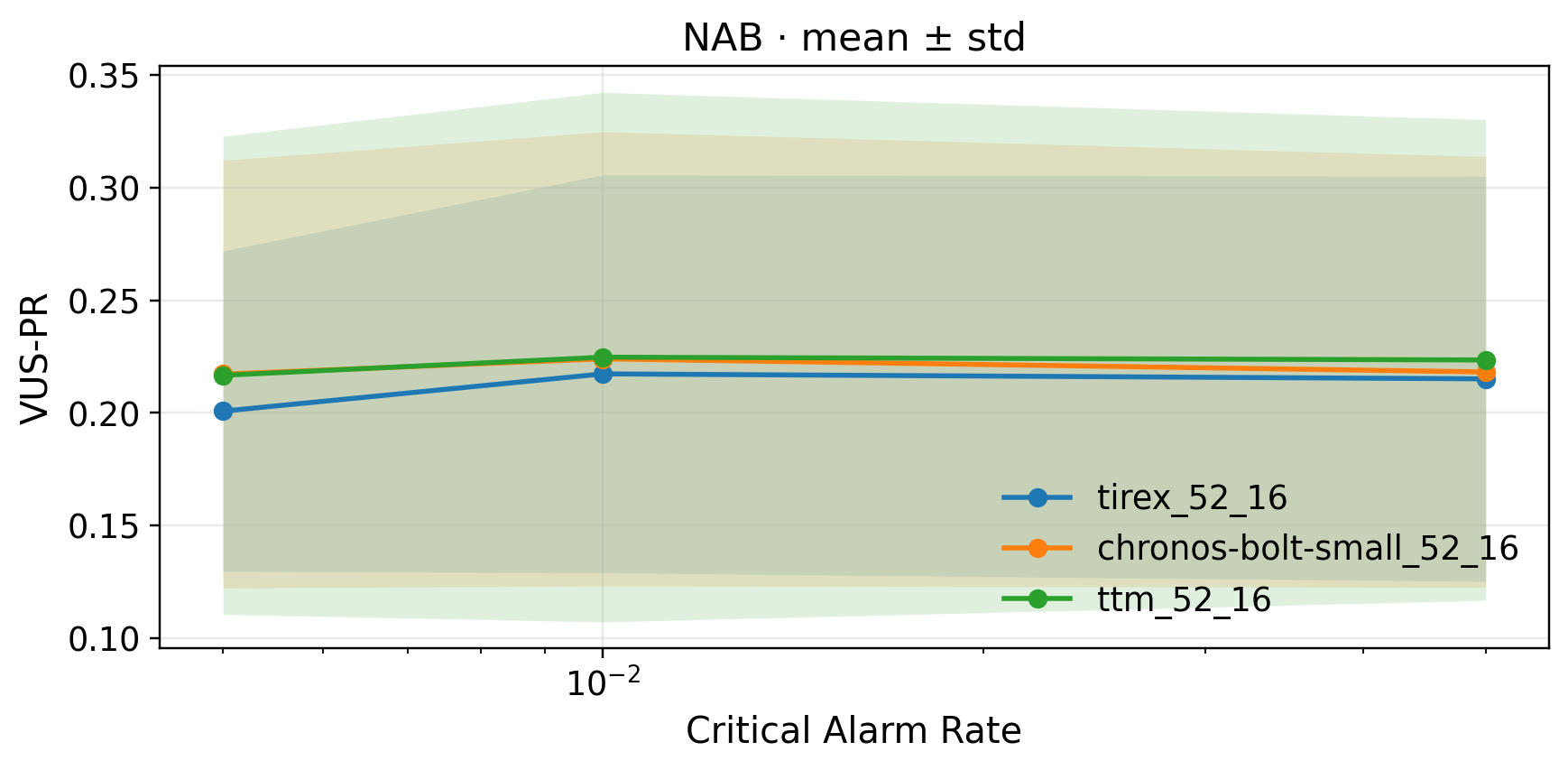}}

  \subfigure[NEK — PA-F1]{\includegraphics[width=0.24\textwidth]{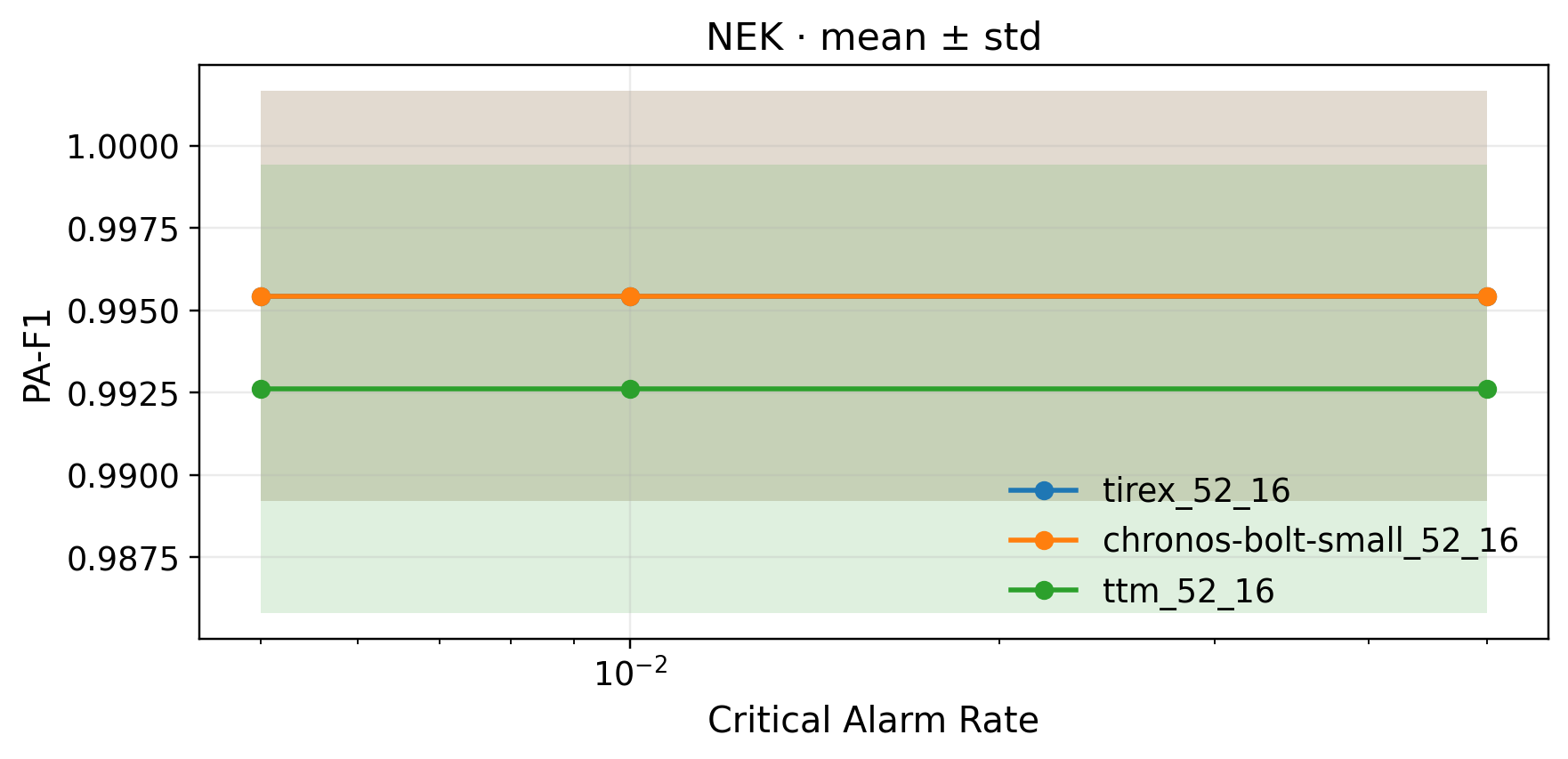}}
  \subfigure[NEK — Affiliation-F]{\includegraphics[width=0.24\textwidth]{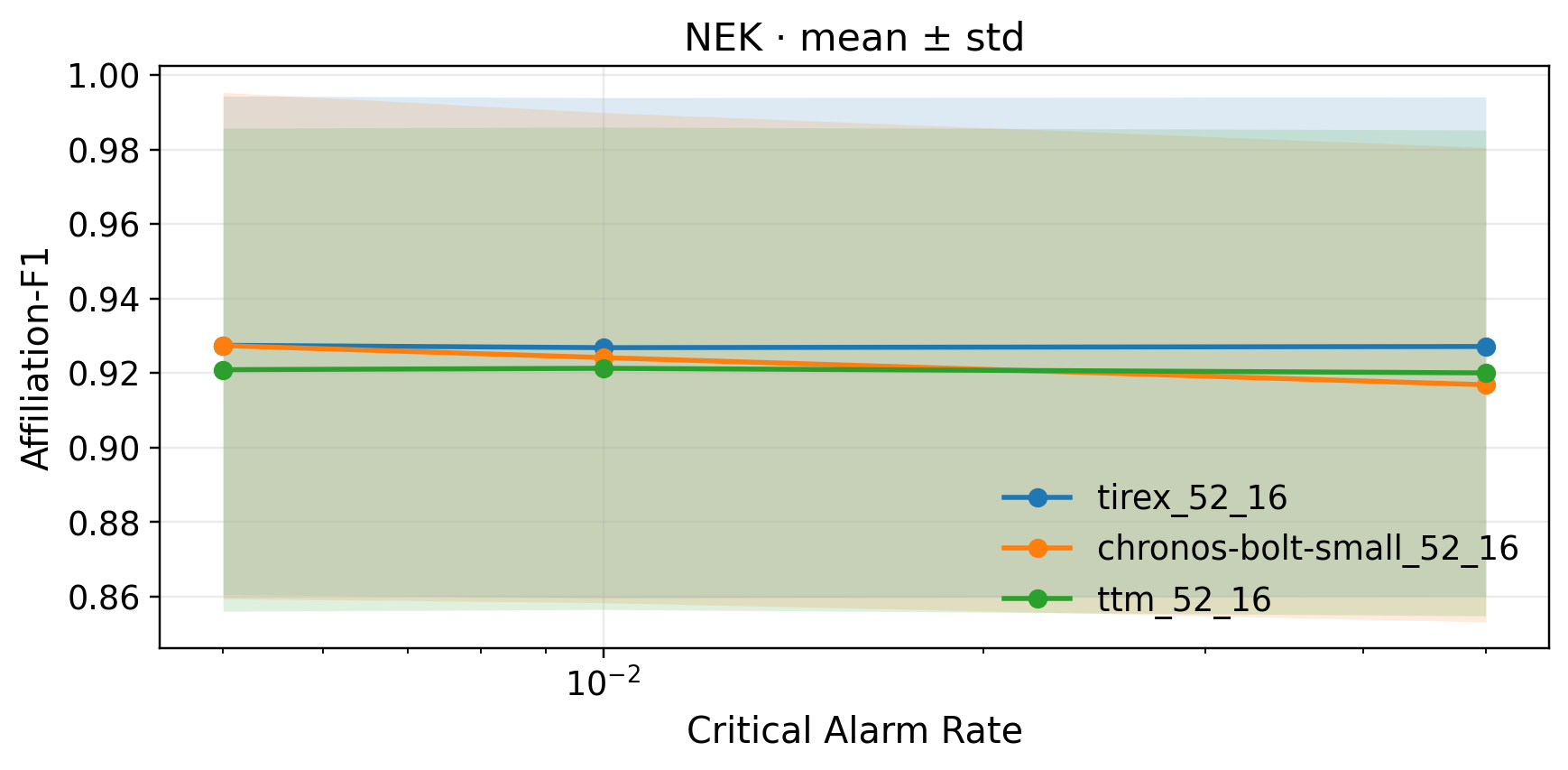}}
  \subfigure[NEK — AUC-PR]{\includegraphics[width=0.24\textwidth]{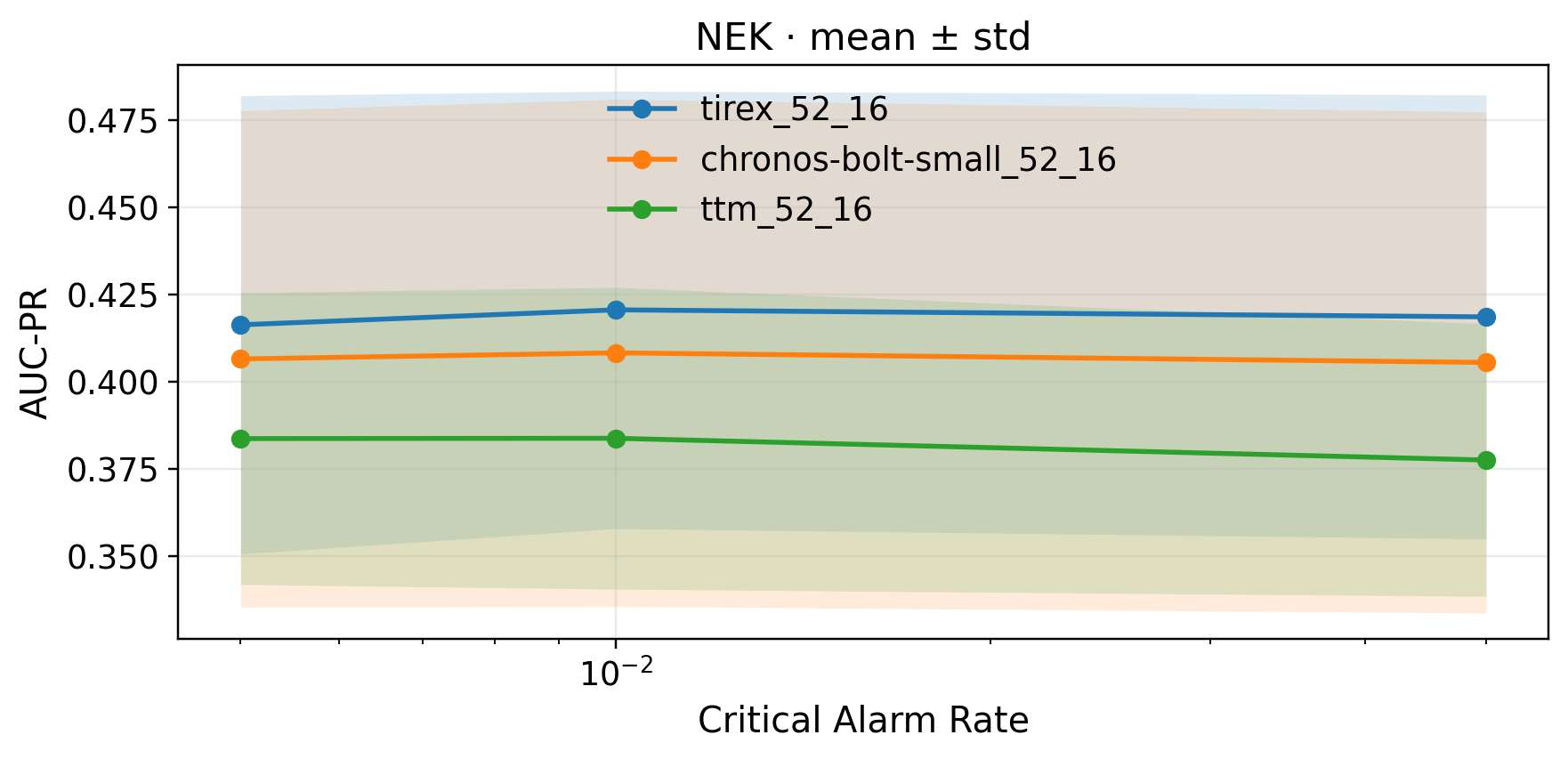}}
  \subfigure[NEK — VUS-PR]{\includegraphics[width=0.24\textwidth]{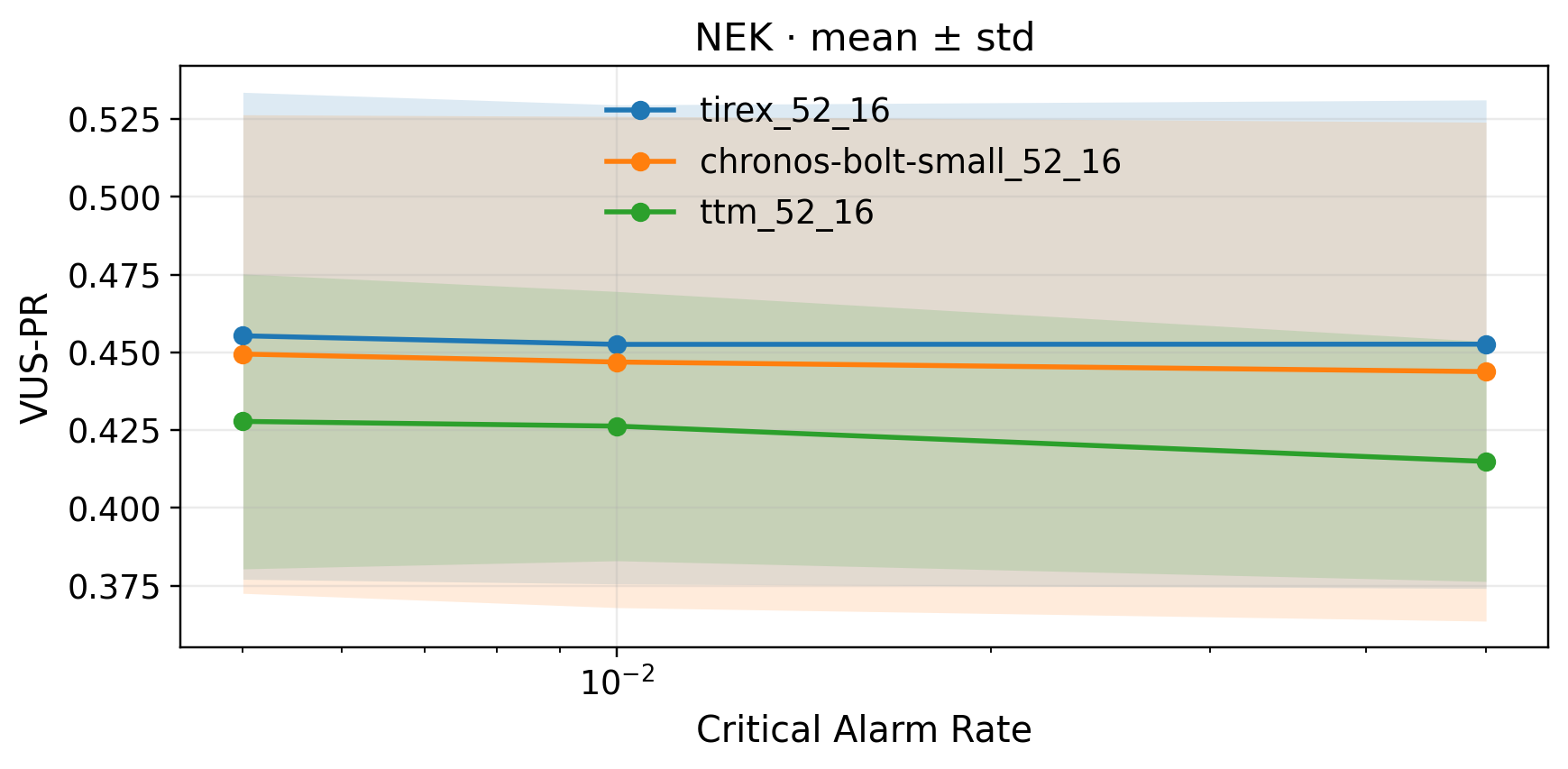}}

  \subfigure[MSL — PA-F1]{\includegraphics[width=0.24\textwidth]{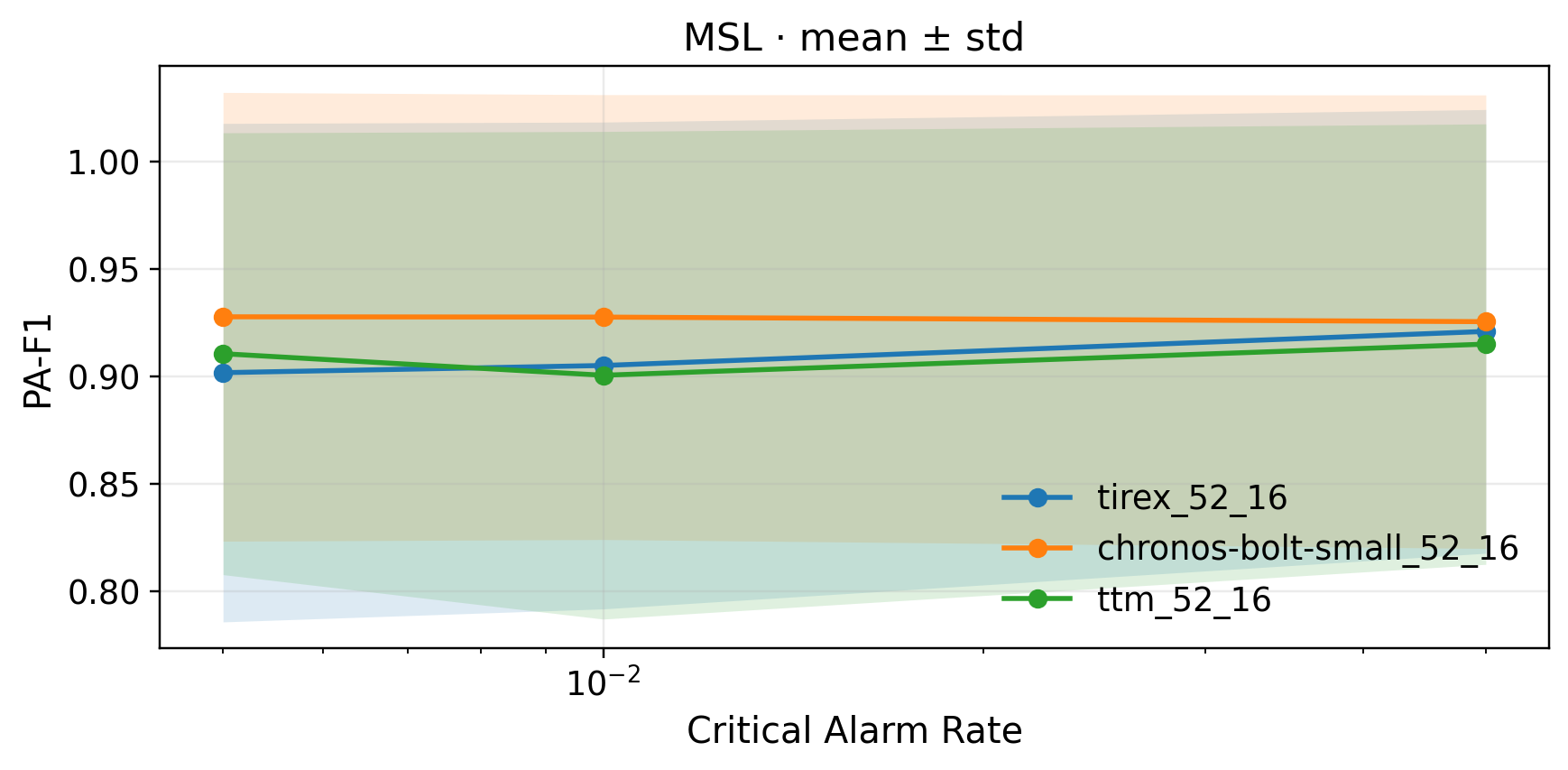}}
  \subfigure[MSL — Affiliation-F]{\includegraphics[width=0.24\textwidth]{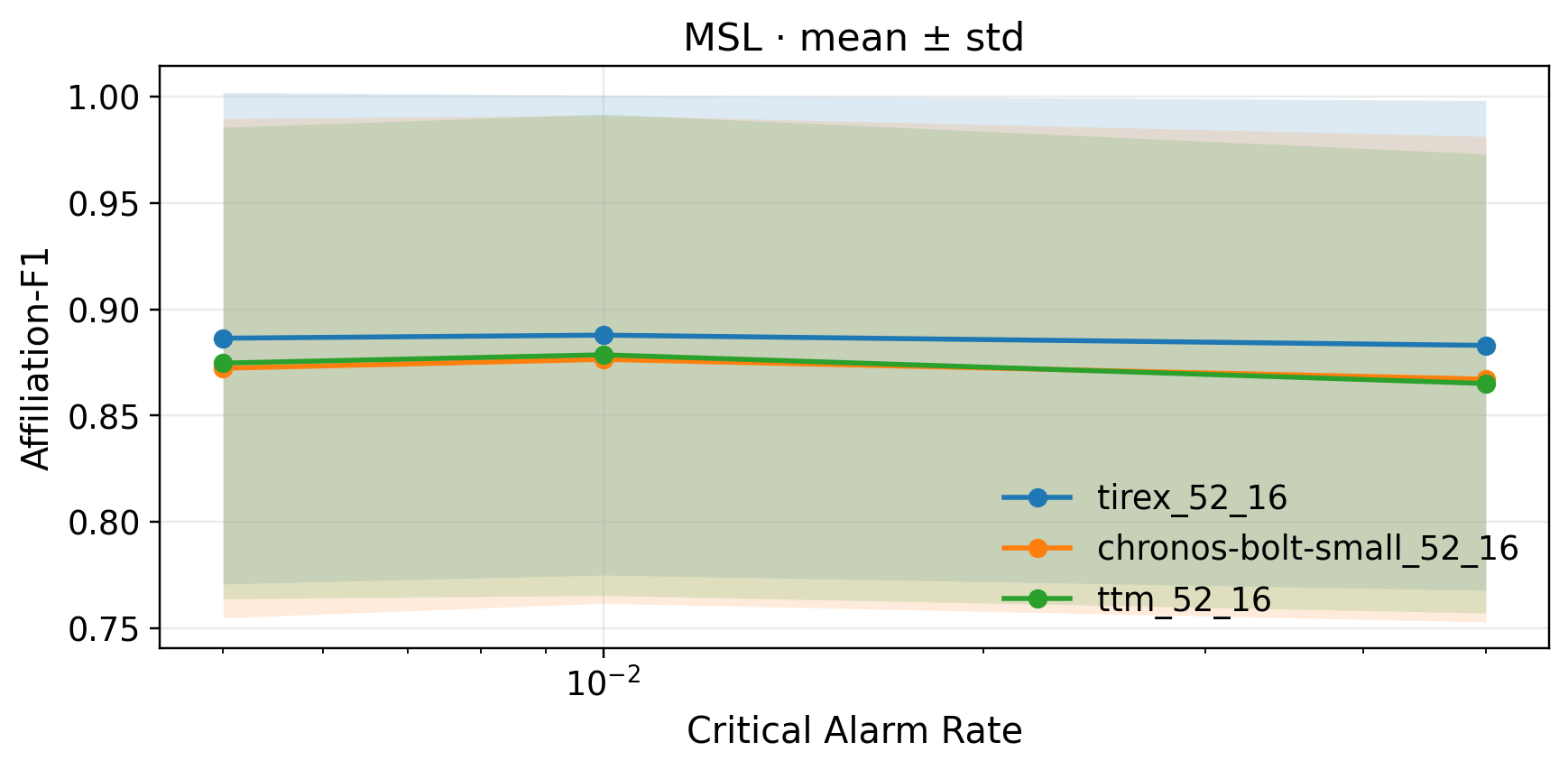}}
  \subfigure[MSL — AUC-PR]{\includegraphics[width=0.24\textwidth]{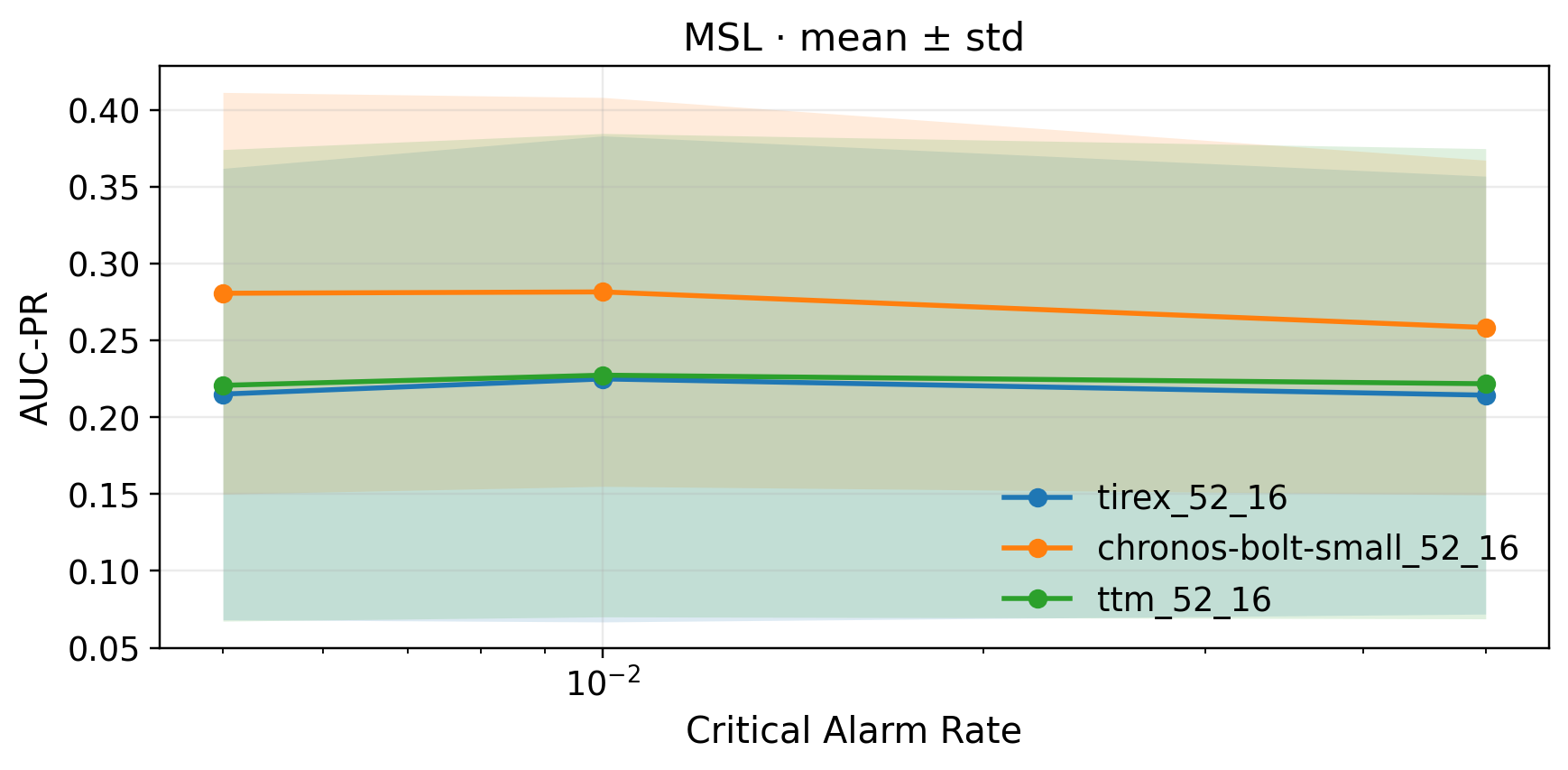}}
  \subfigure[MSL — VUS-PR]{\includegraphics[width=0.24\textwidth]{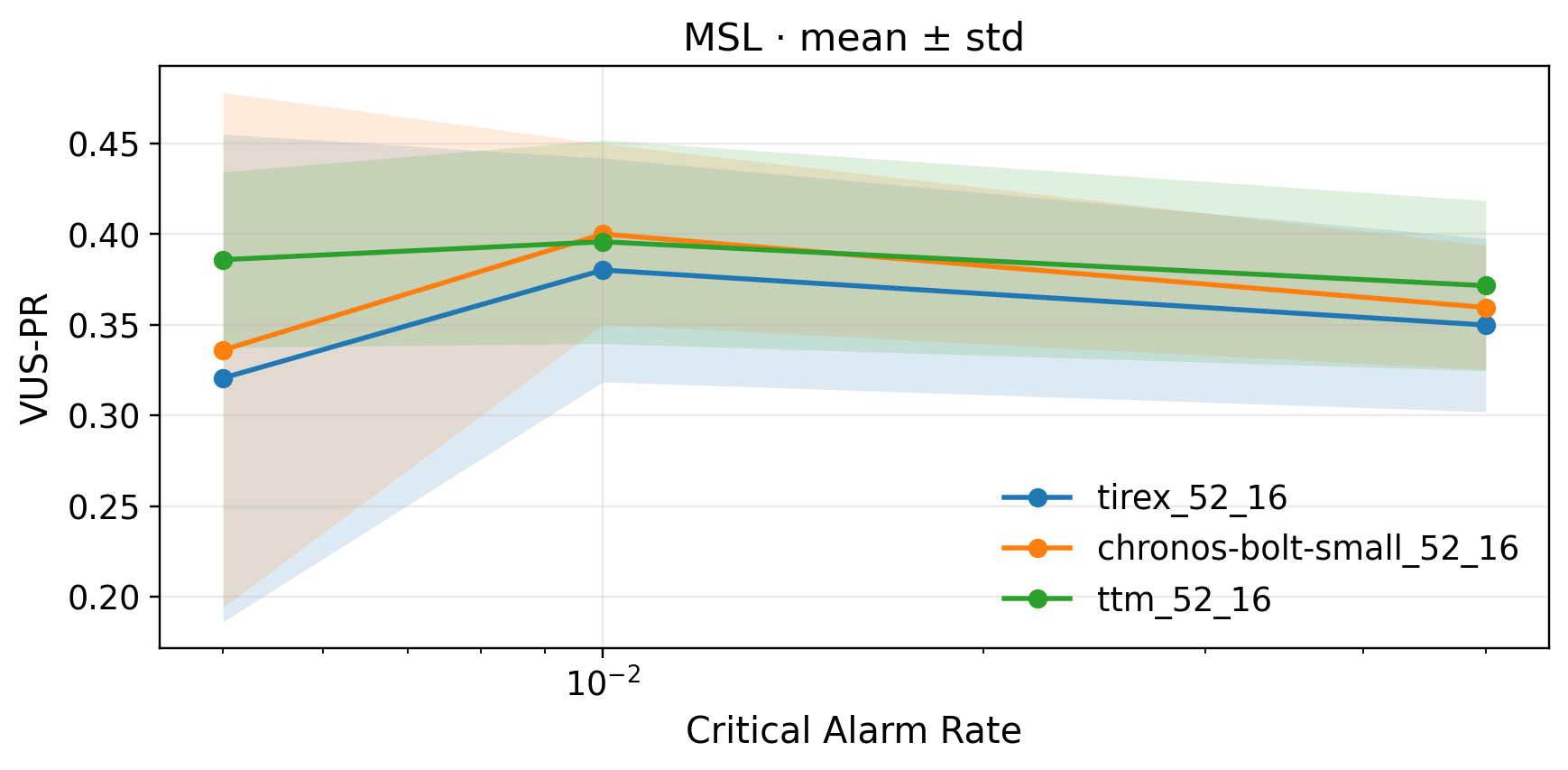}}

  \subfigure[YAHOO — PA-F1]{\includegraphics[width=0.24\textwidth]{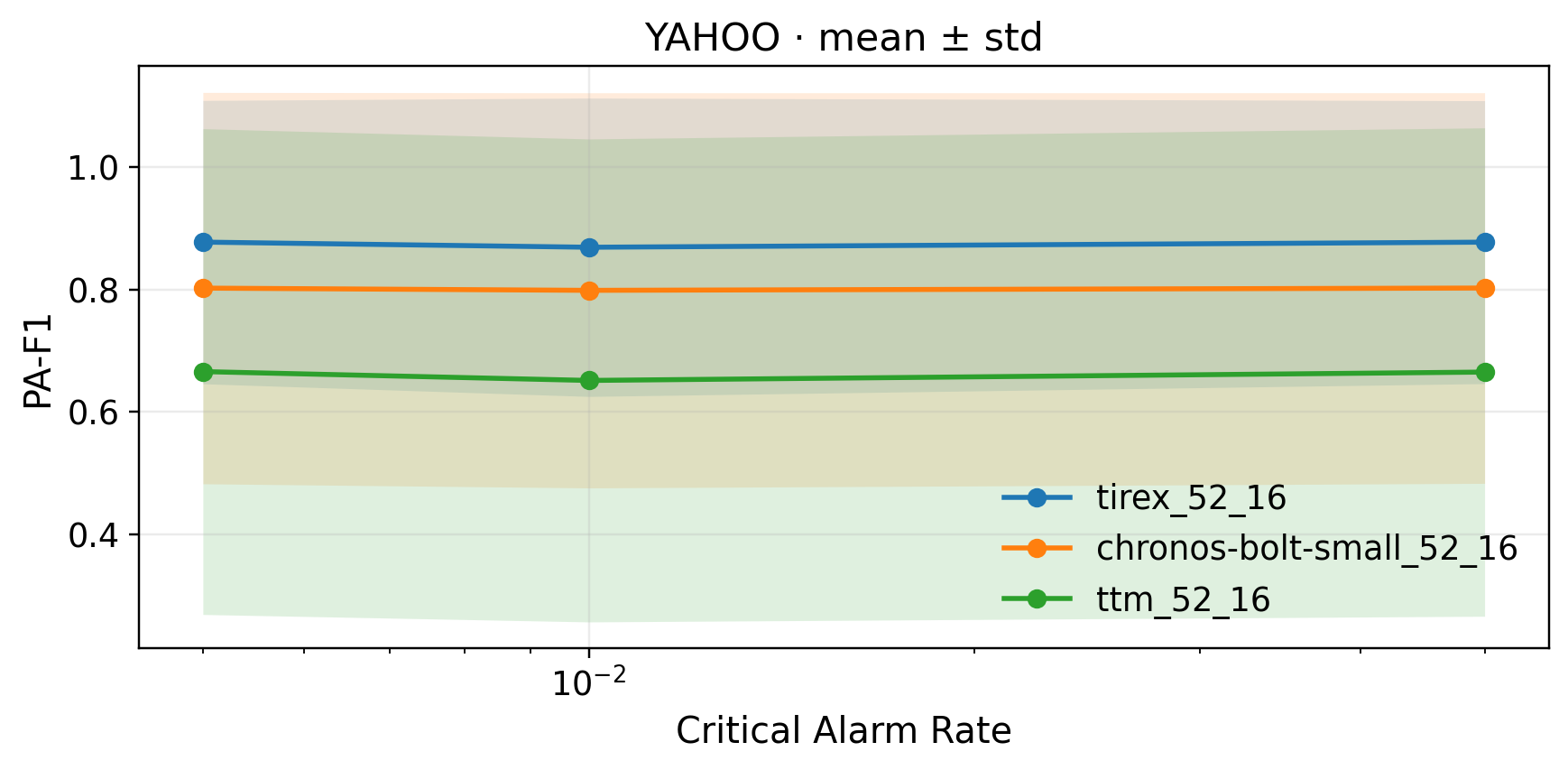}}
  \subfigure[YAHOO-Affiliation-F]{\includegraphics[width=0.24\textwidth]{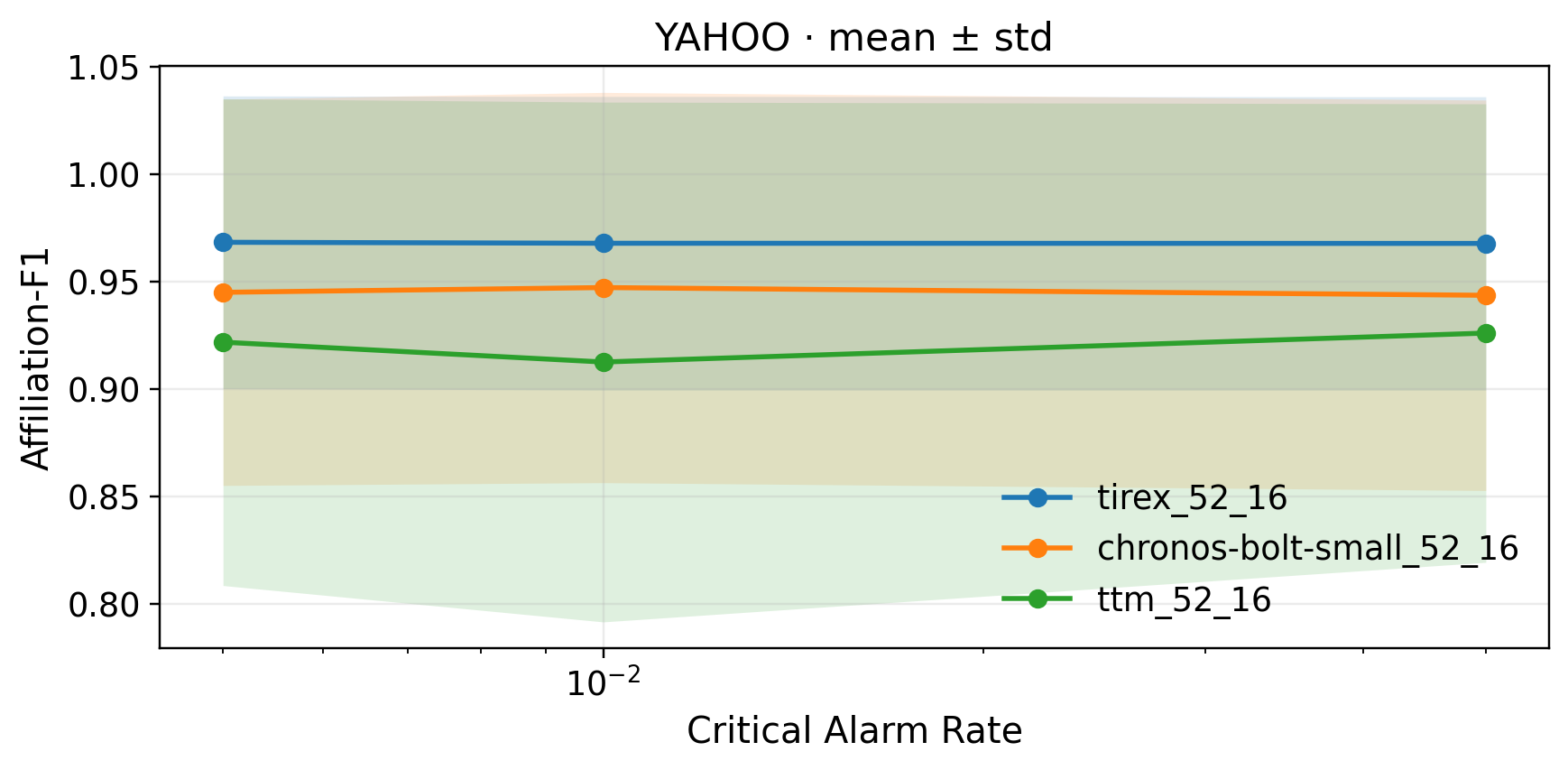}}
  \subfigure[YAHOO — AUC-PR]{\includegraphics[width=0.24\textwidth]{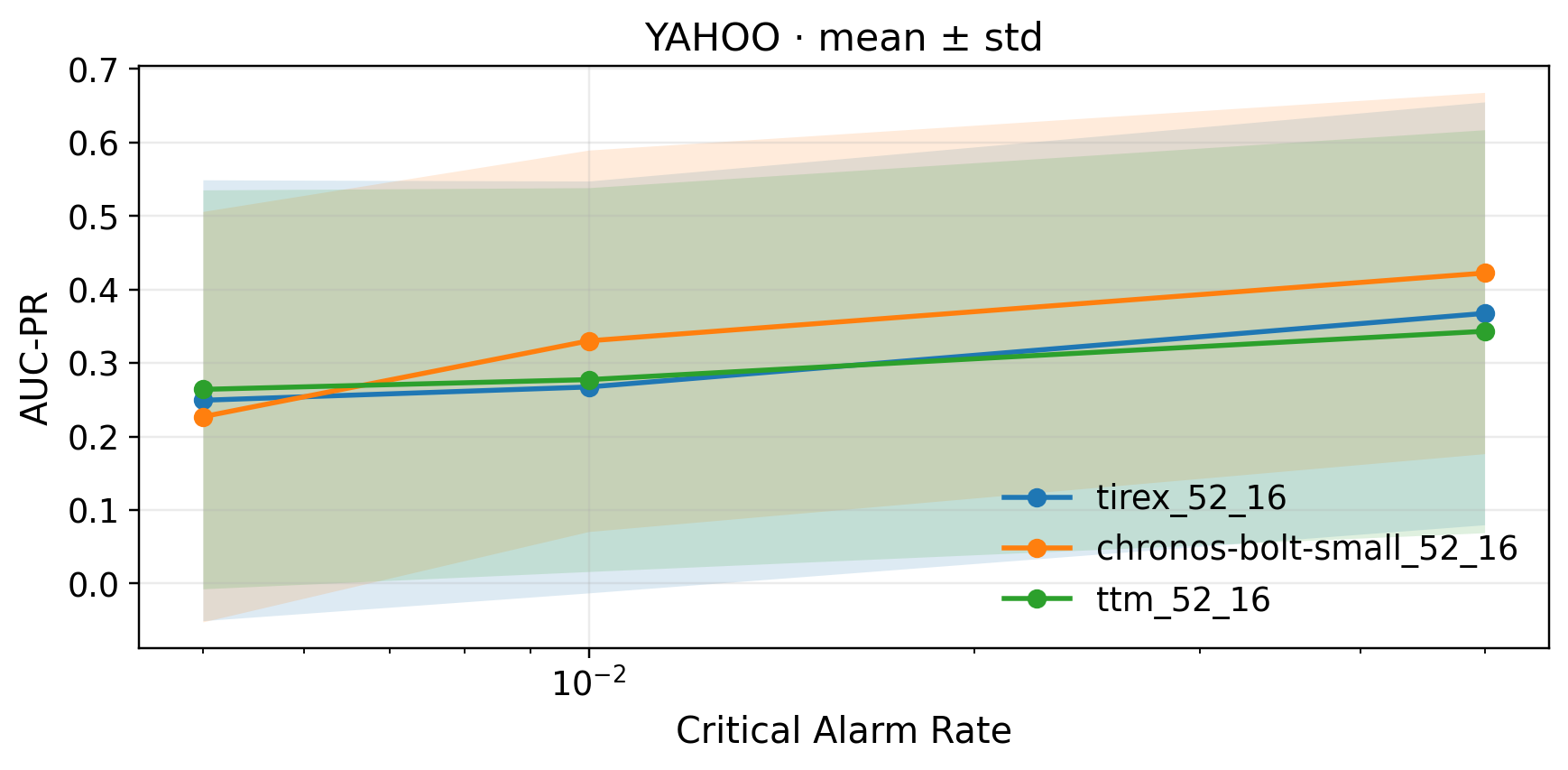}}
  \subfigure[YAHOO — VUS-PR]{\includegraphics[width=0.24\textwidth]{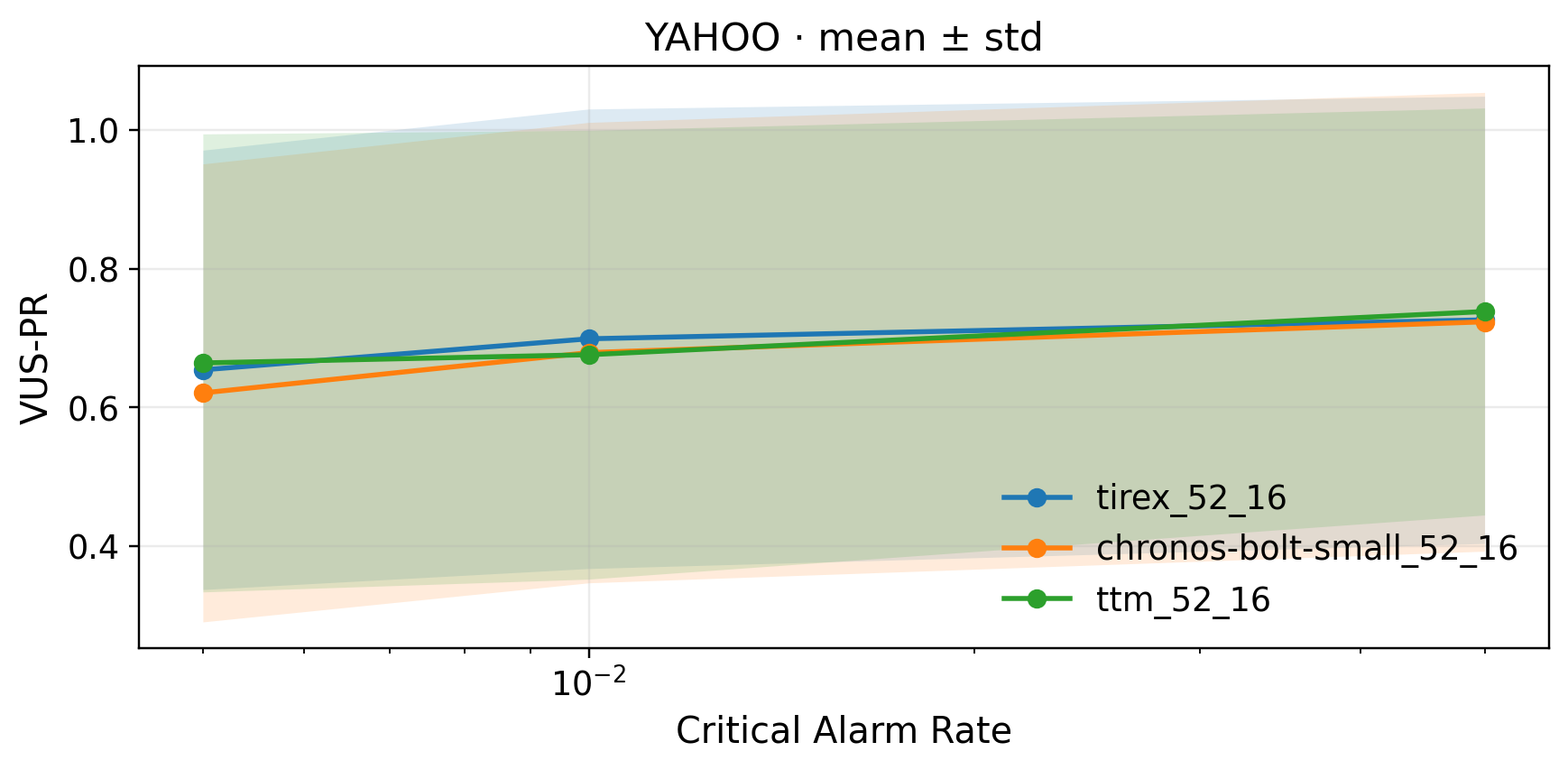}}



  \caption{{Performance of $\mathcal{W}_1$-ACAS when aggregating different critical alarm rate $\alpha_c$. Rows correspond to datasets (NAB, NEK, MSL, YAHOO, Stock, WSD) and columns to metrics (PA-F1, Affiliation-F, AUC-PR, VUS-PR).}}
  \label{fig:w1-acas-steps-grid-alarm}
\end{figure*}

%% file: ICLR2026/sections/appendix_tables.tex
\begin{table}[ht!]
\scriptsize
\begin{tabular}{llllllllll}
\toprule
Dataset & Forecaster & AD Model & PA-F1 \(\uparrow\) & Affiliation-F \(\uparrow\) & FPR \(\downarrow\) & CalErr \(\downarrow\) & AUC-PR \(\uparrow\) & VUC-PR \(\uparrow\) \\
\midrule
YAHOO & -& KShapeAD & 0.523 ± 0.430 & 0.860 ± 0.151 & 0.359 ± 0.437 & 0.119 ± 0.183 & 0.036 ± 0.110 & 0.220 ± 0.225 \\
YAHOO & -& POLY & 0.102 ± 0.217 & 0.831 ± 0.126 & 0.387 ± 0.367 & 0.244 ± 0.240 & 0.037 ± 0.127 & 0.139 ± 0.125 \\
YAHOO & -& Sub-KNN & 0.161 ± 0.273 & 0.895 ± 0.109 & 0.158 ± 0.222 & 0.157 ± 0.161 & 0.016 ± 0.043 & 0.260 ± 0.197 \\
YAHOO & -& Sub-PCA & 0.112 ± 0.261 & 0.750 ± 0.115 & 0.677 ± 0.410 & 0.099 ± 0.163 & 0.056 ± 0.134 & 0.125 ± 0.199 \\
YAHOO & - & SAND & 0.398 ± 0.416 & 0.837 ± 0.147 & 0.409 ± 0.434 & 0.097 ± 0.114 & 0.024 ± 0.071 & 0.198 ± 0.180 \\
\midrule
YAHOO & - & CNN* & 0.596 ± 0.438 & 0.853 ± 0.146 & 0.242 ± 0.407 & 0.240 ± 0.321 & 0.053 ± 0.147 & 0.160 ± 0.258 \\
YAHOO & - & OmniAnomaly* & 0.272 ± 0.381 & 0.791 ± 0.136 & 0.384 ± 0.446 & 0.313 ± 0.318 & 0.195 ± 0.255 & 0.351 ± 0.378 \\
YAHOO & - & USAD* & 0.113 ± 0.287 & 0.736 ± 0.098 & 0.610 ± 0.381 & 0.154 ± 0.187 & 0.068 ± 0.160 & 0.201 ± 0.288 \\
YAHOO & - & MOMENT\_ZS & 0.134 ± 0.222 & 0.832 ± 0.121 & 0.215 ± 0.325 & 0.195 ± 0.193 & 0.086 ± 0.188 & 0.233 ± 0.235 \\
\midrule
YAHOO & Chronos & $\mathcal{W}_1\textsc{-ACAS}$ & 0.798 ± 0.323 & 0.947 ± 0.091 & 0.074 ± 0.253 & 0.007 ± 0.017 & \textbf{0.330 ± 0.259} & 0.679 ± 0.332 \\
YAHOO & Chronos & conformal & 0.652 ± 0.361 & 0.936 ± 0.091 & 0.088 ± 0.258 & 0.015 ± 0.028 & 0.147 ± 0.224 & 0.485 ± 0.347 \\
YAHOO & Chronos & gaussian & 0.317 ± 0.284 & 0.846 ± 0.098 & 0.123 ± 0.265 & 0.044 ± 0.066 & 0.028 ± 0.100 & 0.511 ± 0.345 \\
\midrule
YAHOO & Tirex & $\mathcal{W}_1\textsc{-ACAS}$ & \textbf{0.869 ± 0.244} & \textbf{0.968 ± 0.068} & \textbf{0.069 ± 0.253} & \textbf{0.003 ± 0.007} & 0.267 ± 0.280 & \textbf{0.699 ± 0.331} \\
YAHOO & Tirex & conformal & 0.730 ± 0.310 & 0.928 ± 0.091 & 0.074 ± 0.252 & 0.009 ± 0.015 & 0.176 ± 0.259 & 0.559 ± 0.317 \\
YAHOO & Tirex & gaussian & 0.302 ± 0.269 & 0.825 ± 0.101 & 0.105 ± 0.252 & 0.041 ± 0.062 & 0.030 ± 0.114 & 0.546 ± 0.310 \\
\midrule
YAHOO & TTM & $\mathcal{W}_1\textsc{-ACAS}$ & 0.651 ± 0.395 & 0.912 ± 0.121 & 0.141 ± 0.343 & 0.034 ± 0.057 & 0.277 ± 0.261 & 0.676 ± 0.324 \\
YAHOO & TTM & conformal & 0.607 ± 0.413 & 0.916 ± 0.108 & 0.113 ± 0.273 & 0.052 ± 0.082 & 0.172 ± 0.230 & 0.611 ± 0.350 \\
YAHOO & TTM & gaussian & 0.417 ± 0.334 & 0.870 ± 0.113 & 0.124 ± 0.276 & 0.050 ± 0.104 & 0.028 ± 0.099 & 0.560 ± 0.331 \\
\midrule
\midrule
NEK & -& KShapeAD & 0.602 ± 0.292 & 0.708 ± 0.043 & 0.807 ± 0.284 & 0.077 ± 0.118 & 0.216 ± 0.179 & 0.152 ± 0.138 \\
NEK & -& POLY & 0.848 ± 0.149 & 0.936 ± 0.066 & 0.073 ± 0.058 & 0.478 ± 0.135 & 0.063 ± 0.073 & 0.616 ± 0.162 \\
NEK & -& Sub-KNN & 0.738 ± 0.307 & 0.779 ± 0.098 & 0.561 ± 0.451 & 0.054 ± 0.086 & 0.172 ± 0.068 & 0.321 ± 0.131 \\
NEK & -& Sub-PCA & 0.933 ± 0.107 & \textbf{0.980 ± 0.022} & 0.041 ± 0.068 & 0.393 ± 0.190 & 0.007 ± 0.013 & 0.705 ± 0.230 \\
NEK & - & SAND & 0.718 ± 0.340 & 0.829 ± 0.105 & 0.312 ± 0.338 & 0.201 ± 0.135 & 0.325 ± 0.196 & 0.214 ± 0.203 \\
\midrule
NEK & - & CNN* & 0.996 ± 0.006 & 0.965 ± 0.078 & \textbf{0.000 ± 0.000} & 0.859 ± 0.046 & \textbf{0.438 ± 0.197} & 0.730 ± 0.218 \\
NEK & - & OmniAnomaly* & \textbf{0.998 ± 0.005} & 0.968 ± 0.077 & 0.001 ± 0.001 & 0.875 ± 0.033 & 0.195 ± 0.189 & \textbf{0.872 ± 0.132} \\
NEK & - & USAD* & 0.785 ± 0.295 & 0.933 ± 0.058 & 0.179 ± 0.162 & 0.440 ± 0.124 & 0.006 ± 0.015 & 0.555 ± 0.174 \\
NEK & - & MOMENT\_ZS & 0.849 ± 0.200 & 0.942 ± 0.028 & 0.125 ± 0.095 & 0.496 ± 0.165 & 0.046 ± 0.041 & 0.583 ± 0.138 \\
\midrule
NEK & Chronos & $\mathcal{W}_1\textsc{-ACAS}$ & \underline{0.995 ± 0.006} & 0.924 ± 0.066 & \underline{0.003 ± 0.005} & \textbf{0.004 ± 0.003} & 0.408 ± 0.073 & 0.447 ± 0.079 \\
NEK & Chronos & conformal & 0.979 ± 0.012 & 0.934 ± 0.067 & 0.007 ± 0.006 & 0.007 ± 0.004 & 0.418 ± 0.104 & 0.490 ± 0.092 \\
NEK & Chronos & gaussian & 0.890 ± 0.021 & 0.860 ± 0.069 & 0.047 ± 0.028 & 0.045 ± 0.025 & 0.347 ± 0.054 & {0.519 ± 0.093} \\
\midrule
NEK & Tirex & $\mathcal{W}_1\textsc{-ACAS}$ & 0.995 ± 0.006 & 0.927 ± 0.067 & 0.006 ± 0.015 & 0.005 ± 0.003 & \underline{0.421 ± 0.063} & 0.453 ± 0.077 \\
NEK & Tirex & conformal & 0.971 ± 0.011 & 0.934 ± 0.066 & 0.011 ± 0.007 & 0.009 ± 0.004 & 0.421 ± 0.097 & 0.496 ± 0.097 \\
NEK & Tirex & gaussian & 0.890 ± 0.027 & 0.865 ± 0.064 & 0.044 ± 0.021 & 0.043 ± 0.021 & 0.354 ± 0.056 & 0.513 ± 0.099 \\
\midrule
NEK & TTM & $\mathcal{W}_1\textsc{-ACAS}$ & 0.993 ± 0.007 & 0.921 ± 0.065 & 0.004 ± 0.007 & 0.005 ± 0.003 & 0.384 ± 0.043 & 0.426 ± 0.043 \\
NEK & TTM & conformal & 0.977 ± 0.016 & 0.931 ± 0.067 & 0.008 ± 0.007 & 0.008 ± 0.005 & 0.417 ± 0.047 & 0.471 ± 0.057 \\
NEK & TTM & gaussian & 0.895 ± 0.012 & 0.871 ± 0.068 & 0.059 ± 0.045 & 0.040 ± 0.019 & 0.337 ± 0.060 & 0.501 ± 0.064 \\

\bottomrule
\end{tabular}
\caption{\textbf{Performance Summary per datasets.}
Entries indicate the mean $\pm$ standard deviation computed by averaging within each dataset group. Higher numbers are better for PA-F1, Affiliation-F, AUC-PR, VUS-PR; lower numbers are better for FPR, and calibration error (CalErr). Methods marked with * denote deep learning semi-supervised approaches; the best overall method is shown in \textbf{bold}, and the best non–semi-supervised method is \underline{underlined} when different from the bold one.}
\label{tab:placeholder}
\end{table}

\begin{table}[]
\scriptsize
\begin{tabular}{llllllllll}
\toprule
Dataset & Forecaster & AD Model & PA-F1 \(\uparrow\) & Affiliation-F \(\uparrow\) & FPR \(\downarrow\) & CalErr \(\downarrow\) & AUC-PR \(\uparrow\) & VUC-PR \(\uparrow\) \\
\midrule
MSL & -& KShapeAD & 0.854 ± 0.207 & 0.915 ± 0.113 & 0.116 ± 0.154 & 0.188 ± 0.131 & 0.108 ± 0.119 & 0.260 ± 0.163 \\
MSL & -& POLY & 0.619 ± 0.330 & 0.881 ± 0.114 & 0.248 ± 0.339 & 0.076 ± 0.108 & 0.077 ± 0.106 & 0.353 ± 0.187 \\
MSL & -& Sub-KNN & 0.685 ± 0.379 & 0.835 ± 0.124 & 0.293 ± 0.411 & 0.137 ± 0.099 & 0.132 ± 0.164 & 0.179 ± 0.153 \\
MSL & -& Sub-PCA & 0.683 ± 0.354 & 0.882 ± 0.110 & 0.175 ± 0.248 & 0.145 ± 0.185 & 0.056 ± 0.071 & 0.371 ± 0.329 \\
MSL & - & SAND & 0.655 ± 0.328 & 0.877 ± 0.122 & 0.242 ± 0.251 & 0.176 ± 0.153 & 0.064 ± 0.068 & 0.303 ± 0.179 \\

\midrule
MSL & - & CNN* & 0.826 ± 0.225 & 0.885 ± 0.099 & 0.096 ± 0.217 & 0.460 ± 0.407 & 0.105 ± 0.105 & 0.308 ± 0.264 \\
MSL & - & OmniAnomaly* & 0.818 ± 0.257 & 0.879 ± 0.106 & 0.038 ± 0.063 & 0.588 ± 0.409 & 0.121 ± 0.133 & 0.344 ± 0.262 \\
MSL & - & USAD* & 0.667 ± 0.349 & 0.881 ± 0.108 & 0.133 ± 0.197 & 0.384 ± 0.288 & 0.060 ± 0.095 & 0.415 ± 0.389 \\
MSL & - & MOMENT\_ZS & 0.799 ± 0.300 & \textbf{0.905 ± 0.128} & 0.151 ± 0.375 & 0.429 ± 0.328 & 0.134 ± 0.093 & \textbf{0.501 ± 0.290} \\

\midrule
MSL & Chronos & $\mathcal{W}_1\textsc{-ACAS}$ & \textbf{0.928 ± 0.104} & 0.876 ± 0.115 & 0.033 ± 0.062 & 0.022 ± 0.027 & 0.282 ± 0.127 & 0.400 ± 0.050 \\
MSL & Chronos & conformal & 0.829 ± 0.318 & 0.813 ± 0.122 & 0.310 ± 0.472 & 0.159 ± 0.370 & 0.308 ± 0.159 & 0.306 ± 0.175 \\
MSL & Chronos & gaussian & 0.854 ± 0.126 & 0.842 ± 0.101 & 0.180 ± 0.363 & 0.074 ± 0.151 & 0.262 ± 0.124 & 0.368 ± 0.157 \\
\midrule
MSL & Tirex & $\mathcal{W}_1\textsc{-ACAS}$ & 0.905 ± 0.113 & \underline{0.888 ± 0.113} & 0.152 ± 0.374 & \textbf{0.017 ± 0.020} & 0.225 ± 0.158 & 0.380 ± 0.062 \\
MSL & Tirex & conformal & 0.826 ± 0.315 & 0.816 ± 0.121 & 0.312 ± 0.471 & 0.158 ± 0.371 & 0.226 ± 0.160 & 0.299 ± 0.171 \\
MSL & Tirex & gaussian & 0.856 ± 0.124 & 0.841 ± 0.100 & 0.187 ± 0.362 & 0.131 ± 0.182 & 0.267 ± 0.118 & 0.378 ± 0.170 \\
\midrule
MSL & TTM & $\mathcal{W}_1\textsc{-ACAS}$ & 0.901 ± 0.114 & 0.879 ± 0.113 & \textbf{0.023 ± 0.039} & \textbf{0.017 ± 0.026} & 0.227 ± 0.157 & 0.396 ± 0.056 \\
MSL & TTM & conformal & 0.803 ± 0.308 & 0.812 ± 0.120 & 0.319 ± 0.467 & 0.165 ± 0.368 & 0.243 ± 0.196 & 0.396 ± 0.181 \\
MSL & TTM & gaussian & 0.855 ± 0.118 & 0.849 ± 0.104 & 0.180 ± 0.364 & 0.074 ± 0.133 & 0.286 ± 0.174 & \underline{0.416 ± 0.122} \\

\midrule
\midrule
NAB & -& KShapeAD & 0.835 ± 0.222 & 0.833 ± 0.141 & 0.341 ± 0.424 & 0.195 ± 0.250 & 0.123 ± 0.162 & 0.272 ± 0.220 \\
NAB & -& POLY & 0.863 ± 0.203 & 0.900 ± 0.109 & 0.136 ± 0.251 & 0.232 ± 0.248 & 0.087 ± 0.073 & {0.322 ± 0.189} \\
NAB & -& Sub-KNN & 0.810 ± 0.257 & 0.819 ± 0.124 & 0.272 ± 0.383 & 0.273 ± 0.318 & 0.153 ± 0.150 & 0.304 ± 0.283 \\
NAB & -& Sub-PCA & 0.905 ± 0.185 & 0.923 ± 0.100 & 0.082 ± 0.223 & 0.341 ± 0.320 & 0.199 ± 0.245 & 0.427 ± 0.286 \\
NAB & - & SAND & 0.785 ± 0.245 & 0.809 ± 0.130 & 0.340 ± 0.422 & 0.151 ± 0.140 & 0.130 ± 0.131 & 0.296 ± 0.207 \\
\midrule

NAB & - & CNN* & 0.982 ± 0.054 & \textbf{0.937 ± 0.077} & \textbf{0.008 ± 0.034} & 0.468 ± 0.420 & 0.194 ± 0.115 & 0.260 ± 0.146 \\
NAB & - & OmniAnomaly* & 0.977 ± 0.082 & 0.925 ± 0.091 & 0.068 ± 0.231 & 0.506 ± 0.366 & 0.201 ± 0.071 & 0.274 ± 0.139 \\
NAB & - & USAD* & 0.927 ± 0.139 & 0.926 ± 0.101 & 0.123 ± 0.263 & 0.473 ± 0.314 & 0.200 ± 0.206 & \textbf{0.445 ± 0.237} \\
NAB & - & MOMENT\_ZS & 0.958 ± 0.115 & 0.931 ± 0.103 & 0.129 ± 0.315 & 0.490 ± 0.341 & 0.220 ± 0.218 & 0.407 ± 0.216 \\

\midrule
NAB & Chronos & $\mathcal{W}_1\textsc{-ACAS}$ & 0.983 ± 0.044 & 0.855 ± 0.092 & 0.054 ± 0.208 & 0.012 ± 0.028 & 0.205 ± 0.089 & 0.224 ± 0.101 \\
NAB & Chronos & conformal & 0.978 ± 0.057 & 0.880 ± 0.089 & \underline{0.048 ± 0.208} & 0.013 ± 0.011 & 0.205 ± 0.094 & 0.232 ± 0.108 \\
NAB & Chronos & gaussian & 0.941 ± 0.051 & 0.800 ± 0.103 & 0.149 ± 0.328 & 0.077 ± 0.097 & 0.201 ± 0.094 & 0.223 ± 0.098 \\
\midrule
NAB & Tirex & $\mathcal{W}_1\textsc{-ACAS}$ & \textbf{0.985 ± 0.030} & 0.851 ± 0.095 & 0.049 ± 0.208 & 0.011 ± 0.027 & 0.201 ± 0.087 & 0.217 ± 0.088 \\
NAB & Tirex & conformal & 0.977 ± 0.057 & 0.876 ± 0.090 & 0.052 ± 0.207 & 0.012 ± 0.017 & 0.201 ± 0.097 & 0.228 ± 0.096 \\
NAB & Tirex & gaussian & 0.935 ± 0.059 & 0.792 ± 0.095 & 0.154 ± 0.331 & 0.085 ± 0.105 & 0.195 ± 0.097 & 0.218 ± 0.086 \\
\midrule
NAB & TTM & $\mathcal{W}_1\textsc{-ACAS}$ & 0.980 ± 0.057 & 0.858 ± 0.093 & 0.050 ± 0.207 & \textbf{0.010 ± 0.019} & \textbf{0.220 ± 0.110} & 0.225 ± 0.118 \\
NAB & TTM & conformal & 0.976 ± 0.058 & 0.881 ± 0.087 & 0.050 ± 0.207 & 0.008 ± 0.011 & 0.217 ± 0.115 & 0.231 ± 0.125 \\
NAB & TTM & gaussian & 0.945 ± 0.064 & 0.815 ± 0.103 & 0.140 ± 0.310 & 0.078 ± 0.109 & 0.203 ± 0.104 & 0.235 ± 0.124 \\

\bottomrule
\end{tabular}
\caption{\textbf{Performance Summary per datasets.}
Entries indicate the mean $\pm$ standard deviation computed by averaging within each dataset group. Higher numbers are better for PA-F1, Affiliation-F, AUC-PR, VUS-PR; lower numbers are better for FPR, and calibration error (CalErr). Methods marked with * denote deep learning semi-supervised approaches; the best overall method is shown in \textbf{bold}, and the best non–semi-supervised method is \underline{underlined} when different from the bold one.}
\label{tab:placeholder_ext}
\end{table}

\begin{table}[]
\scriptsize
\begin{tabular}{llllllllll}
\toprule
Dataset & Forecaster & AD Model & PA-F1 \(\uparrow\) & Affiliation-F \(\uparrow\) & FPR \(\downarrow\) & CalErr \(\downarrow\) & AUC-PR \(\uparrow\) & VUC-PR \(\uparrow\) \\
\midrule
WSD & -& KShapeAD & 0.117 ± 0.210 & 0.722 ± 0.084 & 0.469 ± 0.361 & 0.162 ± 0.133 & 0.011 ± 0.023 & 0.061 ± 0.116 \\
WSD & -& POLY & 0.475 ± 0.337 & 0.862 ± 0.138 & 0.199 ± 0.333 & 0.281 ± 0.240 & 0.006 ± 0.010 & 0.226 ± 0.223 \\
WSD & -& Sub-KNN & 0.195 ± 0.237 & 0.755 ± 0.088 & 0.312 ± 0.422 & 0.054 ± 0.071 & 0.026 ± 0.066 & 0.103 ± 0.135 \\
WSD & -& Sub-PCA & 0.208 ± 0.296 & 0.747 ± 0.093 & 0.479 ± 0.393 & 0.205 ± 0.212 & 0.040 ± 0.110 & 0.102 ± 0.135 \\
\midrule
WSD & - & CNN* & \textbf{0.980 ± 0.038} & \textbf{0.970 ± 0.061} & \textbf{0.001 ± 0.001} & 0.712 ± 0.287 & 0.033 ± 0.035 & 0.216 ± 0.200 \\
WSD & - & OmniAnomaly* & 0.414 ± 0.431 & 0.804 ± 0.134 & 0.471 ± 0.470 & 0.328 ± 0.353 & 0.047 ± 0.116 & 0.090 ± 0.116 \\
WSD & - & USAD* & 0.102 ± 0.210 & 0.711 ± 0.061 & 0.602 ± 0.335 & 0.269 ± 0.209 & 0.009 ± 0.011 & 0.041 ± 0.059 \\
WSD & - & MOMENT\_ZS & 0.568 ± 0.238 & 0.944 ± 0.078 & 0.059 ± 0.194 & 0.504 ± 0.284 & 0.030 ± 0.061 & \textbf{0.394 ± 0.248} \\
\midrule
WSD & Chronos & $\mathcal{W}_1\textsc{-ACAS}$ & 0.868 ± 0.175 & 0.890 ± 0.096 & 0.096 ± 0.292 & 0.007 ± 0.016 & 0.224 ± 0.192 & 0.230 ± 0.226 \\
WSD & Chronos & conformal & 0.810 ± 0.193 & 0.882 ± 0.086 & 0.098 ± 0.292 & 0.006 ± 0.009 & 0.105 ± 0.120 & 0.227 ± 0.172 \\
WSD & Chronos & gaussian & 0.387 ± 0.207 & 0.788 ± 0.072 & 0.111 ± 0.283 & 0.025 ± 0.025 & 0.079 ± 0.085 & 0.226 ± 0.173 \\
\midrule
WSD & Tirex & $\mathcal{W}_1\textsc{-ACAS}$ & \underline{0.882 ± 0.159} & \underline{0.891 ± 0.090} & \underline{0.048 ± 0.210} & 0.007 ± 0.022 & 0.222 ± 0.190 & \underline{0.239 ± 0.228} \\
WSD & Tirex & conformal & 0.841 ± 0.173 & 0.886 ± 0.087 & 0.052 ± 0.222 & 0.006 ± 0.006 & 0.119 ± 0.115 & 0.238 ± 0.208 \\
WSD & Tirex & gaussian & 0.393 ± 0.210 & 0.783 ± 0.074 & 0.110 ± 0.283 & 0.023 ± 0.023 & 0.067 ± 0.074 & 0.231 ± 0.202 \\
\midrule
WSD & TTM & $\mathcal{W}_1\textsc{-ACAS}$ & 0.868 ± 0.172 & 0.882 ± 0.089 & 0.064 ± 0.228 & \textbf{0.005 ± 0.012} & \textbf{0.225 ± 0.198} & 0.236 ± 0.229 \\
WSD & TTM & conformal & 0.812 ± 0.174 & 0.879 ± 0.084 & 0.067 ± 0.229 & 0.007 ± 0.006 & 0.191 ± 0.146 & 0.237 ± 0.194 \\
WSD & TTM & gaussian & 0.389 ± 0.212 & 0.782 ± 0.076 & 0.112 ± 0.292 & 0.026 ± 0.032 & 0.063 ± 0.066 & 0.230 ± 0.196 \\
\midrule
\midrule
Stock & -& KShapeAD & 0.135 ± 0.072 & 0.680 ± 0.010 & 0.951 ± 0.095 & 0.081 ± 0.060 & 0.060 ± 0.036 & 0.603 ± 0.342 \\
Stock & -& POLY & 0.201 ± 0.089 & 0.720 ± 0.082 & 0.805 ± 0.360 & 0.245 ± 0.248 & 0.000 ± 0.000 & 0.615 ± 0.350 \\
Stock & -& Sub-KNN & 0.150 ± 0.087 & 0.678 ± 0.008 & 0.979 ± 0.027 & 0.175 ± 0.136 & 0.083 ± 0.067 & 0.627 ± 0.369 \\
Stock & -& Sub-PCA & 0.199 ± 0.086 & 0.726 ± 0.090 & 0.792 ± 0.335 & 0.128 ± 0.199 & 0.117 ± 0.068 & 0.844 ± 0.087 \\
Stock & - & SAND & 0.174 ± 0.100 & 0.687 ± 0.001 & 0.933 ± 0.086 & 0.137 ± 0.192 & 0.071 ± 0.027 & 0.549 ± 0.549 \\
\midrule
Stock & - & CNN* & \textbf{0.996 ± 0.003} & \textbf{0.999 ± 0.001} & \textbf{0.001 ± 0.000} & 0.872 ± 0.066 & 0.900 ± 0.105 & 0.980 ± 0.027 \\
Stock & - & OmniAnomaly* & 0.372 ± 0.038 & 0.886 ± 0.046 & 0.242 ± 0.122 & 0.688 ± 0.160 & 0.284 ± 0.054 & 0.962 ± 0.026 \\
Stock & - & USAD* & 0.146 ± 0.070 & 0.676 ± 0.008 & 0.983 ± 0.019 & \textbf{0.021 ± 0.027} & 0.068 ± 0.043 & 0.747 ± 0.149 \\
Stock & - & MOMENT\_ZS & 0.163 ± 0.071 & 0.680 ± 0.006 & 0.931 ± 0.045 & 0.049 ± 0.024 & 0.093 ± 0.051 & 0.598 ± 0.365 \\
\midrule
Stock & Chronos & $\mathcal{W}_1\textsc{-ACAS}$ & 0.959 ± 0.031 & 0.985 ± 0.008 & 0.009 ± 0.012 & 0.074 ± 0.040 & 0.973 ± 0.020 & \textbf{0.998 ± 0.001} \\
Stock & Chronos & conformal & 0.959 ± 0.039 & \underline{0.990 ± 0.009} & 0.009 ± 0.010 & 0.072 ± 0.043 & 0.841 ± 0.151 & 0.968 ± 0.034 \\
Stock & Chronos & gaussian & 0.958 ± 0.037 & \underline{0.990 ± 0.008} & 0.006 ± 0.007 & 0.175 ± 0.081 & 0.799 ± 0.197 & 0.974 ± 0.027 \\
\midrule
Stock & Tirex & $\mathcal{W}_1\textsc{-ACAS}$ & 0.955 ± 0.027 & 0.983 ± 0.008 & 0.010 ± 0.009 & 0.072 ± 0.043 & \textbf{0.984 ± 0.010} & 0.985 ± 0.024 \\
Stock & Tirex & conformal & 0.947 ± 0.039 & 0.985 ± 0.006 & 0.010 ± 0.007 & 0.079 ± 0.047 & 0.880 ± 0.104 & 0.987 ± 0.017 \\
Stock & Tirex & gaussian & 0.964 ± 0.034 & 0.989 ± 0.007 & 0.006 ± 0.005 & 0.179 ± 0.084 & 0.855 ± 0.133 & 0.986 ± 0.018 \\
\midrule
Stock & TTM & $\mathcal{W}_1\textsc{-ACAS}$ & \underline{0.967 ± 0.028} & 0.989 ± 0.008 & 0.009 ± 0.011 & 0.074 ± 0.041 & 0.963 ± 0.053 & 0.991 ± 0.015 \\
Stock & TTM & conformal & 0.963 ± 0.030 & 0.989 ± 0.007 & 0.008 ± 0.009 & \underline{0.071 ± 0.046} & 0.825 ± 0.184 & 0.975 ± 0.031 \\
Stock & TTM & gaussian & 0.965 ± 0.018 & 0.988 ± 0.005 & \underline{0.004 ± 0.004} & 0.182 ± 0.087 & 0.818 ± 0.164 & 0.974 ± 0.028 \\
\midrule
\midrule

IOPS & -& KShapeAD & 0.365 ± 0.358 & 0.703 ± 0.064 & 0.699 ± 0.337 & 0.171 ± 0.149 & 0.025 ± 0.028 & 0.049 ± 0.044 \\
IOPS & -& POLY & 0.493 ± 0.410 & 0.854 ± 0.113 & 0.214 ± 0.318 & 0.442 ± 0.308 & 0.042 ± 0.067 & 0.230 ± 0.121 \\
IOPS & -& Sub-KNN & 0.334 ± 0.342 & 0.695 ± 0.038 & 0.693 ± 0.378 & 0.137 ± 0.187 & 0.021 ± 0.029 & 0.073 ± 0.095 \\
IOPS & -& Sub-PCA & 0.497 ± 0.432 & 0.780 ± 0.112 & 0.360 ± 0.341 & 0.352 ± 0.236 & 0.059 ± 0.070 & 0.206 ± 0.158 \\
IOPS & - & SAND & 0.052 ± 0.038 & 0.703 ± 0.039 & 0.808 ± 0.196 & 0.091 ± 0.056 & 0.008 ± 0.004 & 0.082 ± 0.055 \\
\midrule
IOPS & - & CNN* & 0.865 ± 0.224 & 0.870 ± 0.091 & 0.018 ± 0.040 & 0.800 ± 0.271 & 0.102 ± 0.071 & 0.285 ± 0.165 \\
IOPS & - & OmniAnomaly* & 0.734 ± 0.275 & 0.803 ± 0.113 & 0.161 ± 0.261 & 0.639 ± 0.286 & 0.044 ± 0.038 & 0.207 ± 0.129 \\
IOPS & - & USAD* & 0.493 ± 0.348 & 0.771 ± 0.113 & 0.387 ± 0.314 & 0.402 ± 0.278 & 0.041 ± 0.044 & 0.130 ± 0.077 \\
IOPS & - & MOMENT\_ZS & 0.565 ± 0.347 & 0.870 ± 0.098 & 0.074 ± 0.124 & 0.665 ± 0.293 & 0.052 ± 0.060 & \textbf{0.337 ± 0.261} \\

\midrule
IOPS & Chronos & $\mathcal{W}_1\textsc{-ACAS}$ & 0.886 ± 0.147 & 0.882 ± 0.062 & 0.004 ± 0.008 & \textbf{0.005 ± 0.014} & 0.158 ± 0.143 & 0.291 ± 0.151 \\
IOPS & Chronos & conformal & 0.850 ± 0.175 & 0.884 ± 0.067 & 0.008 ± 0.013 & 0.006 ± 0.013 & 0.151 ± 0.160 & 0.298 ± 0.208 \\
IOPS & Chronos & gaussian & 0.543 ± 0.217 & 0.811 ± 0.035 & 0.024 ± 0.019 & 0.021 ± 0.020 & 0.109 ± 0.087 & 0.308 ± 0.208 \\
\midrule
IOPS & Tirex & $\mathcal{W}_1\textsc{-ACAS}$ & \textbf{0.921 ± 0.073} & \textbf{0.889 ± 0.061} & \textbf{0.003 ± 0.007} & 0.007 ± 0.014 & \textbf{0.184 ± 0.144} & 0.296 ± 0.167 \\
IOPS & Tirex & conformal & 0.875 ± 0.126 & 0.888 ± 0.061 & 0.007 ± 0.013 & 0.006 ± 0.012 & 0.151 ± 0.140 & 0.301 ± 0.207 \\
IOPS & Tirex & gaussian & 0.528 ± 0.232 & 0.800 ± 0.035 & 0.026 ± 0.020 & 0.023 ± 0.019 & 0.113 ± 0.089 & 0.304 ± 0.207 \\
\midrule
IOPS & TTM & $\mathcal{W}_1\textsc{-ACAS}$ & 0.871 ± 0.203 & 0.870 ± 0.075 & 0.009 ± 0.029 & \textbf{0.005 ± 0.012} & 0.167 ± 0.136 & 0.304 ± 0.153 \\
IOPS & TTM & conformal & 0.826 ± 0.212 & 0.875 ± 0.084 & 0.019 ± 0.050 & 0.007 ± 0.013 & 0.123 ± 0.102 & 0.313 ± 0.224 \\
IOPS & TTM & gaussian & 0.558 ± 0.228 & 0.809 ± 0.045 & 0.032 ± 0.048 & 0.021 ± 0.024 & 0.109 ± 0.092 & \underline{0.309 ± 0.215} \\

\bottomrule
\end{tabular}
\caption{\textbf{Performance Summary per datasets.}
Entries indicate the mean $\pm$ standard deviation computed by averaging within each dataset group. Higher numbers are better for PA-F1, Affiliation-F, AUC-PR, VUS-PR; lower numbers are better for FPR, and calibration error (CalErr). Methods marked with * denote deep learning semi-supervised approaches; the best overall method is shown in \textbf{bold}, and the best non–semi-supervised method is \underline{underlined} when different from the bold one.}
\label{tab:placeholder-v2}
\end{table}

%% file: ICLR2026/sections/appendix_forecast_tables.tex
\begin{table}
\scriptsize
\begin{tabular}{lccccccc}
\toprule
Dataset & IOPS & MSL & NAB & NEK & Stock & WSD & YAHOO \\
Forecaster &  &  &  &  &  &  &  \\
\midrule
MAE &  &  &  &  &  &  & \\
\midrule
Chronos & 1.50 $\pm$ 1.83 & 0.06 $\pm$ 0.06 & 244.96 $\pm$ 1108.10 & 0.38 $\pm$ 0.21 & 6.81 $\pm$ 3.89 & 149.73 $\pm$ 200.81 & 280.75 $\pm$ 232.69 \\
TiRex & 1.45 $\pm$ 1.77 & 0.06 $\pm$ 0.05 & 234.95 $\pm$ 1063.39 & 0.37 $\pm$ 0.22 & 6.85 $\pm$ 4.00 & 138.51 $\pm$ 182.07 & 250.80 $\pm$ 220.62 \\
TTM & 1.45 $\pm$ 1.74 & 0.09 $\pm$ 0.07 & 267.97 $\pm$ 1219.00 & 0.59 $\pm$ 0.41 & 7.59 $\pm$ 4.74 & 139.11 $\pm$ 183.01 & 481.91 $\pm$ 252.17 \\
\midrule
RMSE &  &  &  &  &  &  & \\
\midrule
Chronos & 2.38 $\pm$ 2.87 & 0.20 $\pm$ 0.19 & 341.38 $\pm$ 1487.54 & 0.75 $\pm$ 0.38 & 15.33 $\pm$ 9.43 & 212.32 $\pm$ 286.53 & 461.99 $\pm$ 433.56 \\
TiRex & 2.31 $\pm$ 2.78 & 0.20 $\pm$ 0.19 & 326.81 $\pm$ 1426.95 & 0.75 $\pm$ 0.39 & 15.40 $\pm$ 9.53 & 198.23 $\pm$ 261.90 & 417.93 $\pm$ 411.42 \\
TTM & 2.30 $\pm$ 2.74 & 0.22 $\pm$ 0.21 & 373.15 $\pm$ 1654.63 & 0.94 $\pm$ 0.54 & 15.37 $\pm$ 9.53 & 196.56 $\pm$ 262.49 & 683.50 $\pm$ 429.53 \\

\bottomrule
\end{tabular}
\caption{{\textbf{TSFM Forecasting Performance (MAE and RMSE) per Multivariate Dataset.}
Mean Absolute Error and Root Mean Squared Error for each TSFM model on the anomaly detection datasets, computed using a 15-step-ahead forecast and a context length of 52 past observations. Entries report mean $\pm$ standard deviation across all series within each dataset. Overall, forecasting performance is similar across models; the slightly higher error of TTM on YAHOO aligns with its correspondingly lower AD performance in Table~\ref{tab:placeholder}.}}
\label{tab:uv_tsfm_forecasting_error}
\end{table}

%% file: ICLR2026/sections/appendix_mv_tables.tex
\begin{table}[]
\scriptsize
\begin{tabular}{llllllllll}
\toprule
Dataset & Forecaster & AD Model & PA-F1 \(\uparrow\) & Affiliation-F \(\uparrow\) & FPR \(\downarrow\) & CalErr \(\downarrow\) & AUC-PR \(\uparrow\) & VUC-PR \(\uparrow\) \\
\midrule
TAO & - & CNN* & 0.998 ± 0.001 & 0.999 ± 0.000 & \textbf{0.000 ± 0.000} & 0.612 ± 0.044 & 0.895 ± 0.094 & 0.999 ± 0.001 \\
TAO & - & OmniAnomaly* & 0.377 ± 0.021 & 0.863 ± 0.053 & 0.321 ± 0.153 & 0.497 ± 0.136 & 0.311 ± 0.039 & 0.940 ± 0.051 \\
TAO & - & USAD* & 0.172 ± 0.061 & 0.679 ± 0.006 & 0.986 ± 0.018 & 0.033 ± 0.027 & 0.018 ± 0.005 & 0.097 ± 0.017 \\
\midrule
TAO & Chronos\_allctx & $\mathcal{W}_1\textsc{-ACAS}$-F & \textbf{1.000 ± 0.000} & \textbf{1.000 ± 0.000} & \textbf{0.000 ± 0.000} & 0.029 ± 0.028 & 0.901 ± 0.085 & 0.998 ± 0.003 \\
TAO & Chronos\_allctx & $\mathcal{W}_1\textsc{-ACAS}$-H & 0.999 ± 0.001 & \textbf{1.000 ± 0.000} & \textbf{0.000 ± 0.000} & 0.251 ± 0.110 & \textbf{0.945 ± 0.047} & 0.999 ± 0.001 \\
\midrule
TAO & Tirex\_allctx & $\mathcal{W}_1\textsc{-ACAS}$-F & \textbf{1.000 ± 0.000} & \textbf{1.000 ± 0.000} & \textbf{0.000 ± 0.000} & \textbf{0.026 ± 0.023} & 0.907 ± 0.100 & \textbf{1.000 ± 0.001} \\
TAO & Tirex-allctx & $\mathcal{W}_1\textsc{-ACAS}$-H & 0.999 ± 0.000 & \textbf{1.000 ± 0.000} & \textbf{0.000 ± 0.000} & 0.257 ± 0.109 & 0.938 ± 0.060 & 0.998 ± 0.003 \\
\midrule
\midrule

GECCO & - & CNN* & 0.583 & 0.875 & 0.139 & 0.860 & \textbf{0.294} & 0.152 \\
GECCO & - & OmniAnomaly* & 0.579 & 0.840 & 0.206 & 0.792 & 0.216 & 0.186 \\
GECCO & - & USAD* & 0.561 & 0.772 & 0.353 & 0.643 & 0.038 & 0.091 \\
\midrule
GECCO & Chronos-allctx & $\mathcal{W}_1\textsc{-ACAS}$-F & 0.704 & \textbf{0.882} & \textbf{0.030} & \textbf{0.014} & 0.231 & 0.236 \\
GECCO & Chronos-allctx & $\mathcal{W}_1\textsc{-ACAS}$-H & 0.606 & 0.835 & 0.044 & 0.316 & 0.110 & 0.127 \\
\midrule
GECCO & Tirex-allctx & $\mathcal{W}_1\textsc{-ACAS}$-F & \textbf{0.742} & 0.864 & 0.034 & 0.026 & 0.237 & \textbf{0.241} \\
GECCO & Tirex-allctx & $\mathcal{W}_1\textsc{-ACAS}$-H & 0.587 & 0.842 & 0.039 & 0.320 & 0.124 & 0.136 \\
\midrule
\midrule

LTDB & - &  CNN* & 0.910 ± 0.137 & \textbf{0.837 ± 0.105} & 0.228 ± 0.404 & 0.657 ± 0.368 & \textbf{0.247 ± 0.191} & 0.343 ± 0.272 \\
LTDB & - &  OmniAnomaly* & 0.875 ± 0.184 & 0.830 ± 0.106 & 0.297 ± 0.471 & 0.511 ± 0.445 & 0.211 ± 0.176 & 0.272 ± 0.204 \\
LTDB & - &  USAD* & 0.639 ± 0.380 & 0.866 ± 0.126 & 0.280 ± 0.481 & 0.540 ± 0.363 & 0.178 ± 0.277 & \textbf{0.453 ± 0.385} \\
\midrule
LTDB & Chronos-allctx & $\mathcal{W}_1\textsc{-ACAS}$-F & 0.845 ± 0.142 & 0.792 ± 0.056 & 0.065 ± 0.054 & 0.038 ± 0.039 & 0.227 ± 0.161 & 0.265 ± 0.187 \\
LTDB & Chronos-allctx & $\mathcal{W}_1\textsc{-ACAS}$-H & 0.910 ± 0.054 & 0.816 ± 0.024 & 0.068 ± 0.070 & 0.162 ± 0.119 & 0.170 ± 0.116 & 0.263 ± 0.185 \\
\midrule
LTDB & Tirex-allctx & $\mathcal{W}_1\textsc{-ACAS}$-F & 0.888 ± 0.101 & 0.814 ± 0.096 & \textbf{0.047 ± 0.050} & \textbf{0.026 ± 0.029} & 0.227 ± 0.157 & 0.260 ± 0.183 \\
LTDB & Tirex-allctx & $\mathcal{W}_1\textsc{-ACAS}$-H & \textbf{0.937 ± 0.041} & 0.829 ± 0.067 & 0.060 ± 0.067 & 0.134 ± 0.108 & 0.171 ± 0.114 & 0.263 ± 0.181 \\
\midrule
\midrule

Genesis & - &  CNN* & 0.649 & 0.852 & 0.004 & 0.896 & 0.031 & 0.045 \\
Genesis & - &  OmniAnomaly* & 0.473 & 0.873 & 0.009 & 0.784 & 0.018 & 0.028 \\
Genesis & - &  USAD* & 0.094 & 0.864 & 0.088 & 0.272 & 0.031 & \textbf{0.074} \\
\midrule
Genesis & Chronos-allctx & $\mathcal{W}_1\textsc{-ACAS}$-F & 0.793 & 0.891 & \textbf{0.002} & 0.001 & 0.075 & {0.066} \\
Genesis & Chronos-allctx & $\mathcal{W}_1\textsc{-ACAS}$-H & 0.850 & 0.903 & \textbf{0.002} & 0.058 & \textbf{0.084} & {0.066} \\
\midrule
Genesis & Tirex-allctx & $\mathcal{W}_1\textsc{-ACAS}$-F & 0.850 & \textbf{0.912} & \textbf{0.002} & \textbf{0.000} & 0.039 & 0.029 \\
Genesis & Tirex-allctx & $\mathcal{W}_1\textsc{-ACAS}$-H & \textbf{0.873} & 0.897 & 0.004 & 0.116 & 0.058 & 0.046 \\

\bottomrule
\end{tabular}
\caption{{\textbf{Performance Summary per Multivariate Dataset.}
Entries indicate the mean $\pm$ standard deviation computed by averaging within each dataset group. Higher numbers are better for PA-F1, Affiliation-F, AUC-PR, VUS-PR; lower numbers are better for FPR, and calibration error (CalErr). Standard deviations are omitted for GECCO and Genesis, as each corresponds to a single multivariate time series (9 features, 138k samples for GECCO; 18 features, 16k samples for Genesis).  Methods marked with * denote deep learning semi-supervised approaches; the best overall method is shown in \textbf{bold}.}}
\label{tab:mv_results_ext}
\end{table}